\documentclass{article}
\usepackage{iftex}
\RequirePDFTeX
\usepackage{iclr2027_conference,times}
\usepackage[utf8]{inputenc} 
\usepackage[T1]{fontenc}    
       
\usepackage{url}
\usepackage{xurl}            
\usepackage{array}
\usepackage{booktabs}       
\usepackage{amsfonts}       
\usepackage{nicefrac}       
\usepackage{microtype}      
\usepackage{xcolor}         
\usepackage{amsmath}
\usepackage{subcaption}
\usepackage{graphicx}
\usepackage{amssymb}
\usepackage{multirow} 
\usepackage[ruled]{algorithm2e}
\usepackage{algpseudocode}
\usepackage{hyperref}
\usepackage{tocloft}
\usepackage{titletoc}

\hypersetup{pdfauthor={},pdftitle={OScaR: The Occam's Razor for Extreme KV Cache Quantization in LLMs and Beyond}}
\title{\includegraphics[height=1.25em]{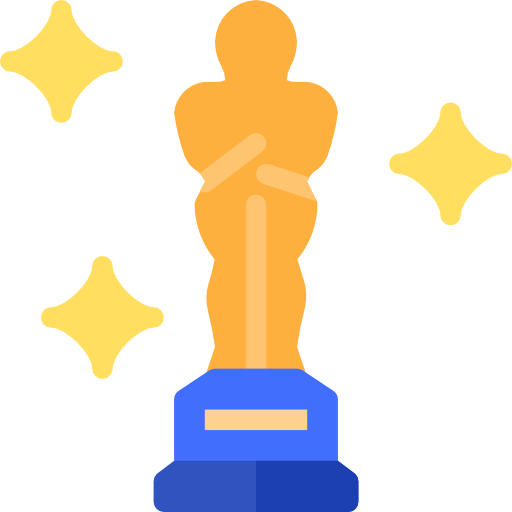}OScaR: The Occam's Razor for Extreme KV Cache Quantization in LLMs and Beyond}
\iclrfinalcopy
\author{
  Zunhai Su$^{1,2}$\enskip\enskip
  Rui Yang$^{2}$\enskip\enskip
  Chao Zhang$^{2}$\enskip\enskip
  Yaxiu Liu$^{3}$\enskip\enskip
  Yifan Zhang$^{2}$\enskip\enskip
  Wei Wu$^{2}$\enskip\enskip\\
  \textbf{Jing Xiong$^{1}$\enskip\enskip
  Dayou Du$^{4}$\enskip\enskip
  Xialie Zhuang$^{1}$\enskip\enskip
  Yulei Qian$^{2}$\enskip\enskip
  Jianchao Tan$^{2}$\enskip\enskip
  Yuchen Xie$^{2}$}\enskip\enskip\\
  \textbf{Yik-Chung Wu$^{1}$\enskip\enskip
  Hongxia Yang$^{5}$\enskip\enskip
  Ngai Wong$^{1}$}\enskip\enskip\\
  $^{1}$The University of Hong Kong \enspace
  $^{2}$Meituan LongCat Team \enspace
  $^{3}$Tsinghua University \enspace\\
  $^{4}$The University of Edinburgh \enspace
  $^{5}$The Hong Kong Polytechnic University \enspace
}
\begin{document}
\brokenpenalty=10000
\maketitle

\begin{abstract}
\vspace{-5mm}
The rapid advancement toward long-context reasoning and multi-modal intelligence has made the memory footprint of the Key-Value (KV) cache a dominant memory bottleneck for efficient deployment.
Extreme low-bit quantization has emerged as a fundamental imperative to reclaim memory efficiency and sustain high-throughput inference. While the established per-channel quantization effectively accommodates intrinsic channel-wise outliers in Key tensors, its efficacy diminishes under extreme compression.
In this work, we revisit the inherent limitations of the per-channel quantization paradigm from both empirical and theoretical perspectives. Our analysis identifies \textit{Token Norm Imbalance (TNI)} as the primary bottleneck to quantization fidelity. We demonstrate that TNI systematically amplifies errors when shared quantization parameters are required to span token groups exhibiting substantial norm disparities.
Instead of relying on intricate quantization pipelines (e.g., TurboQuant), we propose \textit{OScaR (\textbf{O}mni-\textbf{Sca}led Canalized \textbf{R}otation)}, an accurate and lightweight KV cache compression framework for X-LLMs (i.e., text-only, multi-modal, and omni-modal LLMs). 
Advancing the per-channel paradigm, OScaR employs \textit{Canalized Rotation} followed by \textit{Omni-Token Scaling} to mitigate TNI-induced sequence-dimensional variance both effectively and efficiently, further supported by our optimized system design and CUDA kernels.
Extensive evaluations across X-LLMs show that OScaR consistently outperforms existing methods and achieves near-lossless performance under INT2 quantization, establishing it as a robust, low-complexity, and universal framework that defines a new Pareto front.
Compared with the \texttt{BF16 FlashDecoding-v2} baseline, our OScaR implementation achieves a notable up to 3.0$\times$ speedup in decoding, reduces memory footprint by 5.3$\times$, and increases throughput by 4.1$\times$. 
The code for OScaR is publicly available at \href{https://github.com/ZunhaiSu/OScaR-KV-Quant}{\texttt{https://github.com/ZunhaiSu/OScaR-KV-Quant}}.
\vspace{-5mm}
\end{abstract}

\begin{figure}[h]
    \centering
    \vspace{-5mm}
    \includegraphics[width=1\textwidth]{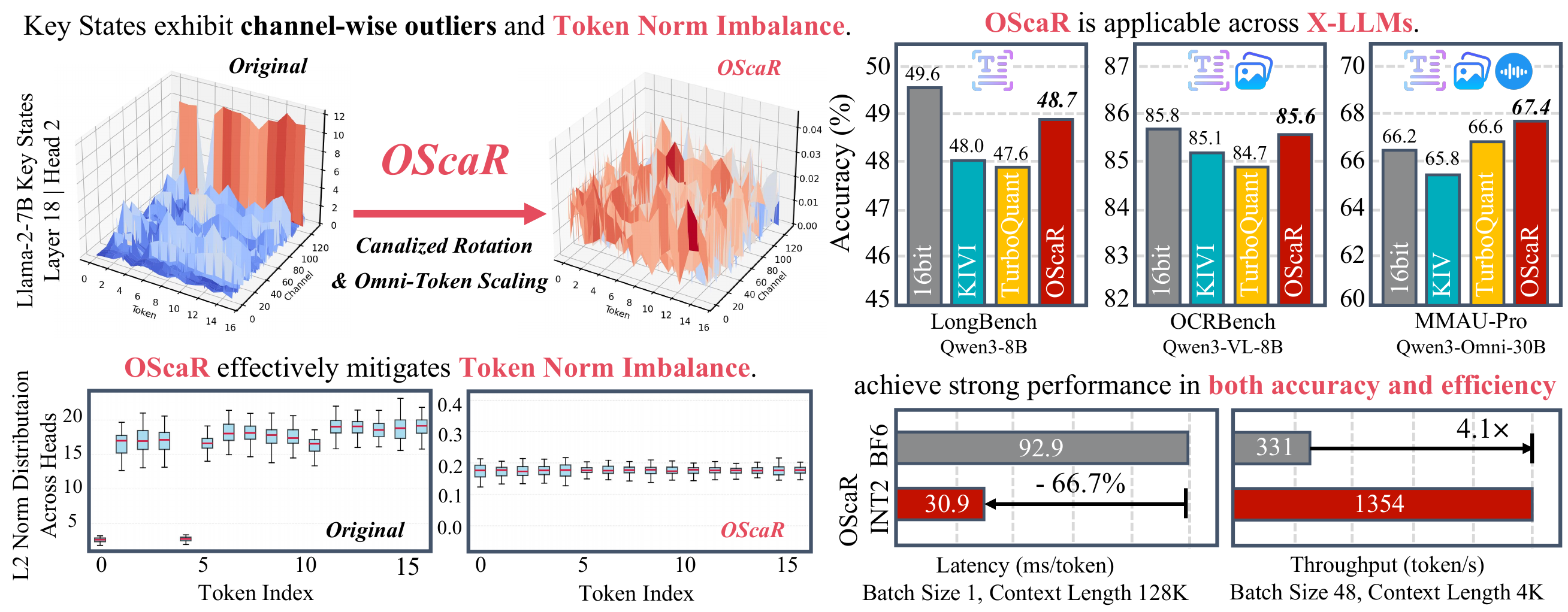}
    \vspace{-4mm}
    \caption{We revisit the per-channel key quantization paradigm and identify its inherent limitation, termed \textit{token norm imbalance (TNI)}. We then propose OScaR, a streamlined framework that applies \textit{Canalized Rotation} followed by \textit{Omni-Token Scaling} to effectively mitigate TNI. Extensive evaluations across X-LLMs demonstrate that OScaR establishes a superior accuracy-efficiency Pareto front.\looseness=-1}
    \vspace{-5mm}
    \label{fig:paper overview}
    
\end{figure}

\section{Introduction}
Recent advancements in large language models (LLMs) and their multi-modal counterparts have demonstrated remarkable capabilities in complex reasoning and multi-modal perception~\citep{team2025longcat1,team2025introducing,team2026longcat1,team2025longcat3,team2025longcat2}, establishing a new foundation for artificial intelligence. To further unlock these emergent abilities, the research frontier is increasingly prioritizing long-context processing, streaming tasks, and long-range audio-video multi-modal understanding~\citep{wang2026longcat,team2025longcat2,team2025longcat4,team2026longcat2}.
However, these trends necessitate handling massive context sequences, causing the memory footprint of the Key-Value (KV) cache to grow linearly and dominate total memory consumption~\citep{li2024survey,liu2025kv}. In memory-bound inference scenarios, the KV cache rapidly exhausts the High Bandwidth Memory (HBM) capacity of modern accelerators, severely restricting batch sizes and hindering efficient large-scale deployment~\citep{liu2024kivi,hooper2024kvquant,ge2023model,liu2024minicache}. Consequently, reclaiming memory efficiency while sustaining high-throughput inference has become a fundamental imperative for next-generation LLMs~\citep{team2025kimi,cao2026qwen3,team2026longcat2,team2025longcat4}.

To address these constraints, KV cache compression has matured into a significant research frontier, with methodologies such as quantization, pruning, and low-rank decomposition being extensively explored~\citep{liu2024kivi,hooper2024kvquant,liu2024minicache,ge2023model,wan2024look,cai2024pyramidkv}. By mapping high-precision tensors to reduced bit-widths, quantization reduces memory overhead without compromising the structural integrity of the KV cache~\citep{li2024survey,liu2024kivi,hooper2024kvquant}.
Within the landscape of KV cache quantization, Key quantization has emerged as a focal point, posing more substantial challenges than Value quantization due to salient channel-wise outliers~\citep{liu2024kivi,hooper2024kvquant,su2025rotatekv,jin2025massive}. Specifically, a sparse subset of channels within Key tensors often exhibits disproportionately large magnitudes. To mitigate this, per-channel Key quantization, which leverages intrinsic distributional characteristics, has proven to be a promising approach~\citep{liu2024kivi,hooper2024kvquant,su2025accurate,tao2025plug,su2026xstreamvggt,zandieh2025qjl} (see also Appendix~\ref{app:related_work}).

Although the per-channel quantization paradigm has achieved notable success, its effectiveness progressively diminishes under extreme compression~\citep{liu2024kivi,duanmu2024skvq,su2025rotatekv,zandieh2025turboquant}. In this study, we revisit the inherent limitations of per-channel quantization. Through a meticulous token-wise norm distribution analysis of KV caches across multiple text-only and multi-modal LLMs, we identify a pervasive structural property, which we term \textit{Token Norm Imbalance (TNI)}.
Intuitively, TNI undermines per-channel quantization because shared quantization parameters must accommodate token groups with highly divergent norms~\citep{nagel2021white}. Our empirical validation confirms that TNI systematically amplifies quantization error. Going beyond empirical exploration, our theoretical analysis further corroborates TNI-induced error amplification within per-channel quantization, revealing TNI as a fundamental vulnerability of the per-channel paradigm.\looseness=-1

Existing KV cache quantization methods often lean heavily on auxiliary mechanisms to suppress quantization errors~\citep{zandieh2025turboquant,pope2026rotorquant,zandieh2025qjl,han2025polarquant}.
These intricate pipelines, coupled with unavoidable on-the-fly quantization, introduce substantial computational overhead and extra parameters, undermining practical viability.
\textbf{Guided by the principle of Occam's Razor, we advocate for elegance and simplicity over intricate, heavy-weight quantization pipelines.} 
To this end, we introduce \textit{OScaR (\textbf{O}mni-\textbf{Sca}led Canalized \textbf{R}otation)}, an accurate and lightweight KV cache quantization framework designed for X-LLMs (i.e., text-only, multi-modal, and omni-modal LLMs).
As discussed in Section~\ref{sec:OScaR}, building upon the established per-channel paradigm, OScaR first applies the Hadamard transform to prevent \textit{Scaling-Induced Outlier Artifacts} from biasing the subsequent token scaling process (\textit{Canalized Rotation}).
Subsequently, \textit{Omni-Token Scaling} performs omnidirectional sequence-level normalization to effectively mitigate the impact of diverse TNI patterns.
The resulting pipeline remains training-free and highly streamlined, with both components being mutually essential.

Our empirical evaluations, along with theoretical complexity analyses across a diverse set of representative methods, demonstrate that OScaR's methodology is both robust and computationally efficient. Moreover, OScaR is built upon our carefully optimized system design and CUDA kernels, ensuring hardware efficiency and immediate deployability.  
Figure~\ref{fig:paper overview} provides a comprehensive overview of our paper.  
\textbf{The main contributions of our work are summarized as follows:}

\begin{itemize}
    \item \textbf{Unveiling TNI as the Structural Bottleneck of Per-Channel Quantization:} We identify Token Norm Imbalance (TNI) as the fundamental bottleneck limiting per-channel quantization in X-LLMs, supported by empirical evaluations and theoretical analysis.
    
    \item \textbf{Streamlined OScaR Framework:} Guided by the principle of Occam's Razor, we introduce OScaR, an accurate and lightweight KV cache quantization framework for X-LLMs. It first applies \textit{Canalized Rotation} to prevent \textit{Scaling-Induced Outlier Artifacts}, followed by \textit{Omni-Token Scaling} to safely mitigate the impact of TNI.
        
    \item \textbf{Redefining the Pareto Front:} Extensive evaluations across X-LLMs demonstrate that OScaR outperforms existing methods while achieving near-lossless performance under INT2 quantization. By preserving high quantization fidelity and maintaining low overall complexity, OScaR establishes an advantageous accuracy-efficiency Pareto front.
        
    \item \textbf{Optimized CUDA Implementations and Efficiency Gains:} We provide a carefully optimized system design and dedicated CUDA kernels that translate theoretical insights into tangible performance improvements. Compared with the \texttt{BF16 FlashDecoding-v2} baseline, our implementation achieves up to a 3.0$\times$ decoding speedup, reduces memory footprint by 5.3$\times$, and increases inference throughput by 4.1$\times$.
\end{itemize}

\begin{figure}[t]
    \vspace{-7mm}
    \centering
    \begin{subfigure}[b]{0.24\textwidth}
        \centering
        \includegraphics[width=\textwidth]{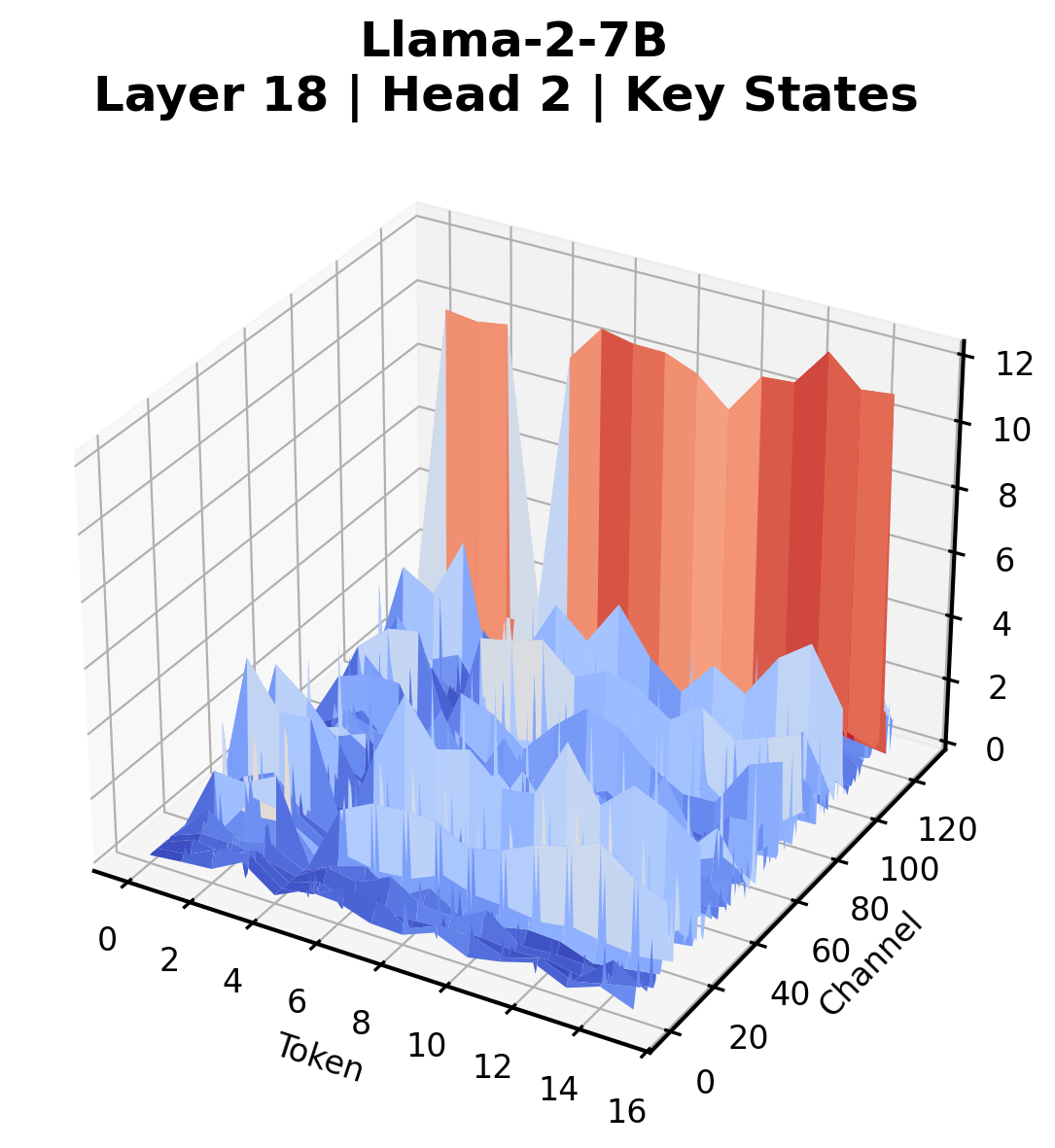}
        \caption{Key distribution.}
        \label{fig:key_magnitude}
    \end{subfigure}
    \hfill
    \begin{subfigure}[b]{0.24\textwidth}
        \centering
        \includegraphics[width=\textwidth]{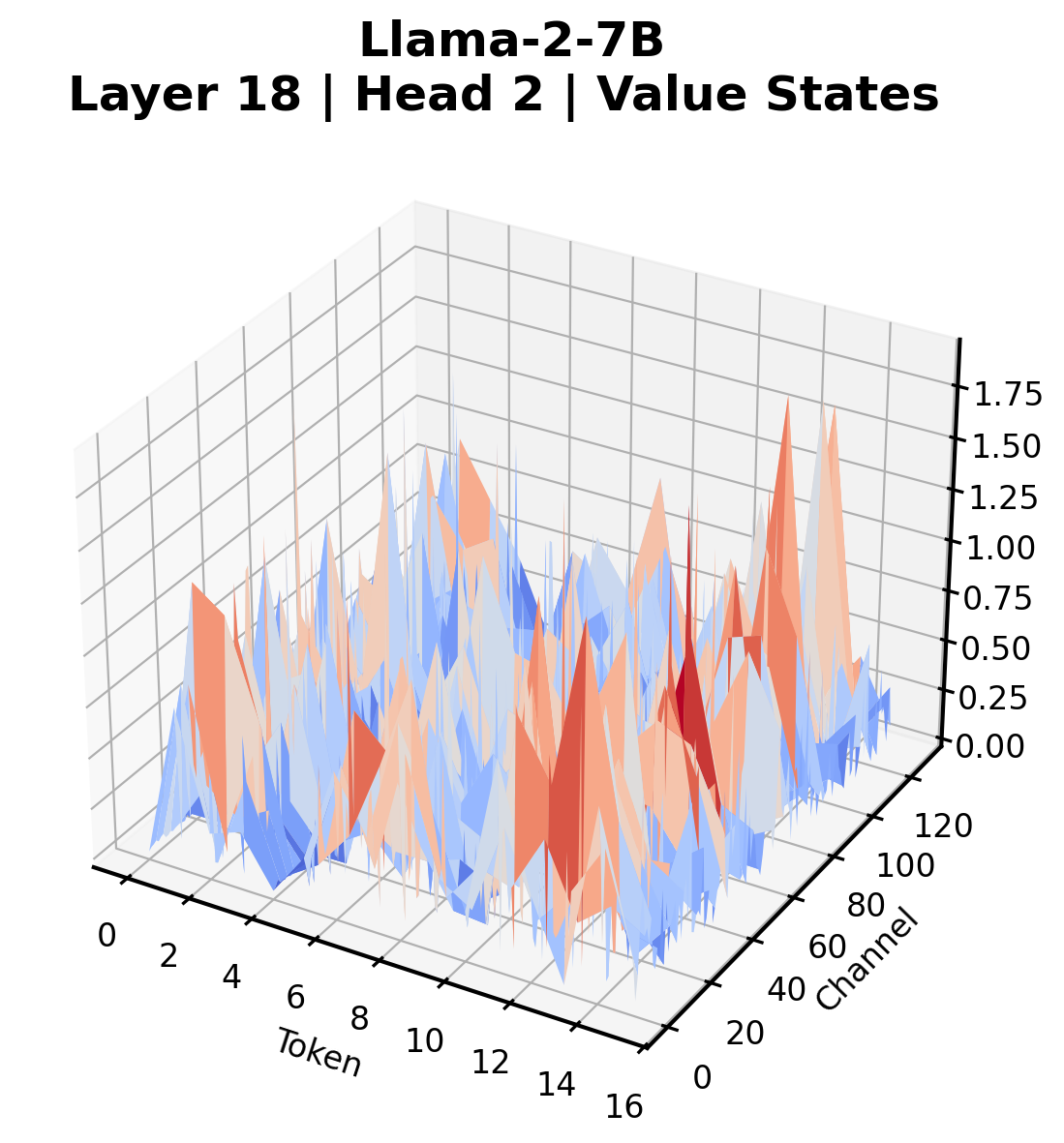}
        \caption{Value distribution.}
        \label{fig:value_magnitude}
    \end{subfigure}
    \hfill
    \begin{subfigure}[b]{0.48\textwidth}
        \centering
        \includegraphics[width=\textwidth]{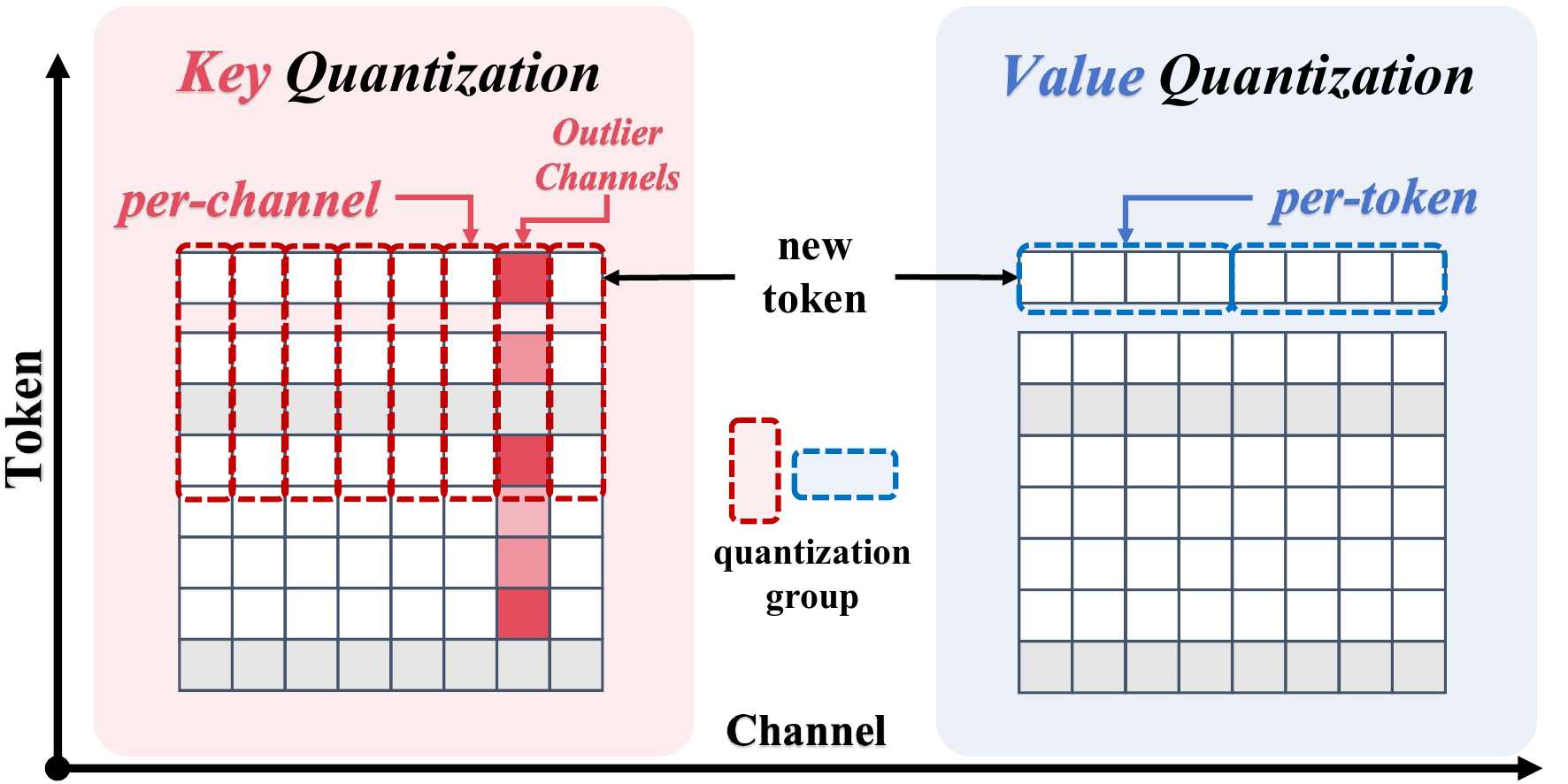}
    \caption{Schematic illustration of KIVI.}
        \label{fig:per_channel_quant}
    \end{subfigure}
    \caption{Key and Value magnitude patterns and the KIVI quantization scheme~\citep{liu2024kivi}. Key states exhibit significant channel-wise outliers, necessitating per-channel quantization. In contrast, Value states have a relatively uniform magnitude distribution and are quantized per-token.}
    \label{fig:kivi_overview}
\end{figure}

\section{Preliminaries}
\subsection{KV Caching in Autoregressive Inference}

LLMs predominantly employ a Transformer decoder-only architecture, where KV caching eliminates redundant computations during autoregressive decoding~\citep{vaswani2017attention,liu2025kv,li2024survey}. In multi-modal configurations, the LLM backbone integrates heterogeneous tokens from modality-specific encoders, projecting them into a shared latent space~\citep{team2025longcat2,team2025longcat4,liu2023visual,liu2024improved}. 
During the \textit{prefill stage}, textual tokens $\mathbf{X}_T$, visual features $\mathbf{X}_V$, and audio embeddings $\mathbf{X}_A$ are concatenated along the sequence dimension to form the prompt sequence $\mathbf{X}_{\text{prompt}} = [\mathbf{X}_T; \mathbf{X}_V; \mathbf{X}_A] \in \mathbb{R}^{S \times D}$, where $S$ is the total sequence length and $D$ the hidden dimension. For each Transformer layer $l \in \{1, \dots, \mathcal{L}\}$, the hidden state $\mathbf{H}^{(l-1)}$ is linearly projected to obtain the Key and Value states forming the initial KV cache:
\begin{equation}
K^{(l)} = \mathbf{H}^{(l-1)} W_K^{(l)}, \quad V^{(l)} = \mathbf{H}^{(l-1)} W_V^{(l)},
\end{equation}
where $\mathbf{H}^{(0)} = \mathbf{X}_{\text{prompt}}$, and $W_K^{(l)}, W_V^{(l)}\in\mathbb{R}^{D\times D_{\mathrm{kv}}}$ are Key/Value projection weights, with $D_{\mathrm{kv}}=H_{\mathrm{kv}}d_h$ for $H_{\mathrm{kv}}$ KV heads of dimension $d_h$.
During the \textit{decoding stage}, for each layer $l \in \{1, \dots, \mathcal{L}\}$ and step $t$, the input $\mathbf{h}_t^{(l-1)} \in \mathbb{R}^{1 \times D}$ is projected to produce the Query, Key, and Value vectors:
\begin{equation}
\mathbf{q}_t^{(l)} = \mathbf{h}_t^{(l-1)} W_Q^{(l)}, \quad 
\mathbf{k}_t^{(l)} = \mathbf{h}_t^{(l-1)} W_K^{(l)}, \quad 
\mathbf{v}_t^{(l)} = \mathbf{h}_t^{(l-1)} W_V^{(l)}.
\end{equation}
The KV cache for each layer is updated by concatenating the new vectors: $K^{(l)} \leftarrow [K^{(l)}; \mathbf{k}_t^{(l)}], \; V^{(l)} \leftarrow [V^{(l)}; \mathbf{v}_t^{(l)}]$. The KV cache memory footprint grows linearly with sequence length, creating a memory-bound bottleneck that motivates compression.

\subsection{Block-Wise Per-Channel Quantization}

Key states exhibit significant channel-wise outliers, while Value states have a relatively uniform magnitude distribution, as shown in Figure~\ref{fig:kivi_overview}. Exploiting these distinct numerical distributions, a range of approaches adopt a hybrid quantization scheme that applies per-channel quantization to Keys while preserving per-token granularity for Values~\citep{liu2024kivi,su2025accurate,su2026xstreamvggt,hooper2024kvquant}. To integrate per-channel quantization into token-wise LLM decoding, the pioneering KIVI framework introduces a block-wise per-channel quantization strategy for the Key cache~\citep{liu2024kivi}.
Specifically, given a Key cache $\mathbf{K} \in \mathbb{R}^{S \times d}$, where $S$ denotes the sequence length and $d$ the head dimension, each channel is partitioned into consecutive blocks of size $G$ for quantization. For the $j$-th channel within block $g$, the quantization step size $\Delta_{j,g}$ and zero-point $z_{j,g}$ are computed as:
\begin{equation}
\Delta_{j,g} = \frac{\max_{i \in g} K_{i,j} - \min_{i \in g} K_{i,j}}{2^b - 1}, \qquad
z_{j,g} = \left\lfloor -\frac{\min_{i \in g} K_{i,j}}{\Delta_{j,g}} \right\rceil.
\end{equation}
Each element $K_{i,j}$ is then quantized and reconstructed as:
\begin{equation}
Q(K_{i,j}) = \text{clamp}\left( \left\lfloor \frac{K_{i,j}}{\Delta_{j,g}} \right\rceil + z_{j,g},\, 0,\, 2^b-1 \right), \qquad
\hat{K}_{i,j} = \Delta_{j,g} \cdot \bigl( Q(K_{i,j}) - z_{j,g} \bigr).
\end{equation}

Importantly, a high-precision residual window mechanism is required to support continuous per-channel quantization during autoregressive generation: newly generated tokens are appended to this buffer, maintained in full precision, and block-wise quantized only once the buffer accumulates the predefined residual number $R$. Background on low-bit quantization is provided in Appendix~\ref{app:lowbit}.

\section{Methodology}
\label{sec:method}

\begin{figure}[t]
    \centering
    \begin{subfigure}[b]{0.325\textwidth}
        \includegraphics[width=\textwidth]{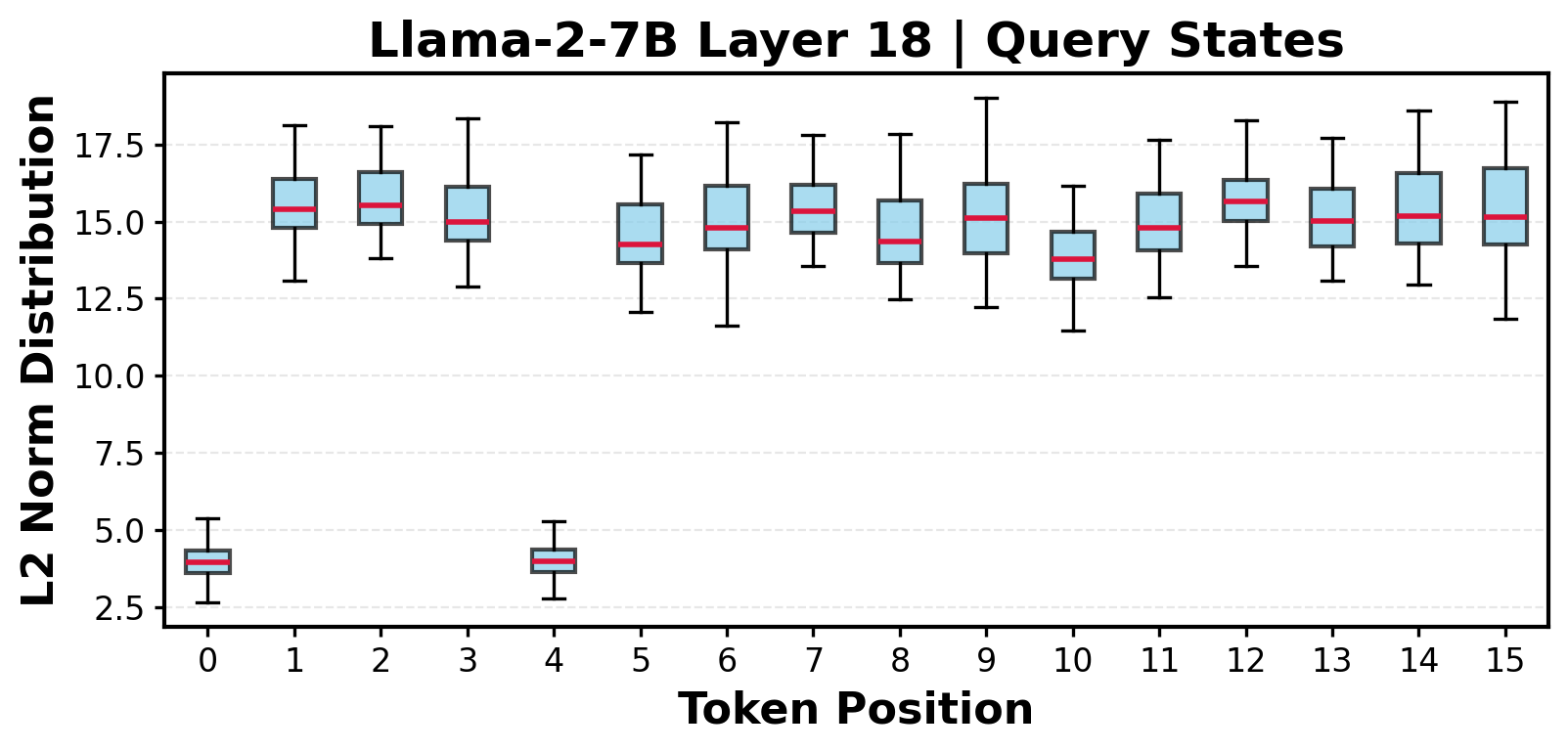}
        \caption{Query L2 norm distribution}
    \end{subfigure}
    \begin{subfigure}[b]{0.325\textwidth}
        \includegraphics[width=\textwidth]{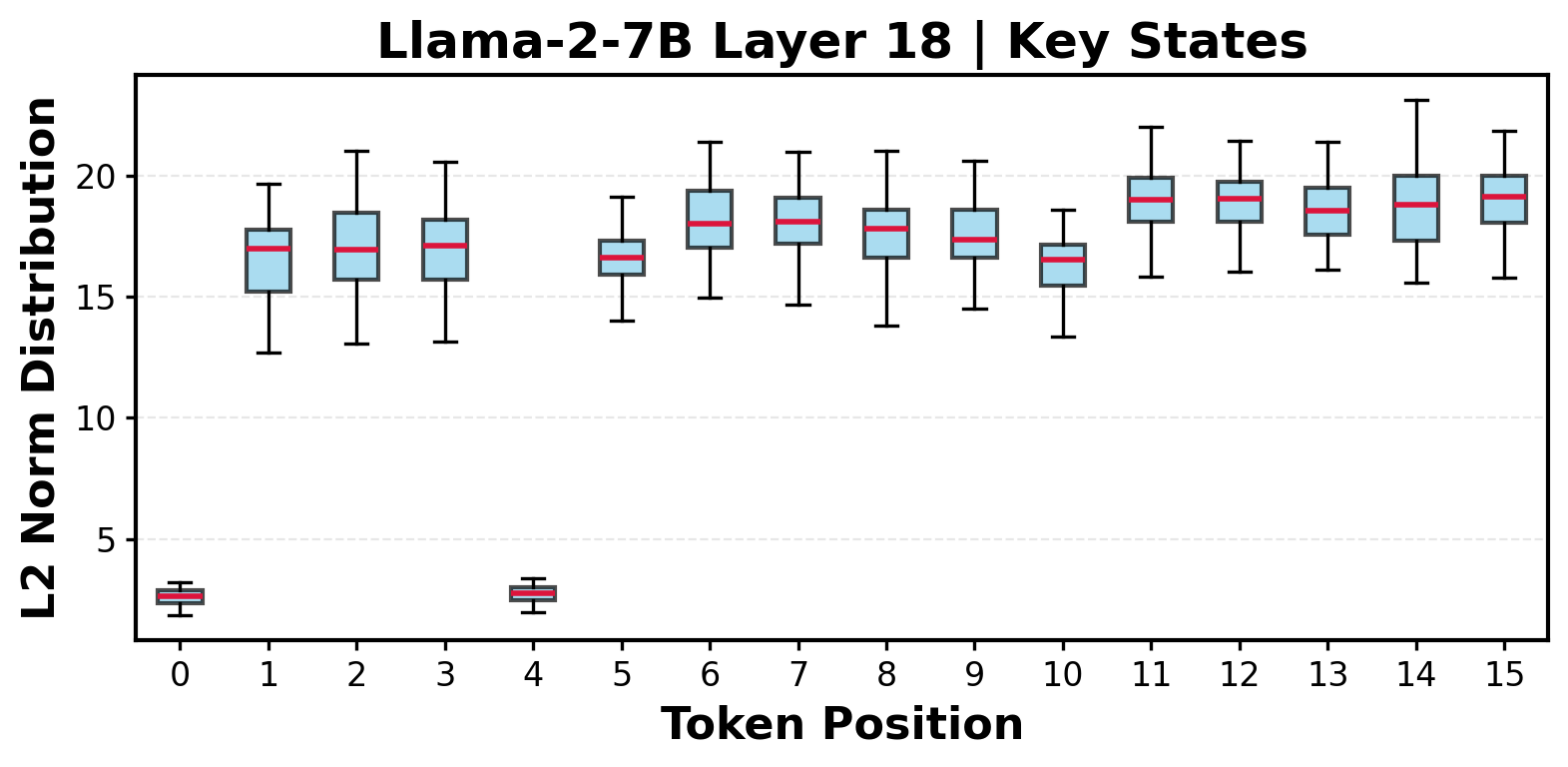}
        \caption{Key L2 norm distribution}
    \end{subfigure}
    \begin{subfigure}[b]{0.325\textwidth}
        \includegraphics[width=\textwidth]{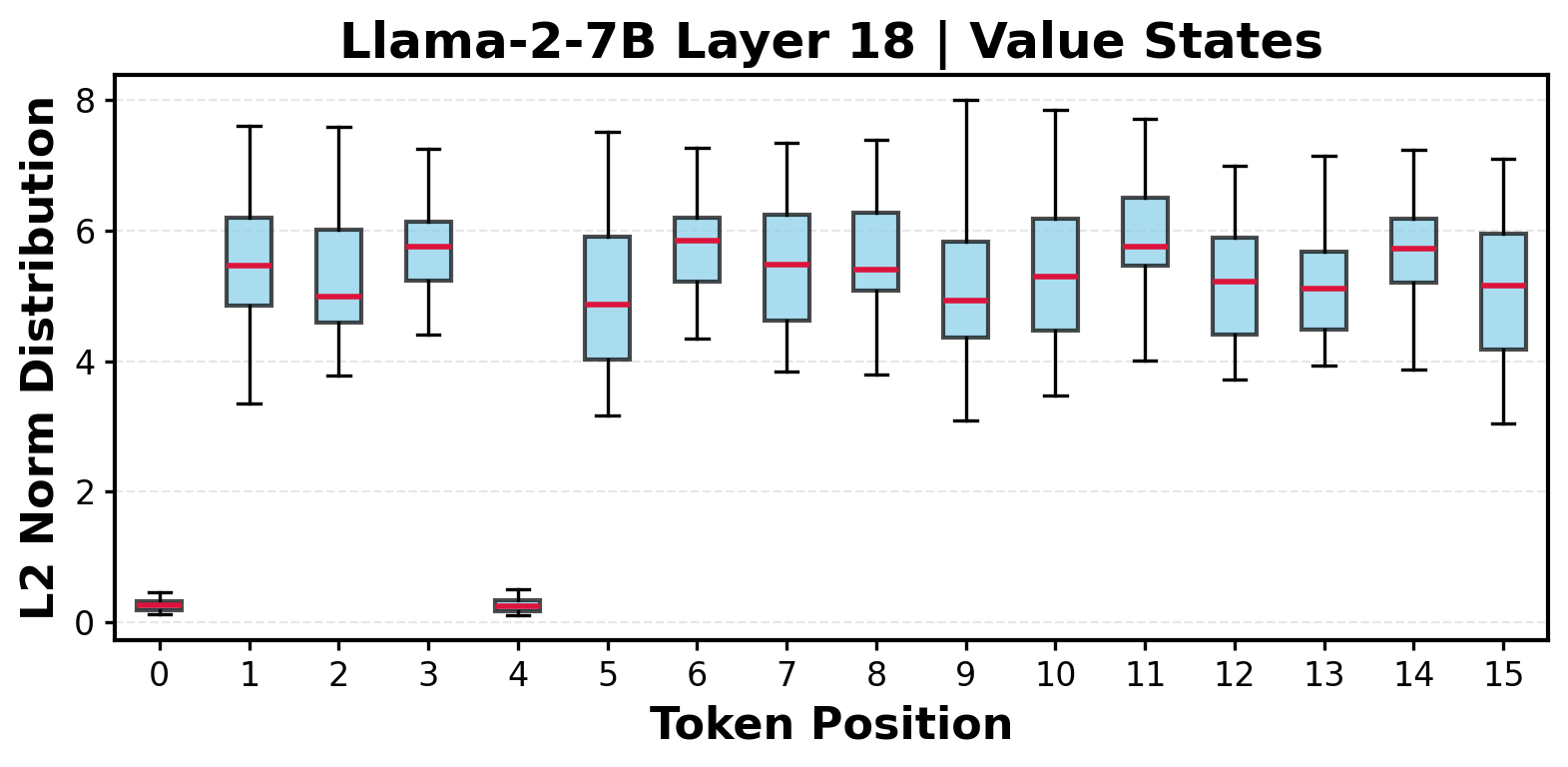}
        \caption{Value L2 norm distribution}
    \end{subfigure}
    
    \vspace{0.1cm}
    
    \begin{subfigure}[b]{0.325\textwidth}
        \centering
        \includegraphics[width=\textwidth]{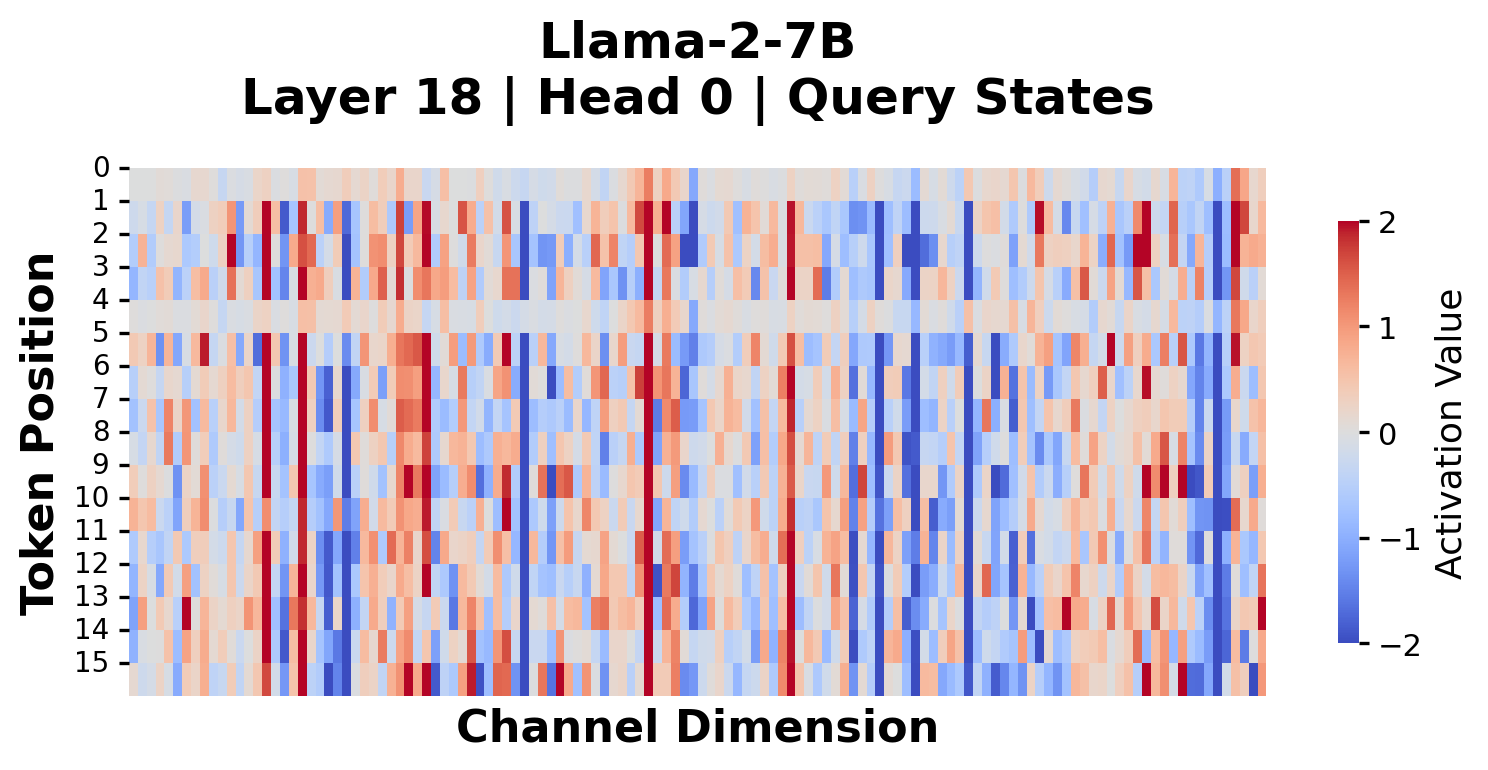}
        \caption{Query heatmap}
    \end{subfigure}
    \begin{subfigure}[b]{0.325\textwidth}
        \centering
        \includegraphics[width=\textwidth]{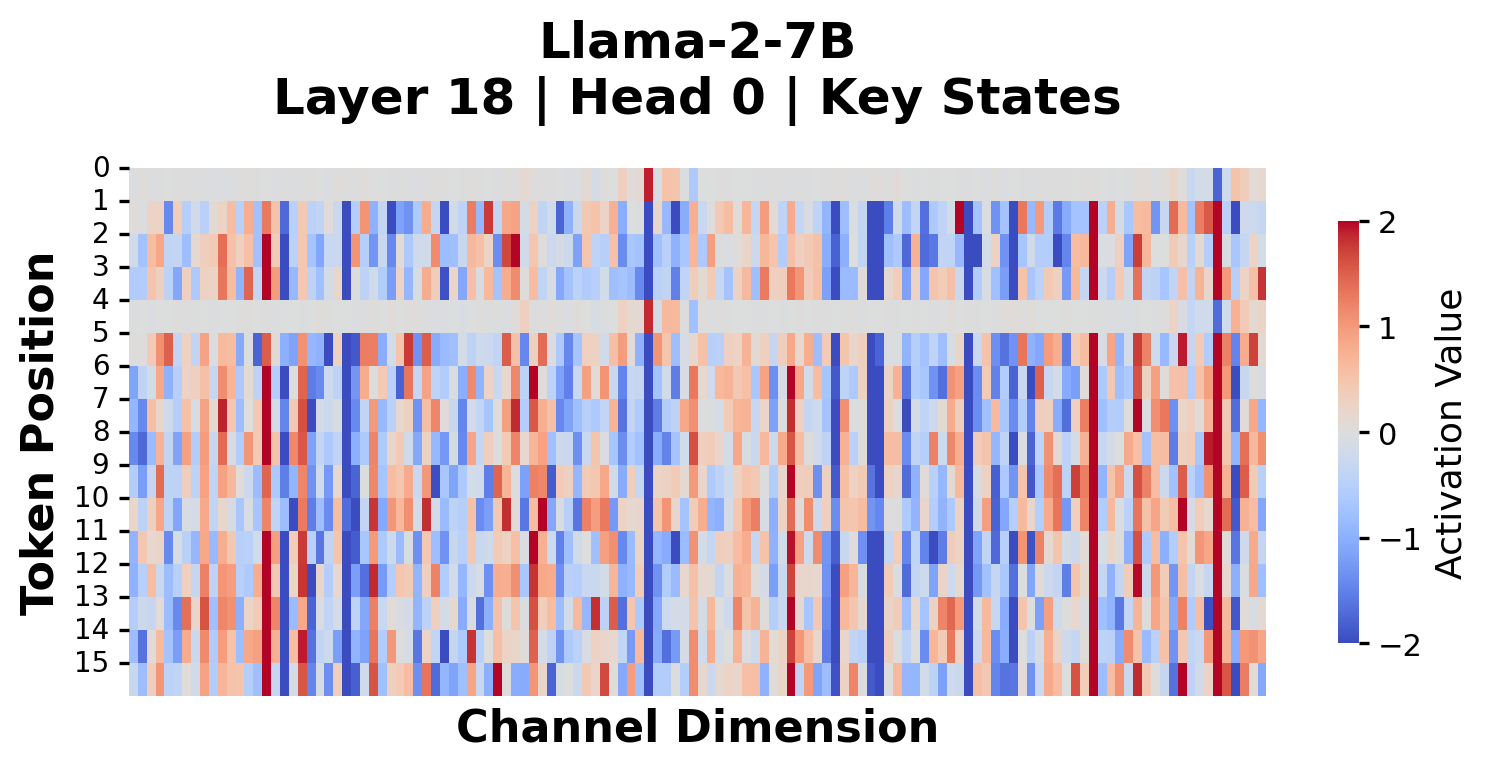}
        \caption{Key heatmap}
    \end{subfigure}
    \begin{subfigure}[b]{0.325\textwidth}
        \centering
        \includegraphics[width=\textwidth]{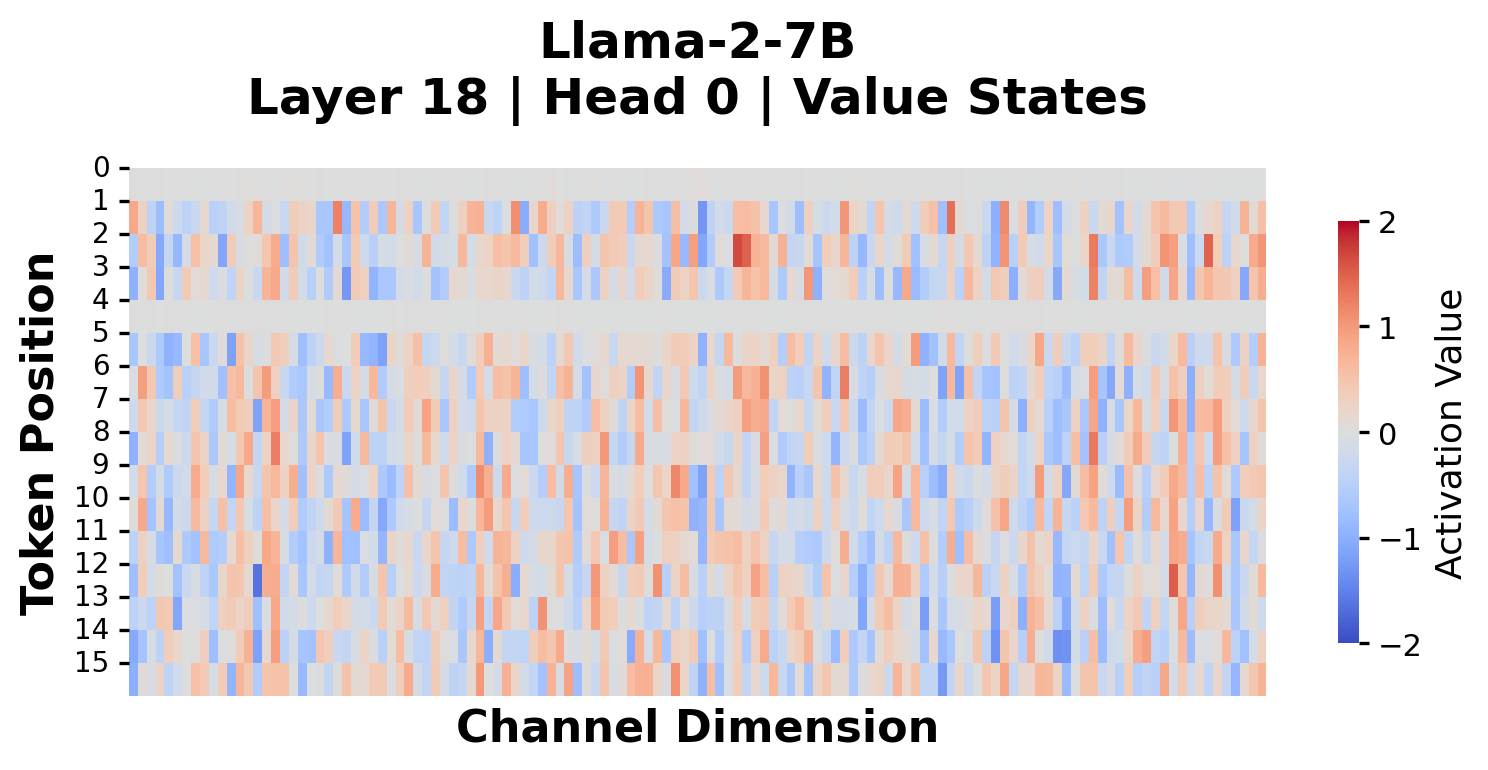}
        \caption{Value heatmap}
    \end{subfigure}
    
    \caption{L2 norm distributions (top row) and heatmaps (bottom row) of Query, Key, and Value states. Each attention state contains a sparse yet consistent subset of tokens with exceptionally low norms.}
    \label{fig:TNI}
\end{figure}

This section is organized into three parts. 
Section~\ref{sec:revisiting} revisits the inherent limitations of per-channel quantization and establishes \textit{Token Norm Imbalance} as the fundamental bottleneck. 
Section~\ref{sec:OScaR} introduces OScaR and its algorithmic design, which comprises \textit{Canalized Rotation} followed by \textit{Omni-Token Scaling}. 
Section~\ref{sec:method-pipeline} presents our efficient system design and CUDA implementations.

\subsection{Revisiting Per-Channel Key Quantization}
\label{sec:revisiting}

While per-channel quantization mitigates channel-wise outliers, it inherently assumes that tokens within a given channel share similar magnitudes. When the within-channel distribution becomes skewed or contains even a few divergent tokens, the shared quantization parameters for that block are severely compromised, causing substantial fidelity degradation~\citep{nagel2021white}. In this subsection, we systematically examine this assumption through (i) empirical observations, (ii) theoretical derivations, and (iii) quantitative error analysis.

\begin{figure}[!t]
    \centering   
    \vspace{-16pt}
    \includegraphics[width=1\linewidth]{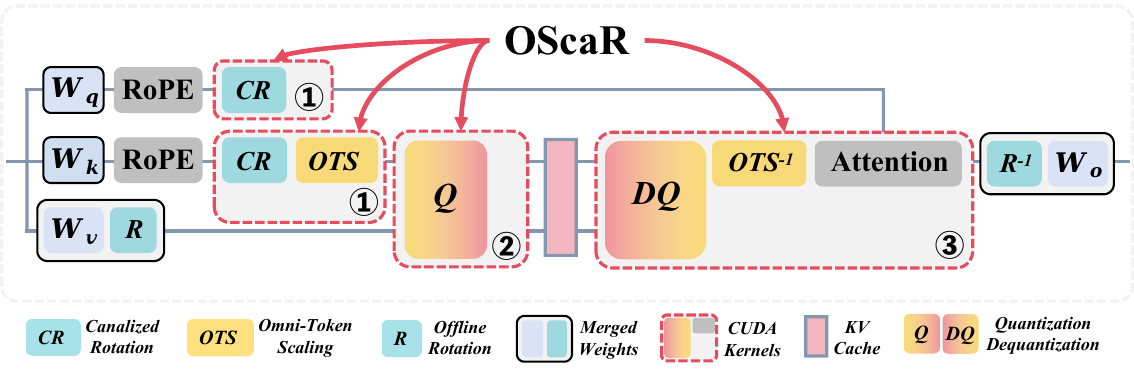}
    
    \captionsetup{skip=5pt}
    \caption{Conceptual overview of OScaR. The detailed algorithm is presented in Algorithm~\ref{algo:oscar}.}
    
    \label{fig:OScaR}
\end{figure}

\paragraph{Empirical Observations.} Our analysis is conducted across multiple mainstream open-source LLMs and multi-modal LLMs with fixed inputs (e.g., prompts, images). Systematic token-wise norm distribution profiling of KV caches consistently reveals substantial inter-token norm disparity, which we term \textit{Token Norm Imbalance (TNI)}. Specifically, our experimental procedure is as follows. For each token position $t$ in a transformer layer, we compute its $\ell_2$ norm across all attention heads for the Query, Key, and Value states. These head-wise norms are aggregated into the set
\begin{equation}
\mathcal{N}_t^{(M)} = \left\{ \|\mathbf{t}_{t,h}^{(M)}\|_2 \;\middle|\; h = 1, \dots, H \right\}, \quad
\|\mathbf{t}_{t,h}^{(M)}\|_2 = \sqrt{\sum_{j=1}^{d_h} \left(s_{t,h,j}^{(M)}\right)^2},
\end{equation}
where $d_h$ is the head dimension and $s_{t,h,j}^{(M)}$ denotes the $j$-th component of the token vector in head $h$ for state $M \in \{\text{Query}, \text{Key}, \text{Value}\}$. The set $\mathcal{N}_t^{(M)}$ captures token variation across attention heads and serves as the basis for boxplot visualizations, where each token is represented by a single box illustrating the distribution of its head-wise norms.

Visualizations based on Llama-2-7B are shown in Figure~\ref{fig:TNI}. Additional results for text-only LLMs (Llama-3.1-8B, Qwen-3-8B) and the prompt used are provided in Appendix~\ref{app:llm}. These results reveal significant outlier tokens as a manifestation of TNI. Specifically, each attention state contains a sparse yet consistent subset of tokens with exceptionally low norms. Their presence expands the quantization dynamic range for the corresponding block, representing the weakest link in the per-channel paradigm. Moreover, these low-norm outlier tokens consistently appear across different attention states and correspond directly to Attention Sink tokens~\citep{su2026attention,xiao2023efficient}, aligning with prior findings~\citep{su2025kvsink}. Appendix~\ref{app:AS} discusses Attention Sink tokens as low-norm outliers.

Beyond text-only LLMs, extensive TNI observations also hold in multi-modal LLMs. In such settings, TNI manifests not only as attention-sink-related outlier tokens but also through several distinct patterns: (i) broader token norm variation relative to text-only LLMs (Figure~\ref{fig:TNI-qwen-3-vl-8b-layer-24}); (ii) inter-modality norm disparities, wherein norms remain smooth within each modality yet diverge substantially across modalities (Figure~\ref{fig:TNI-qwen-3-vl-8b-layer-0}); and (iii) exceptionally large-norm outlier tokens, which contrast with the low-norm Attention Sink (Figure~\ref{fig:TNI-qwen-3-vl-8b-layer-15}). Appendix~\ref{app:mllm} provides representative visualizations.

\paragraph{Theoretical Derivations.}
Building on the empirical observations of TNI across X-LLMs, we provide theoretical derivations of TNI-induced errors in per-channel quantization. Detailed derivations are presented in Appendix~\ref{app:theoretical_derivations}, with an attention-logit analysis in Appendix~\ref{app:logit_error}. As shown in Equation~\ref{eq:core_bound}, the reconstruction error of a per-channel quantization block is fundamentally governed by the range of token norms within the block. Thus, TNI systematically amplifies quantization errors, revealing TNI as a fundamental vulnerability of the per-channel paradigm.

\paragraph{Quantitative Error Analysis.}
We conduct an empirical quantization error analysis under extreme KV cache compression to comprehensively quantify the impact of TNI. As shown in Table~\ref{tab:quantization_error}, TNI significantly affects per-channel Key quantization. For per-token Value quantization, although TNI persists, per-token quantization confines norm variations to individual tokens and avoids cross-token interference. Consequently, the error amplification caused by TNI in per-channel schemes does not manifest under per-token quantization. These analysis results validate our assumption and theoretical derivations. Further details and real-tensor calibration appear in Appendices~\ref{app:quantitative_error_analysis} and~\ref{app:calibration}.

\subsection{The OScaR Framework: Omni-Scaled Canalized Rotation}
\label{sec:OScaR}

In this section, we introduce \textit{OScaR (\textbf{O}mni-\textbf{Sca}led Canalized \textbf{R}otation)}, an accurate and lightweight KV cache compression framework for X-LLMs (i.e., text-only, multi-modal, and omni-modal LLMs). We focus on the algorithmic design herein, while the optimized system design and CUDA kernels are presented in the next subsection. An overview of the OScaR pipeline is provided in Figure~\ref{fig:OScaR}, and the detailed algorithm is given in Algorithm~\ref{algo:oscar}.

Advancing the per-channel paradigm, OScaR introduces two key innovations that together mitigate TNI-induced sequence-dimensional variance in a fully training-free manner:
\begin{itemize}
    \item \textbf{Canalized Rotation:} Direct token-wise scaling, though conceptually straightforward, suffers from the \textit{Scaling-Induced Outlier Artifact} in practice. Applying Canalized Rotation prior to scaling suppresses outlier channels that would otherwise dominate token norms, thereby preventing this artifact from biasing subsequent \textit{Omni-Token Scaling}.
    
    \item \textbf{Omni-Token Scaling:} Addresses TNI through omni-directional sequence-level normalization. Following \textit{Canalized Rotation}, it safely applies token-wise scaling to balance token norms across the sequence dimension, thereby resolving the impact of diverse TNI patterns.
\end{itemize}

\begin{figure}[t]

    \centering
    
    \begin{subfigure}[b]{0.245\textwidth}
        \centering
        \includegraphics[width=\textwidth]{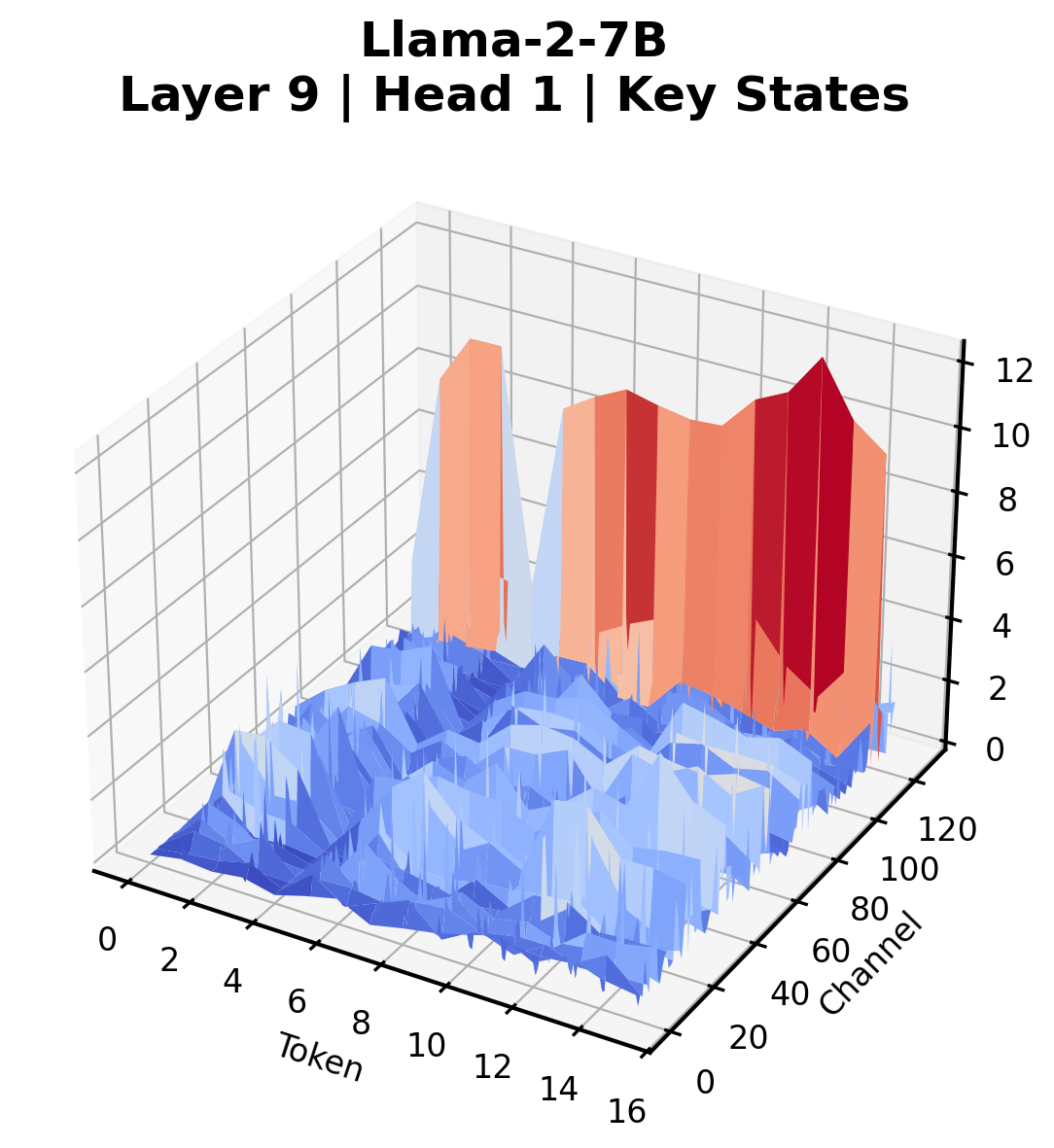}
    \end{subfigure}
    \begin{subfigure}[b]{0.245\textwidth}
        \centering
        \includegraphics[width=\textwidth]{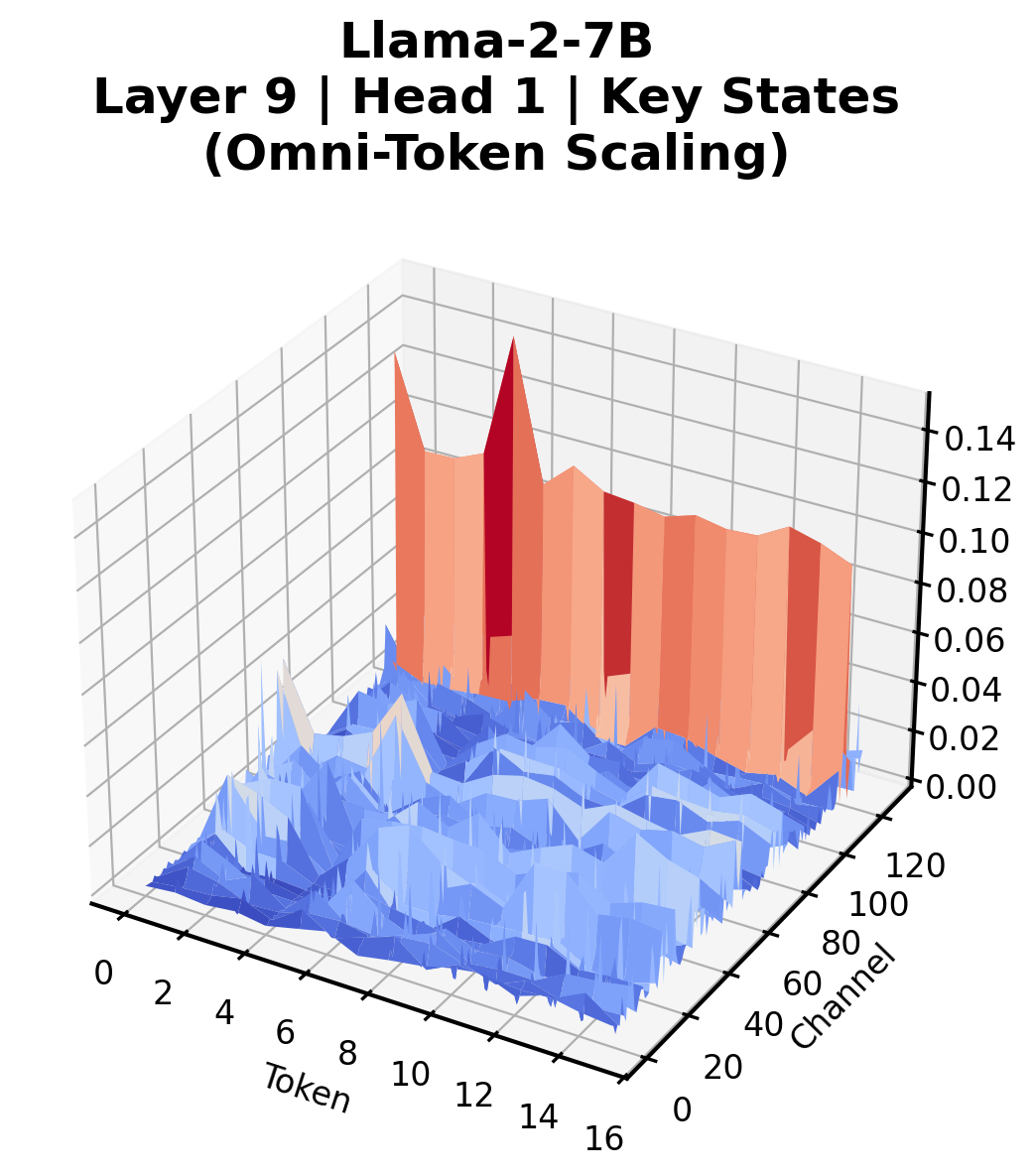}
    \end{subfigure}
    \begin{subfigure}[b]{0.245\textwidth}
        \centering
        \includegraphics[width=\textwidth]{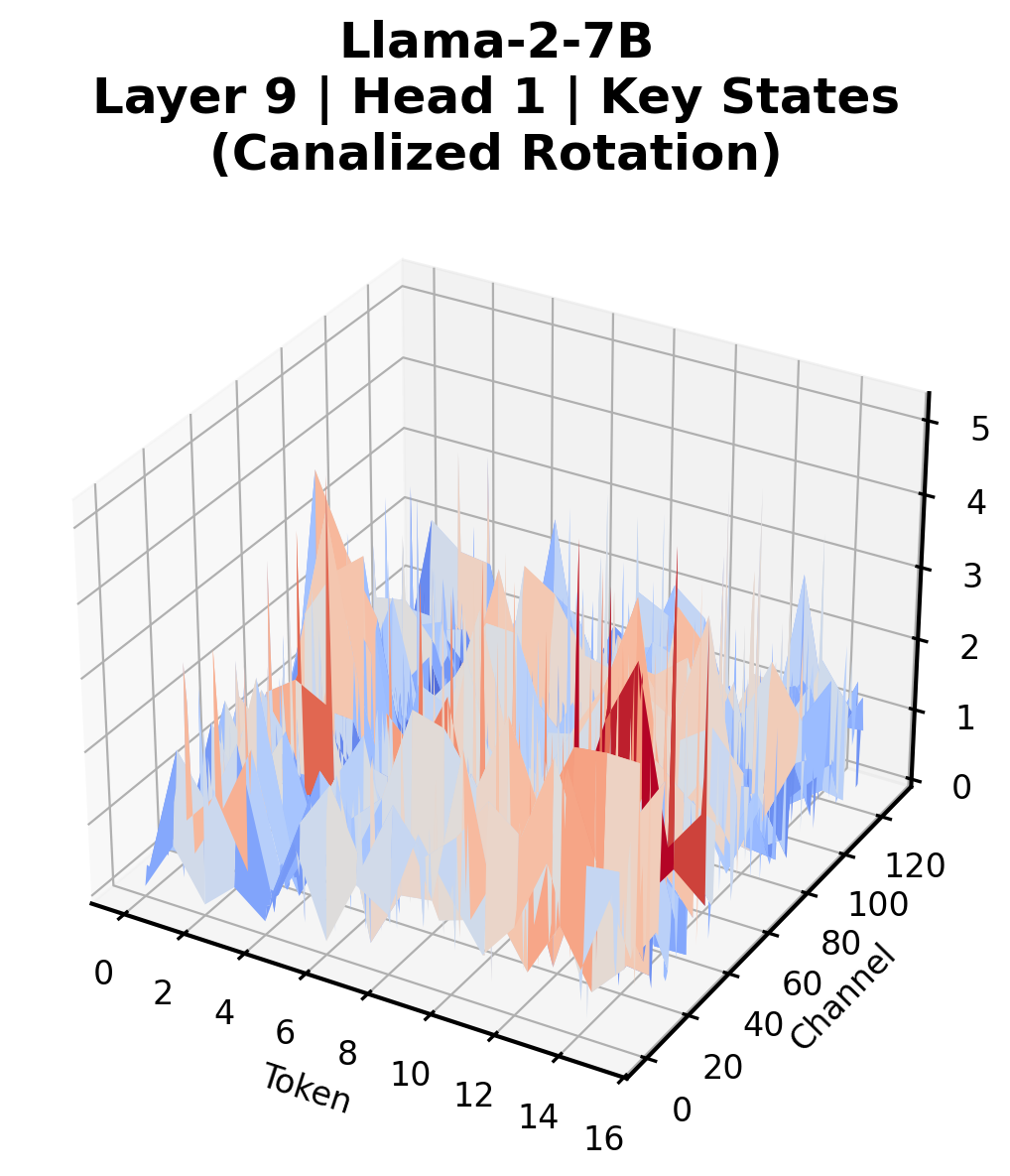}
    \end{subfigure}
    \begin{subfigure}[b]{0.245\textwidth}
        \centering
        \includegraphics[width=\textwidth]{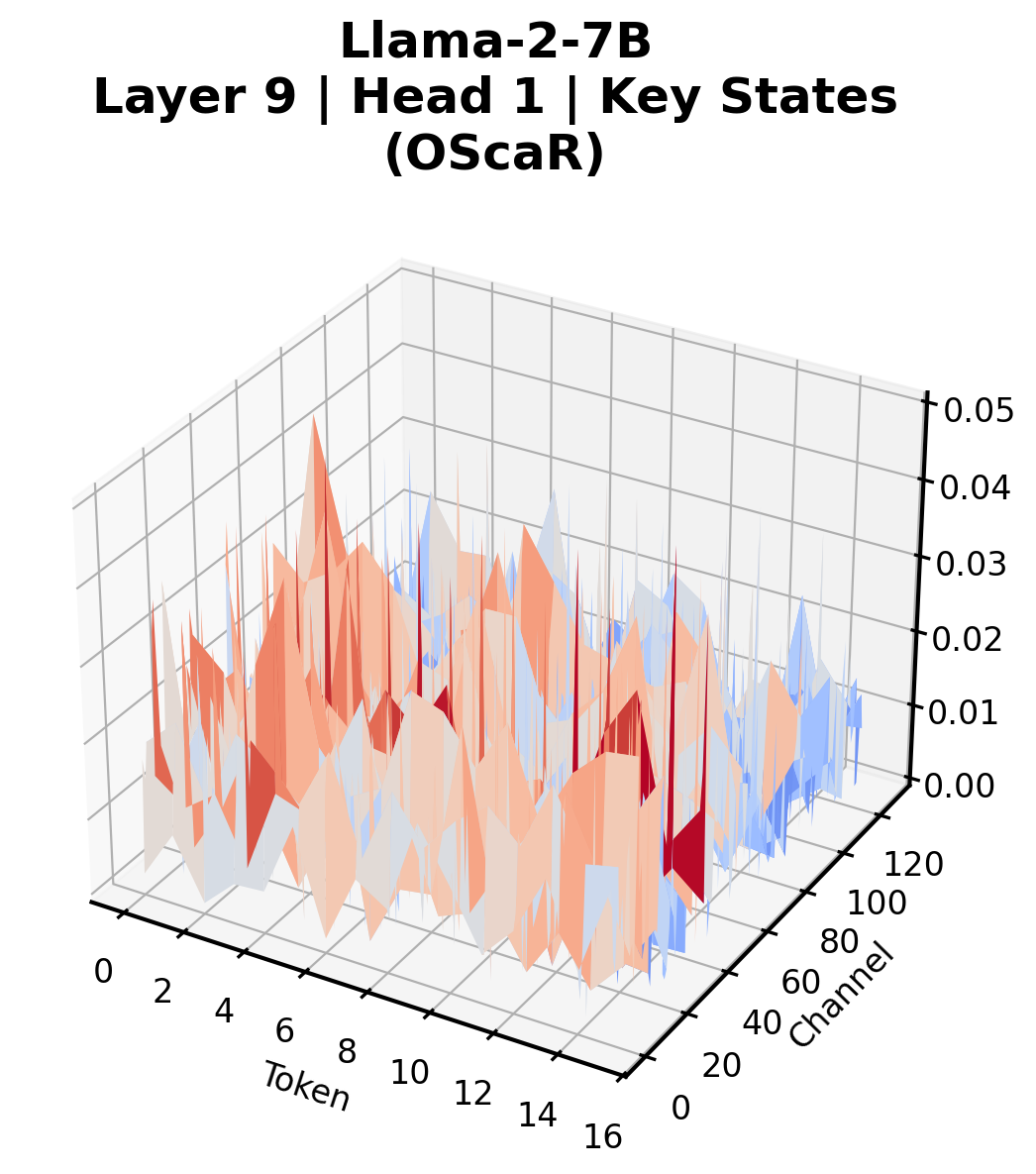}
    \end{subfigure}

    \begin{subfigure}[b]{0.245\textwidth}
        \centering
        \includegraphics[width=\textwidth]{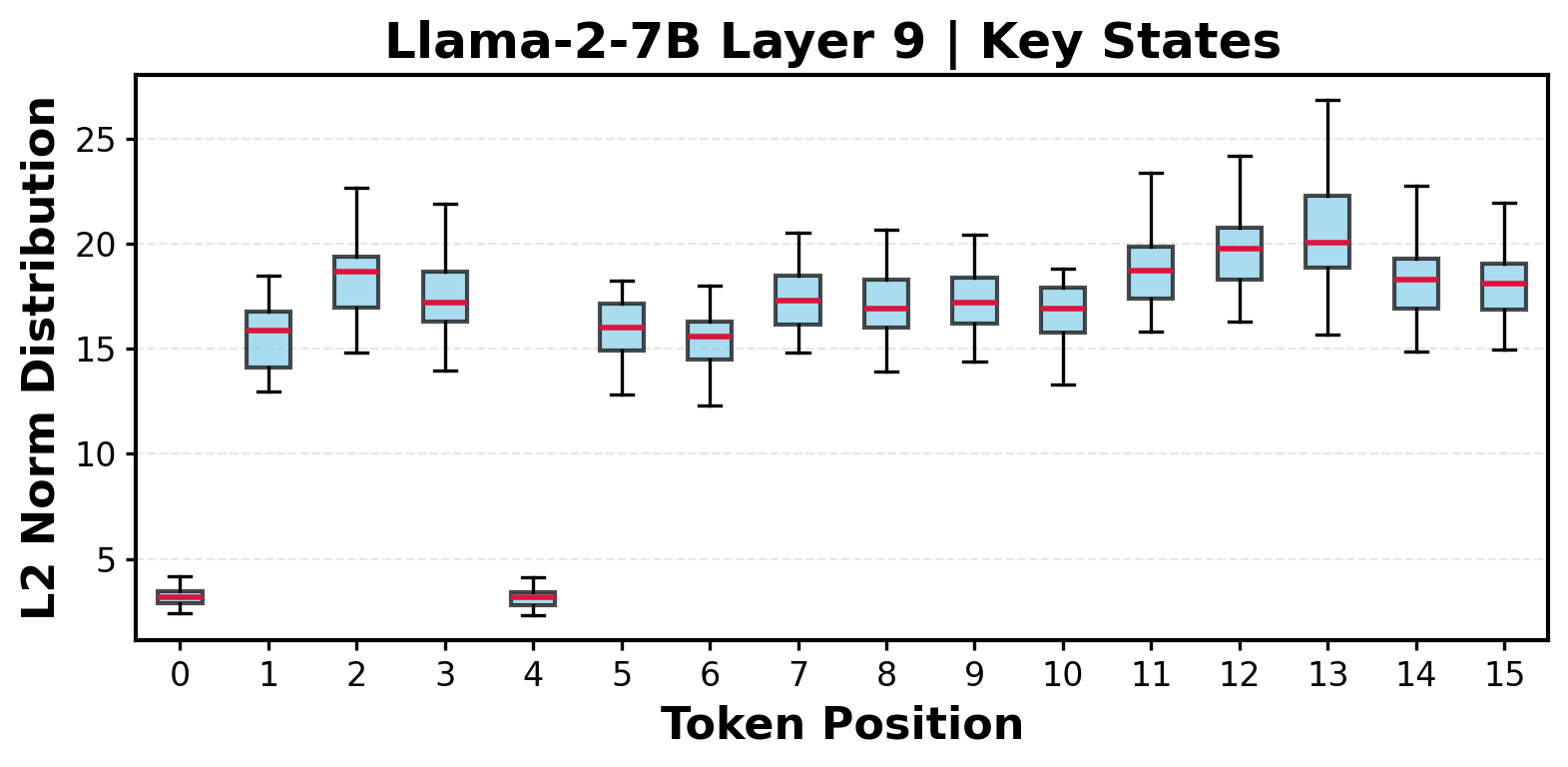}
    \end{subfigure}
    \begin{subfigure}[b]{0.245\textwidth}
        \centering
        \includegraphics[width=\textwidth]{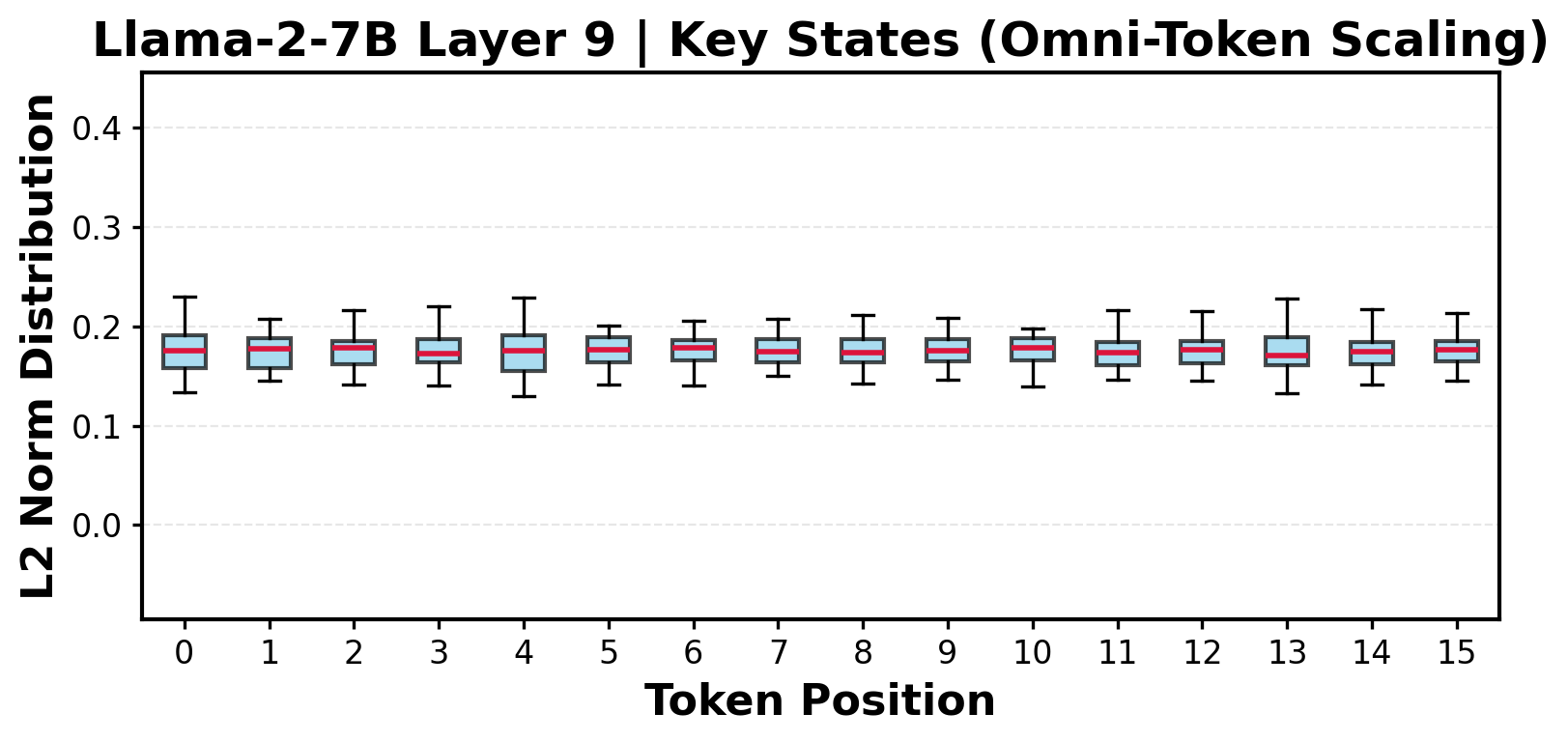}
    \end{subfigure}
    \begin{subfigure}[b]{0.245\textwidth}
        \centering
        \includegraphics[width=\textwidth]{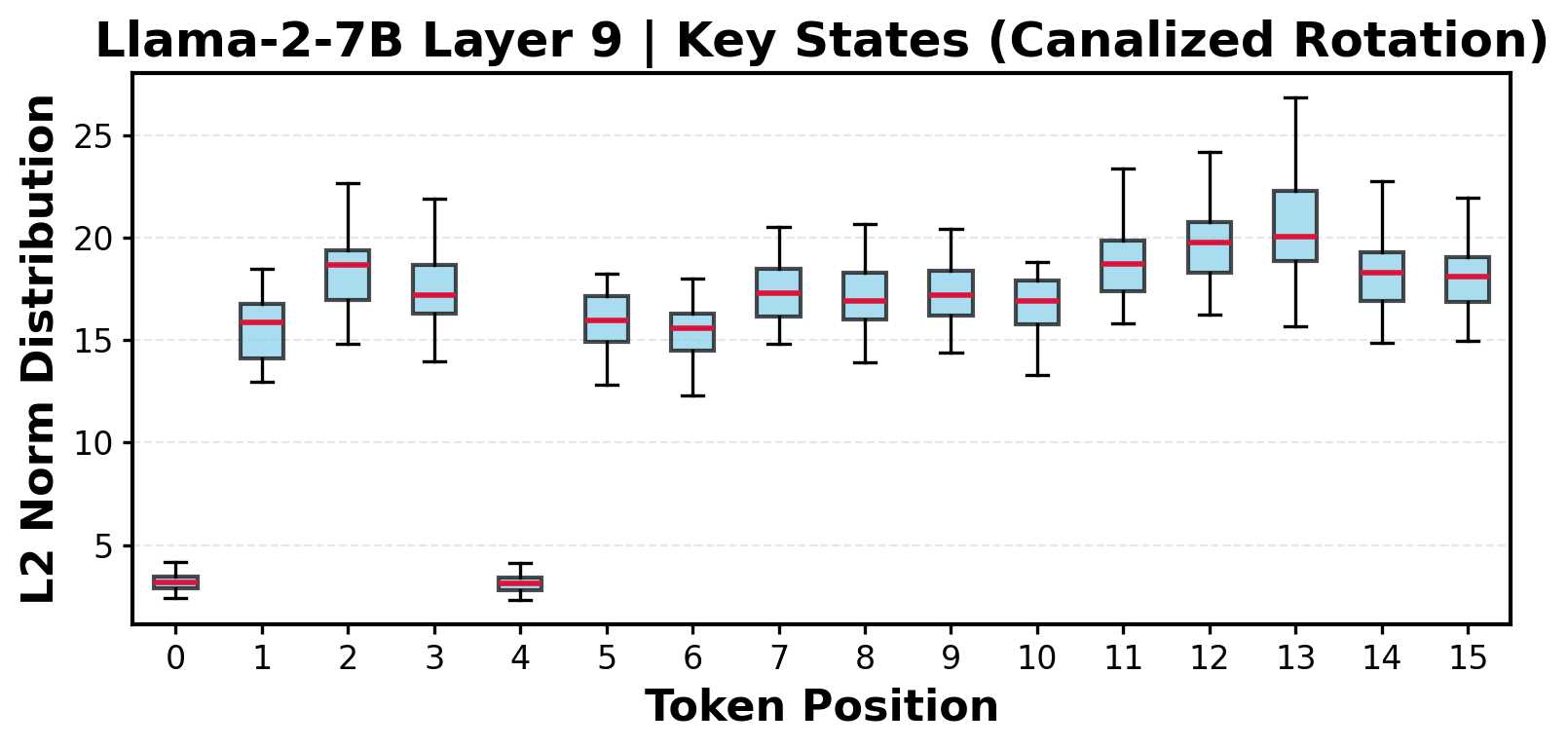}
    \end{subfigure}
    \begin{subfigure}[b]{0.245\textwidth}
        \centering
        \includegraphics[width=\textwidth]{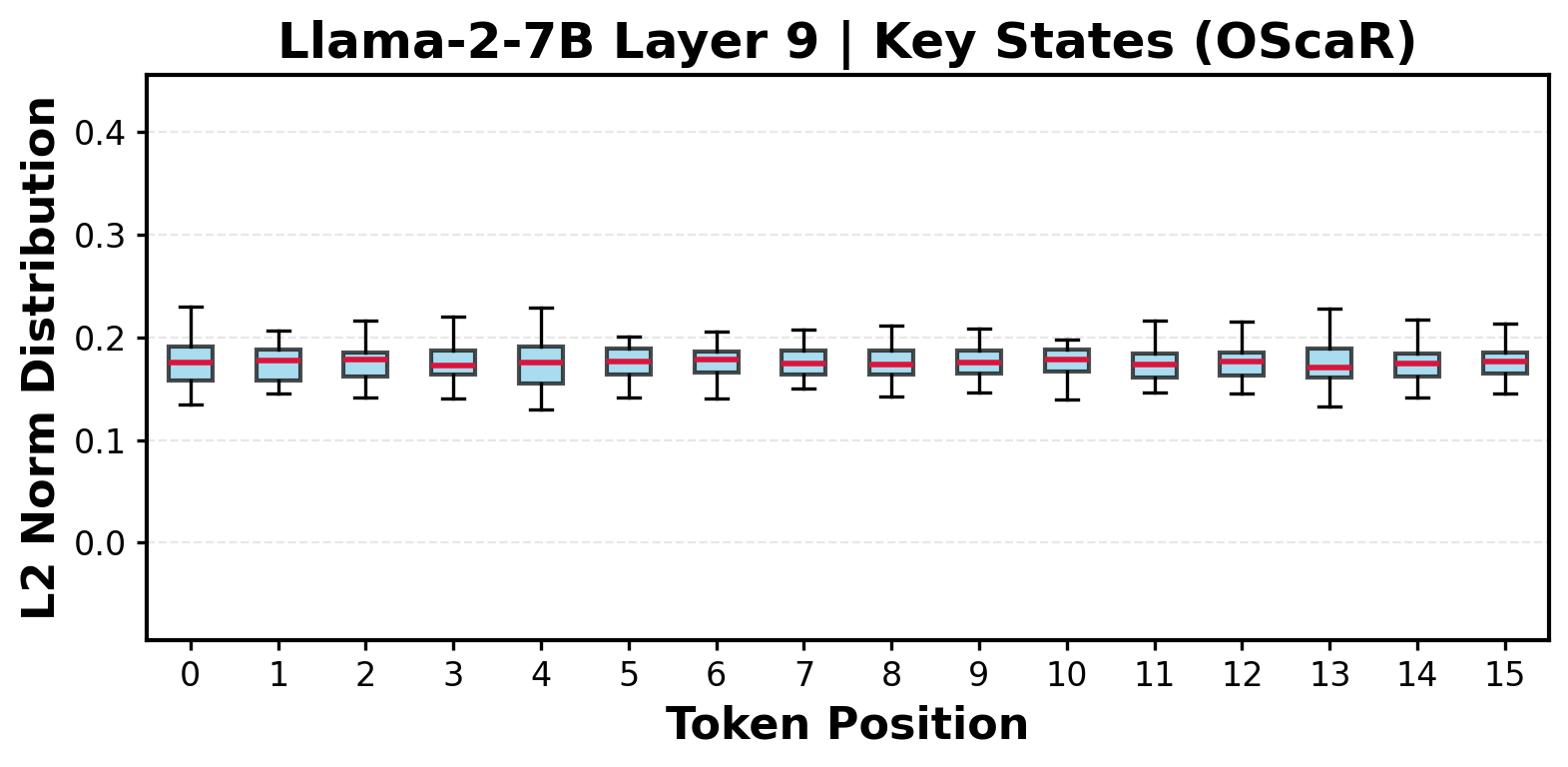}
    \end{subfigure}
    
    \caption{Key magnitude (top row) and L2 norm distribution (bottom row) across different processing stages: Original, after \textit{Omni-Token Scaling} alone, after \textit{Canalized Rotation} alone, and the full OScaR. Direct scaling balances token norms but introduces the \textit{Scaling-Induced Outlier Artifact}. \textit{Canalized Rotation} alone fails to balance token norms. Only the complete OScaR successfully addresses TNI without incurring the artifact. Additional visualizations are provided in Appendix \ref{app:oscar_vis}.}
    \label{fig:direct_scaling}
    
\end{figure}

The effectiveness of these two components is demonstrated in Figure~\ref{fig:direct_scaling}. Guided by Occam's Razor, OScaR avoids complex auxiliary pipelines and instead relies on the two mutually essential components described above to effectively and efficiently mitigate TNI. Below, we detail the design rationales and specific methodologies of the OScaR framework.

\paragraph{On the Failure of Direct Token-Wise Scaling}
A straightforward strategy to mitigate TNI is to apply token-wise scaling directly. However, although it balances token norms, empirical evidence shows that this approach rarely improves quantized models and, in many cases, even leads to degradation. Our analysis attributes this failure to what we term \textit{Scaling-Induced Outlier Artifact}.
Intuitively, consider normal tokens dominated by outlier channels and low-norm outlier tokens with relatively uniformly small entries. When scaled to the same norm, the low-norm tokens are uniformly amplified and become artificial outliers in channels where normal tokens have minimal magnitudes, expanding the per-channel quantization range and degrading precision. This artifact undermines per-channel quantization and cannot be resolved by merely adjusting the scaling target. Therefore, direct token-wise scaling alone is insufficient for handling TNI. A detailed analysis of \textit{Scaling-Induced Outlier Artifact} is provided in Appendix~\ref{app:scaling_artifact}.

\paragraph{Canalized Rotation Followed by Omni-Token Scaling}
To resolve the \textit{Scaling-Induced Outlier Artifact}, OScaR introduces a two-step procedure. First, \textit{Canalized Rotation} applies a Hadamard transform to redistribute the energy of outlier channels across all dimensions. Second, \textit{Omni-Token Scaling} computes the \(\ell_2\) norm of each token across all modalities and applies token-wise scaling to unify these norms. Because \textit{Canalized Rotation} has already smoothed the per-channel distribution, the scaling step can safely balance token norms without introducing artificial outliers. 
As shown in Figure~\ref{fig:direct_scaling}, rotation alone is insufficient for handling TNI. Only when combined with scaling can the framework effectively address TNI while avoiding the \textit{Scaling-Induced Outlier Artifact}.

\paragraph{Occam's Razor for Extreme KV Cache Quantization}
Existing KV cache quantization methods typically rely on online quantization operations augmented with auxiliary mechanisms to mitigate errors~\citep{zandieh2025turboquant,zandieh2025qjl,pope2026rotorquant,ji2026isoquant,han2025polarquant} (see Appendix~\ref{app:mechanism_positioning} for a comparison). These complex pipelines incur substantial computational overhead and additional parameters, limiting both practicality and efficiency. A viable solution to TNI must therefore be concise and essential, ensuring high efficiency in real-world deployments. \textbf{Guided by the principle of Occam's Razor, we advocate simplicity and elegance over intricate, heavyweight quantization pipelines.} Theoretical complexity analysis in Appendix~\ref{app:complexity} demonstrates that OScaR is a highly lightweight approach. Combined with comprehensive benchmarks and efficiency evaluations in Section~\ref{sec:experiment}, OScaR achieves a clearly advantageous position on the accuracy-efficiency Pareto front.

\subsection{Efficient System Design and CUDA Implementations of OScaR}
\label{sec:method-pipeline}

\paragraph{OScaR Pipeline Overview}
As illustrated in Figure~\ref{fig:OScaR} and detailed in Algorithm~\ref{algo:oscar}, OScaR proceeds as follows.
(i) For the Query, an online Fast Hadamard Transform (FHT) is applied to implicitly cancel the FHT applied to the Key during attention computation.
(ii) The Key undergoes an online FHT followed by token-wise norm scaling, after which per-channel quantization is applied. During dequantization, inverse token-wise scaling restores the original token norms.
(iii) For the Value, an offline Hadamard transform is applied to both the Value and the attention output weight matrices prior to inference. Per-token quantization is then applied during inference. This offline transform improves the fidelity of Value quantization without introducing additional \mbox{runtime overhead~\citep{ashkboos2024quarot,su2025rotatekv}}.

\paragraph{System Design and CUDA Kernels}
OScaR is implemented using three CUDA kernels, building upon \texttt{HadaCore} and \texttt{BitDecoding} \citep{agarwal2024hadacore,du2025bitdecoding} with carefully engineered adaptations for high-performance execution on GPU Tensor Cores:  
(i) \textit{Online FHT and Scaling kernel:} performs fused FHT and token scaling for Key, and applies FHT to Query.  
(ii) \textit{Quantization kernel:} performs GPU-efficient quantization for both Key and Value.  
(iii) \textit{Dequantization, De-Scaling, and Attention kernel:} handles dequantization for Key and Value, inverse scaling for Key, and attention computation.\looseness=-1

The FHT is adopted for its computational efficiency over standard matrix multiplication, achieving \(O(d \log d)\) complexity compared to \(O(d^2)\), where \(d\) is the dimension~\citep{ashkboos2024quarot,agarwal2024hadacore}. 
\textit{Omni-Token Scaling} leverages the hardware-accelerated \texttt{rsqrt} instruction, as motivated by our ablation study in Appendix~\ref{app:ablation}. 
Further implementation details of these CUDA kernels are provided in Appendix~\ref{app:cuda}, with storage accounting in Appendix~\ref{app:storage_accounting}. 

\section{Experiments}
\label{sec:experiment}

\begin{table}[t]

\caption{LongBench-E evaluation results. All quantized methods except TurboQuant+ use INT2 quantization with a group size of 32, whereas TurboQuant+ uses 2.5-bit. TurboQuant is based on TurboQuant+ \citep{turney2026turboquantplus}; QJL is excluded as it degrades performance. Implementation details are in Appendix~\ref{app:turboquant}.}
\label{tab:longbench}
\renewcommand{\arraystretch}{1.6}
\resizebox{\textwidth}{!}{
\begin{tabular}{@{}llccccccccc@{}}
\toprule
\textbf{Model} & \textbf{Method} & \textbf{\begin{tabular}[c]{@{}c@{}}Hotpot\\ QA\end{tabular}} & \textbf{Qasper} & \textbf{\begin{tabular}[c]{@{}c@{}}Gov\\ Report\end{tabular}} & \textbf{\begin{tabular}[c]{@{}c@{}}Multi\\ News\end{tabular}} & \textbf{Trec} & \textbf{\begin{tabular}[c]{@{}c@{}}Trivia\\ QA\end{tabular}} & \textbf{\begin{tabular}[c]{@{}c@{}}Passage\\ Retrieval\end{tabular}} & \textbf{\begin{tabular}[c]{@{}c@{}}Repo\\ Bench-P\end{tabular}} & \textbf{Avg.} \\ \midrule
\multirow{7}{*}{Llama-3.1-8B} & 16bit & 13.76 & 11.86 & 29.88 & 5.89 & 72.67 & 90.15 & 43.57 & 65.78 & 41.70 \\ \cmidrule(l){2-11} 
 & QuaRot & 13.73 & 9.24 & 18.23 & 1.75 & 73.00 & 89.68 & 42.46 & 55.46 & 37.94 \\
 & RotateKV & 13.13 & 10.97 & 17.82 & 1.28 & 72.33 & \textbf{90.56} & 38.83 & 58.95 & 37.98 \\
 & KIVI & \textbf{14.66} & 10.71 & 24.41 & 2.70 & 73.00 & 89.83 & 38.95 & 64.42 & 39.84 \\
 & OTT & 13.27 & 10.70 & \textbf{29.45} & 4.65 & 73.00 & 90.02 & 39.39 & \textbf{65.45} & 40.74 \\
 & TurboQuant+ & 13.85 & 10.04 & 26.60 & 3.18 & 72.33 & 89.91 & 41.51 & 62.82 & 40.03 \\
 & \textit{\textbf{OScaR (ours)}} & 13.68 & \textbf{11.61} & \textbf{29.45} & \textbf{5.68} & \textbf{73.33} & \textbf{90.41} & \textbf{44.46} & 65.39 & \textbf{41.75} \\ \midrule
\multirow{7}{*}{Qwen3-8B} & 16bit & 13.10 & 10.27 & 29.56 & 21.96 & 71.33 & 91.95 & 95.81 & 62.47 & 49.56 \\ \cmidrule(l){2-11} 
 & QuaRot & 11.70 & 7.47 & 20.13 & 15.01 & 33.00 & 89.52 & 91.33 & 52.88 & 40.13 \\
 & RotateKV & \textbf{12.74} & 6.78 & 14.11 & 10.59 & 71.00 & 88.81 & 91.57 & 48.00 & 42.95 \\
 & KIVI & 11.86 & 9.03 & 28.12 & 21.58 & 71.00 & 91.09 & 91.92 & 58.96 & 47.95 \\
 & OTT & 11.86 & 9.26 & \textbf{28.56} & \textbf{21.82} & 71.00 & 91.36 & 92.39 & 59.41 & 48.21 \\
 & TurboQuant+ & 12.53 & 8.35 & 28.47 & 21.66 & 71.33 & 90.95 & 88.92 & 58.23 & 47.56 \\
 & \textit{\textbf{OScaR (ours)}} & 12.35 & \textbf{9.94} & 28.33 & 21.46 & \textbf{72.00} & \textbf{91.46} & \textbf{92.78} & \textbf{61.57} & \textbf{48.74} \\ \bottomrule
\end{tabular}
}

\end{table}

\subsection{Experimental Setup}

\paragraph{Models and Tasks}
To comprehensively evaluate OScaR, we select three categories of LLMs: 
(i) text-only LLMs, including Llama-3.1-8B and Qwen3-8B \citep{grattafiori2024llama,yang2025qwen3}; 
(ii) multi-modal LLMs, including LLaVA-v1.6-vicuna-7B and Qwen3-VL-4B/8B-Instruct \citep{liu2023visual,liu2024improved,bai2025qwen3}; and 
(iii) omni-modal LLMs, including Qwen3-Omni-30B-A3B \citep{xu2025qwen3}. 
These models represent a diverse set of open-source families and scales. 
To ensure a rigorous evaluation, most experiments focus on tasks requiring extreme long-context processing. Specifically, text-only LLMs are evaluated on LongBench-E and the Needle-in-a-Haystack (NIAH) benchmark \citep{bai2024longbench,gkamradt2023needle}; multi-modal LLMs on OCRBench and DocVQA \citep{liu2024ocrbench,mathew2021docvqa}; and omni-modal LLMs on MMAU-Pro \citep{kumar2026mmaupro}. Detailed descriptions of these tasks are provided in Appendix~\ref{app:dataset}.
Extensive ablation studies are provided in Appendix~\ref{app:ablation}, including analyses of the proposed innovations and alternative normalization strategies for \textit{Omni-Token Scaling}. In addition, we provide visual comparisons of token norm distributions before and after applying OScaR in Appendix~\ref{app:oscar_vis}. A Pareto front analysis of accuracy and efficiency is presented in Appendix~\ref{app:pareto}. A comprehensive efficiency evaluation is provided in Section~\ref{sec:efficiency} and Appendix~\ref{app:turboquant_decoding_efficiency}, assessing decoding speedup, memory savings, and throughput improvements.\looseness=-1

\paragraph{Baselines}
We compare OScaR against several strong baselines, which fall into three categories: 
(i) per-channel Key quantization, including KIVI and OTT \citep{liu2024kivi,su2025accurate}; 
(ii) rotation-based per-token Key quantization, such as QuaRot and RotateKV \citep{ashkboos2024quarot,su2025rotatekv}; and 
(iii) LUT-based methods, represented by TurboQuant \citep{zandieh2025turboquant}. 
Among these, OTT and RotateKV employ high-precision protection for outlier tokens.
These baselines span diverse strategies, enabling a comprehensive evaluation.
To ensure a fair comparison, we carefully align the configuration of each method. For QuaRot, we adopt only its KV cache quantization component. For KIVI, OTT, and OScaR, the residual length for per-channel quantization is uniformly set to 128. For OTT, the number of high-precision outlier tokens is set to 5.
We also account for the average bit overhead introduced by quantization parameters. 
Specifically, TurboQuant employs 2.5-bit quantization, assigning higher bit-widths to outlier channels while using 2-bit for regular channels, whereas all other methods adopt INT2 quantization. 
Since TurboQuant does not provide an official code release, we use TurboQuant+ \citep{turney2026turboquantplus}, a widely adopted open-source implementation. 
Further implementation details of TurboQuant+ are provided in Appendix~\ref{app:turboquant}.\looseness=-1

\subsection{Main Experimental Results}

\paragraph{Results on Text-Only LLMs}
The LongBench-E results are presented in Table~\ref{tab:longbench}, and the NIAH experiment is shown in Figure~\ref{fig:needle_results} of Appendix~\ref{app:niah}. On LongBench-E, OScaR achieves the highest average accuracy among all competing quantized methods, outperforming the second-best method by 1.01 percentage points (41.75\% vs. 40.74\%). Compared to the 16-bit baseline, OScaR incurs only a negligible accuracy drop of 1.7\% on Qwen3-8B. 
In the NIAH task, OScaR achieves 96.5\% retrieval accuracy, significantly exceeding the second-best method (92.7\%) and slightly surpassing the 16-bit baseline (96.0\%), demonstrating its robustness in long-context retrieval scenarios.

\paragraph{Results on Multi-Modal and Omni-Modal LLMs}  
Table~\ref{tab:ocrbench} reports the OCRBench results, Table~\ref{tab:docvqa} summarizes the DocVQA results, and Table~\ref{tab:mmaupro_results} presents the MMAU-Pro results for omni-modal LLMs. Across all benchmarks, OScaR maintains strong model performance under 2-bit quantization, frequently approaching the 16-bit baseline. On OCRBench, OScaR achieves a 0.7 percentage point improvement over the second-best method on Qwen3-VL-4B. On MMAU-Pro, it attains the highest scores among quantized methods across open-ended QA, Good Rate, and audio instruction following, surpassing the next-best method by 0.8, 2.8, and 4.6 percentage points, respectively. These results indicate that OScaR effectively preserves model capabilities in both multi-modal and omni-modal contexts. Further analyses appear in Appendices~\ref{app:ocrbench}, \ref{app:docvqa}, and~\ref{app:mmau-pro}.

\subsection{Efficiency Analysis}
\label{sec:efficiency}

\begin{figure}[t]
    \centering
    \begin{subfigure}[b]{0.49\textwidth}
        \centering
        \includegraphics[width=\textwidth]{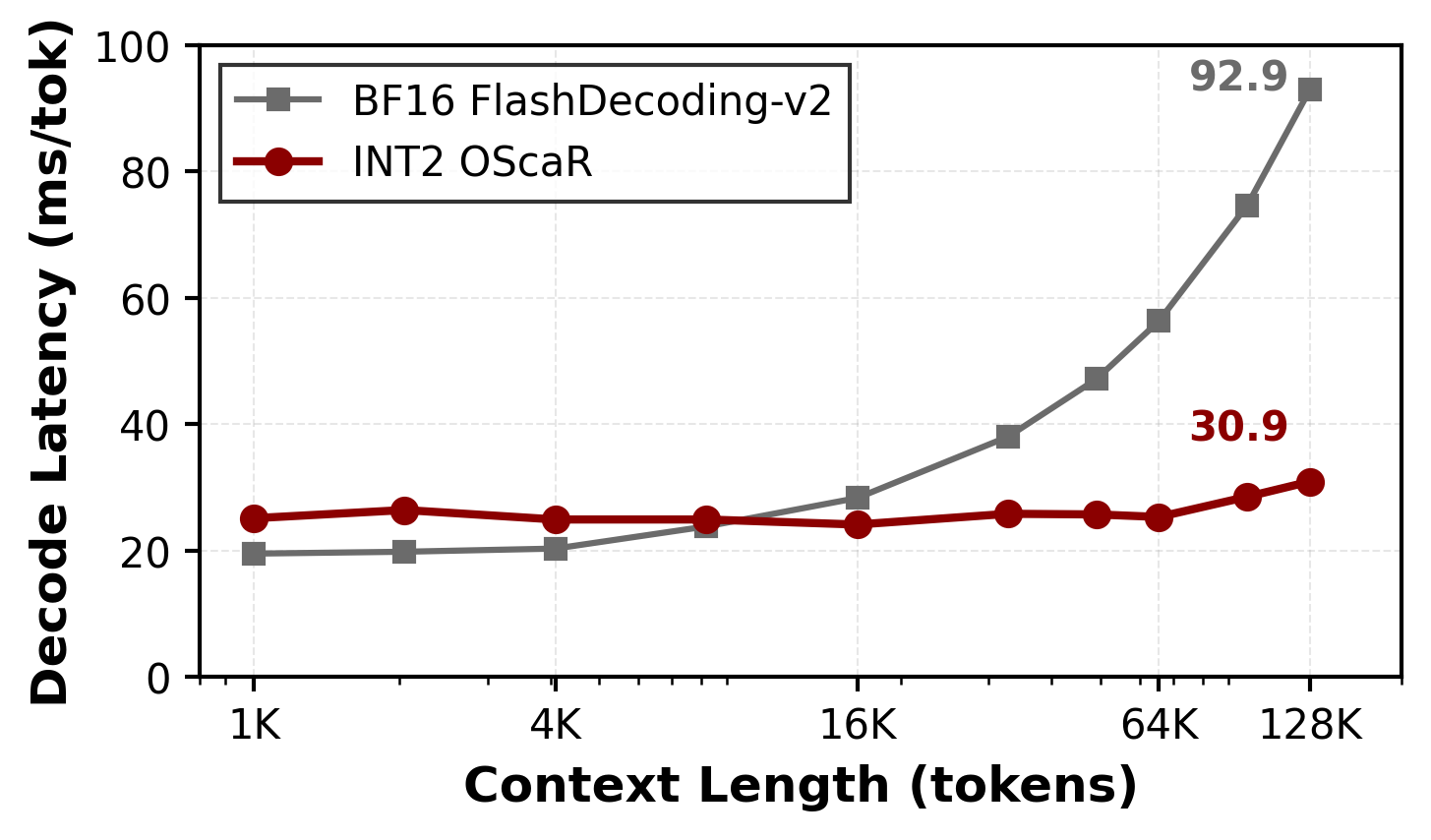}
        \caption{Decoding latency across context lengths.}
        \label{fig:single_batch}
    \end{subfigure}
    \hfill
    \begin{subfigure}[b]{0.485\textwidth}
        \centering
        \includegraphics[width=\textwidth]{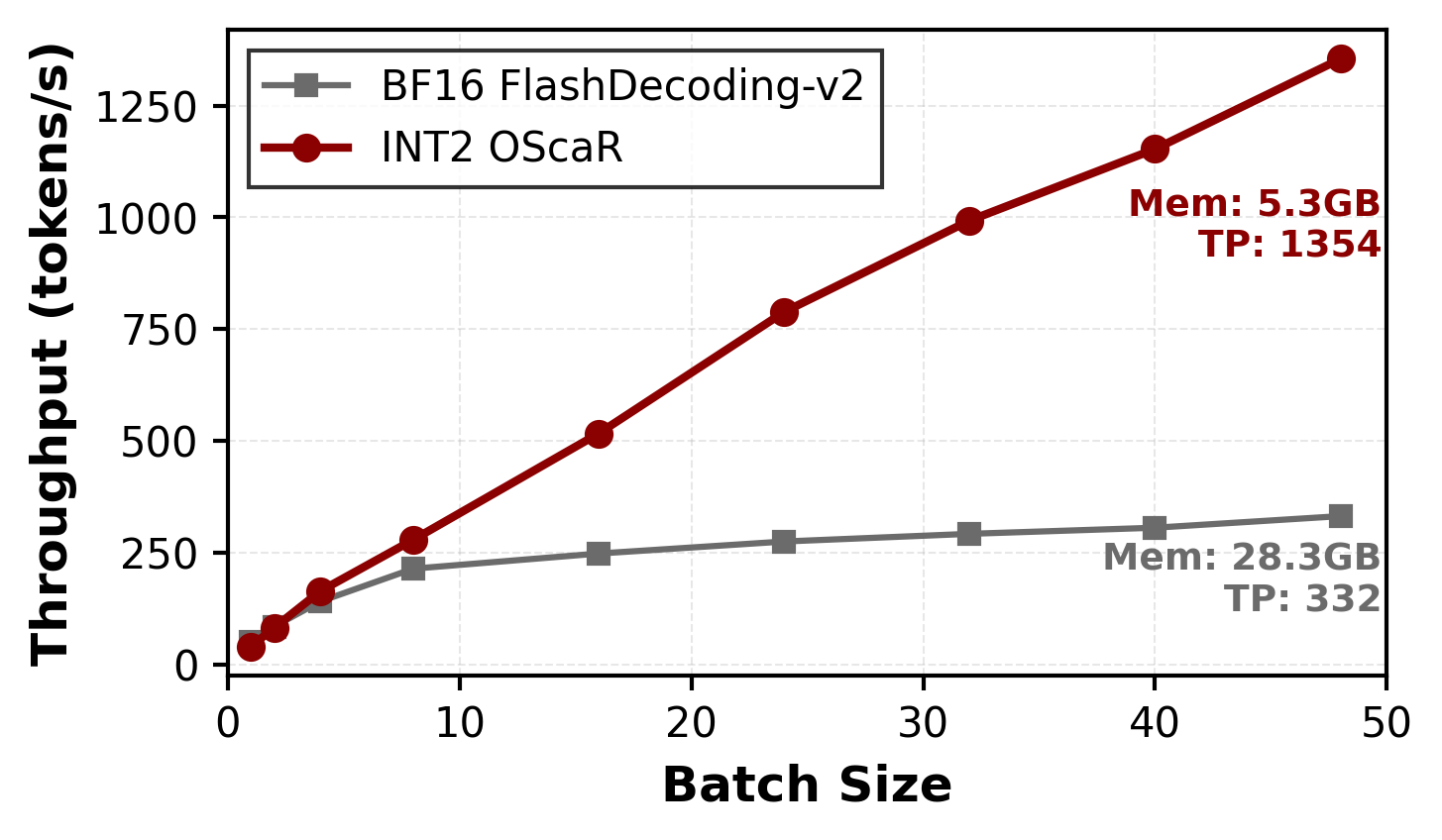}
        \caption{Decoding throughput across batch sizes.}
        \label{fig:multi_throughput}
    \end{subfigure}
    \caption{Efficiency analysis of OScaR against \texttt{BF16 FlashDecoding-v2}. Annotations show OScaR at 128K context length (latency) and batch size 48 (throughput and memory).}
    
    \label{fig:efficiency_comparison}
\end{figure}

In this section, we evaluate the efficiency of OScaR. Experiments are conducted on a single H20 GPU (141GB) using Qwen3-8B, with \texttt{BF16 FlashDecoding-V2} as the baseline. The evaluation consists of two parts: (i) measuring decoding latency under varying context lengths in a single-batch setting, and (ii) fixing the context length at 4K while increasing batch size to assess memory savings and corresponding throughput improvements.

As illustrated in Figure~\ref{fig:efficiency_comparison}, OScaR achieves substantial efficiency gains, reflecting both its low computational complexity and the advantages of our system-level design. Specifically, at a context length of 128K, OScaR attains up to a 3.0$\times$ decoding speedup over the baseline. With a batch size of 48, it reduces the decoding memory footprint by 5.3$\times$ and increases throughput by 4.1$\times$. Additional decoding efficiency comparisons with TurboQuant+ and controlled latency measurements are provided in Appendices~\ref{app:turboquant_decoding_efficiency} and~\ref{app:controlled_latency}, respectively.

\section{Conclusion}

In this work, we revisited the fundamental limitations of per-channel KV cache quantization under extreme compression and identified TNI as a primary structural bottleneck that systematically amplifies quantization error. Motivated by this insight, we proposed \textit{OScaR}, a lightweight and training-free KV cache compression framework for X-LLMs. By integrating \textit{Canalized Rotation} with \textit{Omni-Token Scaling}, OScaR effectively mitigates TNI-induced sequence-dimensional variance. 
We hope that OScaR can serve as a critical framework for efficient LLM inference and provide valuable guidance for KV cache quantization in LLMs and beyond.
\clearpage
\subsection*{AI use statement}
Generative AI tools were used solely for language editing, including grammar, spelling, word choice, and minor improvements to clarity and consistency. All technical content, experimental results, analyses, and conclusions were developed and verified by the authors. The authors take full responsibility for the final manuscript.

\subsection*{Ethics statement}
This work studies efficient inference through KV-cache compression of existing language and multimodal models. Compression does not remove the biases, privacy risks, or potential for harmful outputs inherited from those models. Use of the models, datasets, and third-party software remains subject to their applicable licenses and access conditions.


\bibliography{references}

@article{team2025longcat1,
  title={{LongCat-Flash Technical Report}},
  author={{Meituan LongCat Team} and Bayan and Li, Bei and Lei, Bingye and Wang, Bo and Rong, Bolin and Wang, Chao and Zhang, Chao and Gao, Chen and Zhang, Chen and others},
  journal={arXiv preprint arXiv:2509.01322},
  year={2025}
}

@article{team2025longcat2,
  title={{LongCat-Video Technical Report}},
  author={{Meituan LongCat Team} and Cai, Xunliang and Huang, Qilong and Kang, Zhuoliang and Li, Hongyu and Liang, Shijun and Ma, Liya and Ren, Siyu and Wei, Xiaoming and Xie, Rixu and others},
  journal={arXiv preprint arXiv:2510.22200},
  year={2025}
}

@article{team2025longcat3,
  title={{LongCat-Image Technical Report}},
  author={{Meituan LongCat Team} and Ma, Hanghang and Tan, Haoxian and Huang, Jiale and Wu, Junqiang and He, Jun-Yan and Gao, Lishuai and Xiao, Songlin and Wei, Xiaoming and Ma, Xiaoqi and others},
  journal={arXiv preprint arXiv:2512.07584},
  year={2025}
}

@article{team2025longcat4,
  title={{LongCat-Flash-Omni Technical Report}},
  author={{Meituan LongCat Team} and Wang, Bairui and Bayan and Xiao, Bin and Zhang, Bo and Rong, Bolin and Chen, Borun and Wan, Chang and Zhang, Chao and Huang, Chen and others},
  journal={arXiv preprint arXiv:2511.00279},
  year={2025}
}

@article{team2025introducing,
  title={{Introducing LongCat-Flash-Thinking: A Technical Report}},
  author={{Meituan LongCat Team} and Gui, Anchun and Li, Bei and Tao, Bingyang and Zhou, Bole and Chen, Borun and Zhang, Chao and Zhang, Chao and Han, Chengcheng and Yang, Chenhui and others},
  journal={arXiv preprint arXiv:2509.18883},
  year={2025}
}

@article{wang2026longcat,
  title={{LongCat-Flash-Prover: Advancing Native Formal Reasoning via Agentic Tool-Integrated Reinforcement Learning}},
  author={Wang, Jianing and Zhang, Jianfei and Guo, Qi and Guo, Linsen and Li, Rumei and Zhang, Chao and Peng, Chong and Wang, Cunguang and Zhao, Dengchang and Shi, Jiarong and others},
  journal={arXiv preprint arXiv:2603.21065},
  year={2026}
}

@article{team2026longcat1,
  title={{LongCat-Flash-Thinking-2601 Technical Report}},
  author={{Meituan LongCat Team} and Gui, Anchun and Li, Bei and Tao, Bingyang and Zhou, Bole and Chen, Borun and Zhang, Chao and Zhang, Chao and Gao, Chen and Zhang, Chen and others},
  journal={arXiv preprint arXiv:2601.16725},
  year={2026}
}

@article{team2026longcat2,
  title={{LongCat-Next: Lexicalizing Modalities as Discrete Tokens}},
  author={{Meituan LongCat Team} and Xiao, Bin and Wang, Chao and Li, Chengjiang and Zhang, Chi and Peng, Chong and Yu, Hang and Yang, Hao and Yan, Haonan and Sun, Haoze and others},
  journal={arXiv preprint arXiv:2603.27538},
  year={2026}
}

@inproceedings{su2025rotatekv,
  title={{RotateKV: Accurate and Robust 2-Bit KV Cache Quantization for LLMs via Outlier-Aware Adaptive Rotations}},
  author={Zunhai Su and Hanyu Wei and Zhe Chen and Wang Shen and Linge Li and Huangqi Yu and Kehong Yuan},
  booktitle={Proceedings of the Thirty-Fourth International Joint Conference on Artificial Intelligence, {IJCAI-25}},
  pages={6200--6208},
  year={2025},
  doi={10.24963/ijcai.2025/690},
  url={https://doi.org/10.24963/ijcai.2025/690}
}

@inproceedings{su2025akvq,
  title={{AKVQ-VL: Attention-Aware KV Cache Adaptive 2-Bit Quantization for Vision-Language Models}},
  author={Su, Zunhai and Shen, Wang and Li, Linge and Chen, Zhe and Wei, Hanyu and Yu, Huangqi and Yuan, Kehong},
  booktitle={2025 IEEE International Conference on Multimedia and Expo (ICME)},
  year={2025},
  doi={10.1109/icme59968.2025.11209367},
  pages={1--6}
}

@inproceedings{su2025unveiling,
  author={Su, Zunhai and Li, Qingyuan and Zhang, Hao and Ye, Weihao and Xue, Qibo and Qian, Yulei and Wong, Ngai and Yuan, Kehong},
  booktitle={International Conference on Learning Representations},
  pages={118323--118354},
  title={{Unveiling Super Experts in Mixture-of-Experts Large Language Models}},
  url={https://proceedings.iclr.cc/paper_files/paper/2026/file/c01b90b643d0e877cff0f34cbe27410f-Paper-Conference.pdf},
  volume={2026},
  year={2026}
}

@inproceedings{su2025kvsink,
  title={{KVSink: Understanding and Enhancing the Preservation of Attention Sinks in KV Cache Quantization for LLMs}},
  author={Su, Zunhai and Yuan, Kehong},
  year={2025},
  booktitle={Second Conference on Language Modeling},
  url={https://openreview.net/forum?id=gIqb6zWZoO}
}

@article{su2026xstreamvggt,
  title={{XStreamVGGT: Extremely Memory-Efficient Streaming Vision Geometry Grounded Transformer With KV Cache Compression}},
  author={Su, Zunhai and Ye, Weihao and Feng, Hansen and Fan, Keyu and Zhang, Jing and Yu, Dahai and Liu, Zhengwu and Wong, Ngai},
  journal={Journal of the Society for Information Display},
  year={2026},
  doi={10.1002/jsid.70086},
  volume={34},
  number={5},
  pages={535--545}
}

@article{zhuo2025streaming,
  title={{Streaming 4D Visual Geometry Transformer}},
  author={Zhuo, Dong and Zheng, Wenzhao and Guo, Jiahe and Wu, Yuqi and Zhou, Jie and Lu, Jiwen},
  journal={arXiv preprint arXiv:2507.11539},
  year={2025}
}

@article{li2025memory,
  title={{Memory-Efficient Visual Autoregressive Modeling with Scale-Aware KV Cache Compression}},
  author={Li, Kunjun and Chen, Zigeng and Yang, Cheng-Yen and Hwang, Jenq-Neng},
  journal={arXiv preprint arXiv:2505.19602},
  year={2025}
}

@inproceedings{qin2026head,
  title={{Head-Aware KV Cache Compression for Efficient Visual Autoregressive Modeling}},
  author={Ziran Qin and Youru Lv and Mingbao Lin and Hang Guo and Zeren Zhang and Danping Zou and Weiyao Lin},
  booktitle={Proceedings of the AAAI Conference on Artificial Intelligence},
  year={2026},
  volume={40},
  doi={10.1609/aaai.v40i30.39686},
  pages={24982--24990},
  url={https://ojs.aaai.org/index.php/AAAI/article/view/39686}
}

@article{wu2025fast,
  title={{Fast-dLLM: Training-free Acceleration of Diffusion LLM by Enabling KV Cache and Parallel Decoding}},
  author={Wu, Chengyue and Zhang, Hao and Xue, Shuchen and Liu, Zhijian and Diao, Shizhe and Zhu, Ligeng and Luo, Ping and Han, Song and Xie, Enze},
  journal={arXiv preprint arXiv:2505.22618},
  year={2025}
}

@inproceedings{xiao2025exploring,
  title={{Exploring Layer-wise Information Effectiveness for Post-Training Quantization in Small Language Models}},
  author={He Xiao and Qingyao Yang and Dirui Xie and Wendong XU and Zunhai Su and Runming Yang and Haobo Liu and Wenyong Zhou and Zhengwu Liu and Ngai Wong},
  year={2026},
  booktitle={Findings of the Association for Computational Linguistics: ACL 2026},
  doi={10.18653/v1/2026.findings-acl.771},
  pages={15753--15764},
  url={https://aclanthology.org/2026.findings-acl.771/}
}

@article{su2026attention,
  title={{Attention Sink in Transformers: A Survey on Utilization, Interpretation, and Mitigation}},
  author={Su, Zunhai and Zhang, Hengyuan and Wu, Wei and Zhang, Yifan and Liu, Yaxiu and Xiao, He and Yang, Qingyao and Sun, Yuxuan and Yang, Rui and Zhang, Chao and others},
  journal={arXiv preprint arXiv:2604.10098},
  year={2026}
}

@article{zhang2026locate,
  title={{Locate, steer, and improve: A practical survey of actionable mechanistic interpretability in large language models}},
  author={Zhang, Hengyuan and Zhang, Zhihao and Wang, Mingyang and Su, Zunhai and Wang, Yiwei and Wang, Qianli and Yuan, Shuzhou and Nie, Ercong and Duan, Xufeng and Xue, Qibo and others},
  journal={Computer Science Review},
  year={2026},
  doi={10.1016/j.cosrev.2026.101011},
  volume={62},
  pages={101011}
}

@article{zhang2026beyond,
  title={{Beyond Outliers: A Data-Free Layer-wise Mixed-Precision Quantization Approach Driven by Numerical and Structural Dual-Sensitivity}},
  author={Zhang, Hengyuan and Chen, Xinrong and Su, Zunhai and Liang, Xiao and Xiong, Jing and Xu, Wendong and Xiao, He and Tao, Chaofan and Zhang, Wei and Xie, Ruobing and others},
  journal={arXiv preprint arXiv:2603.17354},
  year={2026}
}

@article{li2024survey,
  title={{A Survey on Large Language Model Acceleration based on KV Cache Management}},
  author={Li, Haoyang and Li, Yiming and Tian, Anxin and Tang, Tianhao and Xu, Zhanchao and Chen, Xuejia and Hu, Nicole and Dong, Wei and Li, Qing and Chen, Lei},
  journal={Transactions on Machine Learning Research},
  year={2025},
  url={https://openreview.net/forum?id=z3JZzu9EA3}
}

@inproceedings{liu2025kv,
  title={{KV Cache Compression for Inference Efficiency in LLMs: A Review}},
  author={Liu, Yanyu and Fu, Jingying and Liu, Sixiang and Zou, Yitian and Zhang, Shouhua and Zhou, Jiehan},
  booktitle={Proceedings of the 4th International Conference on Artificial Intelligence and Intelligent Information Processing},
  year={2025},
  doi={10.1145/3778534.3778567},
  pages={207--212}
}

@article{liu2024kivi,
  title={{KIVI: A Tuning-Free Asymmetric 2bit Quantization for KV Cache}},
  author={Liu, Zirui and Yuan, Jiayi and Jin, Hongye and Zhong, Shaochen and Xu, Zhaozhuo and Braverman, Vladimir and Chen, Beidi and Hu, Xia},
  journal={arXiv preprint arXiv:2402.02750},
  year={2024}
}

@inproceedings{hooper2024kvquant,
  author={Hooper, Coleman and Kim, Sehoon and Mohammadzadeh, Hiva and Mahoney, Michael W. and Shao, Yakun S. and Keutzer, Kurt and Gholami, Amir},
  booktitle={Advances in Neural Information Processing Systems},
  doi={10.52202/079017-0040},
  pages={1270--1303},
  title={{KVQuant: Towards 10 Million Context Length LLM Inference with KV Cache Quantization}},
  url={https://proceedings.neurips.cc/paper_files/paper/2024/file/028fcbcf85435d39a40c4d61b42c99a4-Paper-Conference.pdf},
  volume={37},
  year={2024}
}

@article{ge2023model,
  title={{Model Tells You What to Discard: Adaptive KV Cache Compression for LLMs}},
  author={Ge, Suyu and Zhang, Yunan and Liu, Liyuan and Zhang, Minjia and Han, Jiawei and Gao, Jianfeng},
  journal={arXiv preprint arXiv:2310.01801},
  year={2023}
}

@inproceedings{liu2024minicache,
  author={Liu, Akide and Liu, Jing and Pan, Zizheng and He, Yefei and Haffari, Gholamreza and Zhuang, Bohan},
  booktitle={Advances in Neural Information Processing Systems},
  doi={10.52202/079017-4443},
  pages={139997--140031},
  title={{MiniCache: KV Cache Compression in Depth Dimension for Large Language Models}},
  url={https://proceedings.neurips.cc/paper_files/paper/2024/file/fd0705710bf01b88a60a3d479ea341d9-Paper-Conference.pdf},
  volume={37},
  year={2024}
}

@article{team2025kimi,
  title={{Kimi Linear: An Expressive, Efficient Attention Architecture}},
  author={{Kimi Team} and Zhang, Yu and Lin, Zongyu and Yao, Xingcheng and Hu, Jiaxi and Meng, Fanqing and Liu, Chengyin and Men, Xin and Yang, Songlin and Li, Zhiyuan and others},
  journal={arXiv preprint arXiv:2510.26692},
  year={2025}
}

@article{cao2026qwen3,
  title={{Qwen3-Coder-Next Technical Report}},
  author={Cao, Ruisheng and Chen, Mouxiang and Chen, Jiawei and Cui, Zeyu and Feng, Yunlong and Hui, Binyuan and Jing, Yuheng and Li, Kaixin and Li, Mingze and Lin, Junyang and others},
  journal={arXiv preprint arXiv:2603.00729},
  year={2026}
}

@article{cai2024pyramidkv,
  title={{PyramidKV: Dynamic KV Cache Compression based on Pyramidal Information Funneling}},
  author={Cai, Zefan and Zhang, Yichi and Gao, Bofei and Liu, Yuliang and Li, Yucheng and Liu, Tianyu and Lu, Keming and Xiong, Wayne and Dong, Yue and Hu, Junjie and others},
  journal={arXiv preprint arXiv:2406.02069},
  year={2024}
}

@inproceedings{wan2024look,
  title={{{LOOK}-{M}: Look-Once Optimization in {KV} Cache for Efficient Multimodal Long-Context Inference}},
  author={Zhongwei Wan and Ziang Wu and Che Liu and Jinfa Huang and Zhihong Zhu and Peng Jin and Longyue Wang and Li Yuan},
  booktitle={Findings of the Association for Computational Linguistics: EMNLP 2024},
  year={2024},
  url={https://aclanthology.org/2024.findings-emnlp.235/},
  doi={10.18653/v1/2024.findings-emnlp.235},
  pages={4065--4078}
}

@article{nagel2021white,
  title={{A White Paper on Neural Network Quantization}},
  author={Nagel, Markus and Fournarakis, Marios and Amjad, Rana Ali and Bondarenko, Yelysei and van Baalen, Mart and Blankevoort, Tijmen},
  journal={arXiv preprint arXiv:2106.08295},
  year={2021}
}

@inproceedings{su2025accurate,
  title={{Accurate {KV} Cache Quantization with Outlier Tokens Tracing}},
  author={Yi Su and Yuechi Zhou and Quantong Qiu and Juntao Li and Qingrong Xia and Ping Li and Xinyu Duan and Zhefeng Wang and Min Zhang},
  booktitle={Proceedings of the 63rd Annual Meeting of the Association for Computational Linguistics (Volume 1: Long Papers)},
  year={2025},
  url={https://aclanthology.org/2025.acl-long.631/},
  doi={10.18653/v1/2025.acl-long.631},
  pages={12895--12915}
}

@inproceedings{he2024zipcache,
  author={He, Yefei and Zhang, Luoming and Wu, Weijia and Liu, Jing and Zhou, Hong and Zhuang, Bohan},
  booktitle={Advances in Neural Information Processing Systems},
  doi={10.52202/079017-2181},
  pages={68287--68307},
  title={{ZipCache: Accurate and Efficient KV Cache Quantization with Salient Token Identification}},
  url={https://proceedings.neurips.cc/paper_files/paper/2024/file/7e57131fdeb815764434b65162c88895-Paper-Conference.pdf},
  volume={37},
  year={2024}
}

@article{tao2025plug,
  title={{Plug-and-Play 1.x-Bit KV Cache Quantization for Video Large Language Models}},
  author={Tao, Keda and You, Haoxuan and Sui, Yang and Qin, Can and Wang, Huan},
  journal={arXiv preprint arXiv:2503.16257},
  year={2025}
}

@inproceedings{zandieh2025qjl,
  title={{QJL: 1-Bit Quantized JL Transform for KV Cache Quantization with Zero Overhead}},
  author={Amir Zandieh and Majid Daliri and Insu Han},
  booktitle={Proceedings of the AAAI Conference on Artificial Intelligence},
  volume={39},
  pages={25805--25813},
  year={2025},
  doi={10.1609/aaai.v39i24.34773},
  url={https://ojs.aaai.org/index.php/AAAI/article/view/34773}
}

@article{duanmu2024skvq,
  title={{SKVQ: Sliding-window Key and Value Cache Quantization for Large Language Models}},
  author={Duanmu, Haojie and Yuan, Zhihang and Li, Xiuhong and Duan, Jiangfei and Zhang, Xingcheng and Lin, Dahua},
  journal={arXiv preprint arXiv:2405.06219},
  year={2024}
}

@article{zandieh2025turboquant,
  title={{TurboQuant: Online Vector Quantization with Near-optimal Distortion Rate}},
  author={Zandieh, Amir and Daliri, Majid and Hadian, Majid and Mirrokni, Vahab},
  journal={arXiv preprint arXiv:2504.19874},
  year={2025}
}

@article{ji2026isoquant,
  title={{IsoQuant: Hardware-Aligned SO(4) Isoclinic Rotations for LLM KV Cache Compression}},
  author={Ji, Zhongping},
  journal={arXiv preprint arXiv:2603.28430},
  year={2026}
}

@misc{pope2026rotorquant,
  title={{RotorQuant: Clifford Algebra Vector Quantization for LLM KV Cache Compression}},
  author={Pope, John D},
  year={2026},
  howpublished={Technical report},
  url={https://www.scrya.com/rotorquant.pdf}
}

@article{frantar2022gptq,
  title={{GPTQ: Accurate Post-Training Quantization for Generative Pre-trained Transformers}},
  author={Frantar, Elias and Ashkboos, Saleh and Hoefler, Torsten and Alistarh, Dan},
  journal={arXiv preprint arXiv:2210.17323},
  year={2022}
}

@inproceedings{lin2024awq,
  author={Lin, Ji and Tang, Jiaming and Tang, Haotian and Yang, Shang and Chen, Wei-Ming and Wang, Wei-Chen and Xiao, Guangxuan and Dang, Xingyu and Gan, Chuang and Han, Song},
  booktitle={Proceedings of Machine Learning and Systems},
  pages={87--100},
  title={{AWQ: Activation-aware Weight Quantization for On-Device LLM Compression and Acceleration}},
  url={https://proceedings.mlsys.org/paper_files/paper/2024/file/42a452cbafa9dd64e9ba4aa95cc1ef21-Paper-Conference.pdf},
  volume={6},
  year={2024}
}

@inproceedings{xiao2023smoothquant,
  title={{SmoothQuant: Accurate and Efficient Post-Training Quantization for Large Language Models}},
  author={Guangxuan Xiao and Ji Lin and Mickael Seznec and Hao Wu and Julien Demouth and Song Han},
  booktitle={International Conference on Machine Learning},
  pages={38087--38099},
  year={2023},
  organization={PMLR},
  url={https://proceedings.mlr.press/v202/xiao23c.html}
}

@inproceedings{ashkboos2024quarot,
  author={Ashkboos, Saleh and Mohtashami, Amirkeivan and Croci, Maximilian L. and Li, Bo and Cameron, Pashmina and Jaggi, Martin and Alistarh, Dan and Hoefler, Torsten and Hensman, James},
  booktitle={Advances in Neural Information Processing Systems},
  doi={10.52202/079017-3180},
  pages={100213--100240},
  title={{QuaRot: Outlier-Free 4-Bit Inference in Rotated LLMs}},
  url={https://proceedings.neurips.cc/paper_files/paper/2024/file/b5b939436789f76f08b9d0da5e81af7c-Paper-Conference.pdf},
  volume={37},
  year={2024}
}

@article{han2025polarquant,
  title={{PolarQuant: Quantizing KV Caches with Polar Transformation}},
  author={Han, Insu and Kacham, Praneeth and Karbasi, Amin and Mirrokni, Vahab and Zandieh, Amir},
  journal={arXiv preprint arXiv:2502.02617},
  year={2025}
}

@article{sun2024massive,
  title={{Massive Activations in Large Language Models}},
  author={Sun, Mingjie and Chen, Xinlei and Kolter, J. Zico and Liu, Zhuang},
  journal={arXiv preprint arXiv:2402.17762},
  year={2024}
}

@inproceedings{wei2023outlier,
  title={{Outlier Suppression+: Accurate quantization of large language models by equivalent and effective shifting and scaling}},
  author={Xiuying Wei and Yunchen Zhang and Yuhang Li and Xiangguo Zhang and Ruihao Gong and Jinyang Guo and Xianglong Liu},
  booktitle={Proceedings of the 2023 Conference on Empirical Methods in Natural Language Processing},
  year={2023},
  url={https://aclanthology.org/2023.emnlp-main.102/},
  doi={10.18653/v1/2023.emnlp-main.102},
  pages={1648--1665}
}

@article{jin2025massive,
  title={{Massive Values in Self-Attention Modules are the Key to Contextual Knowledge Understanding}},
  author={Jin, Mingyu and Mei, Kai and Xu, Wujiang and Sun, Mingjie and Tang, Ruixiang and Du, Mengnan and Liu, Zirui and Zhang, Yongfeng},
  journal={arXiv preprint arXiv:2502.01563},
  year={2025}
}

@article{guo2024active,
  title={{Active-Dormant Attention Heads: Mechanistically Demystifying Extreme-Token Phenomena in LLMs}},
  author={Guo, Tianyu and Pai, Druv and Bai, Yu and Jiao, Jiantao and Jordan, Michael I. and Mei, Song},
  journal={arXiv preprint arXiv:2410.13835},
  year={2024}
}

@inproceedings{guo2024attention,
  title={{Attention Score is not All You Need for Token Importance Indicator in {KV} Cache Reduction: Value Also Matters}},
  author={Zhiyu Guo and Hidetaka Kamigaito and Taro Watanabe},
  booktitle={Proceedings of the 2024 Conference on Empirical Methods in Natural Language Processing},
  year={2024},
  url={https://aclanthology.org/2024.emnlp-main.1178/},
  doi={10.18653/v1/2024.emnlp-main.1178},
  pages={21158--21166}
}

@inproceedings{bondarenko2023quantizable,
  author={Bondarenko, Yelysei and Nagel, Markus and Blankevoort, Tijmen},
  booktitle={Advances in Neural Information Processing Systems},
  doi={10.52202/075280-3282},
  pages={75067--75096},
  title={{Quantizable Transformers: Removing Outliers by Helping Attention Heads Do Nothing}},
  url={https://proceedings.neurips.cc/paper_files/paper/2023/file/edbcb7583fd8921dad78adecfe06a99b-Paper-Conference.pdf},
  volume={36},
  year={2023}
}

@inproceedings{lin2025qserve,
  author={Lin, Yujun and Tang, Haotian and Yang, Shang and Zhang, Zhekai and Xiao, Guangxuan and Gan, Chuang and Han, Song},
  booktitle={Proceedings of Machine Learning and Systems},
  title={{QServe: W4A8KV4 Quantization and System Co-design for Efficient LLM Serving}},
  url={https://proceedings.mlsys.org/paper_files/paper/2025/file/fbe2b2f74a2ece8070d8fb073717bda6-Paper-Conference.pdf},
  volume={7},
  year={2025}
}

@inproceedings{vaswani2017attention,
  author={Vaswani, Ashish and Shazeer, Noam and Parmar, Niki and Uszkoreit, Jakob and Jones, Llion and Gomez, Aidan N. and Kaiser, {\L{}}ukasz and Polosukhin, Illia},
  booktitle={Advances in Neural Information Processing Systems},
  title={{Attention is All you Need}},
  url={https://proceedings.neurips.cc/paper_files/paper/2017/file/3f5ee243547dee91fbd053c1c4a845aa-Paper.pdf},
  volume={30},
  year={2017}
}

@article{an2025systematic,
  title={{Systematic Outliers in Large Language Models}},
  author={An, Yongqi and Zhao, Xu and Yu, Tao and Tang, Ming and Wang, Jinqiao},
  journal={arXiv preprint arXiv:2502.06415},
  year={2025}
}

@inproceedings{liu2023visual,
  author={Liu, Haotian and Li, Chunyuan and Wu, Qingyang and Lee, Yong Jae},
  booktitle={Advances in Neural Information Processing Systems},
  doi={10.52202/075280-1516},
  pages={34892--34916},
  title={{Visual Instruction Tuning}},
  url={https://proceedings.neurips.cc/paper_files/paper/2023/file/6dcf277ea32ce3288914faf369fe6de0-Paper-Conference.pdf},
  volume={36},
  year={2023}
}

@inproceedings{liu2024improved,
  title={{Improved Baselines with Visual Instruction Tuning}},
  author={Liu, Haotian and Li, Chunyuan and Li, Yuheng and Lee, Yong Jae},
  booktitle={Proceedings of the IEEE/CVF Conference on Computer Vision and Pattern Recognition},
  pages={26296--26306},
  year={2024},
  url={https://openaccess.thecvf.com/content/CVPR2024/html/Liu_Improved_Baselines_with_Visual_Instruction_Tuning_CVPR_2024_paper.html}
}

@article{xiao2023efficient,
  title={{Efficient Streaming Language Models with Attention Sinks}},
  author={Xiao, Guangxuan and Tian, Yuandong and Chen, Beidi and Han, Song and Lewis, Mike},
  journal={arXiv preprint arXiv:2309.17453},
  year={2023}
}

@article{agarwal2024hadacore,
  title={{HadaCore: Tensor Core Accelerated Hadamard Transform Kernel}},
  author={Agarwal, Krish and Astra, Rishi and Hoque, Adnan and Srivatsa, Mudhakar and Ganti, Raghu and Wright, Less and Chen, Sijia},
  journal={arXiv preprint arXiv:2412.08832},
  year={2024}
}

@article{du2025bitdecoding,
  title={{BitDecoding: Unlocking Tensor Cores for Long-Context LLMs with Low-Bit KV Cache}},
  author={Du, Dayou and Cao, Shijie and Cheng, Jianyi and Mai, Luo and Cao, Ting and Yang, Mao},
  journal={arXiv preprint arXiv:2503.18773},
  year={2025}
}

@article{yang2025qwen3,
  title={{Qwen3 Technical Report}},
  author={Yang, An and Li, Anfeng and Yang, Baosong and Zhang, Beichen and Hui, Binyuan and Zheng, Bo and Yu, Bowen and Gao, Chang and Huang, Chengen and Lv, Chenxu and others},
  journal={arXiv preprint arXiv:2505.09388},
  year={2025}
}

@article{grattafiori2024llama,
  title={{The Llama 3 Herd of Models}},
  author={Grattafiori, Aaron and Dubey, Abhimanyu and Jauhri, Abhinav and Pandey, Abhinav and Kadian, Abhishek and Al-Dahle, Ahmad and Letman, Aiesha and Mathur, Akhil and Schelten, Alan and Vaughan, Alex and others},
  journal={arXiv preprint arXiv:2407.21783},
  year={2024}
}

@article{xu2025qwen3,
  title={{Qwen3-Omni Technical Report}},
  author={Xu, Jin and Guo, Zhifang and Hu, Hangrui and Chu, Yunfei and Wang, Xiong and He, Jinzheng and Wang, Yuxuan and Shi, Xian and He, Ting and Zhu, Xinfa and others},
  journal={arXiv preprint arXiv:2509.17765},
  year={2025}
}

@inproceedings{bai2024longbench,
  title={{{L}ong{B}ench: A Bilingual, Multitask Benchmark for Long Context Understanding}},
  author={Yushi Bai and Xin Lv and Jiajie Zhang and Hongchang Lyu and Jiankai Tang and Zhidian Huang and Zhengxiao Du and Xiao Liu and Aohan Zeng and Lei Hou and others},
  booktitle={Proceedings of the 62nd Annual Meeting of the Association for Computational Linguistics (Volume 1: Long Papers)},
  year={2024},
  url={https://aclanthology.org/2024.acl-long.172/},
  doi={10.18653/v1/2024.acl-long.172},
  pages={3119--3137}
}

@misc{gkamradt2023needle,
  author={Kamradt, Greg},
  title={{LLMTest\_NeedleInAHaystack}},
  year={2023},
  howpublished={GitHub},
  url={https://github.com/gkamradt/LLMTest_NeedleInAHaystack}
}

@article{liu2024ocrbench,
  title={{OCRBench: on the hidden mystery of OCR in large multimodal models}},
  author={Liu, Yuliang and Li, Zhang and Huang, Mingxin and Yang, Biao and Yu, Wenwen and Li, Chunyuan and Yin, Xu-Cheng and Liu, Cheng-Lin and Jin, Lianwen and Bai, Xiang},
  journal={Science China Information Sciences},
  year={2024},
  doi={10.1007/s11432-024-4235-6},
  volume={67},
  number={12},
  pages={220102}
}

@article{hong2025worldsense,
  title={{WorldSense: Evaluating Real-world Omnimodal Understanding for Multimodal LLMs}},
  author={Hong, Jack and Yan, Shilin and Cai, Jiayin and Jiang, Xiaolong and Hu, Yao and Xie, Weidi},
  journal={arXiv preprint arXiv:2502.04326},
  year={2025}
}

@article{bai2025qwen3,
  title={{Qwen3-VL Technical Report}},
  author={Bai, Shuai and Cai, Yuxuan and Chen, Ruizhe and Chen, Keqin and Chen, Xionghui and Cheng, Zesen and Deng, Lianghao and Ding, Wei and Gao, Chang and Ge, Chunjiang and others},
  journal={arXiv preprint arXiv:2511.21631},
  year={2025}
}

@misc{turney2026turboquantplus,
  author={Turney, Tom and {Contributors}},
  title={{TurboQuant+}},
  howpublished={GitHub repository},
  year={2026},
  note={Accessed 2026-09-25},
  url={https://github.com/TheTom/turboquant_plus}
}

@inproceedings{mathew2021docvqa,
  title={{DocVQA: A Dataset for VQA on Document Images}},
  author={Mathew, Minesh and Karatzas, Dimosthenis and Jawahar, C.V.},
  booktitle={Proceedings of the IEEE/CVF Winter Conference on Applications of Computer Vision},
  pages={2200--2209},
  year={2021},
  url={https://openaccess.thecvf.com/content/WACV2021/html/Mathew_DocVQA_A_Dataset_for_VQA_on_Document_Images_WACV_2021_paper.html}
}

@inproceedings{qiu2025gated,
  author={Qiu, Zihan and Wang, Zekun and Zheng, Bo and Huang, Zeyu and Wen, Kaiyue and Yang, Songlin and Men, Rui and Yu, Le and Huang, Fei and Huang, Suozhi and others},
  booktitle={Advances in Neural Information Processing Systems},
  doi={10.52202/085713-3345},
  pages={100092--100118},
  title={{Gated Attention for Large Language Models: Non-linearity, Sparsity, and Attention-Sink-Free}},
  url={https://proceedings.neurips.cc/paper_files/paper/2025/file/904e89bb4e632e75fb47f093b620b257-Paper-Conference.pdf},
  volume={38, Main Conference},
  year={2025}
}

@article{liang2025tweo,
  title={{TWEO: Transformers Without Extreme Outliers Enables FP8 Training And Quantization For Dummies}},
  author={Liang, Guang and Shao, Jie and Tang, Ningyuan and Liu, Xinyao and Wu, Jianxin},
  journal={arXiv preprint arXiv:2511.23225},
  year={2025}
}

@inproceedings{kang2025see,
  author={Kang, Seil and Kim, Jinyeong and Kim, Junhyeok and Hwang, Seong Jae},
  booktitle={International Conference on Learning Representations},
  pages={87676--87703},
  title={{See What You Are Told: Visual Attention Sink in Large Multimodal Models}},
  url={https://proceedings.iclr.cc/paper_files/paper/2025/file/da8a39bc39ae1c89dd6ebb1e3bcbb3f3-Paper-Conference.pdf},
  volume={2025},
  year={2025}
}

@article{tu2026attention,
  title={{Attention Reallocation: Towards Zero-cost and Controllable Hallucination Mitigation of MLLMs}},
  author={Tu, Chongjun and Ye, Peng and Zhou, Dongzhan and Bai, Lei and Yu, Gang and Chen, Tao and Ouyang, Wanli},
  journal={International Journal of Computer Vision},
  year={2026},
  doi={10.1007/s11263-025-02607-z},
  volume={134},
  number={1},
  pages={22}
}

@article{zuhri2025softpick,
  title={{Softpick: No Attention Sink, No Massive Activations with Rectified Softmax}},
  author={Zuhri, Zayd M. K. and Fuadi, Erland Hilman and Aji, Alham Fikri},
  journal={arXiv preprint arXiv:2504.20966},
  year={2025}
}

@article{xiong2025dope,
  title={{DoPE: Denoising Rotary Position Embedding}},
  author={Xiong, Jing and Fan, Liyang and Shen, Hui and Su, Zunhai and Yang, Min and Kong, Lingpeng and Wong, Ngai},
  journal={arXiv preprint arXiv:2511.09146},
  year={2025}
}

@article{agarwal2025gpt,
  title={{gpt-oss-120b \& gpt-oss-20b Model Card}},
  author={{OpenAI}},
  journal={arXiv preprint arXiv:2508.10925},
  year={2025}
}

@inproceedings{kumar2026mmaupro,
  title={{MMAU-Pro: A Challenging and Comprehensive Benchmark for Holistic Evaluation of Audio General Intelligence}},
  author={Sonal Kumar and {\v{S}}imon Sedlá{\v{c}}ek and Vaibhavi Lokegaonkar and Fernando López and Wenyi Yu and Nishit Anand and Hyeonggon Ryu and Lichang Chen and Maxim Pli{\v{c}}ka and Miroslav Hlavá{\v{c}}ek and others},
  booktitle={Proceedings of the AAAI Conference on Artificial Intelligence},
  volume={40},
  pages={22688--22697},
  year={2026},
  doi={10.1609/aaai.v40i27.39430},
  url={https://ojs.aaai.org/index.php/AAAI/article/view/39430}
}
\bibliographystyle{iclr2027_conference}

\newpage
\appendix

\section*{Appendix Contents}  
\startcontents[appendix] 
\printcontents[appendix]{}{1}{\section*{}}
\newpage

\section{Limitations and Future Directions}
\label{app:limitations}

Although OScaR imposes lower overhead than most existing quantization frameworks, its online rotation and token-wise scaling operations incur non-trivial computational costs relative to plain per-channel quantization.
Specifically, \textit{Canalized Rotation} requires online computation in the presence of RoPE, which precludes offline fusion of Query and Key Hadamard transforms with weight matrices~\citep{ashkboos2024quarot,su2025rotatekv}. 
In our system design, we mitigate this overhead via two key strategies: (i) employing HadaCore~\citep{agarwal2024hadacore} for efficient token-wise FHT computation on Tensor Cores; and (ii) fusing kernels to consolidate (FHT + scaling) and (dequantization + de-scaling + attention) into dedicated CUDA operators.
Future work may explore alternative \textit{Canalized Rotation} that further reduce online overhead or enable more efficient hardware-aware implementations.

Furthermore, OScaR represents a highly general framework for KV cache quantization, applicable to LLMs and beyond (i.e., multi-modal and omni-modal LLMs). However, our current experiments are primarily conducted on models with LLM backbones.
We posit that OScaR can be applied to autoregressive inference beyond LLMs that requires KV caching, including streaming vision models such as StreamVGGT~\citep{su2026xstreamvggt,zhuo2025streaming}, visual autoregressive models~\citep{li2025memory,qin2026head}, and diffusion LLMs with KV cache~\citep{wu2025fast}. However, these models often exhibit architectural characteristics that differ substantially from standard LLM backbones, and we have not yet conducted extensive experiments in these domains. Moreover, KV cache compression in such models remains an emerging area. We leave a more thorough evaluation across diverse model families to future work.

\section{Related Work}
\label{app:related_work}

\subsection{KV Cache Quantization}

Quantization is essential for efficient deployment of LLMs, with seminal works such as GPTQ, AWQ, and SmoothQuant establishing effective methods for weight and activation compression~\citep{frantar2022gptq,lin2024awq,xiao2023smoothquant,xiao2025exploring,zhang2026beyond}. As context lengths increase, the KV cache has emerged as the dominant memory bottleneck during decoding, necessitating specialized quantization strategies~\citep{liu2024kivi,li2024survey}.
Existing approaches can be broadly categorized by their quantization granularity: per-token, per-channel, and per-element paradigms. Per-token quantization aligns with the incremental dynamics of auto-regressive decoding but remains vulnerable to persistent channel-wise outliers in Key tensors~\citep{liu2024kivi,hooper2024kvquant}. To address this, methods such as QuaRot, RotateKV, and ZipCache employ transformations including rotation and smoothing to redistribute outlier energy~\citep{ashkboos2024quarot,su2025rotatekv,he2024zipcache,duanmu2024skvq}. Per-channel approaches, including KIVI, KVQuant, and OTT, exploit intrinsic channel-wise outlier distributions to reduce quantization difficulty~\citep{liu2024kivi,hooper2024kvquant,su2025accurate}. Recently, per-element paradigms such as TurboQuant and its extensions leverage randomized rotations combined with residual error correction to achieve KV cache compression~\citep{zandieh2025turboquant,pope2026rotorquant,ji2026isoquant,zandieh2025qjl,han2025polarquant}. While these methods provide rigorous theoretical guarantees, their complex pipelines often result in high implementation overhead and practical deviations during deployment.
Despite these advancements, accurate and lightweight KV cache compression at extreme bit-widths remains a challenging problem. Moreover, specialized studies on multi-modal and omni-modal LLMs are still limited.

\subsection{Outliers in Large Language Models}

Outliers in LLMs fundamentally disrupt numerical precision and pose a critical challenge for high-fidelity quantization~\citep{nagel2021white,wei2023outlier,sun2024massive,su2025kvsink,su2026attention,zhang2026locate}. These outliers can be broadly categorized as channel-wise and token-wise based on their distributional characteristics.
Channel-wise outliers exhibit disproportionately large magnitudes in specific feature dimensions, predominantly appearing in Key and Query tensors while remaining comparatively subdued in Value tensors~\citep{liu2024kivi,hooper2024kvquant,jin2025massive}. Token-level outliers manifest in two distinct forms. The first consists of systematic activation outlier tokens arising from the outputs of down-projection layers and inter-block hidden states, which can reach magnitudes tens of thousands of times larger than the median, severely destabilizing activation quantization~\citep{sun2024massive,su2025unveiling,ashkboos2024quarot,an2025systematic}. The second consists of attention outlier tokens, where specific tokens exhibit markedly reduced norms across Query, Key, and Value tensors~\citep{su2025kvsink,bondarenko2023quantizable,guo2024attention,guo2024active}.
Both channel-wise outliers and the second form of token-level outliers are closely associated with representational collapse under extreme KV cache compression. While per-channel paradigms and equivalent transformations can effectively mitigate channel-wise impacts~\citep{xiao2023smoothquant,ashkboos2024quarot,duanmu2024skvq,lin2025qserve}, existing methods often inadequately address token-level outliers. Techniques such as OTT and RotateKV trace and preserve a small number of outlier tokens with high precision in text-only LLMs, but they introduce hardware fragmentation and mixed-precision overheads, limiting the achievable effective compression~\citep{su2025accurate,su2025rotatekv,su2025kvsink,duanmu2024skvq,su2025akvq,hooper2024kvquant}.
In this work, we further characterize TNI across X-LLMs. OScaR addresses TNI through \textit{Canalized Rotation} and \textit{Omni-Token Scaling}, enabling uniform and efficient mitigation of TNI, including principled handling of outlier tokens.

\section{Algorithm of OScaR}
\label{app:algorithm}

This section presents the OScaR algorithm discussed in Section~\ref{sec:method}. As shown in Algorithm~\ref{algo:oscar}, we first describe the preprocessing step that applies Hadamard transforms to $\mathbf{W}_V$ and $\mathbf{W}_O$, followed by the pseudocode for OScaR during attention computation in both the prefill and decoding phases.
\section{Preliminaries on Low-Bit Quantization}
\label{app:lowbit}

Low-bit quantization is an effective method to reduce the memory footprint of the KV cache~\citep{liu2024kivi,li2024survey,liu2025kv}. Consider the widely adopted asymmetric uniform quantization scheme, which maps floating-point values to $b$-bit integers. This scheme is parameterized by the quantization step size $\Delta$ and zero-point $z$, computed from the dynamic range $[x_{\min}, x_{\max}]$:
\begin{equation}
\Delta = \frac{x_{\max} - x_{\min}}{2^b - 1}, \quad z = \left\lfloor -\frac{x_{\min}}{\Delta} \right\rceil.
\end{equation}
The step size $\Delta$ determines the numerical resolution, with larger values yielding coarser representations and higher quantization error~\citep{nagel2021white}. The quantized integer $Q(x)$ and its reconstruction $\hat{x}$ are given by:
\begin{equation}
Q(x) = \text{clamp}\left( \left\lfloor \frac{x}{\Delta} \right\rceil + z, 0, 2^b - 1 \right), \quad \hat{x} = (Q(x) - z) \cdot \Delta.
\end{equation}

For extreme low-bit settings (e.g., $b \leq 4$), the reduced numerical resolution often leads to severe accuracy degradation, particularly in the presence of outliers. This necessitates outlier-aware strategies, including rotation-based energy redistribution~\citep{ashkboos2024quarot,su2025rotatekv}, residual error correction~\citep{zandieh2025turboquant}, or mixed-precision preservation of salient tokens~\citep{liu2024kivi,su2025accurate}.
\section{Token Norm Imbalance in Text-Only LLMs}
\label{app:llm}

As discussed in Section~\ref{sec:revisiting}, we visualize the L2 norm distributions and heatmaps of Query, Key, and Value states across multiple text-only LLMs using the following prompt:

\texttt{Summer is warm.\string\n Winter is cold.\string\n Spring is mild.\string\n Autumn is crisp.}

\begin{figure}[t]
    \centering
    \begin{subfigure}[b]{0.325\textwidth}
        \includegraphics[width=0.9\textwidth]{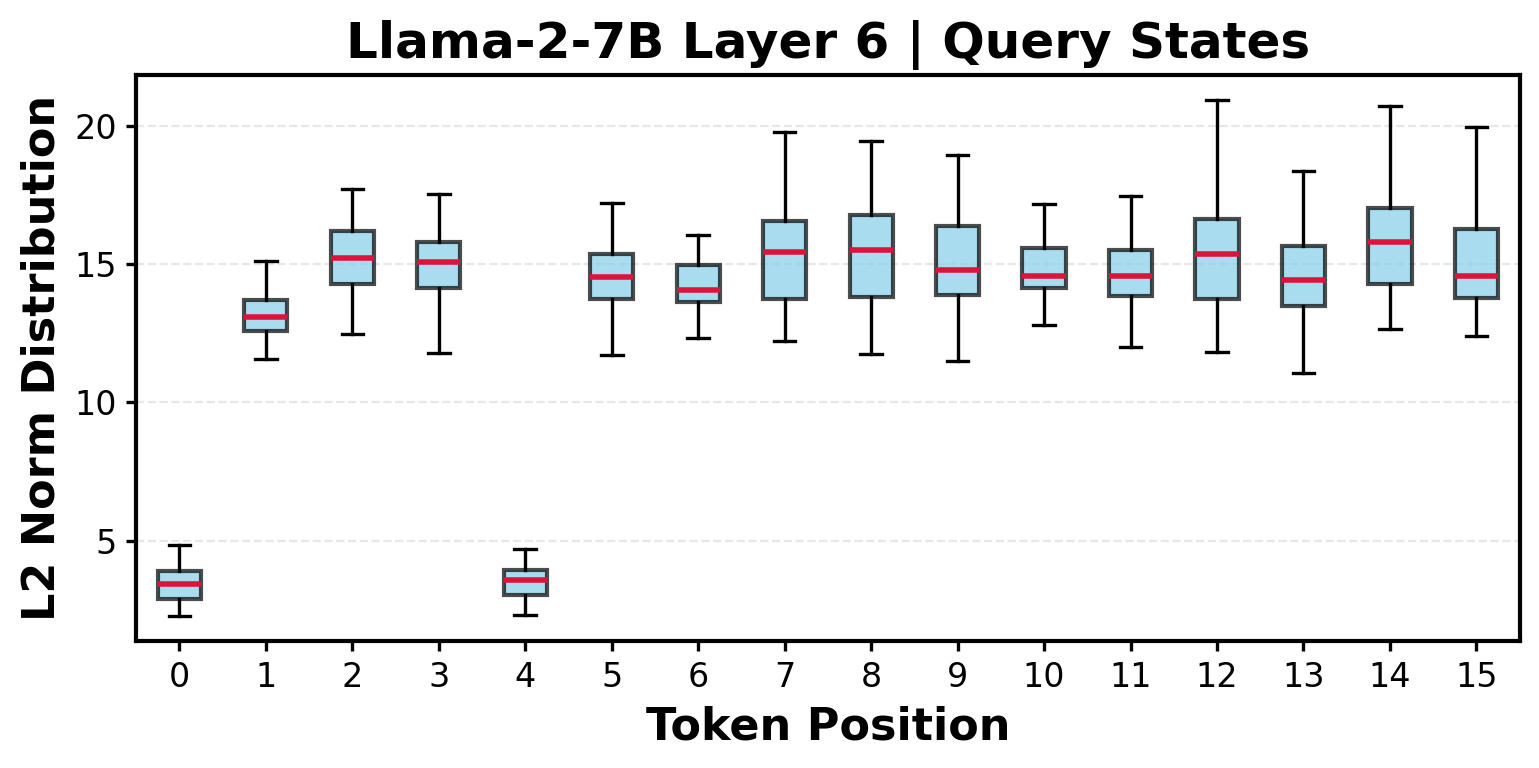}
        \caption{Query L2 norm distribution}
    \end{subfigure}
    \begin{subfigure}[b]{0.325\textwidth}
        \includegraphics[width=0.9\textwidth]{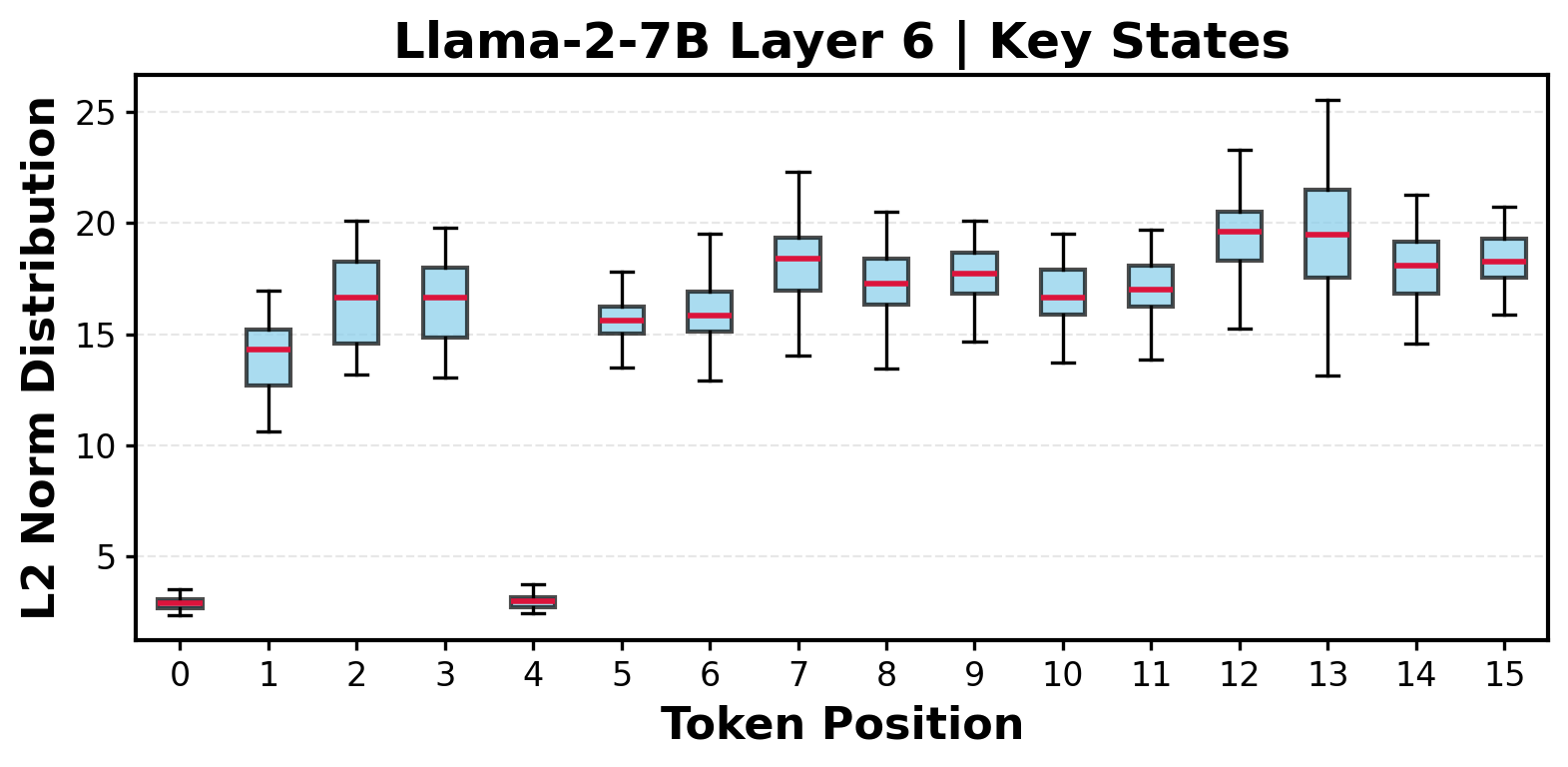}
        \caption{Key L2 norm distribution}
    \end{subfigure}
    \begin{subfigure}[b]{0.325\textwidth}
        \includegraphics[width=0.9\textwidth]{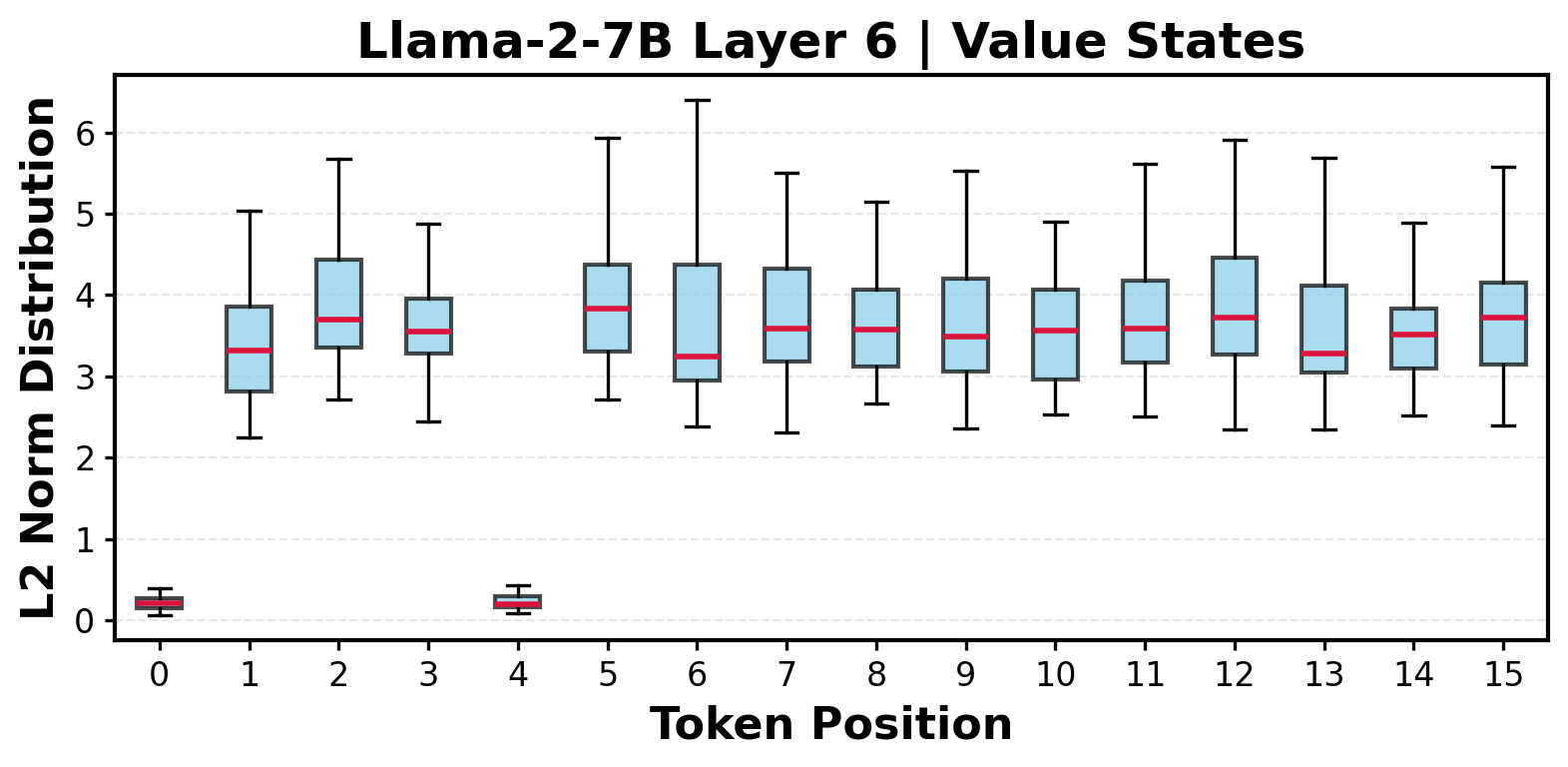}
        \caption{Value L2 norm distribution}
    \end{subfigure}
    \vspace{0.1cm}
    \begin{subfigure}[b]{0.325\textwidth}
        \centering
        \includegraphics[width=\textwidth]{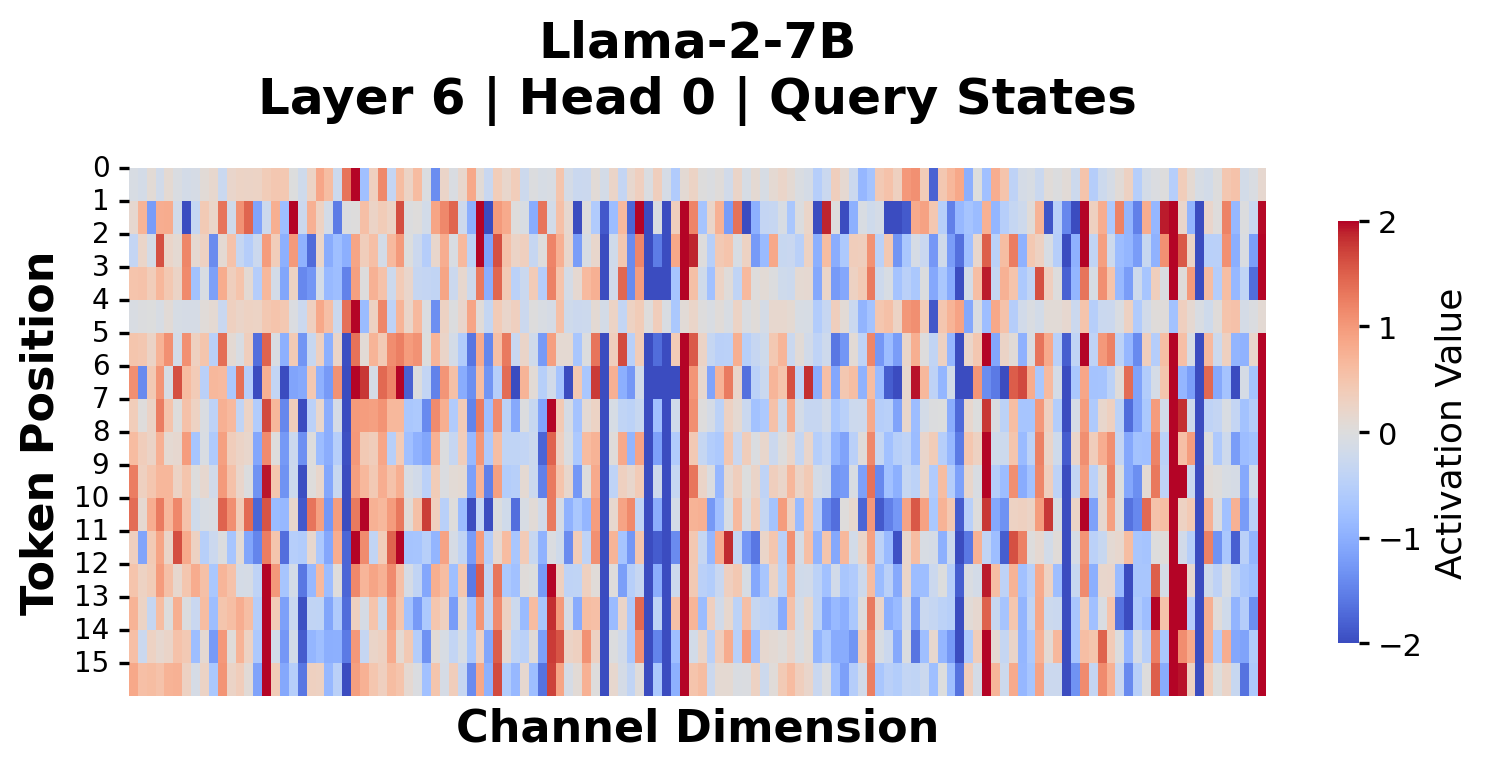}
        \caption{Query heatmap}
    \end{subfigure}
    \begin{subfigure}[b]{0.325\textwidth}
        \centering
        \includegraphics[width=\textwidth]{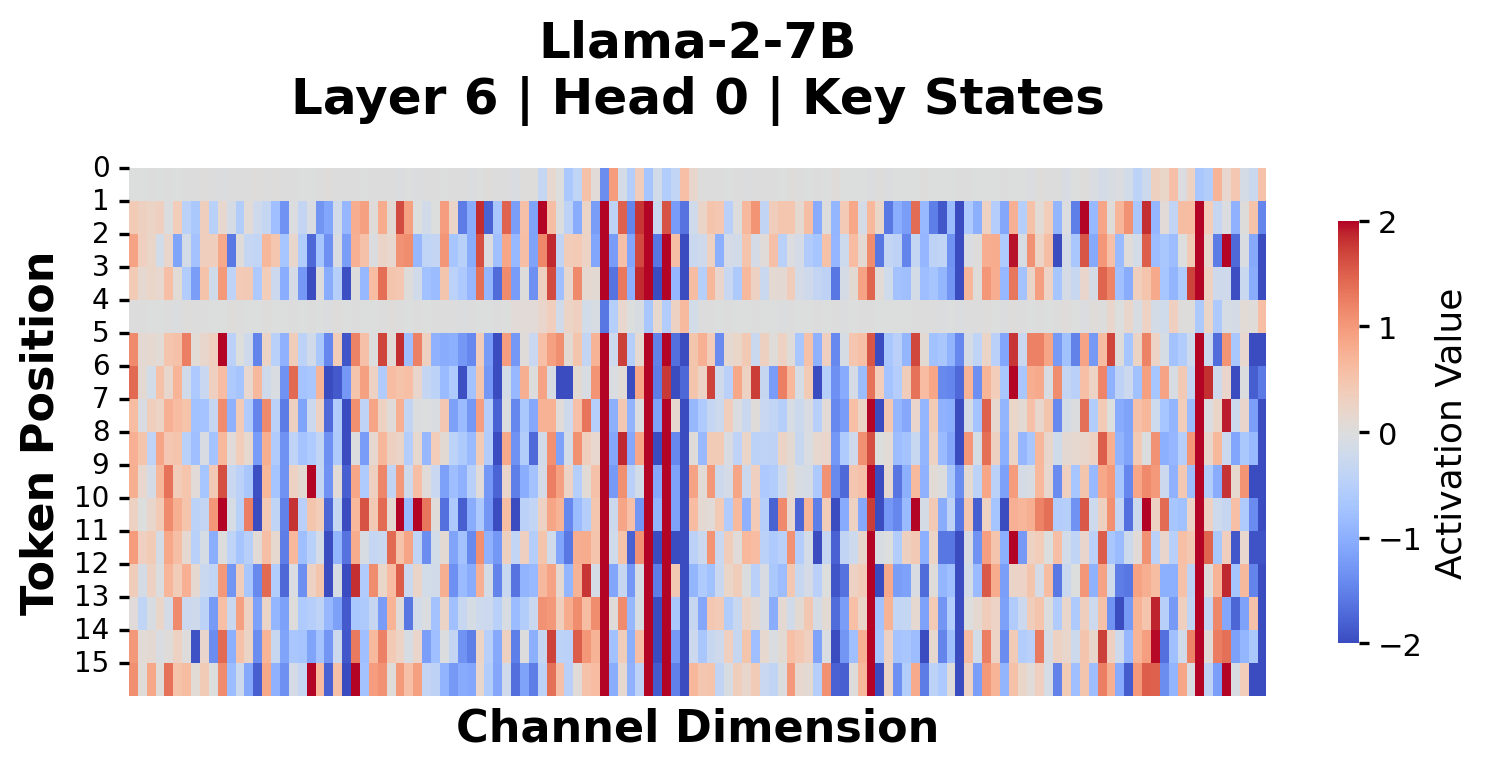}
        \caption{Key heatmap}
    \end{subfigure}
    \begin{subfigure}[b]{0.325\textwidth}
        \centering
        \includegraphics[width=\textwidth]{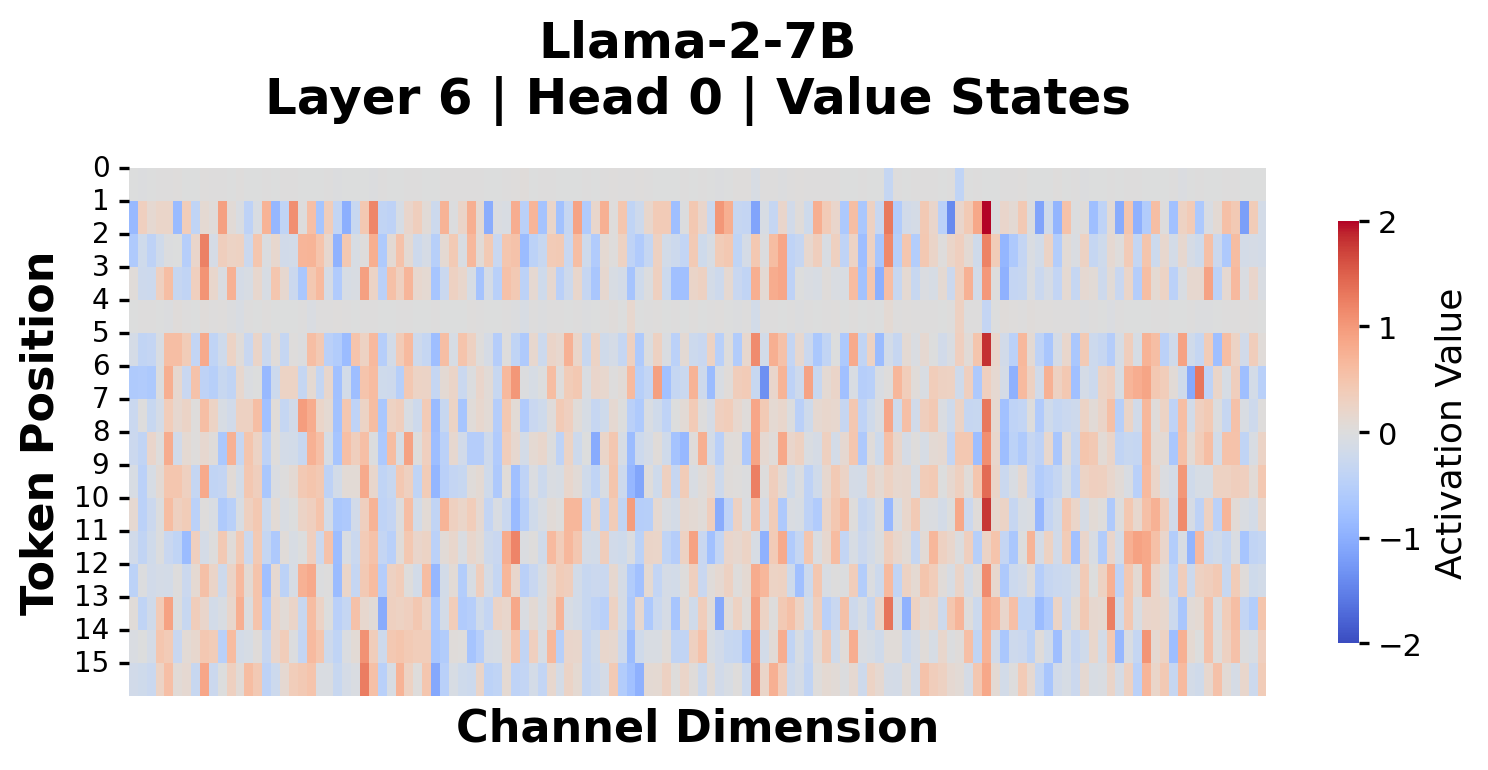}
        \caption{Value heatmap}
    \end{subfigure}
    
    \caption{L2 norm distributions (top row) and value heatmaps (bottom row) of Query, Key, and Value states in Layer 6 of Llama-2-7B.}
    \label{fig:TNI-llama-2-7b-layer-6}
\end{figure}

\begin{figure}[t]
    \centering
    \begin{subfigure}[b]{0.325\textwidth}
        \includegraphics[width=0.9\textwidth]{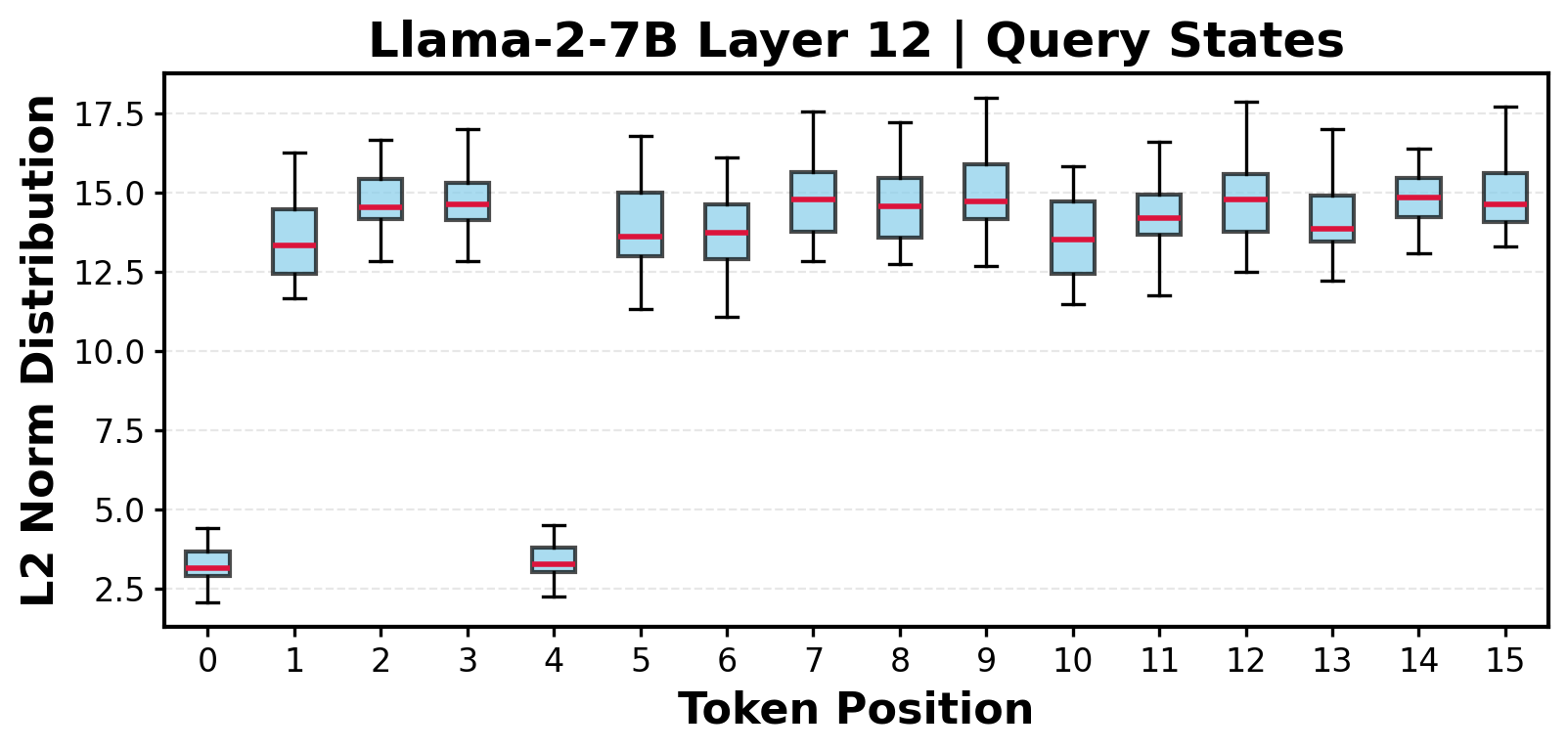}
        \caption{Query L2 norm distribution}
    \end{subfigure}
    \begin{subfigure}[b]{0.325\textwidth}
        \includegraphics[width=0.9\textwidth]{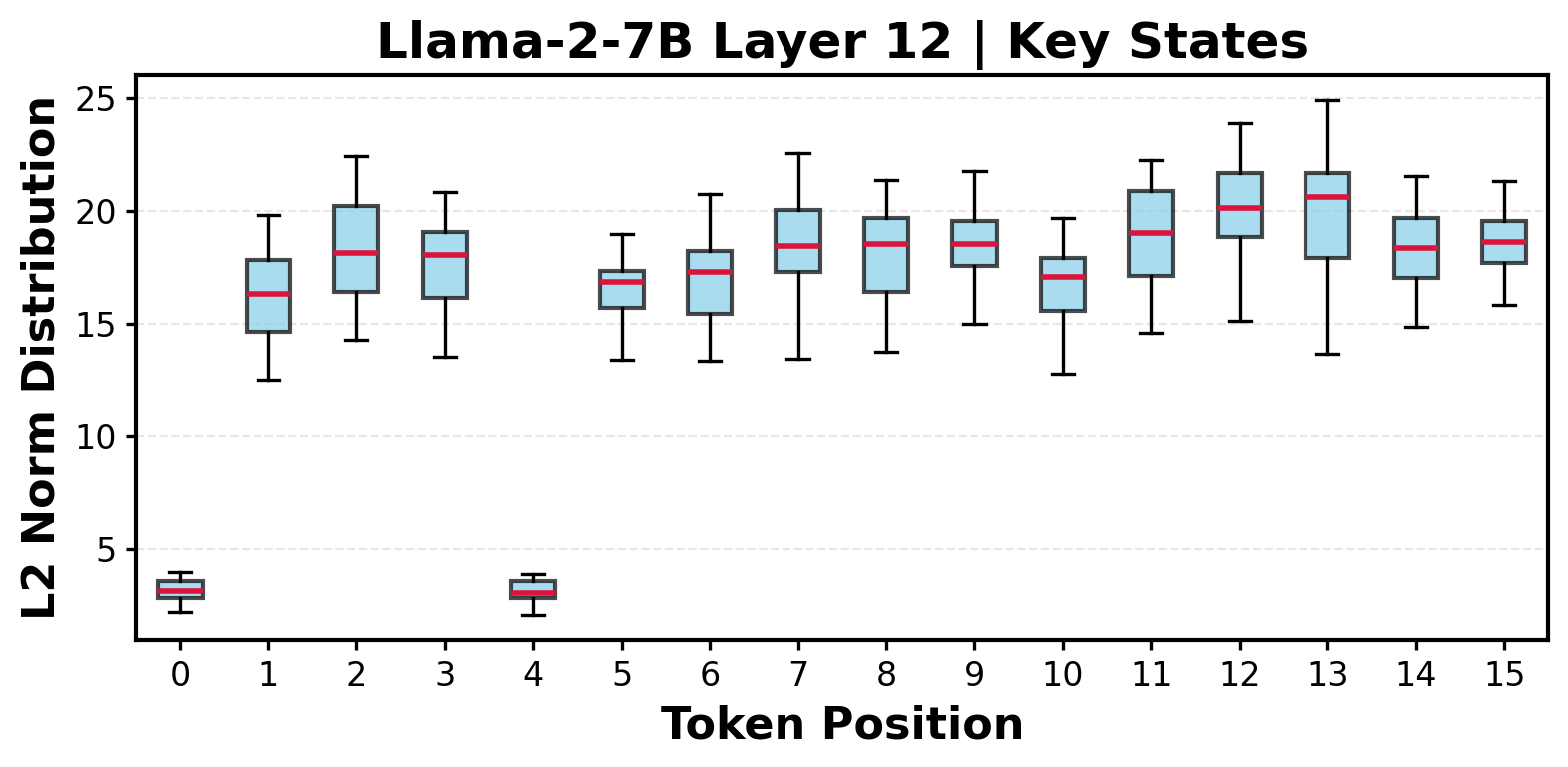}
        \caption{Key L2 norm distribution}
    \end{subfigure}
    \begin{subfigure}[b]{0.325\textwidth}
        \includegraphics[width=0.9\textwidth]{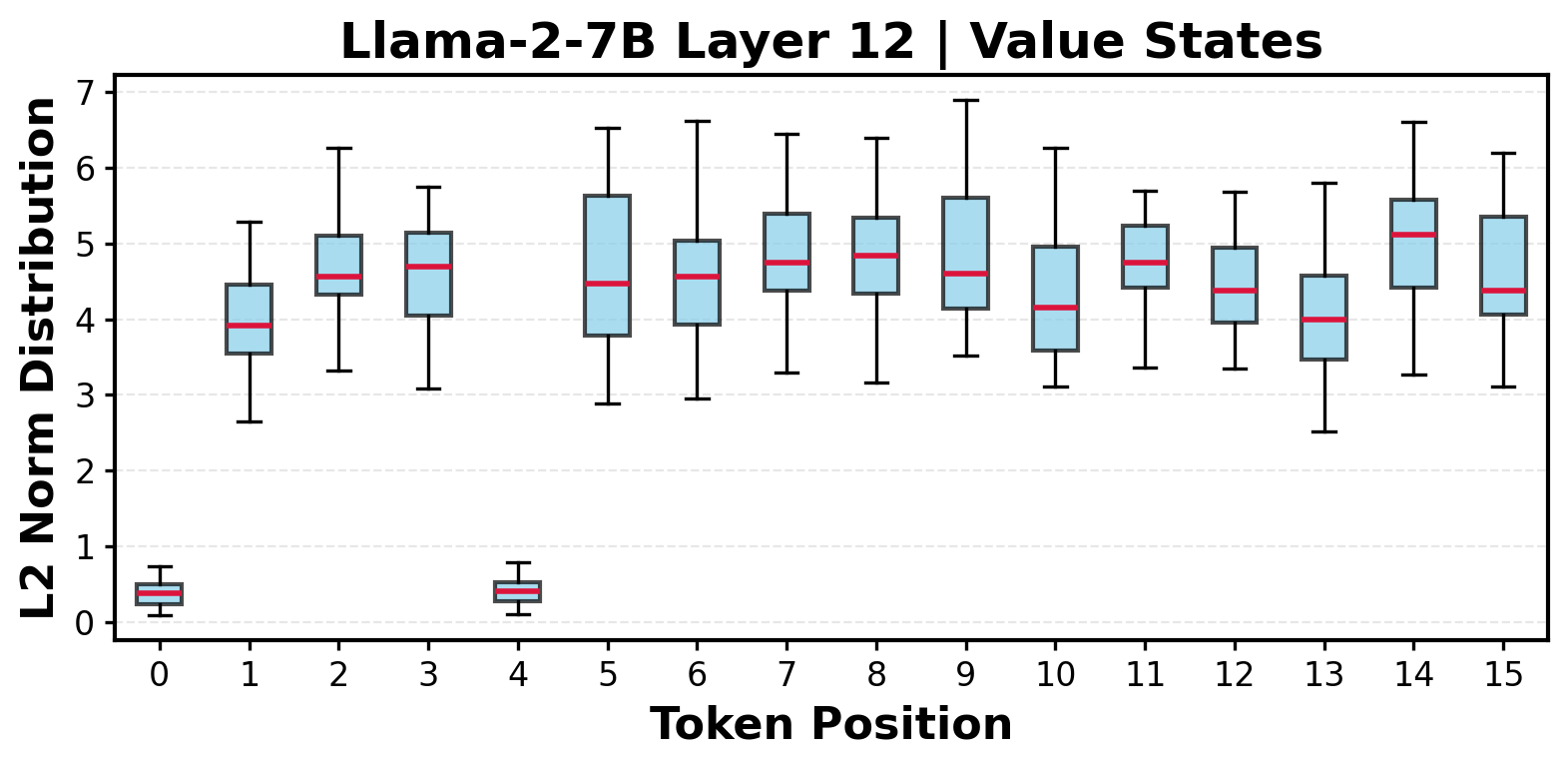}
        \caption{Value L2 norm distribution}
    \end{subfigure}
    \vspace{0.1cm}
    \begin{subfigure}[b]{0.325\textwidth}
        \centering
        \includegraphics[width=\textwidth]{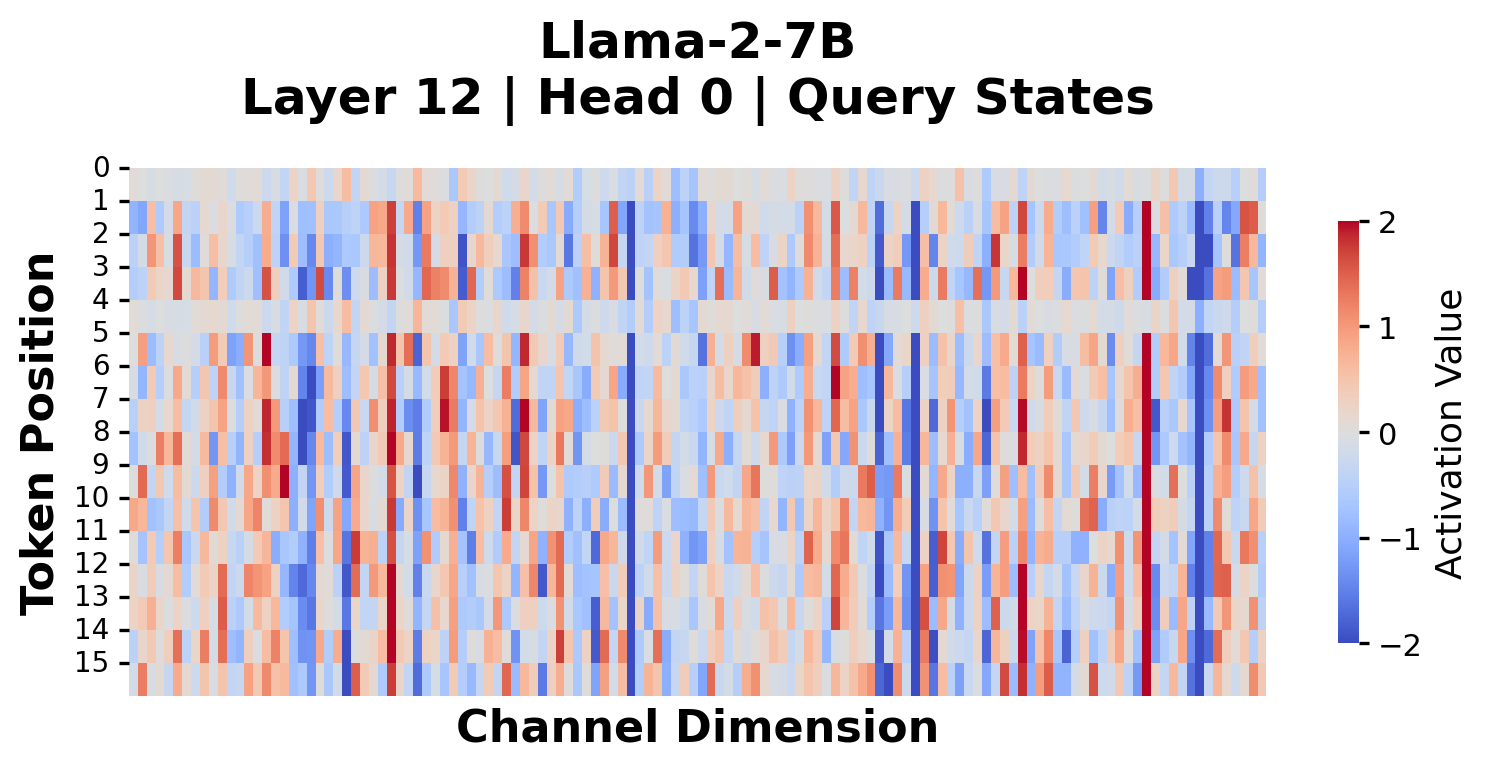}
        \caption{Query heatmap}
    \end{subfigure}
    \begin{subfigure}[b]{0.325\textwidth}
        \centering
        \includegraphics[width=\textwidth]{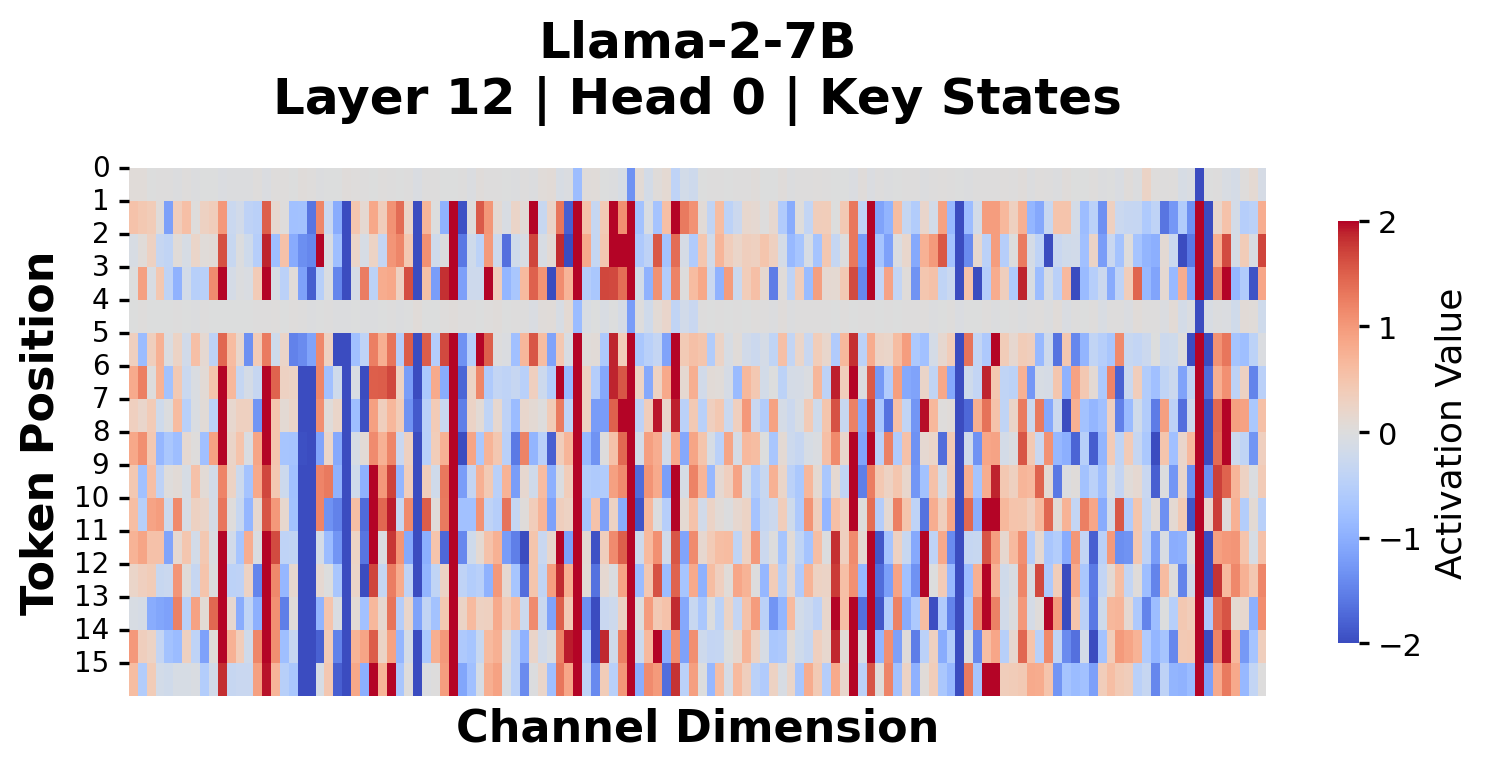}
        \caption{Key heatmap}
    \end{subfigure}
    \begin{subfigure}[b]{0.325\textwidth}
        \centering
        \includegraphics[width=\textwidth]{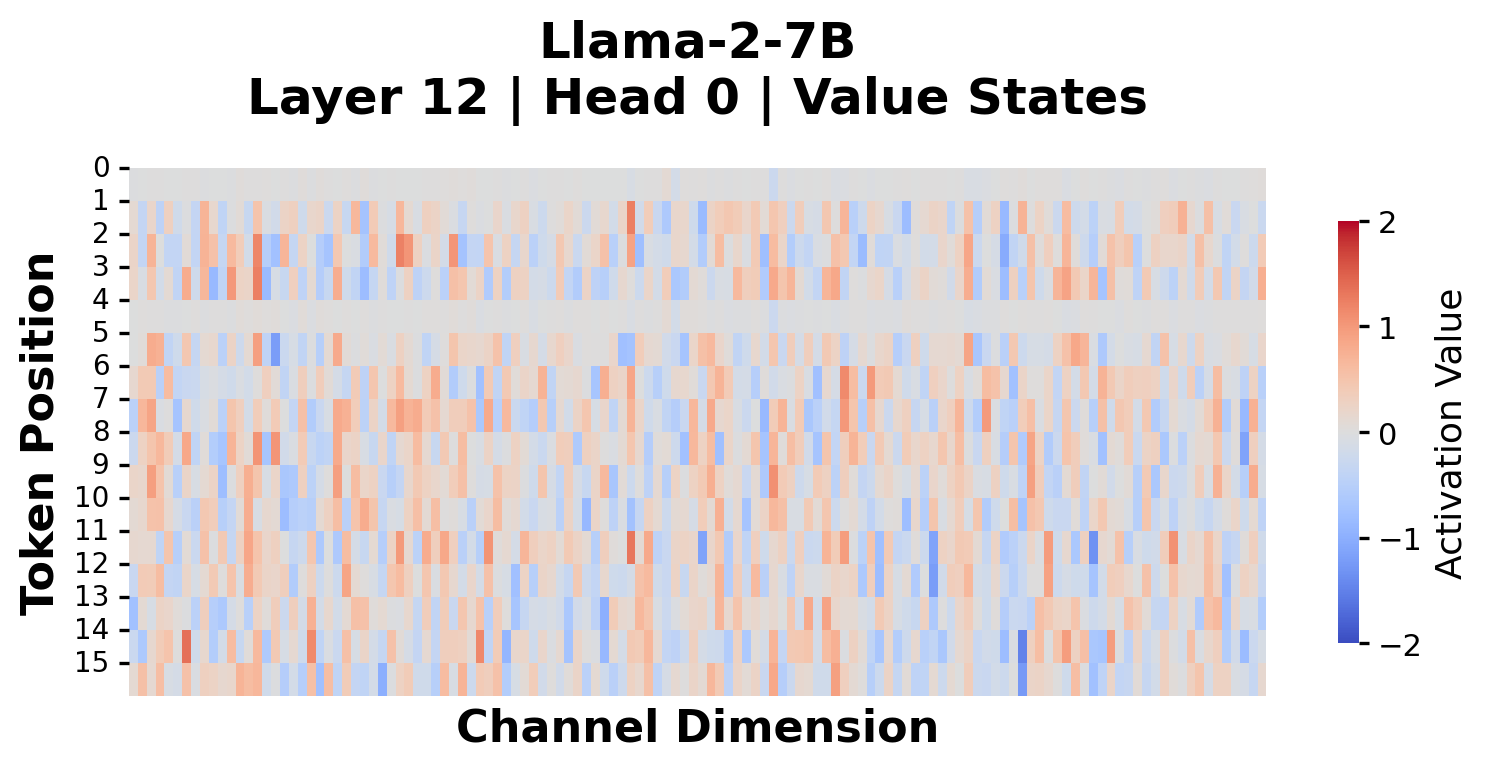}
        \caption{Value heatmap}
    \end{subfigure}
    
    \caption{L2 norm distributions (top row) and value heatmaps (bottom row) of Query, Key, and Value states in Layer 12 of Llama-2-7B.}
    \label{fig:TNI-llama-2-7b-layer-12}
\end{figure}

\begin{figure}[t]
    \centering
    \begin{subfigure}[b]{0.325\textwidth}
        \includegraphics[width=0.9\textwidth]{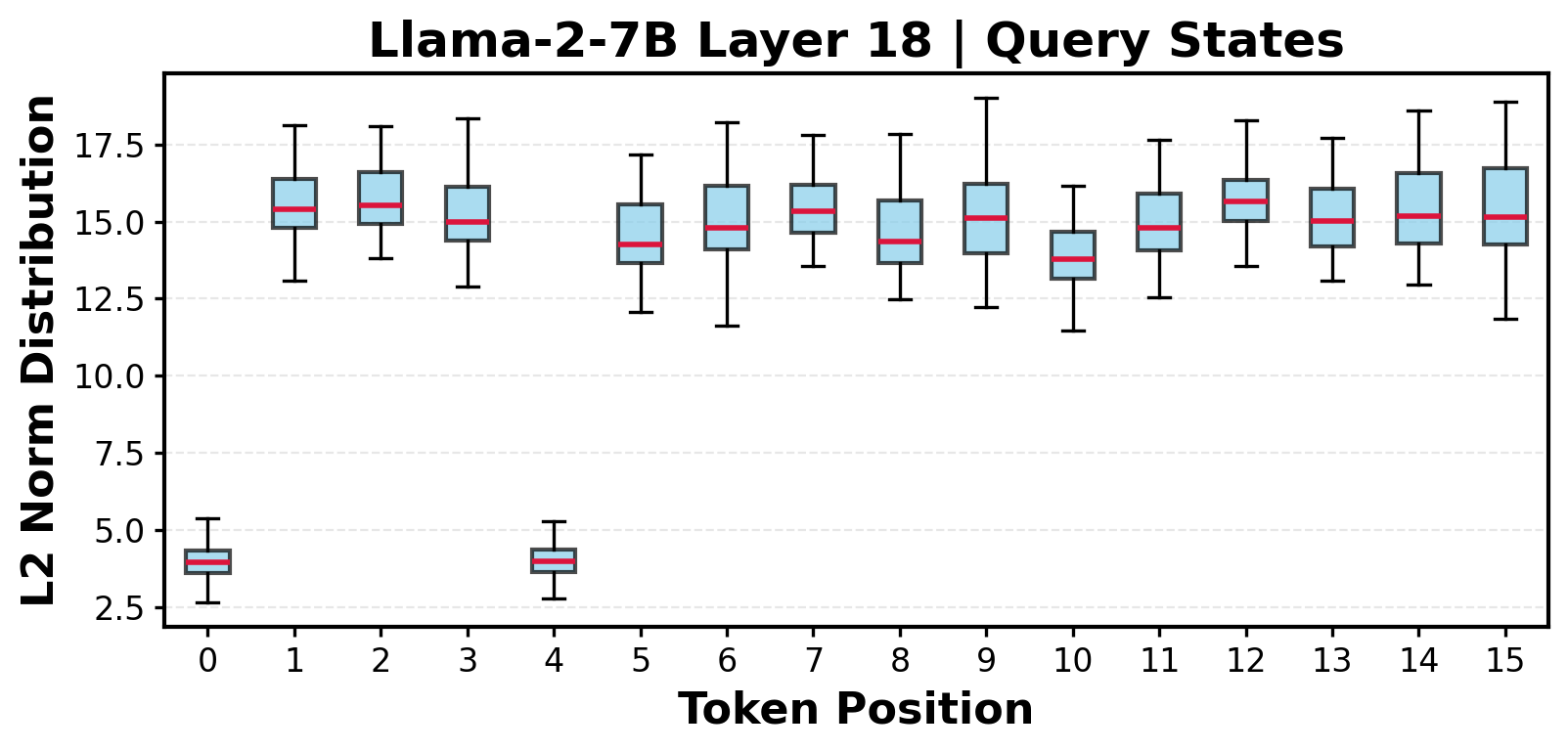}
        \caption{Query L2 norm distribution}
    \end{subfigure}
    \begin{subfigure}[b]{0.325\textwidth}
        \includegraphics[width=0.9\textwidth]{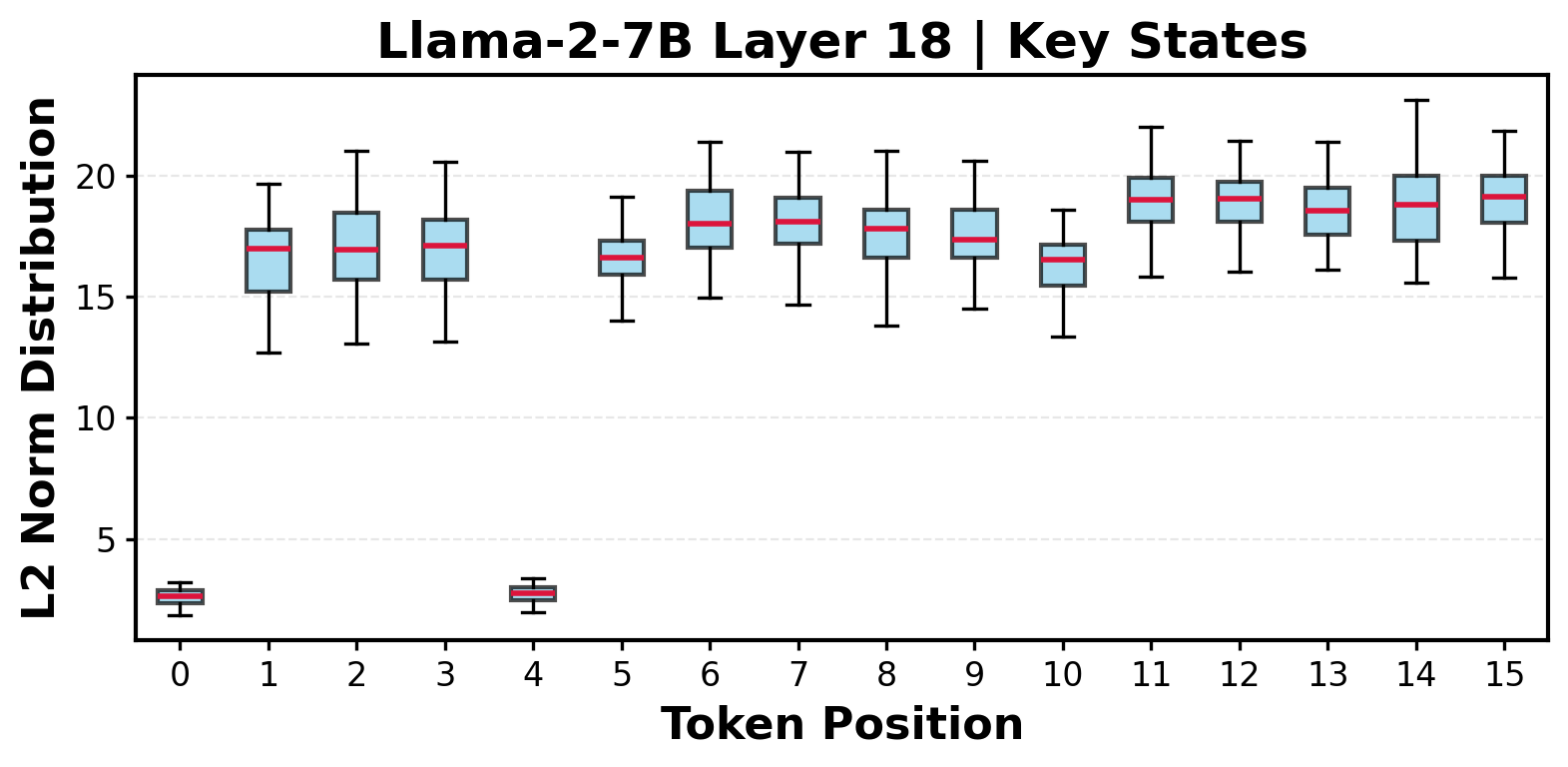}
        \caption{Key L2 norm distribution}
    \end{subfigure}
    \begin{subfigure}[b]{0.325\textwidth}
        \includegraphics[width=0.9\textwidth]{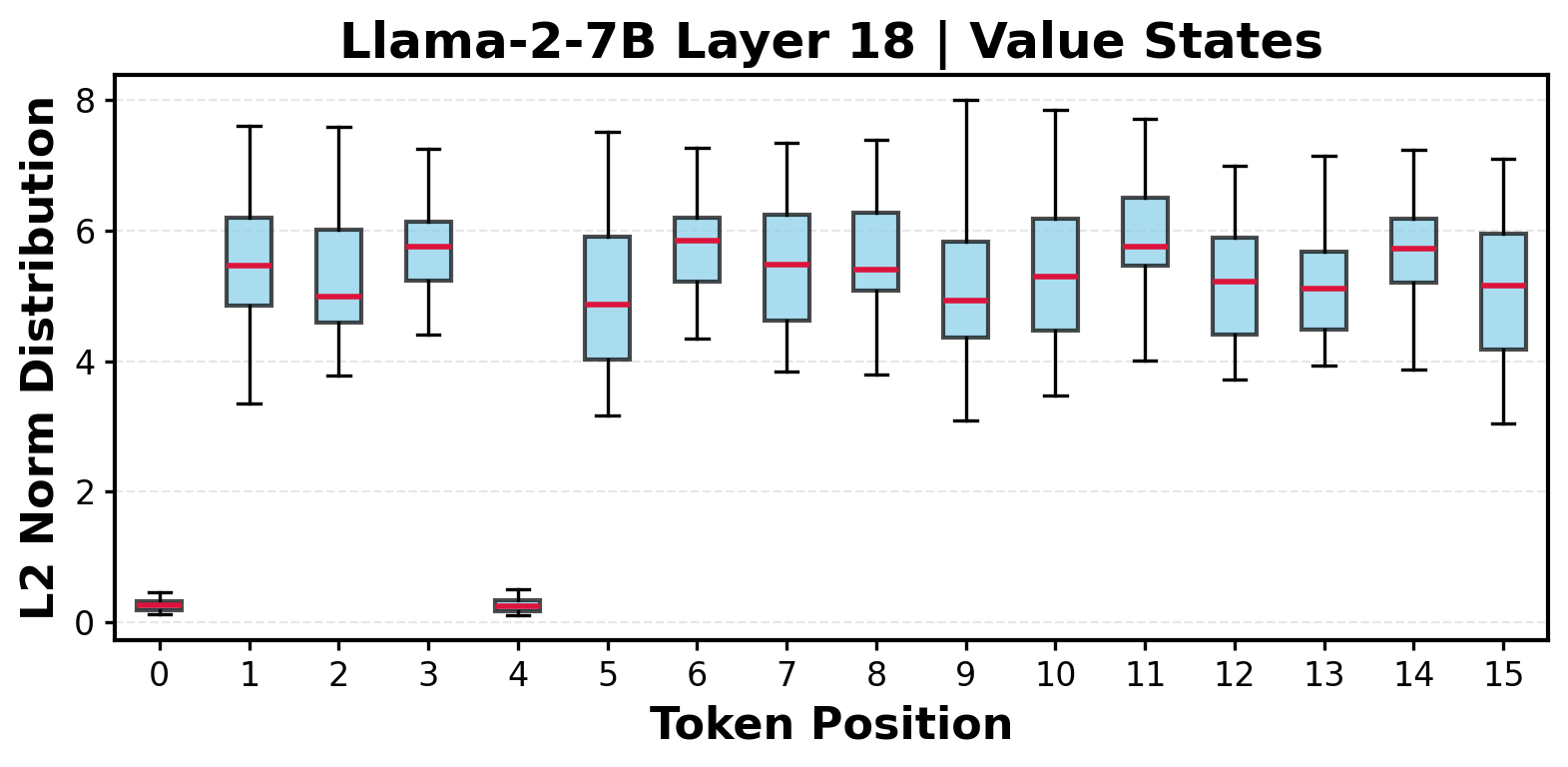}
        \caption{Value L2 norm distribution}
    \end{subfigure}
    \vspace{0.1cm}
    \begin{subfigure}[b]{0.325\textwidth}
        \centering
        \includegraphics[width=\textwidth]{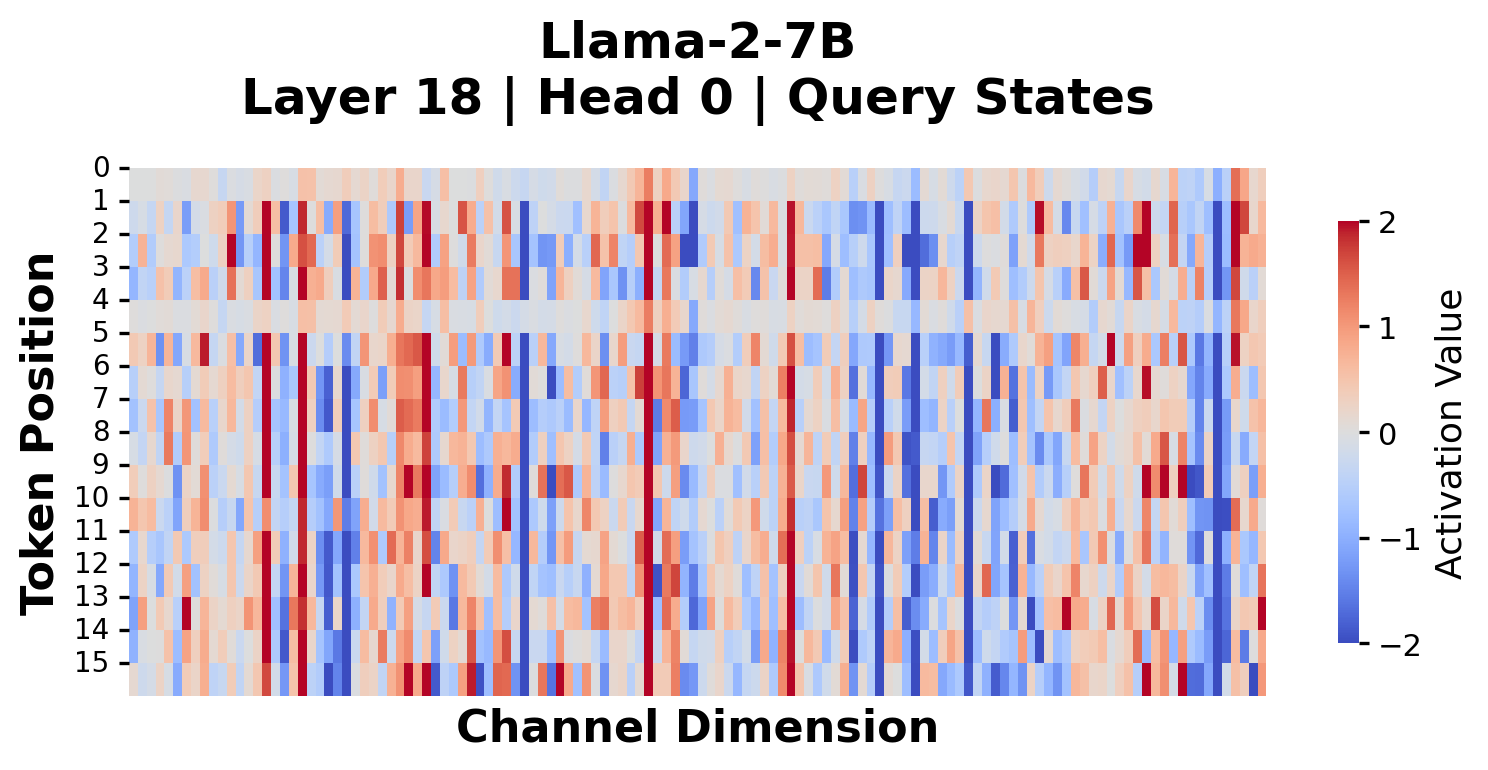}
        \caption{Query heatmap}
    \end{subfigure}
    \begin{subfigure}[b]{0.325\textwidth}
        \centering
        \includegraphics[width=\textwidth]{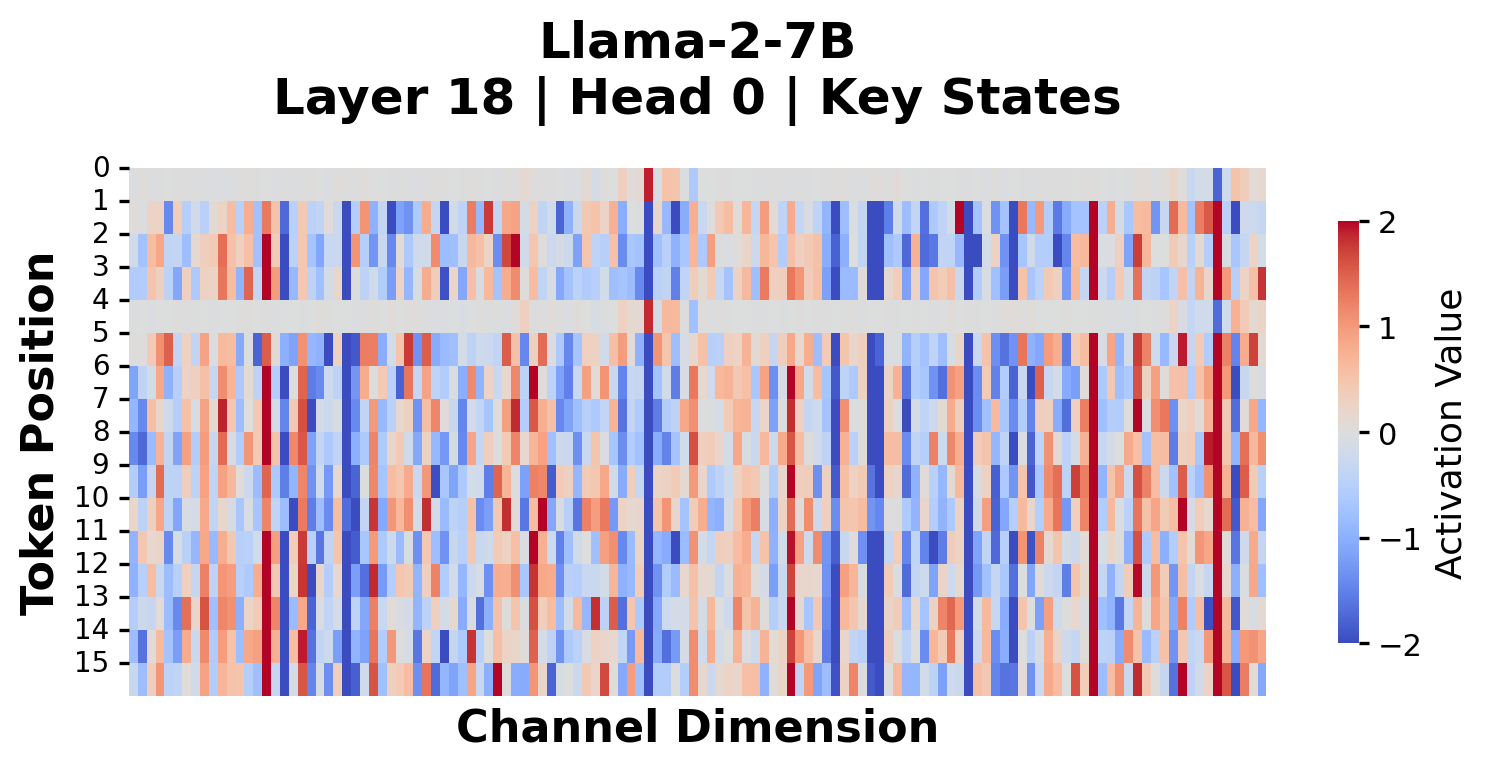}
        \caption{Key heatmap}
    \end{subfigure}
    \begin{subfigure}[b]{0.325\textwidth}
        \centering
        \includegraphics[width=\textwidth]{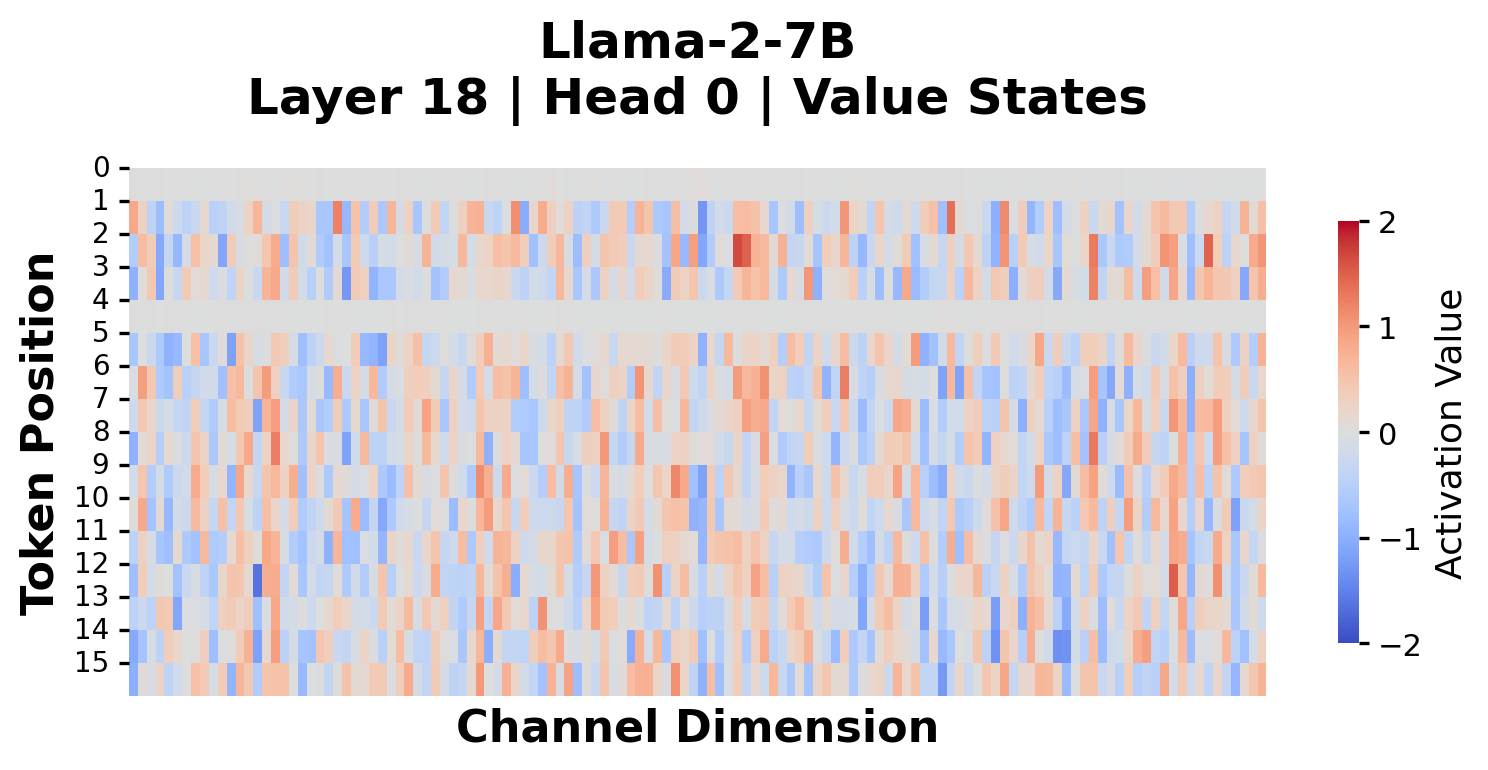}
        \caption{Value heatmap}
    \end{subfigure}
    
    \caption{L2 norm distributions (top row) and value heatmaps (bottom row) of Query, Key, and Value states in Layer 18 of Llama-2-7B.}
    \label{fig:TNI-llama-2-7b-layer-18}
\end{figure}

\begin{figure}[t]
    \centering
    \begin{subfigure}[b]{0.325\textwidth}
        \includegraphics[width=0.9\textwidth]{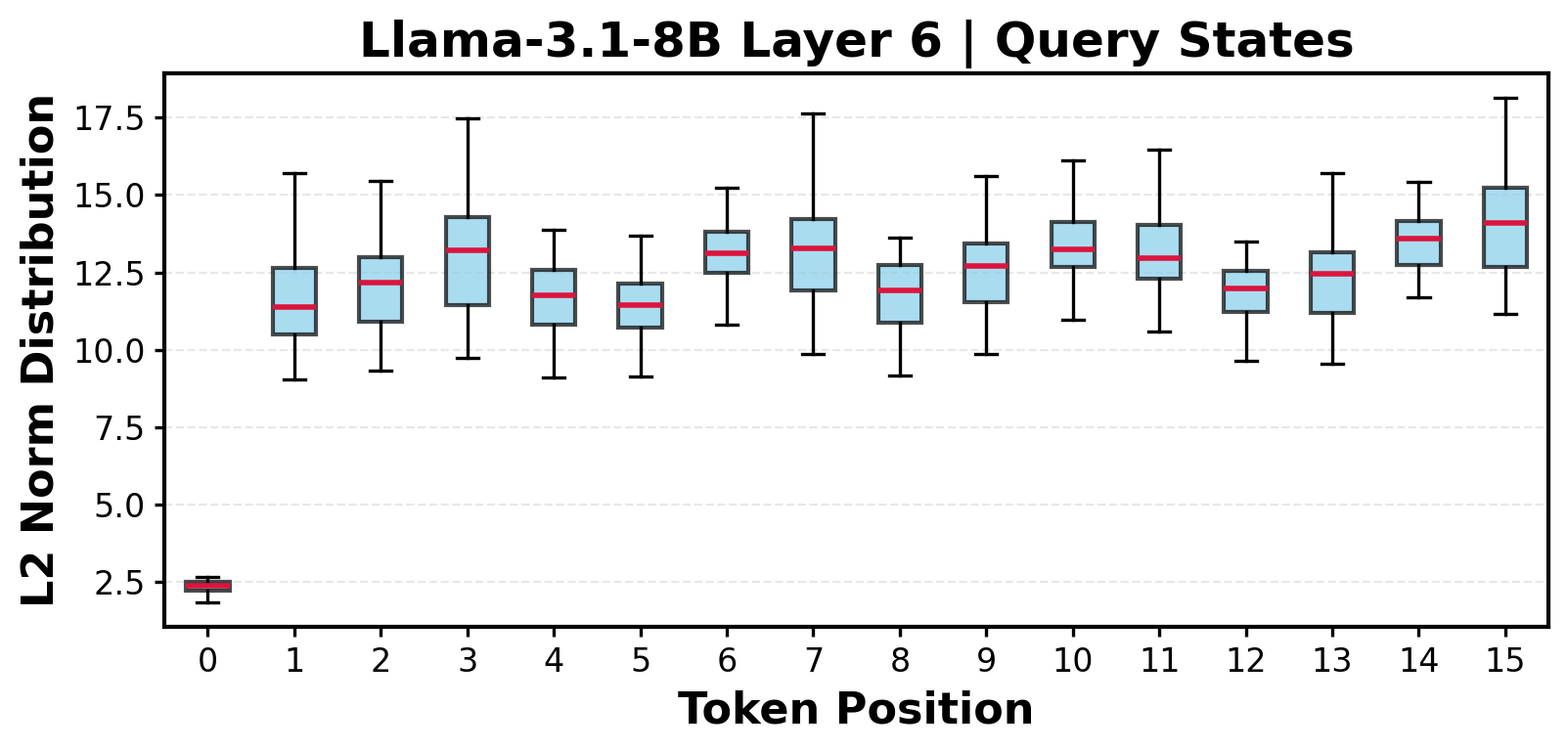}
        \caption{Query L2 norm distribution}
    \end{subfigure}
    \begin{subfigure}[b]{0.325\textwidth}
        \includegraphics[width=0.9\textwidth]{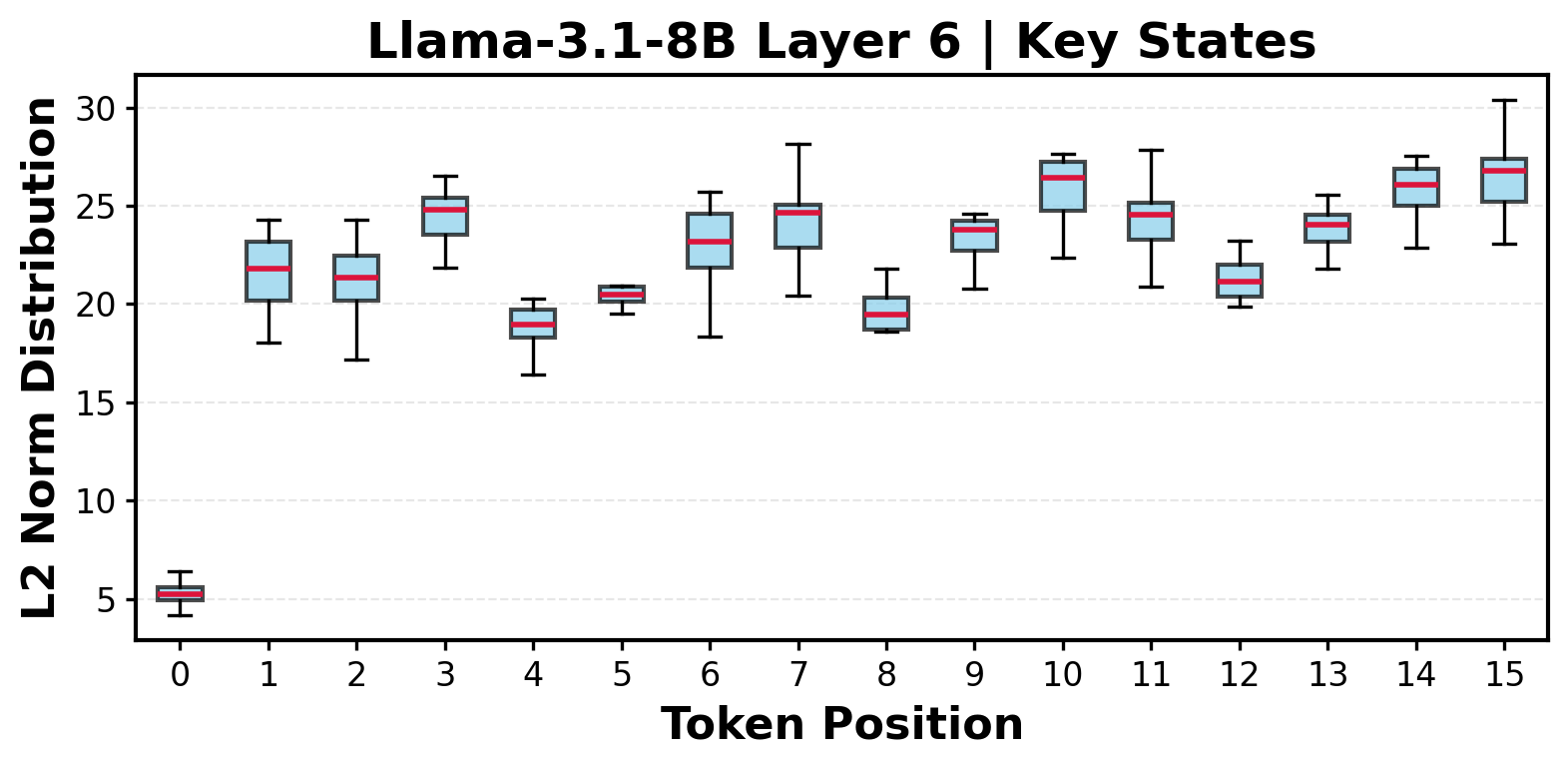}
        \caption{Key L2 norm distribution}
    \end{subfigure}
    \begin{subfigure}[b]{0.325\textwidth}
        \includegraphics[width=0.9\textwidth]{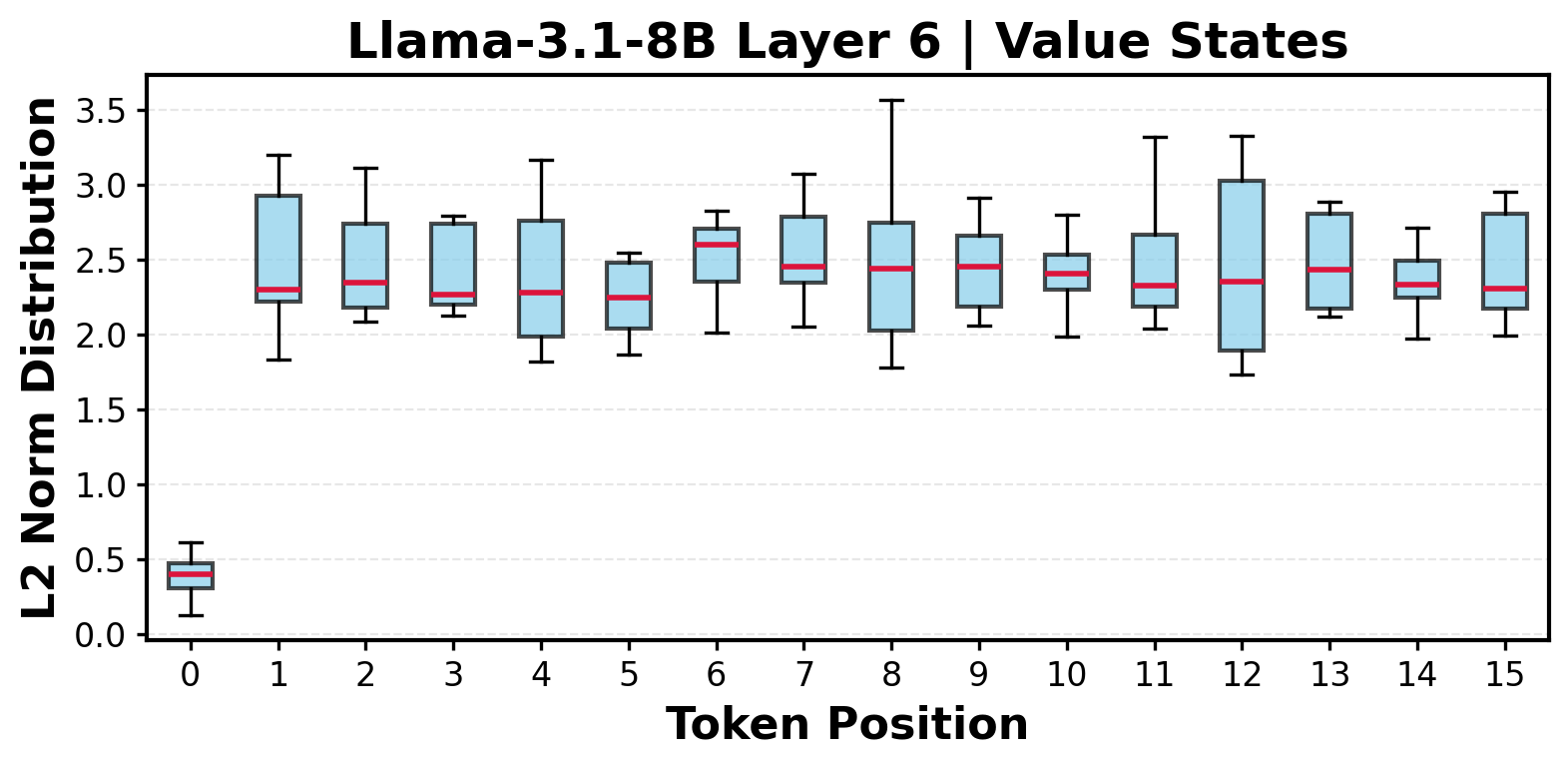}
        \caption{Value L2 norm distribution}
    \end{subfigure}
    \vspace{0.1cm}
    \begin{subfigure}[b]{0.325\textwidth}
        \centering
        \includegraphics[width=\textwidth]{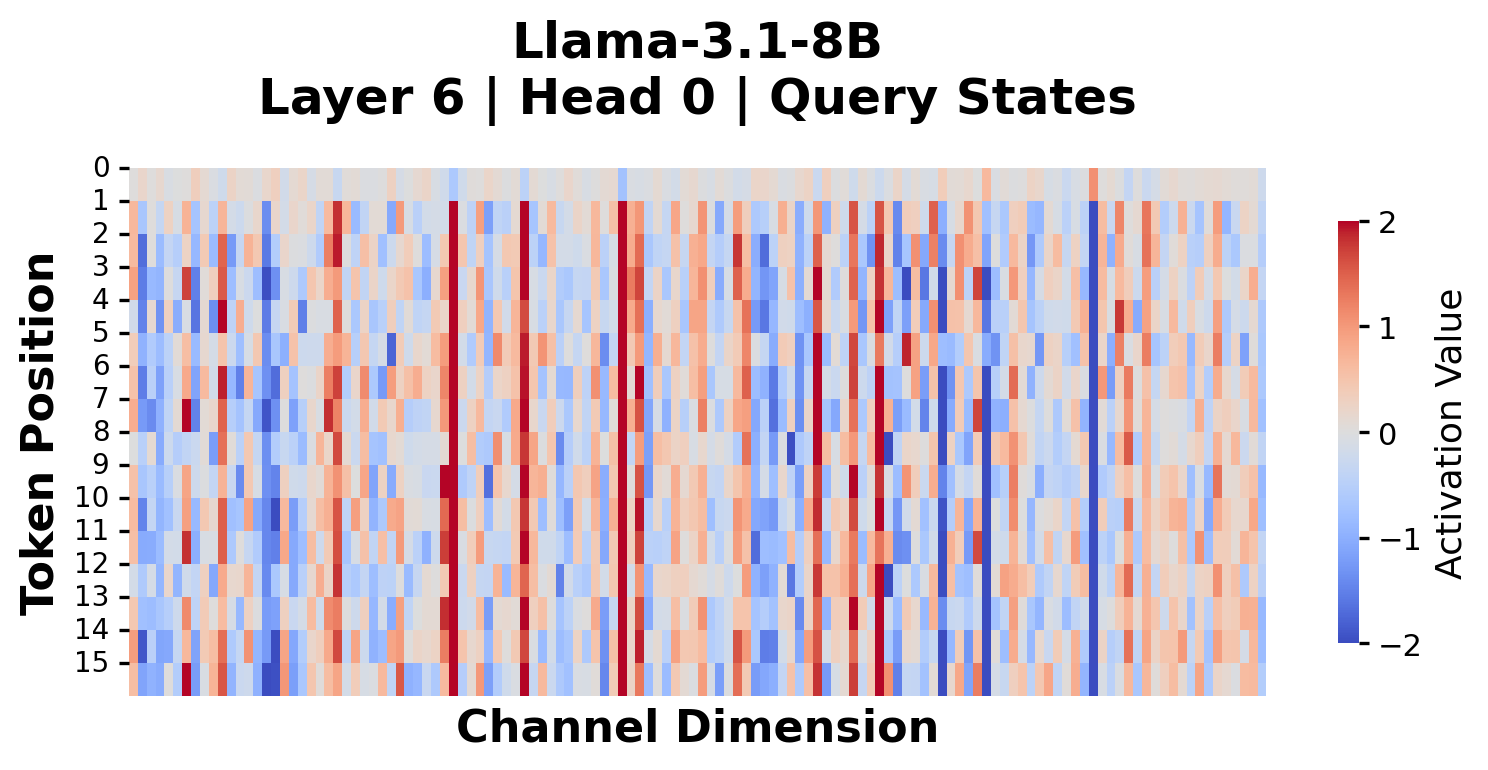}
        \caption{Query heatmap}
    \end{subfigure}
    \begin{subfigure}[b]{0.325\textwidth}
        \centering
        \includegraphics[width=\textwidth]{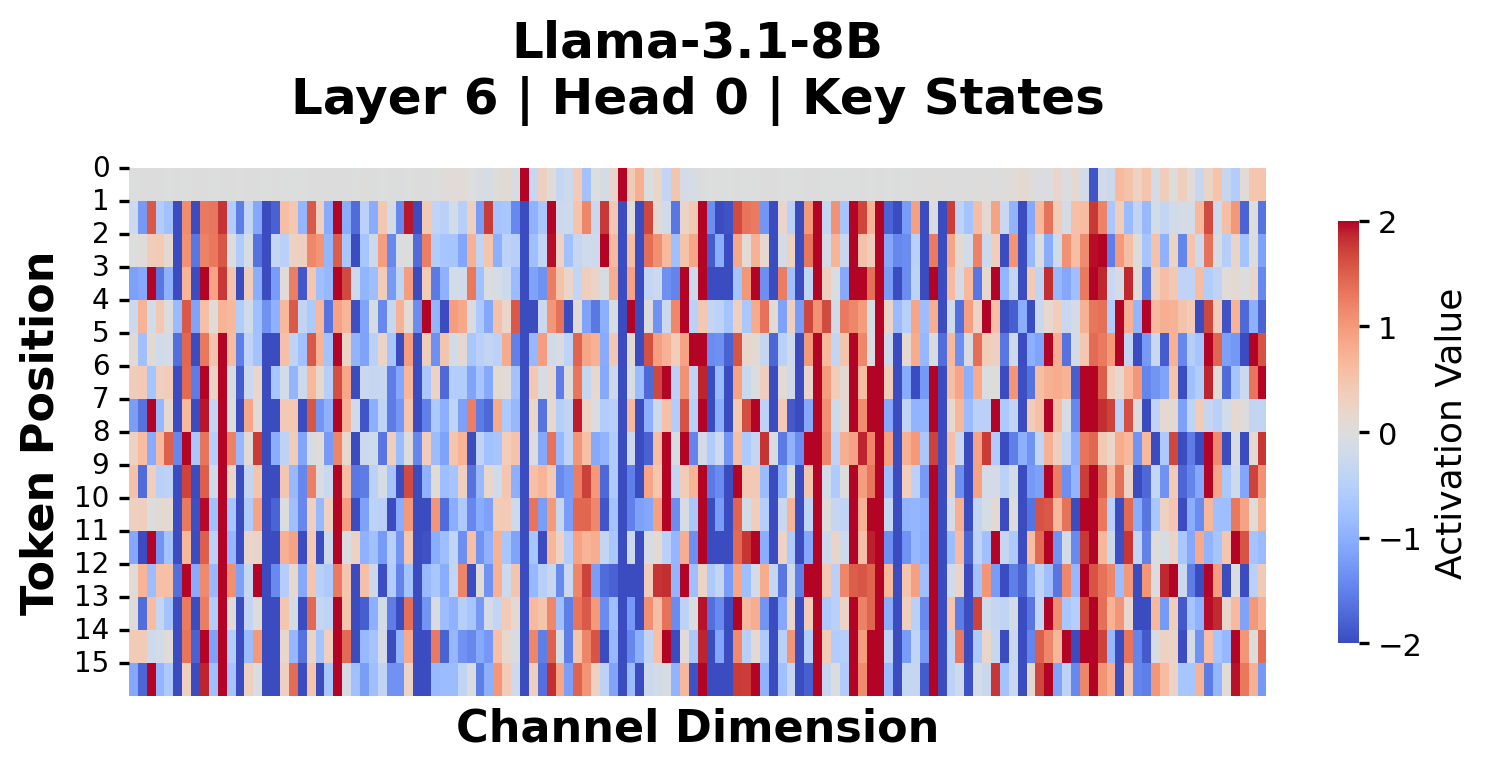}
        \caption{Key heatmap}
    \end{subfigure}
    \begin{subfigure}[b]{0.325\textwidth}
        \centering
        \includegraphics[width=\textwidth]{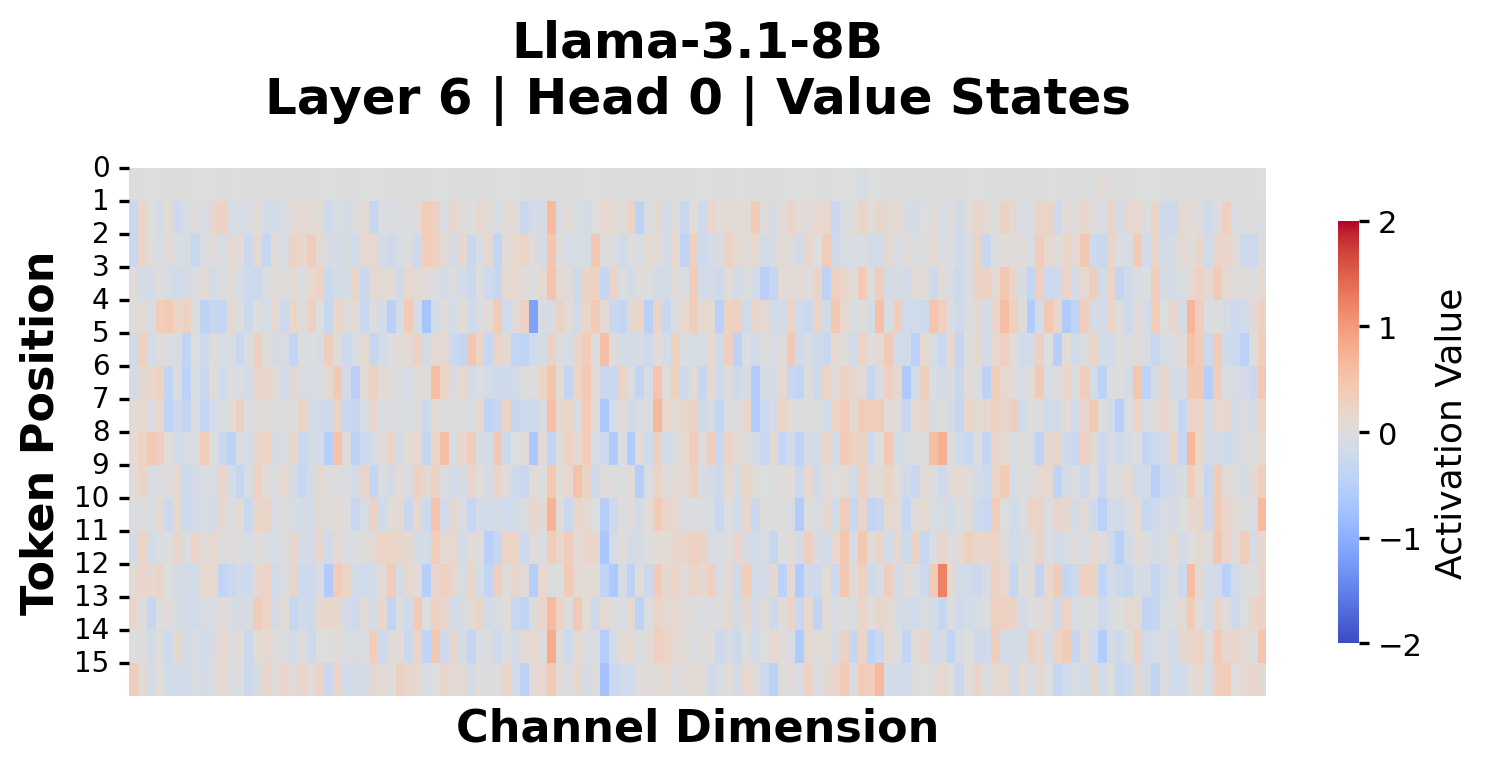}
        \caption{Value heatmap}
    \end{subfigure}
    
    \caption{L2 norm distributions (top row) and value heatmaps (bottom row) of Query, Key, and Value states in Layer 6 of Llama-3.1-8B.}
    \label{fig:TNI-llama-3-8b-layer-6}
\end{figure}

\begin{figure}[t]
    \centering
    \begin{subfigure}[b]{0.325\textwidth}
        \includegraphics[width=0.9\textwidth]{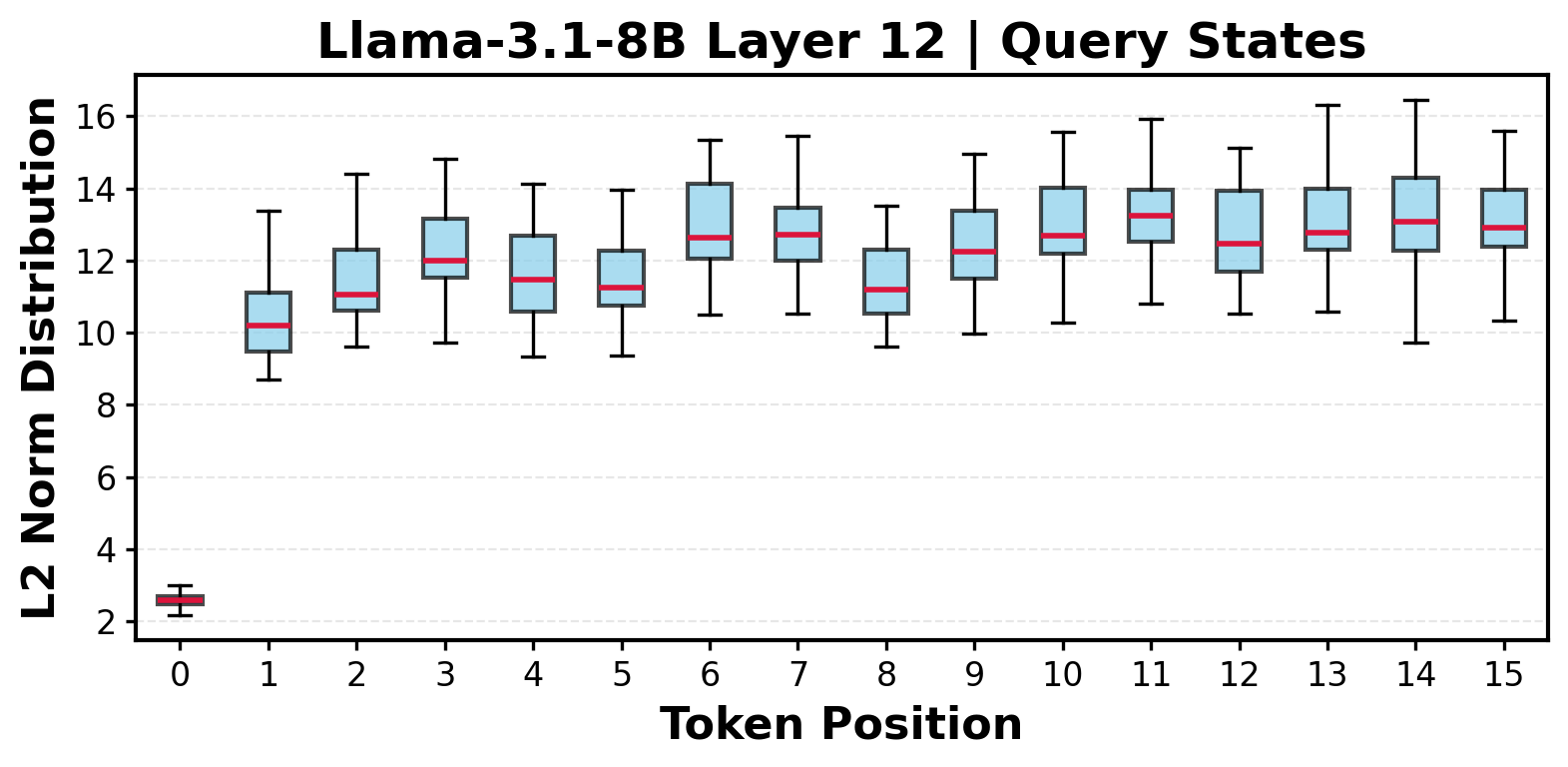}
        \caption{Query L2 norm distribution}
    \end{subfigure}
    \begin{subfigure}[b]{0.325\textwidth}
        \includegraphics[width=0.9\textwidth]{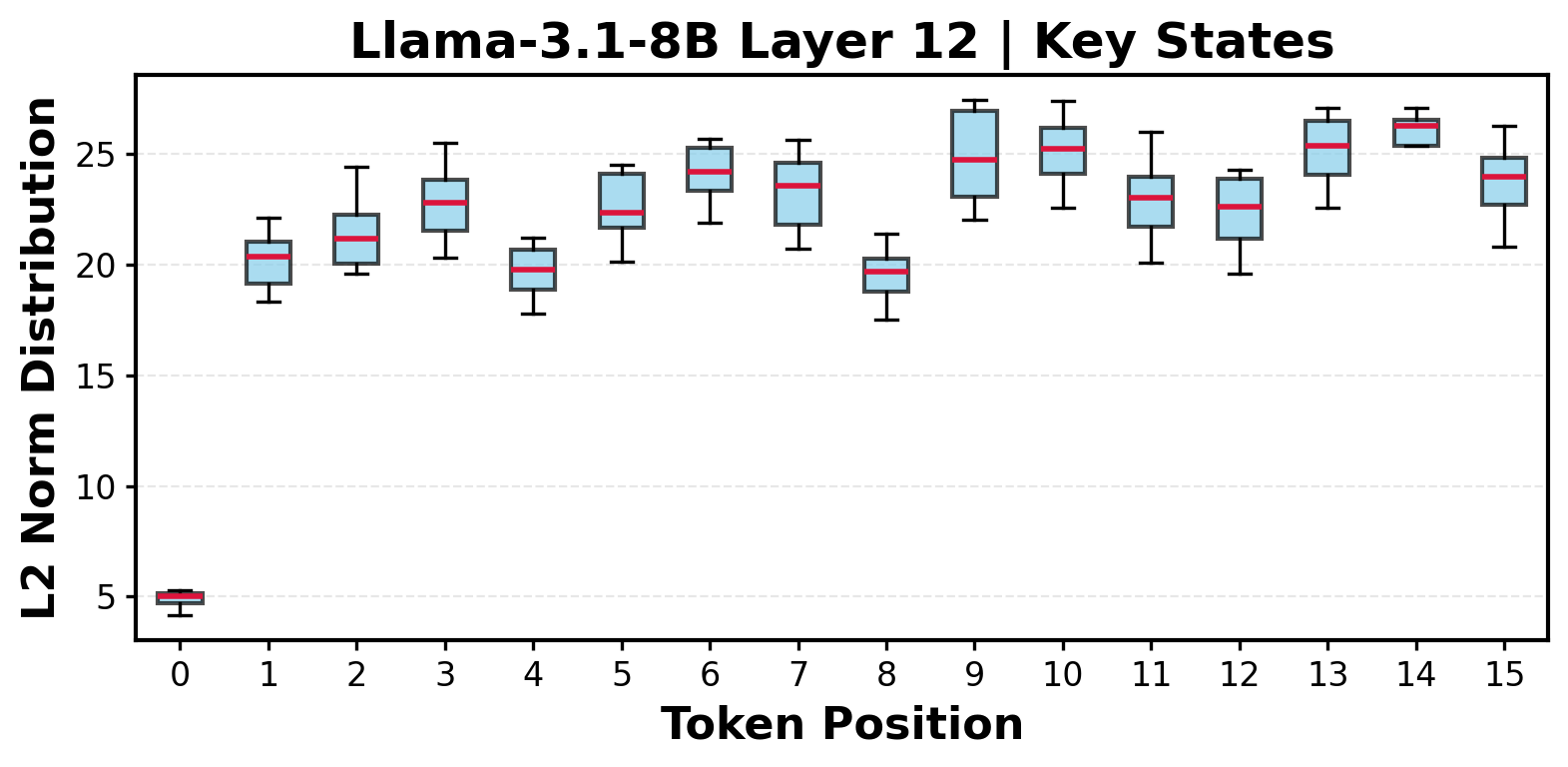}
        \caption{Key L2 norm distribution}
    \end{subfigure}
    \begin{subfigure}[b]{0.325\textwidth}
        \includegraphics[width=0.9\textwidth]{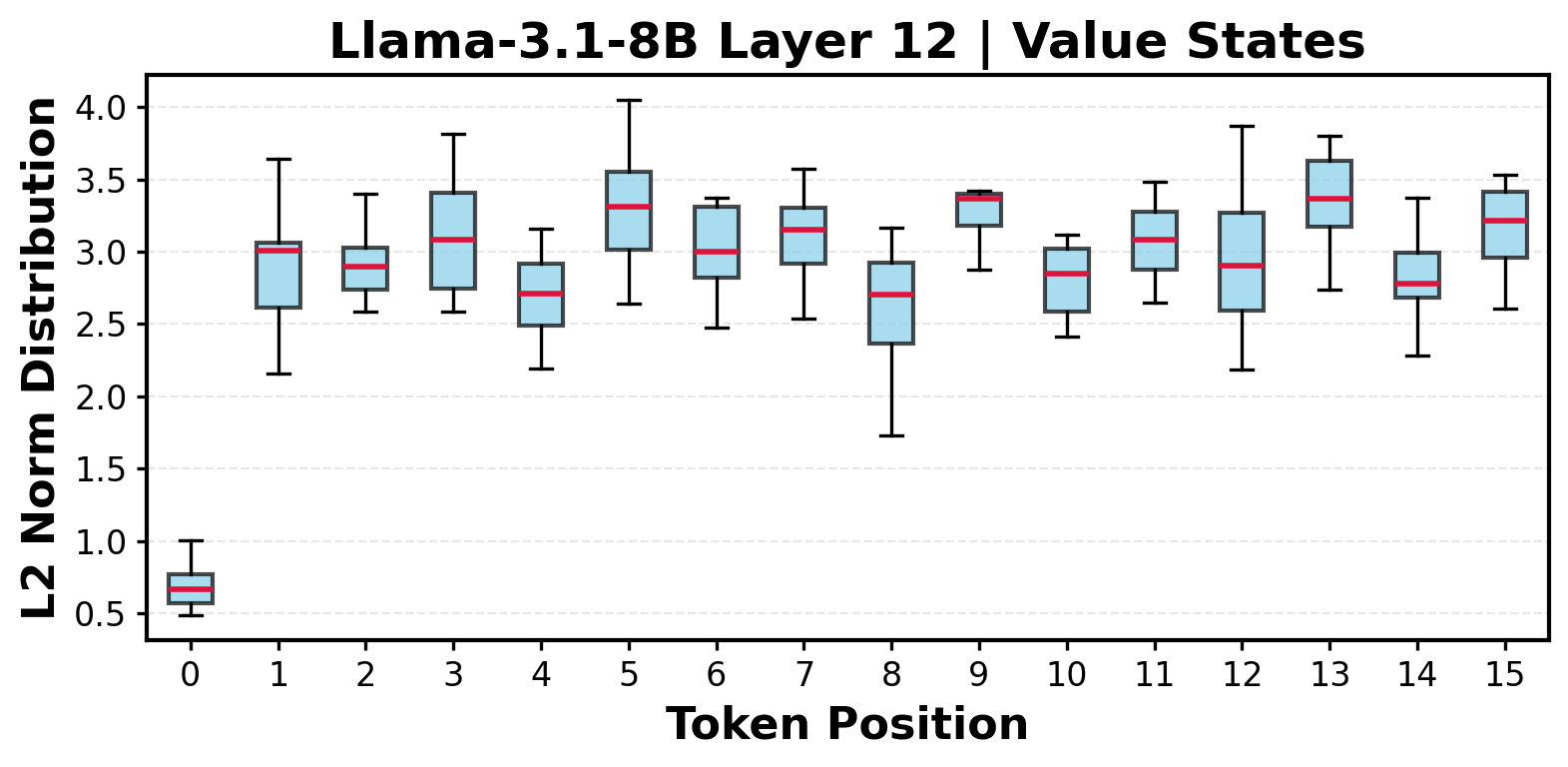}
        \caption{Value L2 norm distribution}
    \end{subfigure}
    \vspace{0.1cm}
    \begin{subfigure}[b]{0.325\textwidth}
        \centering
        \includegraphics[width=\textwidth]{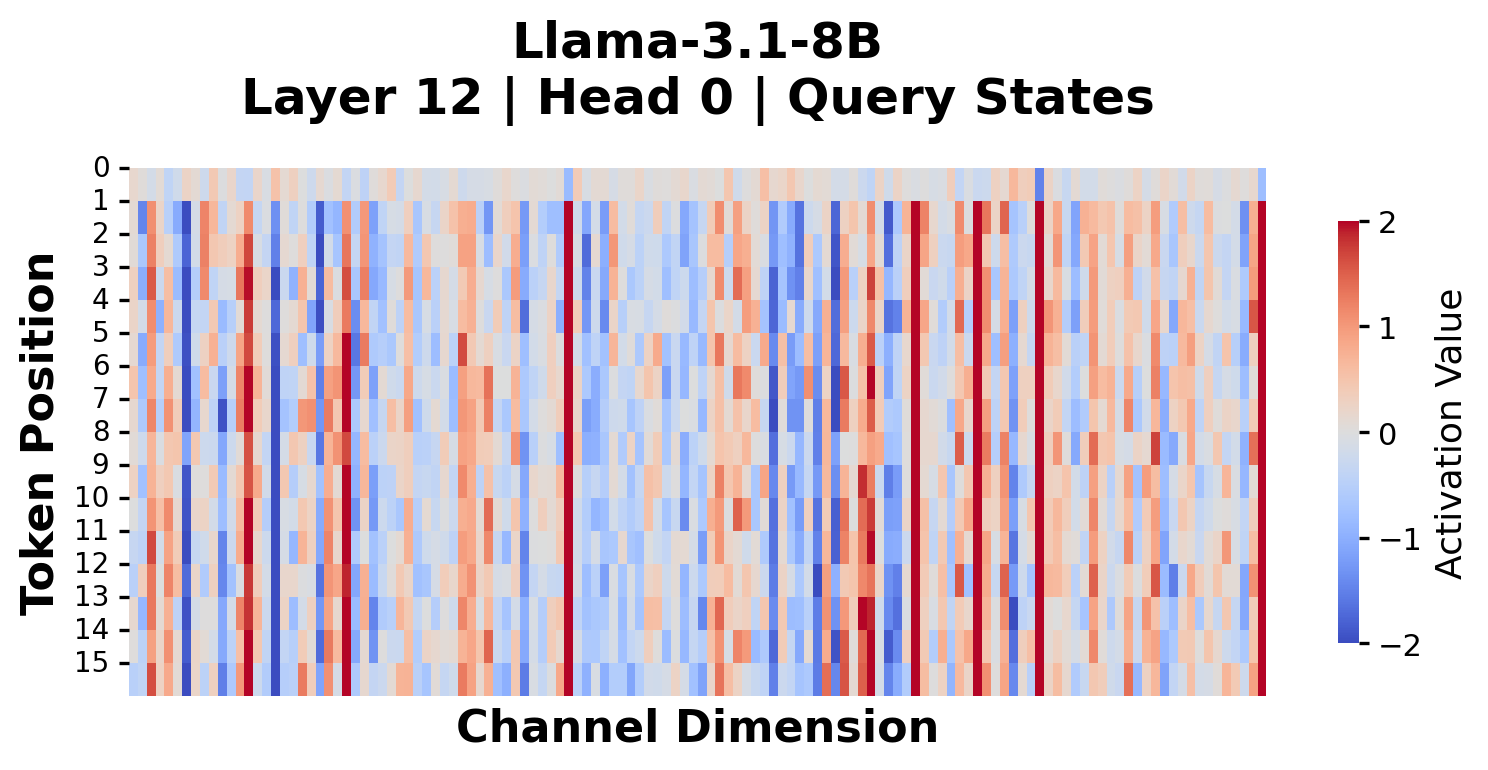}
        \caption{Query heatmap}
    \end{subfigure}
    \begin{subfigure}[b]{0.325\textwidth}
        \centering
        \includegraphics[width=\textwidth]{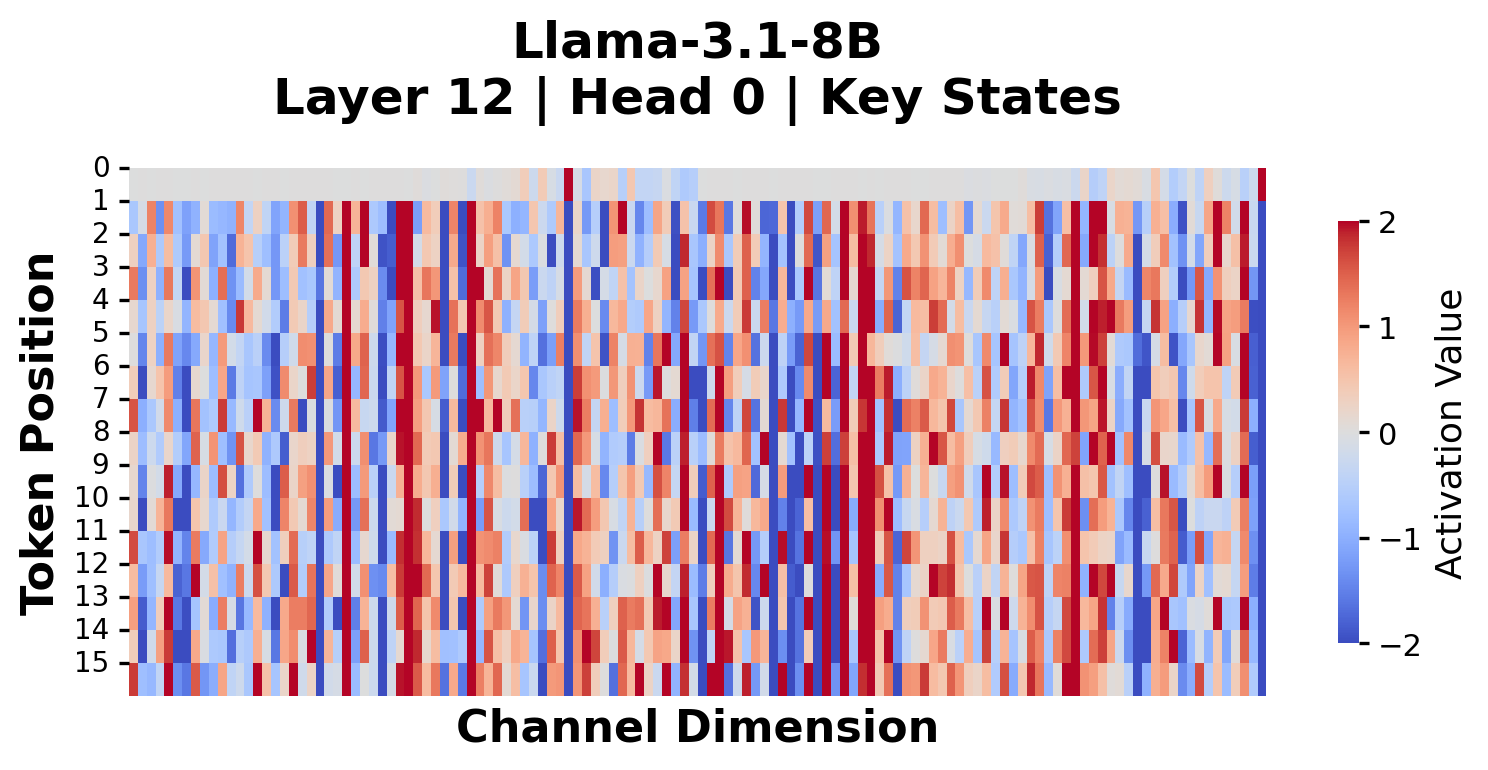}
        \caption{Key heatmap}
    \end{subfigure}
    \begin{subfigure}[b]{0.325\textwidth}
        \centering
        \includegraphics[width=\textwidth]{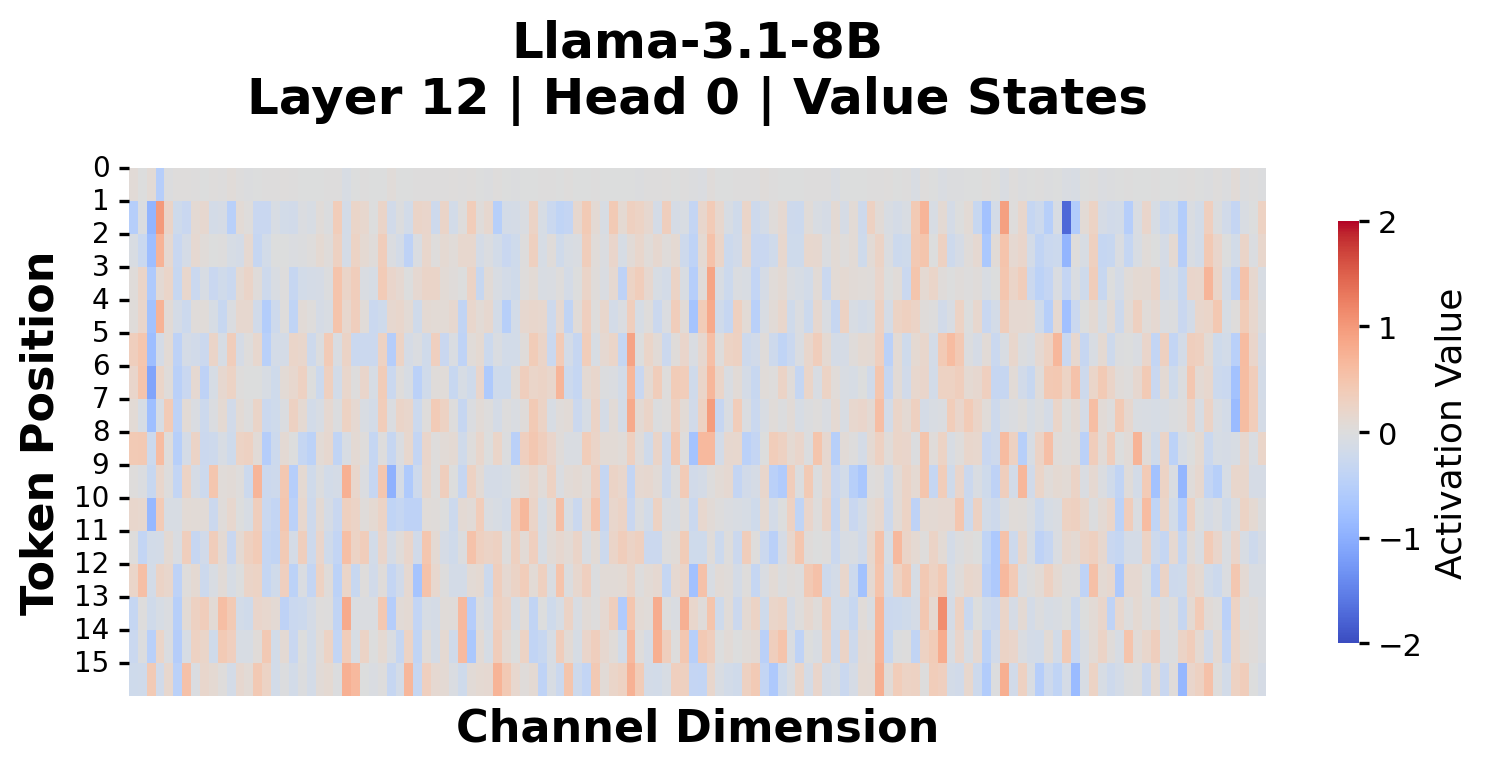}
        \caption{Value heatmap}
    \end{subfigure}
    
    \caption{L2 norm distributions (top row) and value heatmaps (bottom row) of Query, Key, and Value states in Layer 12 of Llama-3.1-8B.}
    \label{fig:TNI-llama-3-8b-layer-12}
\end{figure}

\begin{figure}[t]
    \centering
    \begin{subfigure}[b]{0.325\textwidth}
        \includegraphics[width=0.9\textwidth]{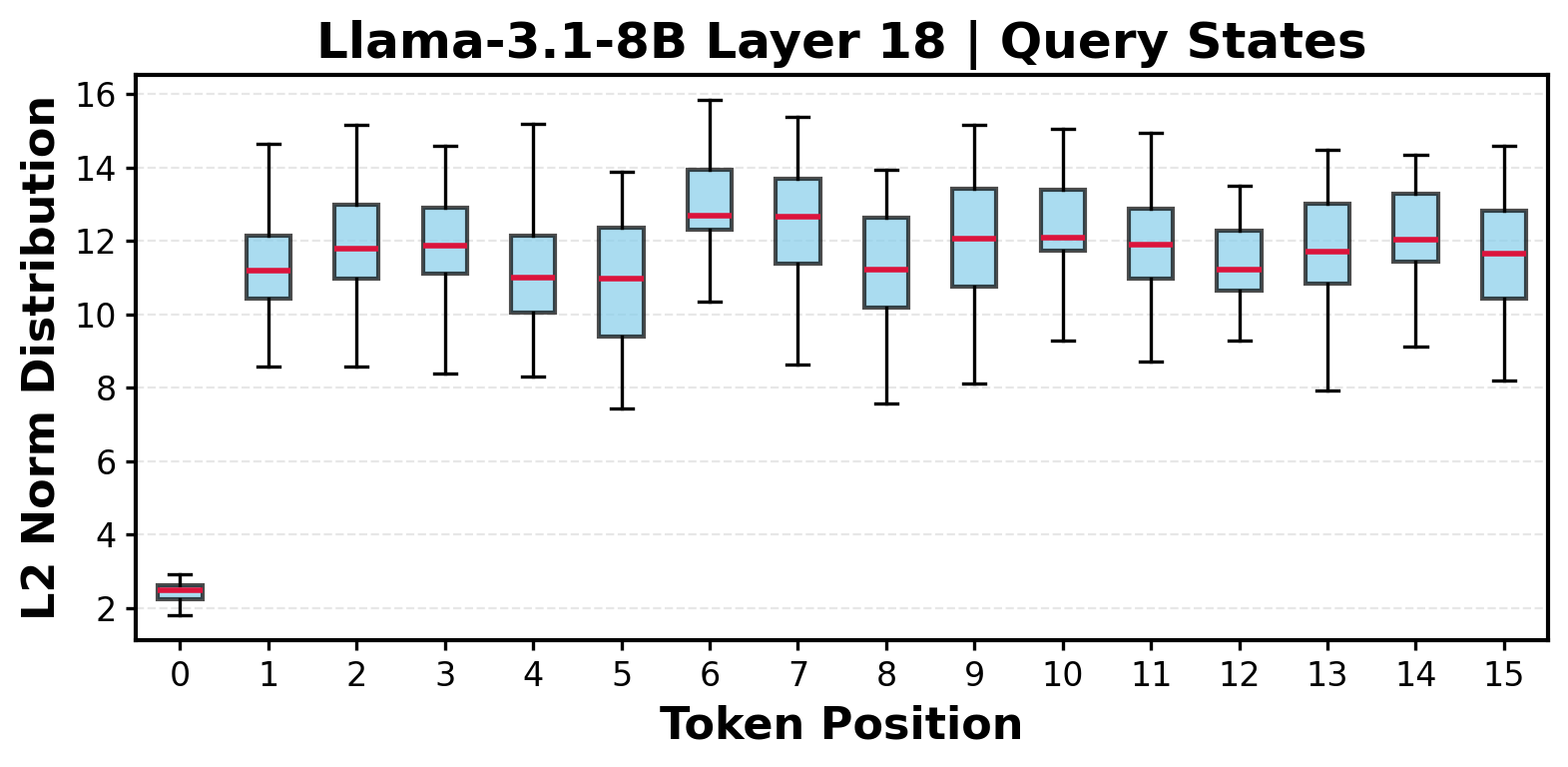}
        \caption{Query L2 norm distribution}
    \end{subfigure}
    \begin{subfigure}[b]{0.325\textwidth}
        \includegraphics[width=0.9\textwidth]{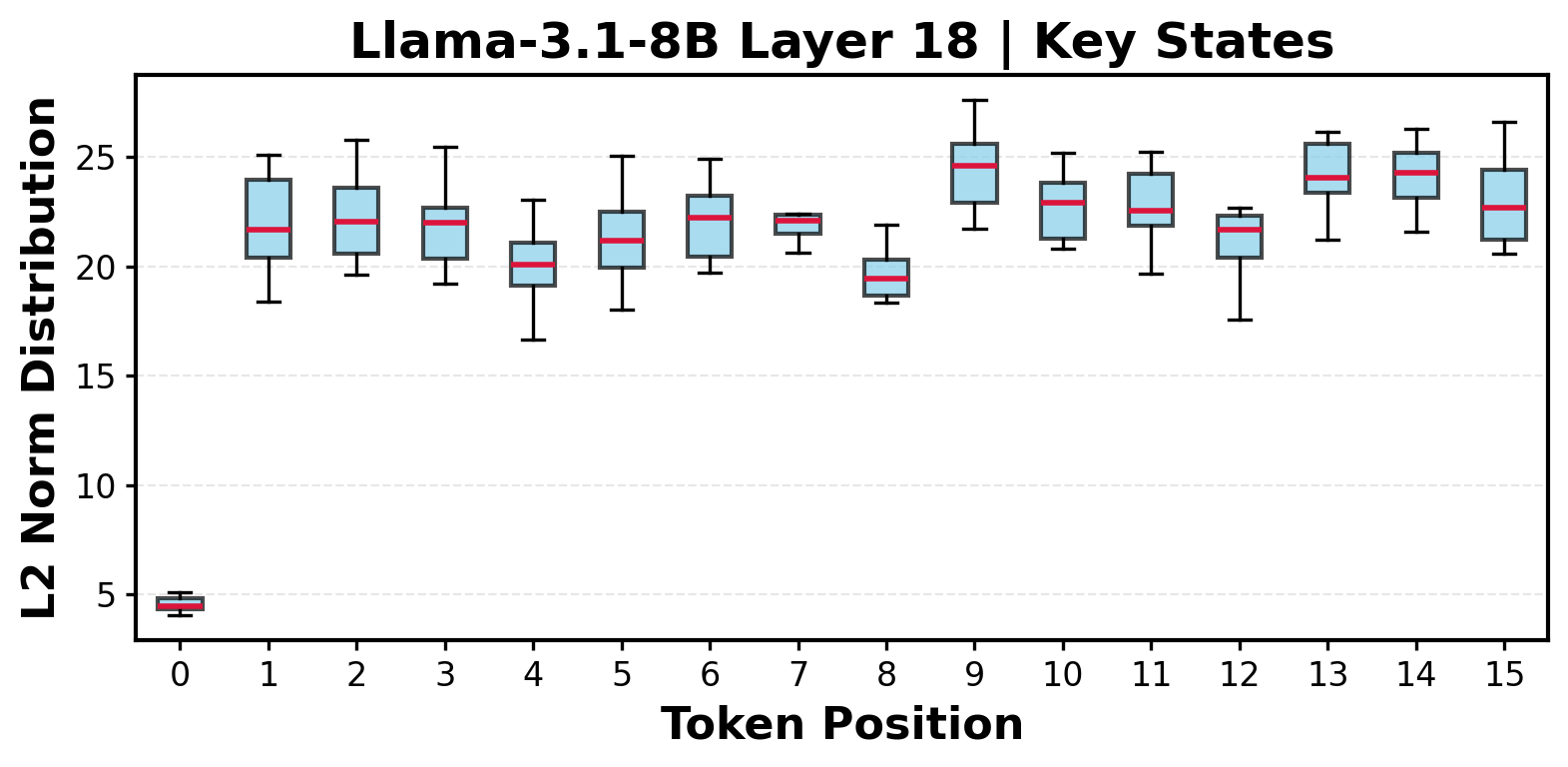}
        \caption{Key L2 norm distribution}
    \end{subfigure}
    \begin{subfigure}[b]{0.325\textwidth}
        \includegraphics[width=0.9\textwidth]{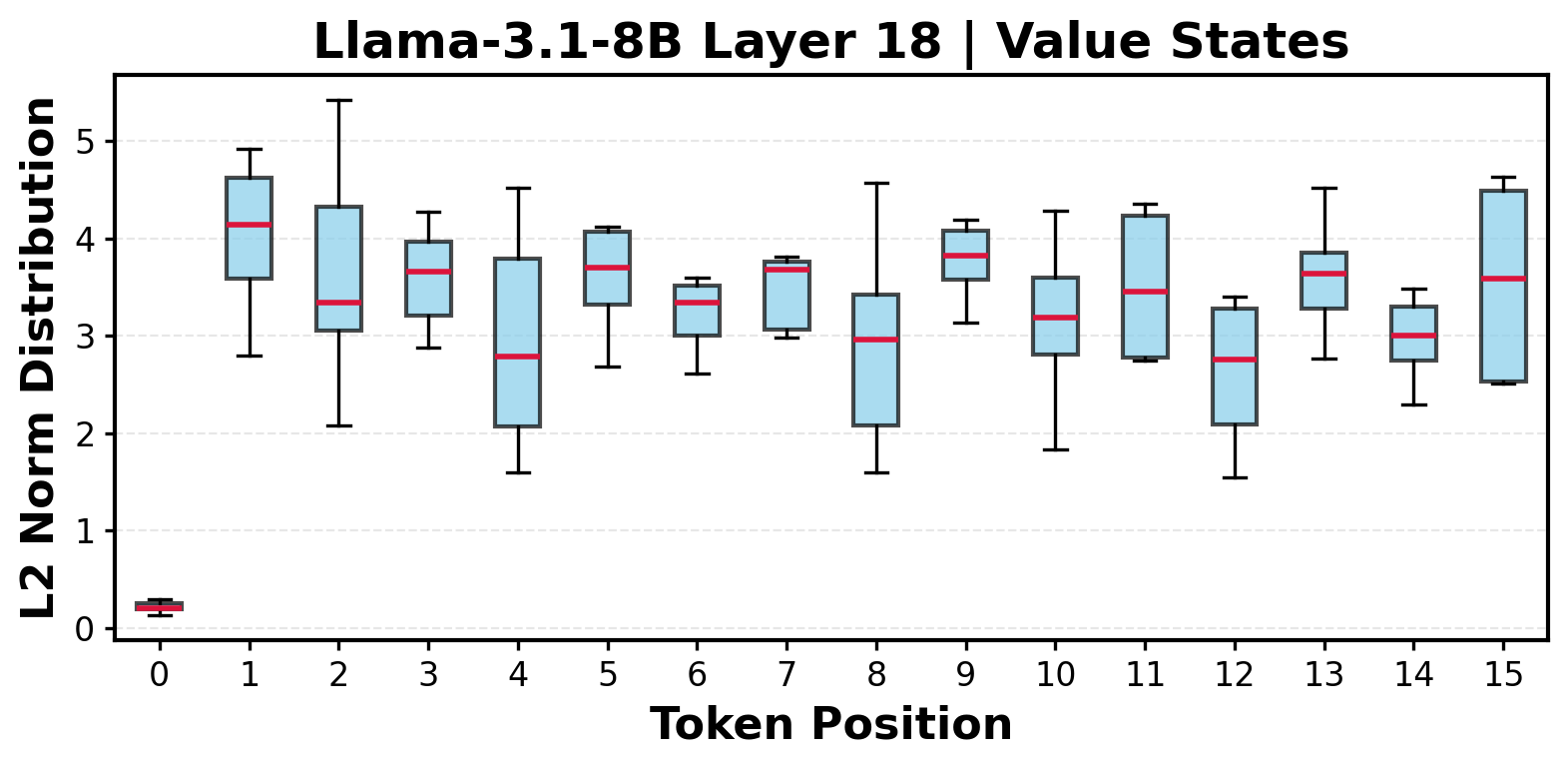}
        \caption{Value L2 norm distribution}
    \end{subfigure}
    \vspace{0.1cm}
    \begin{subfigure}[b]{0.325\textwidth}
        \centering
        \includegraphics[width=\textwidth]{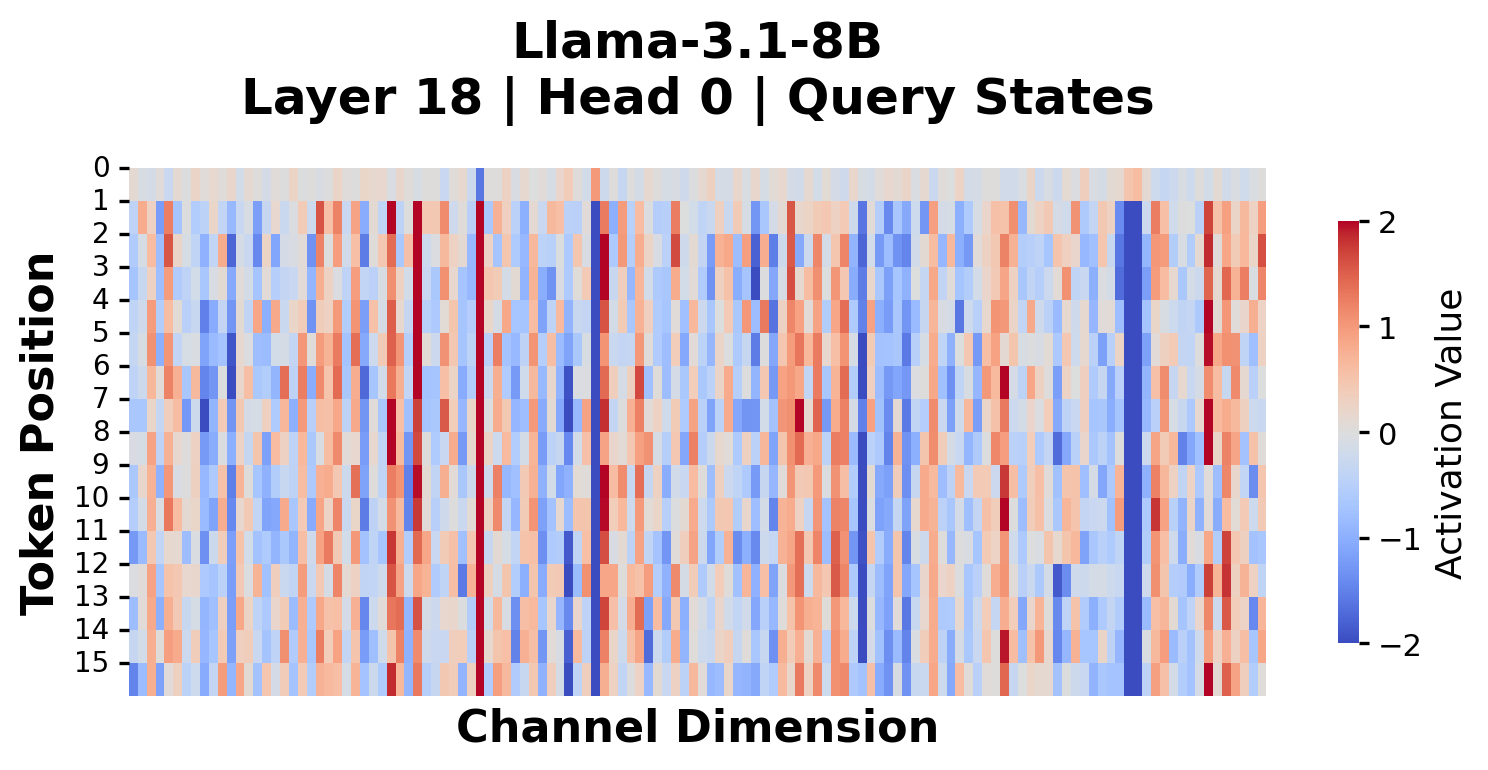}
        \caption{Query heatmap}
    \end{subfigure}
    \begin{subfigure}[b]{0.325\textwidth}
        \centering
        \includegraphics[width=\textwidth]{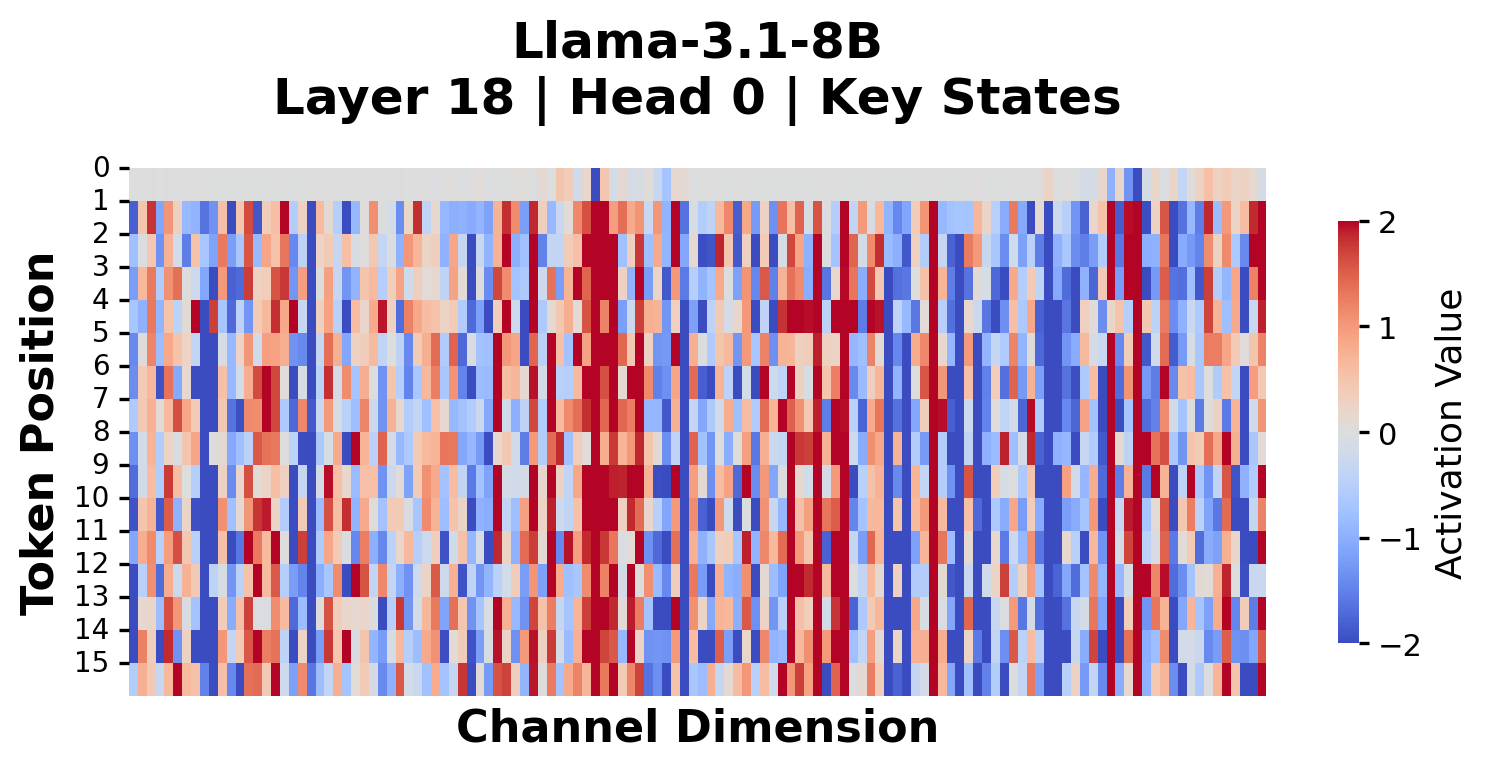}
        \caption{Key heatmap}
    \end{subfigure}
    \begin{subfigure}[b]{0.325\textwidth}
        \centering
        \includegraphics[width=\textwidth]{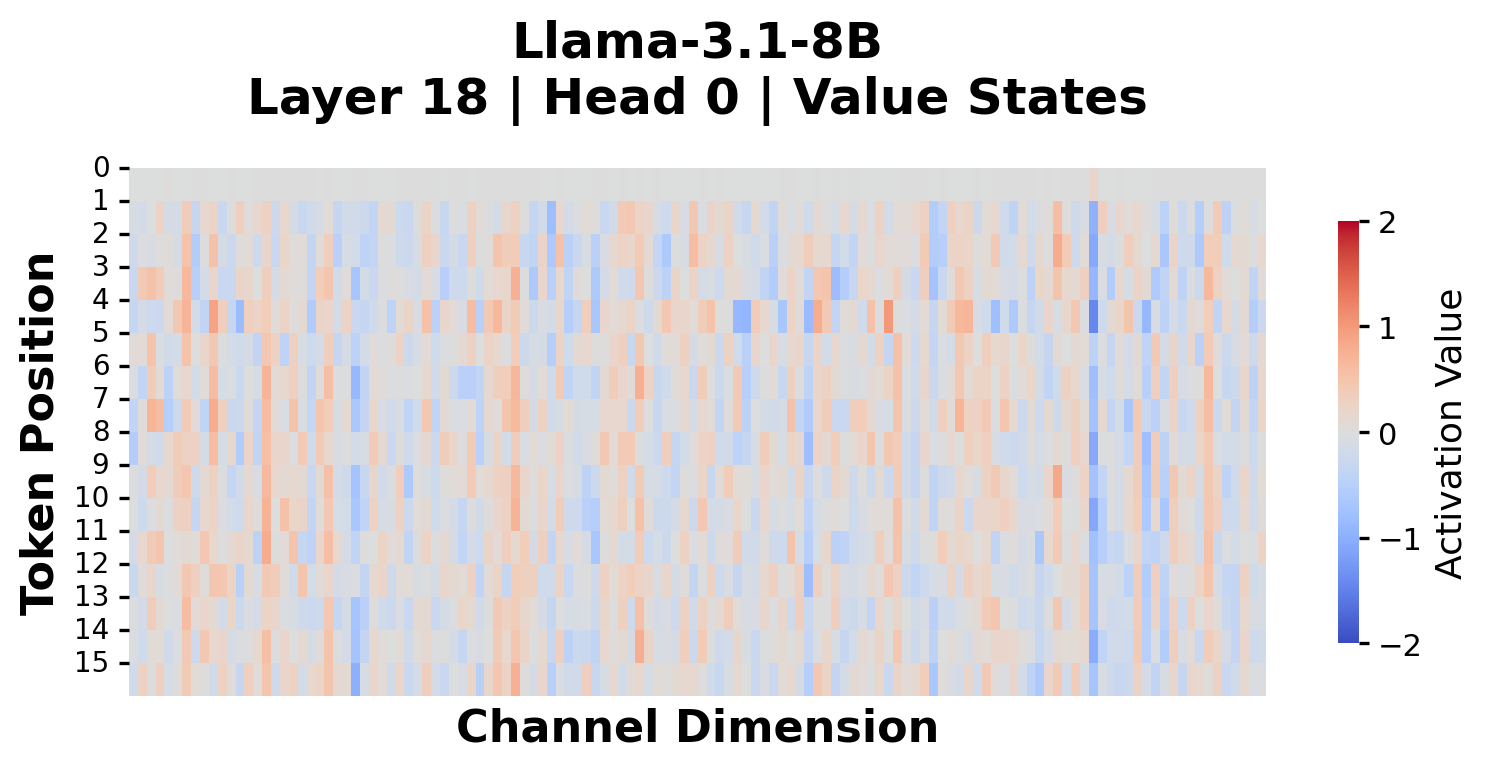}
        \caption{Value heatmap}
    \end{subfigure}
    
    \caption{L2 norm distributions (top row) and value heatmaps (bottom row) of Query, Key, and Value states in Layer 18 of Llama-3.1-8B.}
    \label{fig:TNI-llama-3-8b-layer-18}
\end{figure}

\begin{figure}[t]
    \centering
    \begin{subfigure}[b]{0.325\textwidth}
        \includegraphics[width=0.9\textwidth]{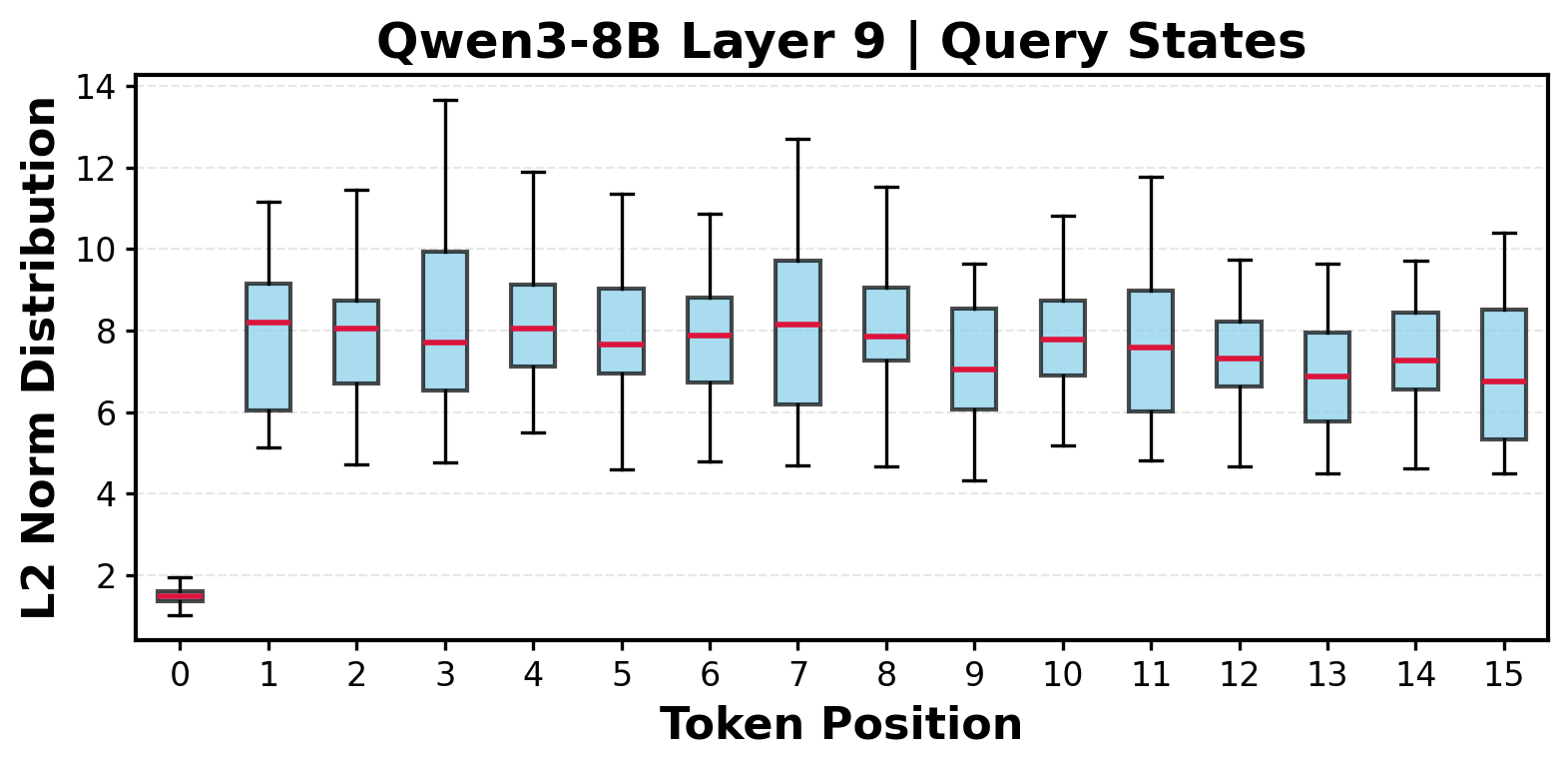}
        \caption{Query L2 norm distribution}
    \end{subfigure}
    \begin{subfigure}[b]{0.325\textwidth}
        \includegraphics[width=0.9\textwidth]{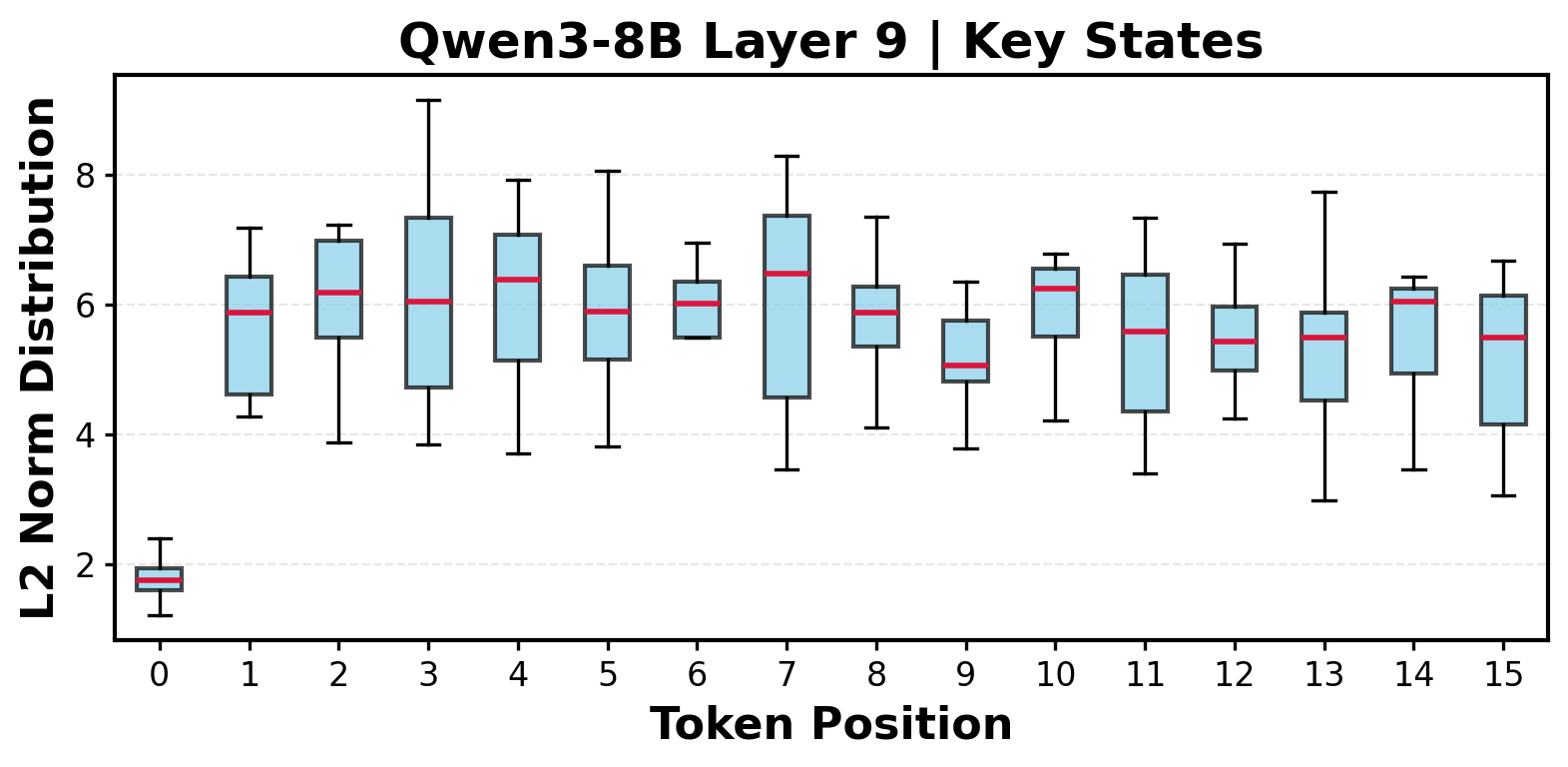}
        \caption{Key L2 norm distribution}
    \end{subfigure}
    \begin{subfigure}[b]{0.325\textwidth}
        \includegraphics[width=0.9\textwidth]{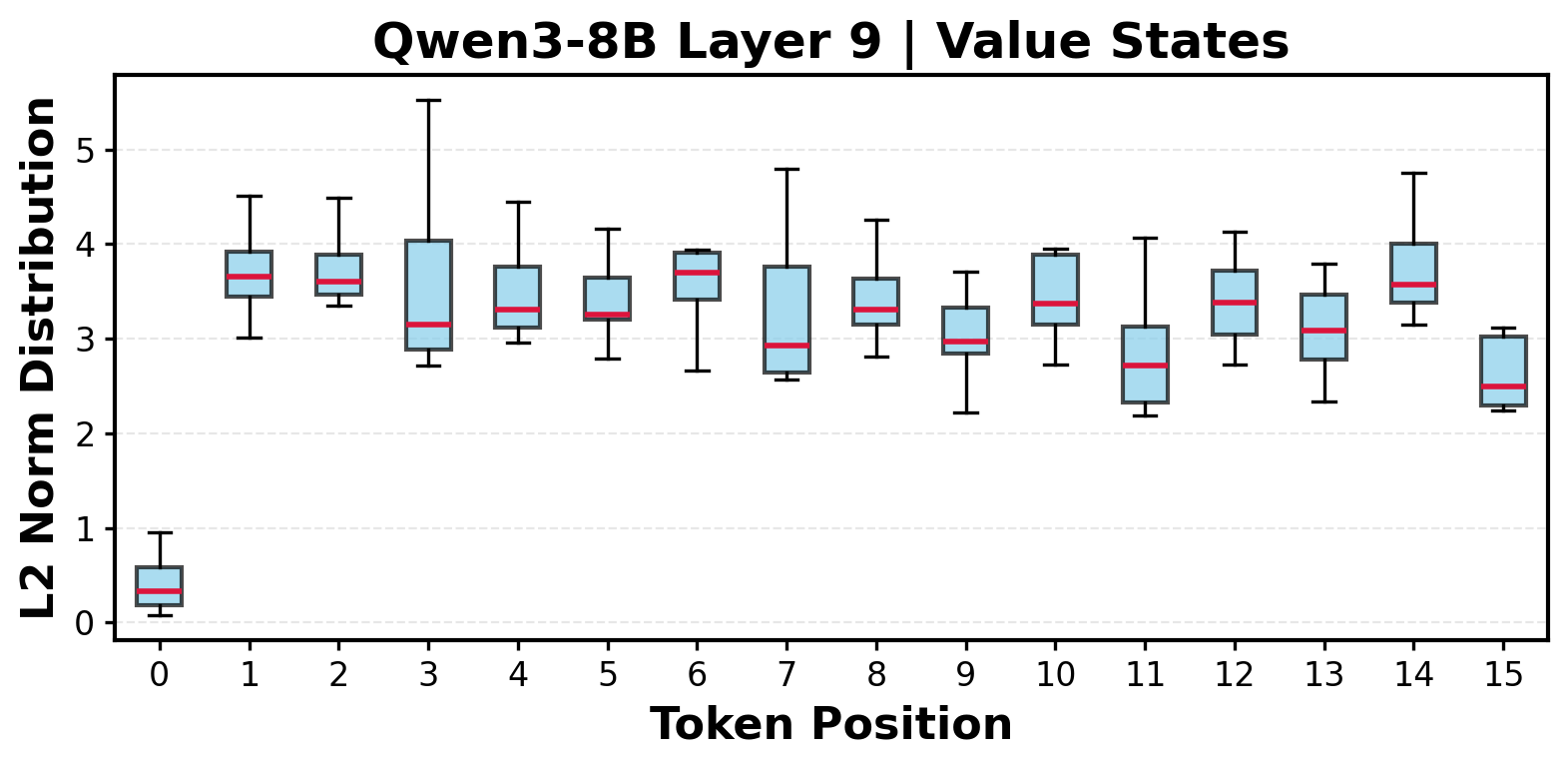}
        \caption{Value L2 norm distribution}
    \end{subfigure}
    \vspace{0.1cm}
    \begin{subfigure}[b]{0.325\textwidth}
        \centering
        \includegraphics[width=\textwidth]{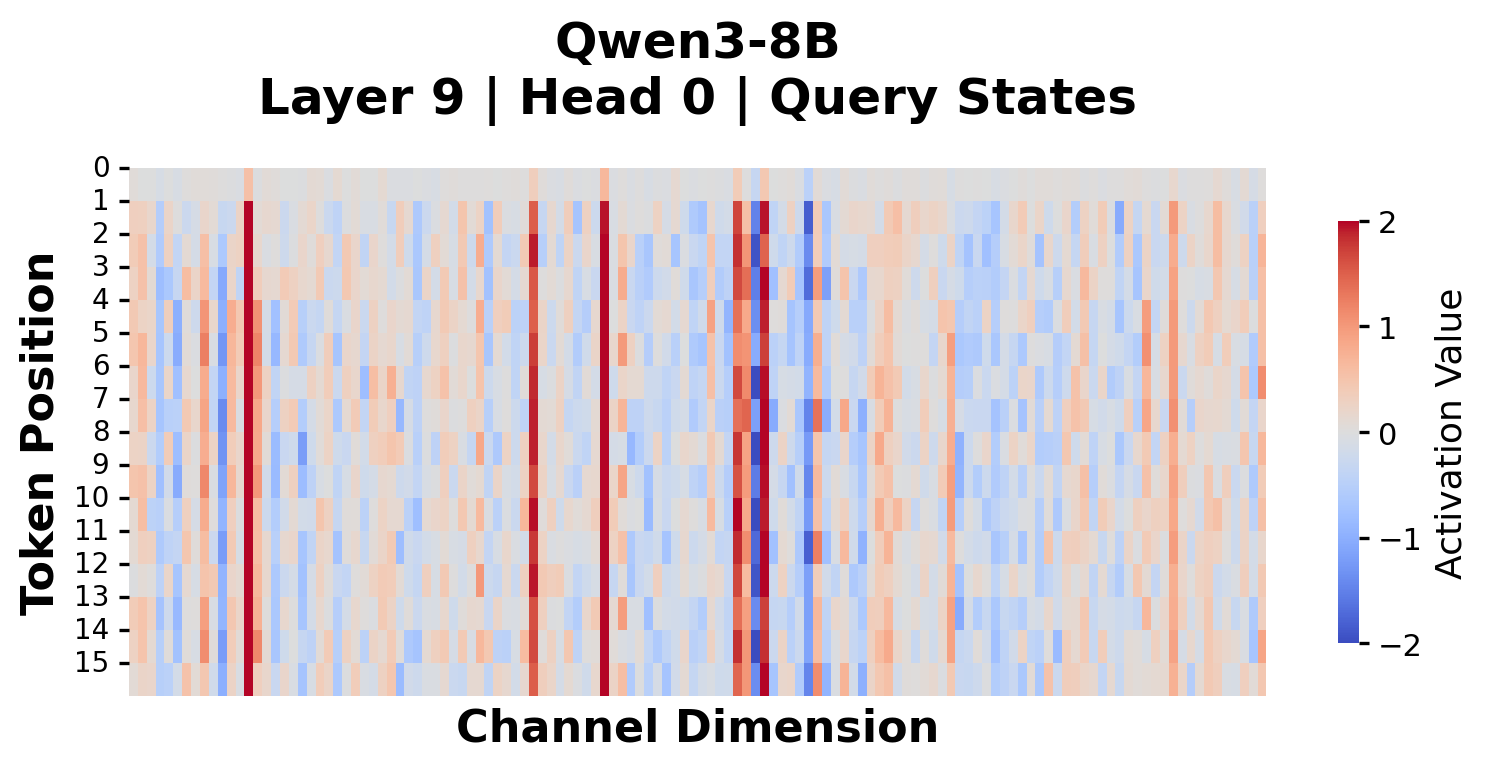}
        \caption{Query heatmap}
    \end{subfigure}
    \begin{subfigure}[b]{0.325\textwidth}
        \centering
        \includegraphics[width=\textwidth]{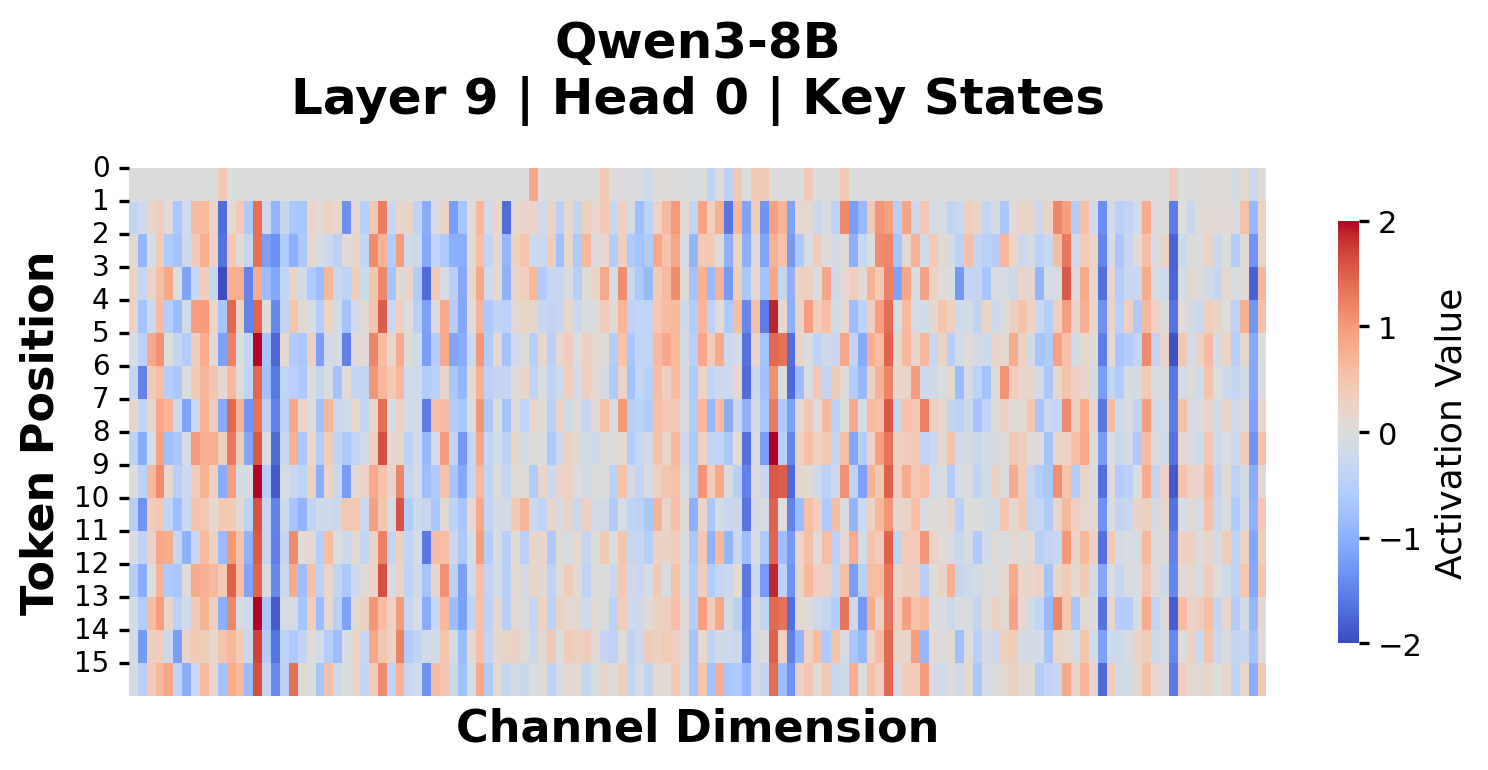}
        \caption{Key heatmap}
    \end{subfigure}
    \begin{subfigure}[b]{0.325\textwidth}
        \centering
        \includegraphics[width=\textwidth]{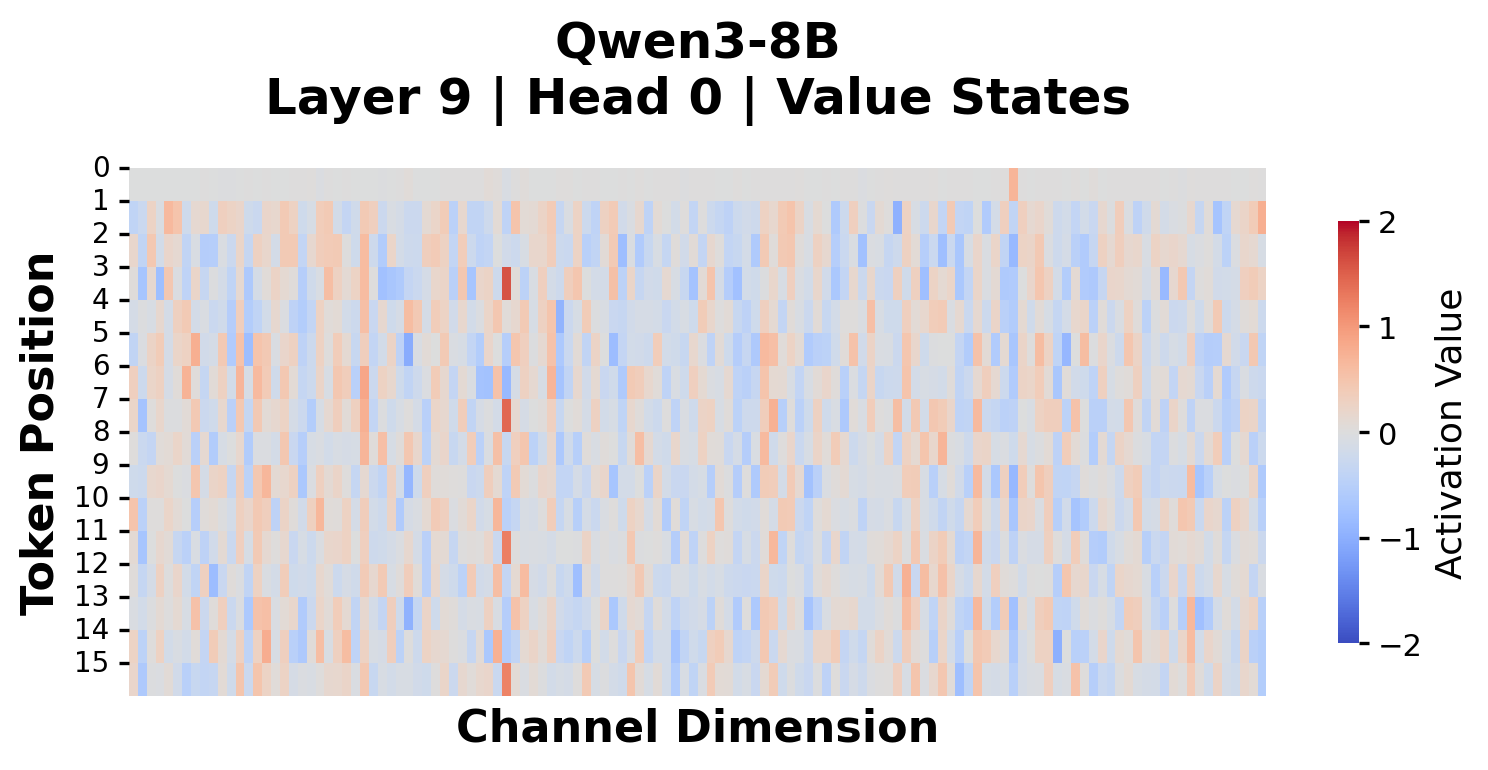}
        \caption{Value heatmap}
    \end{subfigure}
    
    \caption{L2 norm distributions (top row) and value heatmaps (bottom row) of Query, Key, and Value states in Layer 9 of Qwen-3-8B.}
    \label{fig:TNI-qwen-3-8b-layer-9}
\end{figure}

\begin{figure}[t]
    \centering
    \begin{subfigure}[b]{0.325\textwidth}
        \includegraphics[width=0.9\textwidth]{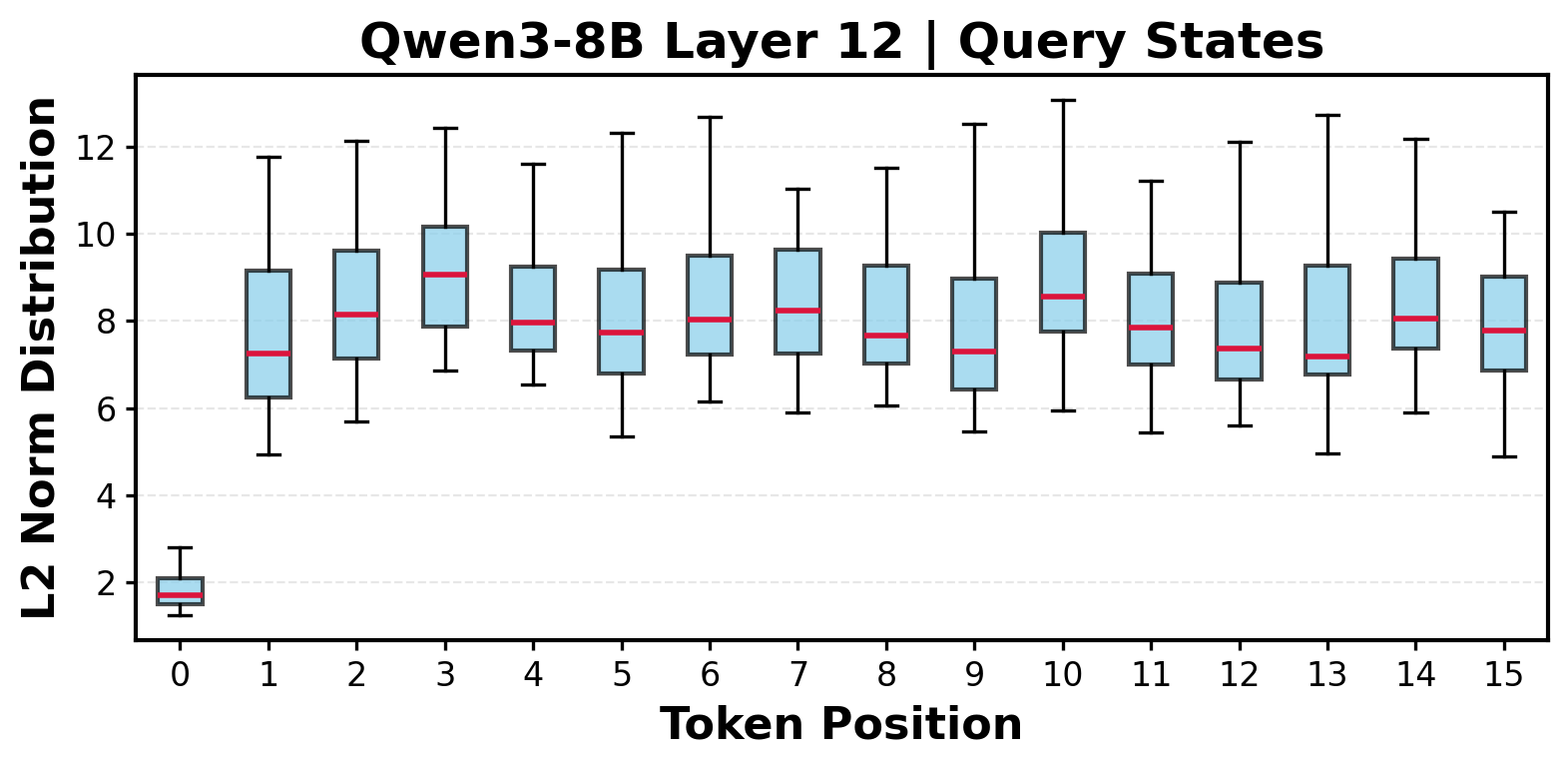}
        \caption{Query L2 norm distribution}
    \end{subfigure}
    \begin{subfigure}[b]{0.325\textwidth}
        \includegraphics[width=0.9\textwidth]{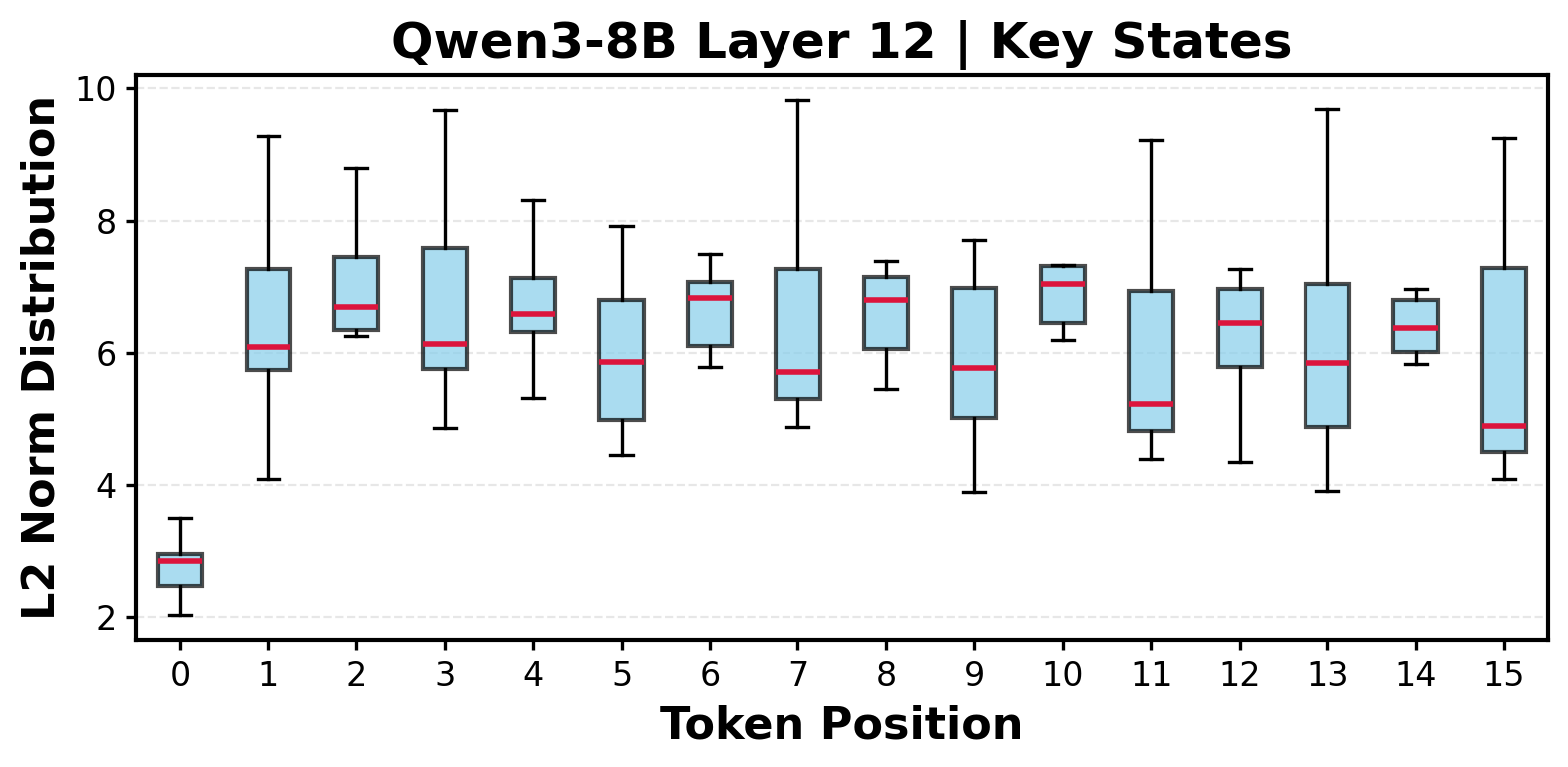}
        \caption{Key L2 norm distribution}
    \end{subfigure}
    \begin{subfigure}[b]{0.325\textwidth}
        \includegraphics[width=0.9\textwidth]{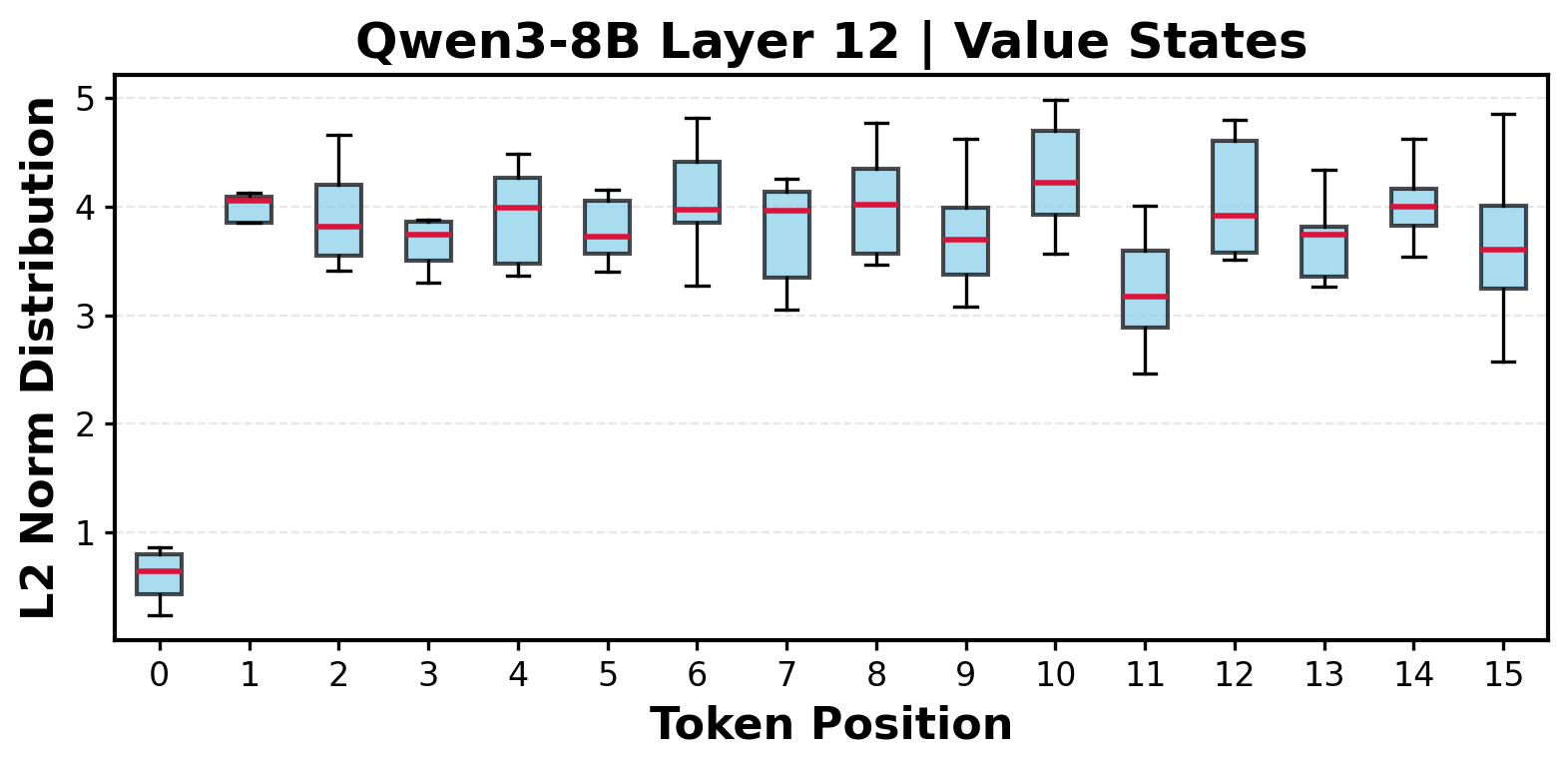}
        \caption{Value L2 norm distribution}
    \end{subfigure}
    \vspace{0.1cm}
    \begin{subfigure}[b]{0.325\textwidth}
        \centering
        \includegraphics[width=\textwidth]{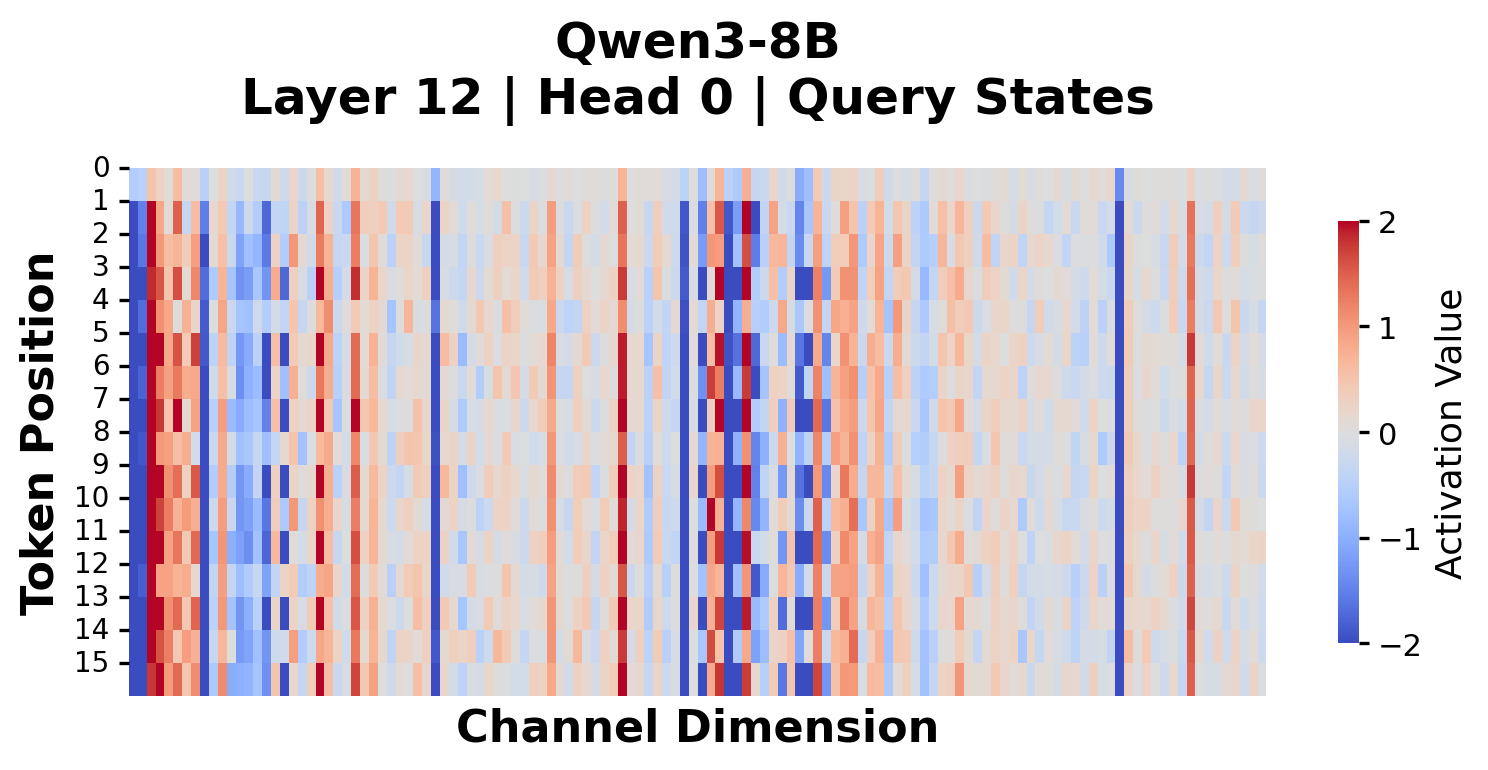}
        \caption{Query heatmap}
    \end{subfigure}
    \begin{subfigure}[b]{0.325\textwidth}
        \centering
        \includegraphics[width=\textwidth]{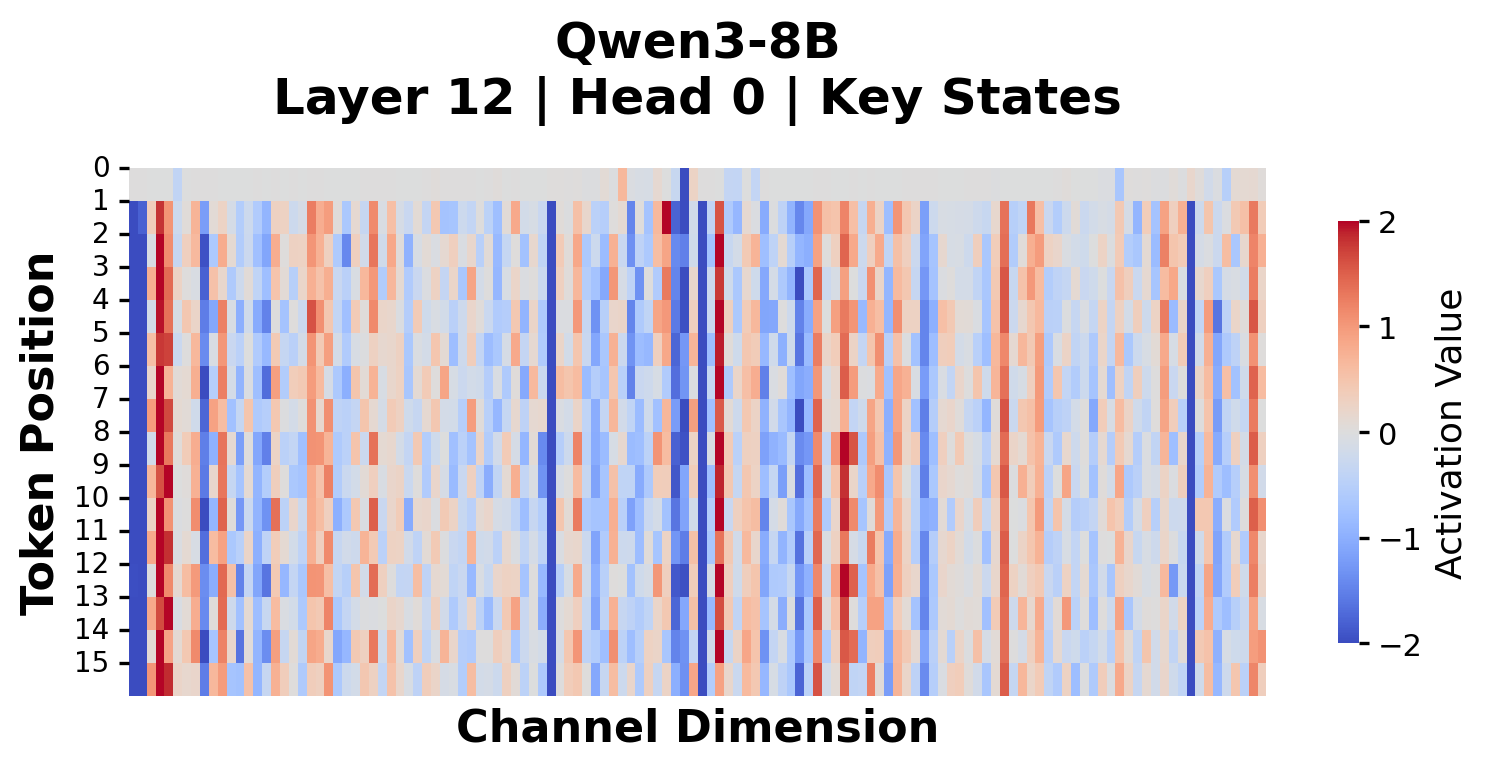}
        \caption{Key heatmap}
    \end{subfigure}
    \begin{subfigure}[b]{0.325\textwidth}
        \centering
        \includegraphics[width=\textwidth]{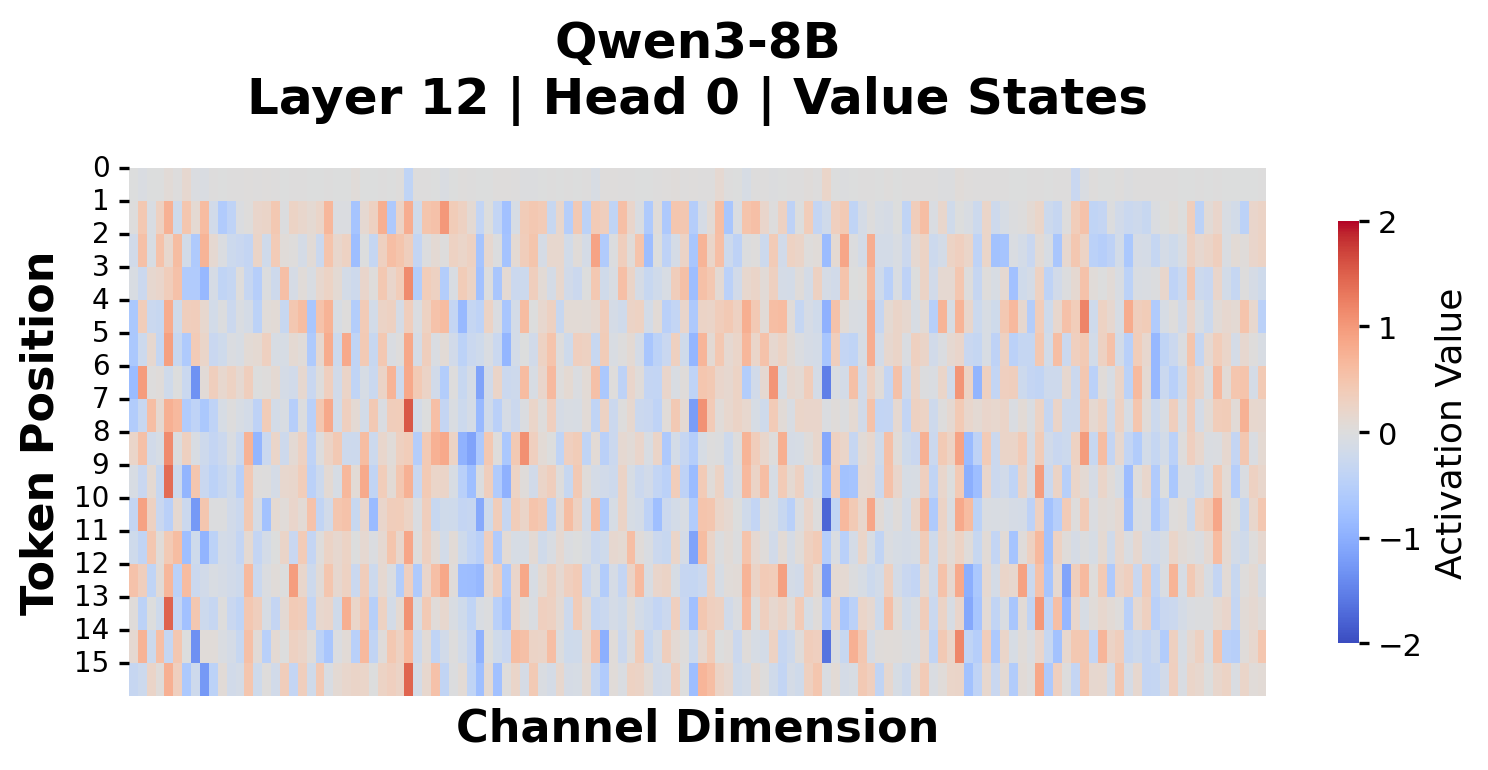}
        \caption{Value heatmap}
    \end{subfigure}
    
    \caption{L2 norm distributions (top row) and value heatmaps (bottom row) of Query, Key, and Value states in Layer 12 of Qwen-3-8B.}
    \label{fig:TNI-qwen-3-8b-layer-12}
\end{figure}

\begin{figure}[t]
    \centering
    \begin{subfigure}[b]{0.325\textwidth}
        \includegraphics[width=0.9\textwidth]{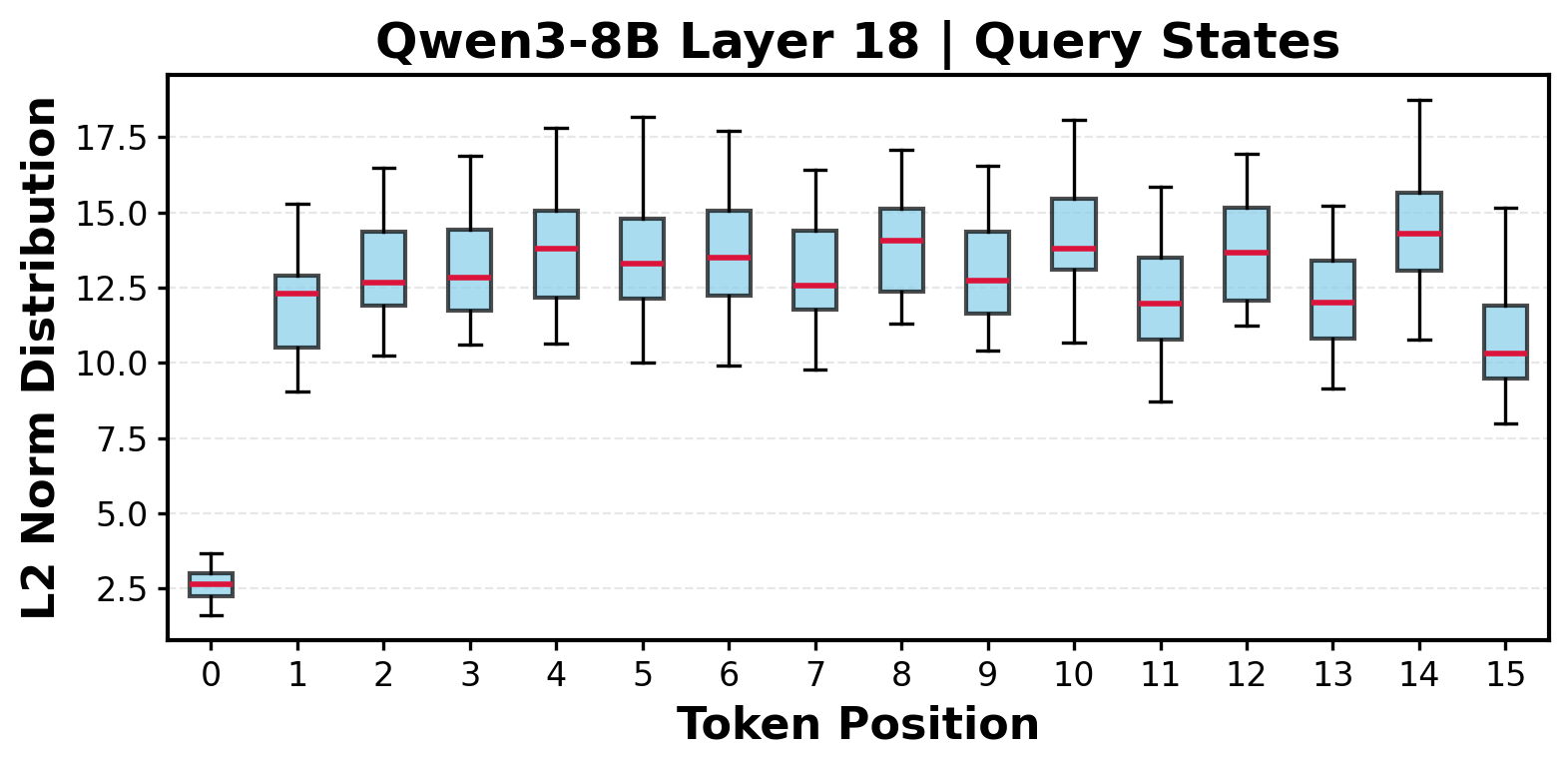}
        \caption{Query L2 norm distribution}
    \end{subfigure}
    \begin{subfigure}[b]{0.325\textwidth}
        \includegraphics[width=0.9\textwidth]{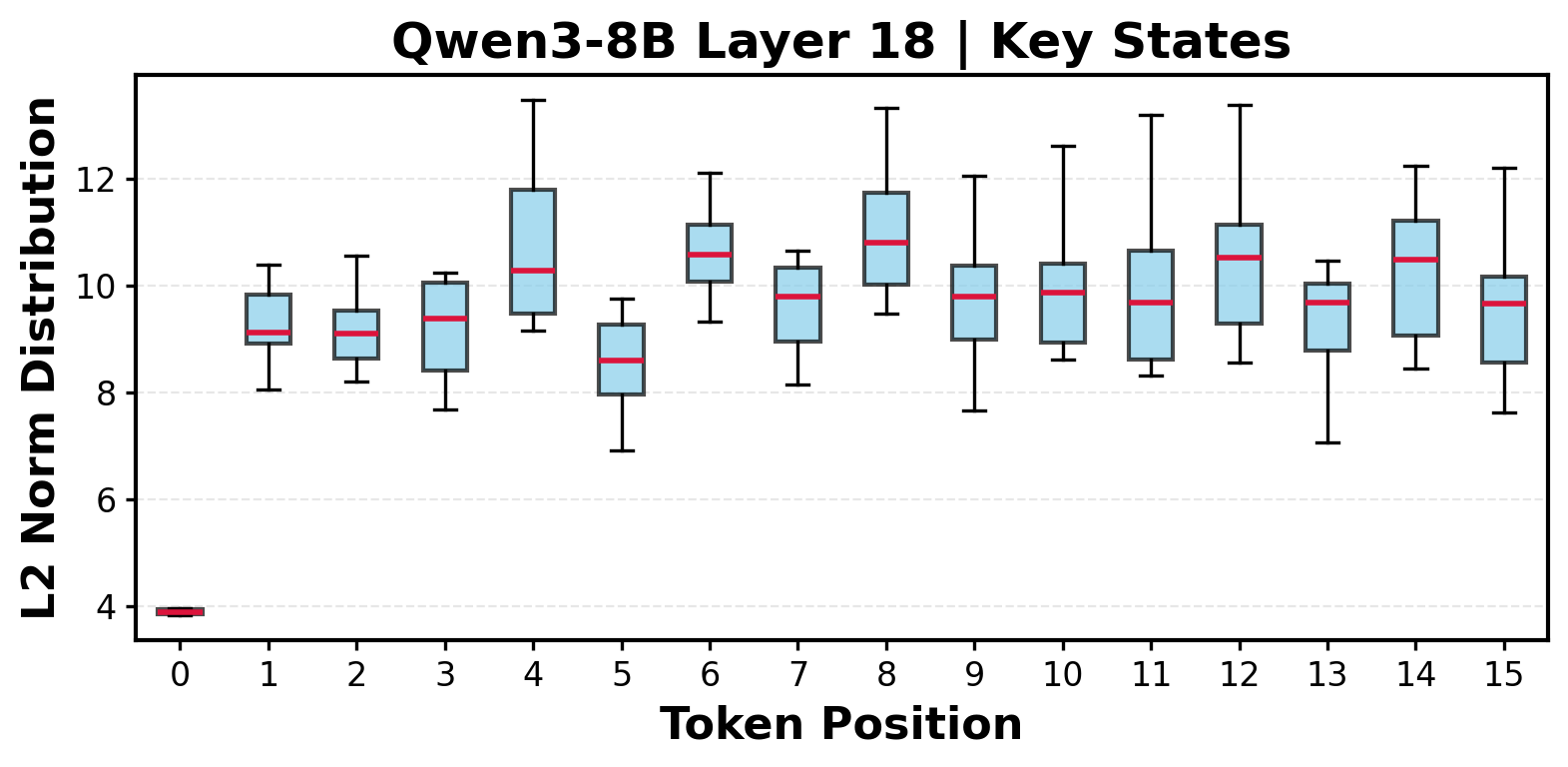}
        \caption{Key L2 norm distribution}
    \end{subfigure}
    \begin{subfigure}[b]{0.325\textwidth}
        \includegraphics[width=0.9\textwidth]{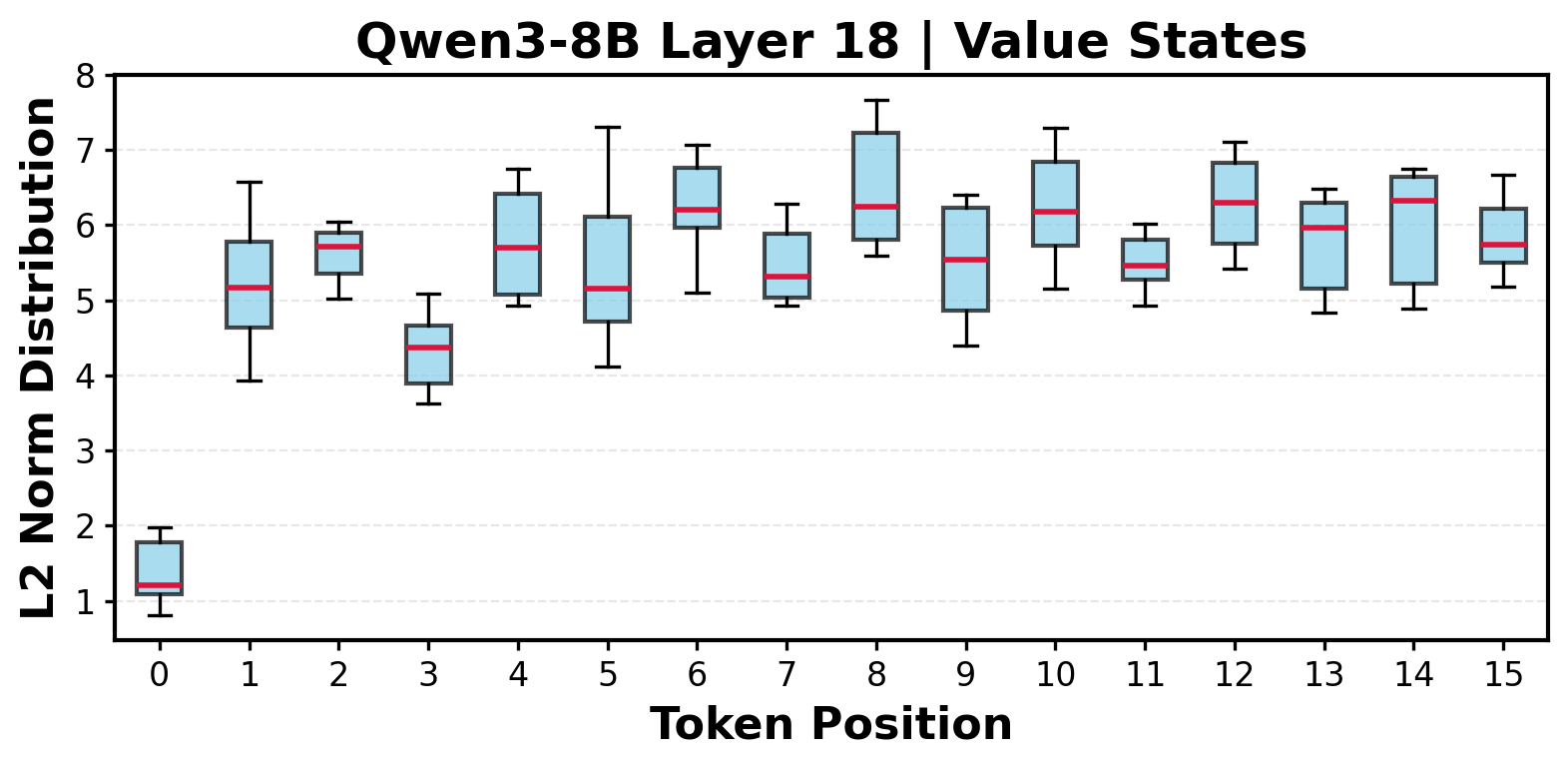}
        \caption{Value L2 norm distribution}
    \end{subfigure}
    \vspace{0.1cm}
    \begin{subfigure}[b]{0.325\textwidth}
        \centering
        \includegraphics[width=\textwidth]{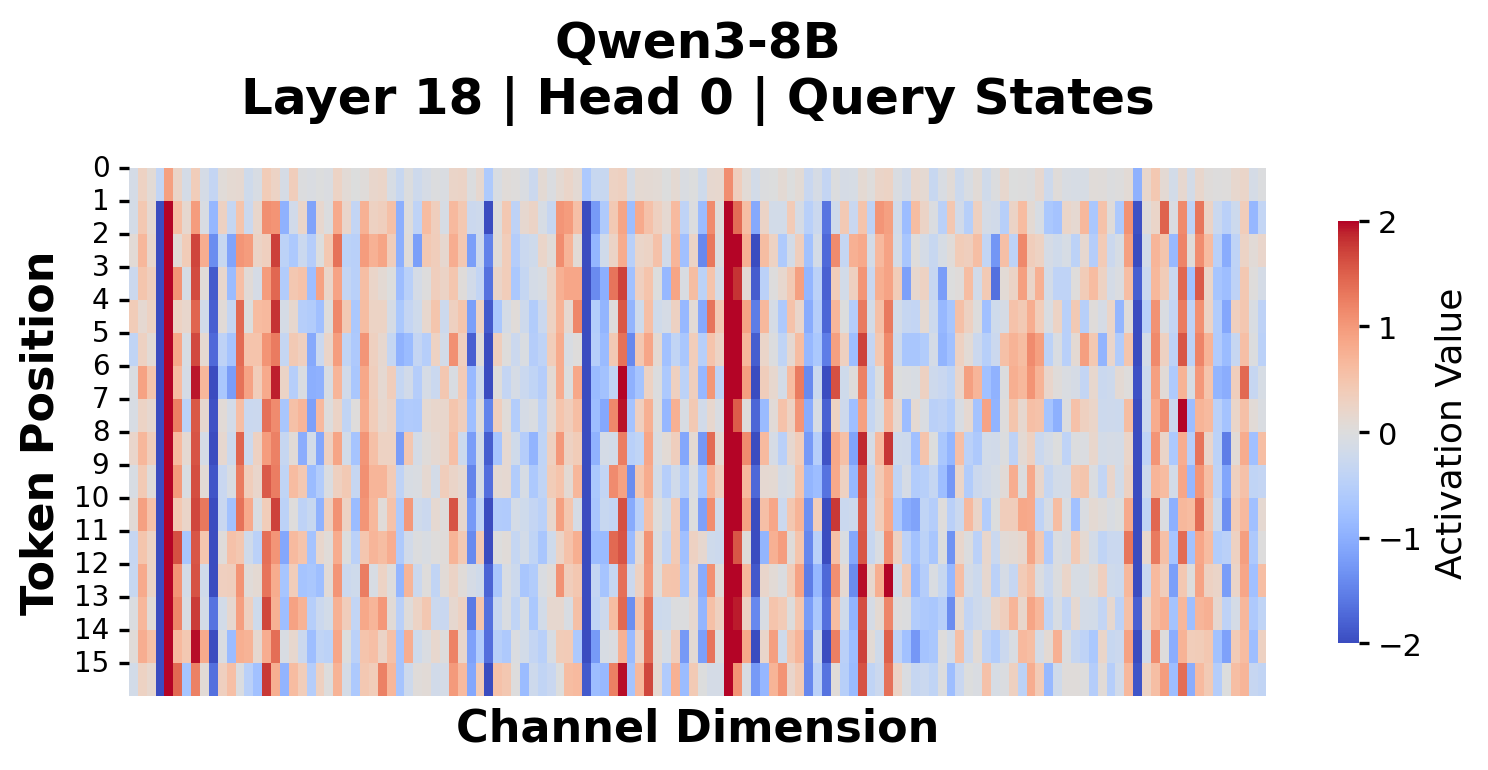}
        \caption{Query heatmap}
    \end{subfigure}
    \begin{subfigure}[b]{0.325\textwidth}
        \centering
        \includegraphics[width=\textwidth]{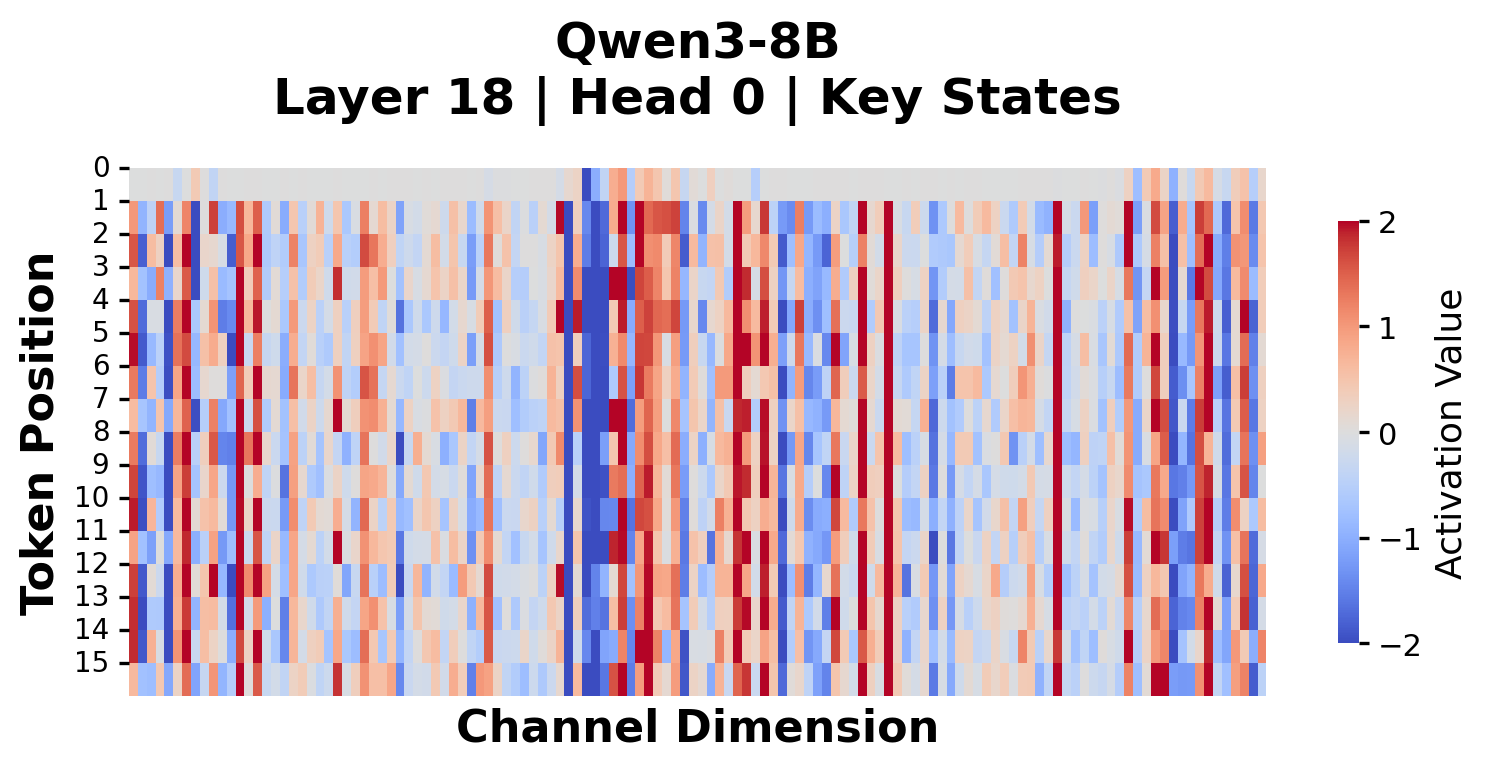}
        \caption{Key heatmap}
    \end{subfigure}
    \begin{subfigure}[b]{0.325\textwidth}
        \centering
        \includegraphics[width=\textwidth]{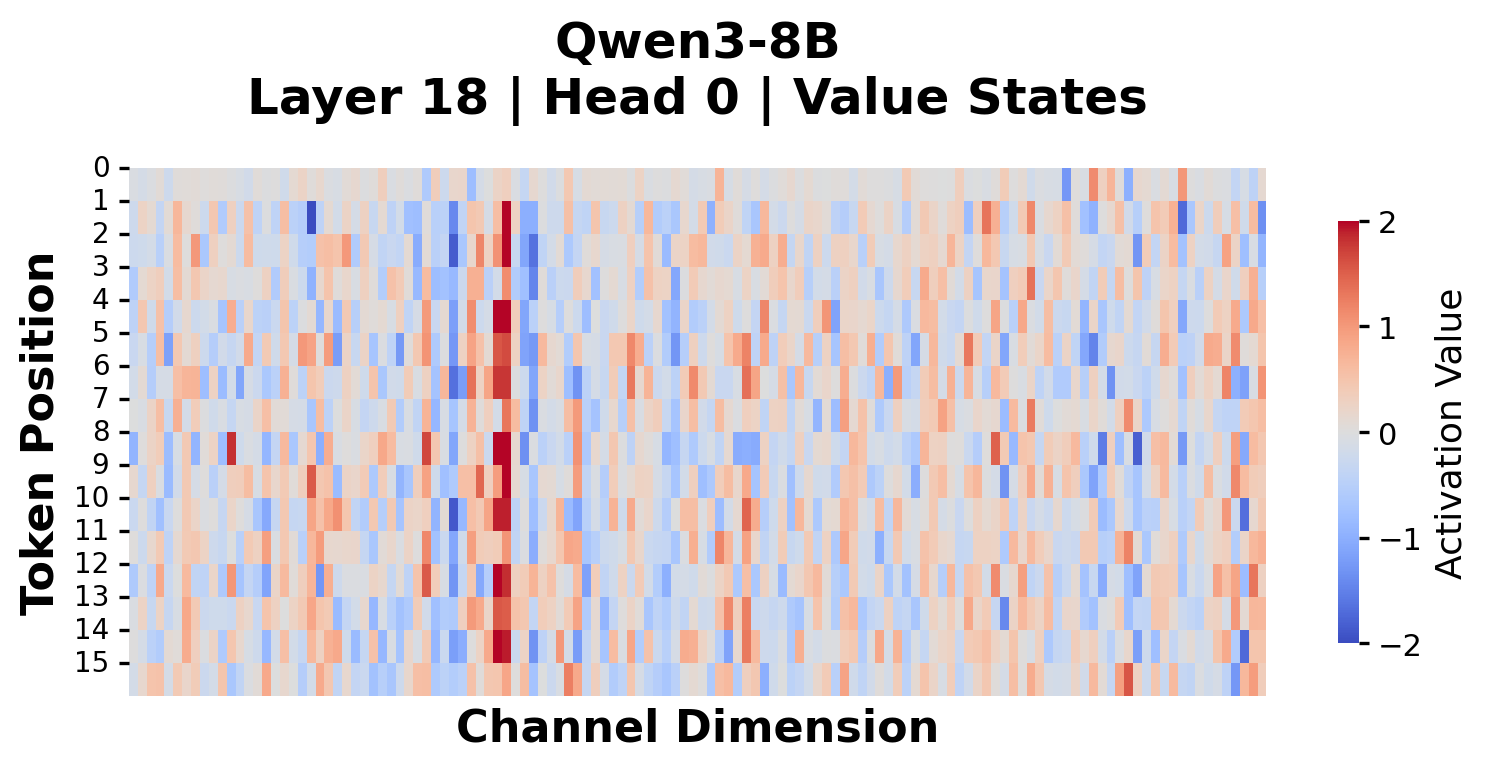}
        \caption{Value heatmap}
    \end{subfigure}
    
    \caption{L2 norm distributions (top row) and value heatmaps (bottom row) of Query, Key, and Value states in Layer 18 of Qwen-3-8B.}
    \label{fig:TNI-qwen-3-8b-layer-18}
\end{figure}

The results are presented in Figures~\ref{fig:TNI-llama-2-7b-layer-6}, \ref{fig:TNI-llama-2-7b-layer-12}, \ref{fig:TNI-llama-2-7b-layer-18}, \ref{fig:TNI-llama-3-8b-layer-6}, \ref{fig:TNI-llama-3-8b-layer-12}, \ref{fig:TNI-llama-3-8b-layer-18}, \ref{fig:TNI-qwen-3-8b-layer-9}, \ref{fig:TNI-qwen-3-8b-layer-12}, and \ref{fig:TNI-qwen-3-8b-layer-18}.
Across all cases, low-norm outlier tokens are clearly observed in Q, K, and V states. These tokens are sparse yet consistent across attention states. Under per-channel quantization, these outliers inflate the per-channel scaling factors, leading to substantial quantization error, resulting in substantial fidelity degradation.

\section{Outlier Tokens and Attention Sinks}
\label{app:AS}

\begin{figure}[h]
    \centering
    \begin{subfigure}[b]{0.325\textwidth}
        \includegraphics[width=0.9\textwidth]{Figure/norm_vis/llm/q_proj_layer_18_boxplot.png}
        \caption{Query L2 norm distribution}
    \end{subfigure}
    \begin{subfigure}[b]{0.325\textwidth}
        \includegraphics[width=0.9\textwidth]{Figure/norm_vis/llm/k_proj_layer_18_boxplot.png}
        \caption{Key L2 norm distribution}
    \end{subfigure}
    \begin{subfigure}[b]{0.325\textwidth}
        \includegraphics[width=0.9\textwidth]{Figure/norm_vis/llm/v_proj_layer_18_boxplot.png}
        \caption{Value L2 norm distribution}
    \end{subfigure}
    
    \vspace{0.1cm}
    
    \begin{subfigure}[b]{0.325\textwidth}
        \centering
        \includegraphics[width=\textwidth]{Figure/norm_vis/llm/q_proj_layer_18_head_0_heatmap.png}
        \caption{Query heatmap}
    \end{subfigure}
    \begin{subfigure}[b]{0.325\textwidth}
        \centering
        \includegraphics[width=\textwidth]{Figure/norm_vis/llm/k_proj_layer_18_head_0_heatmap.png}
        \caption{Key heatmap}
    \end{subfigure}
    \begin{subfigure}[b]{0.325\textwidth}
        \centering
        \includegraphics[width=\textwidth]{Figure/norm_vis/llm/v_proj_layer_18_head_0_heatmap.png}
        \caption{Value heatmap}
    \end{subfigure}

    \vspace{0.1cm}
    
    \begin{subfigure}[b]{0.325\textwidth}
        \centering
        \includegraphics[width=\textwidth]{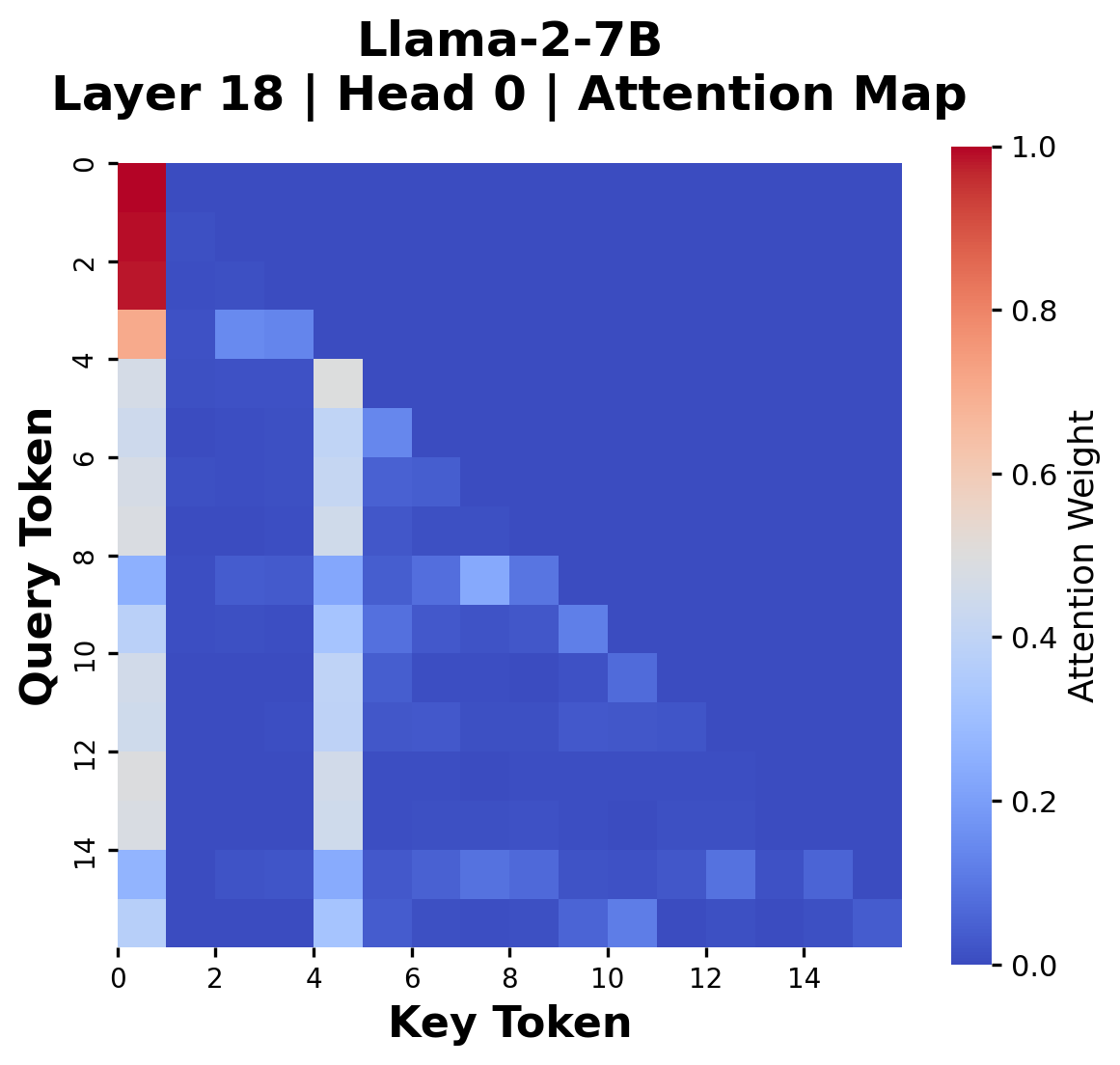}
        \caption{Attention map (Head 0)}
    \end{subfigure}
    \begin{subfigure}[b]{0.325\textwidth}
        \centering
        \includegraphics[width=\textwidth]{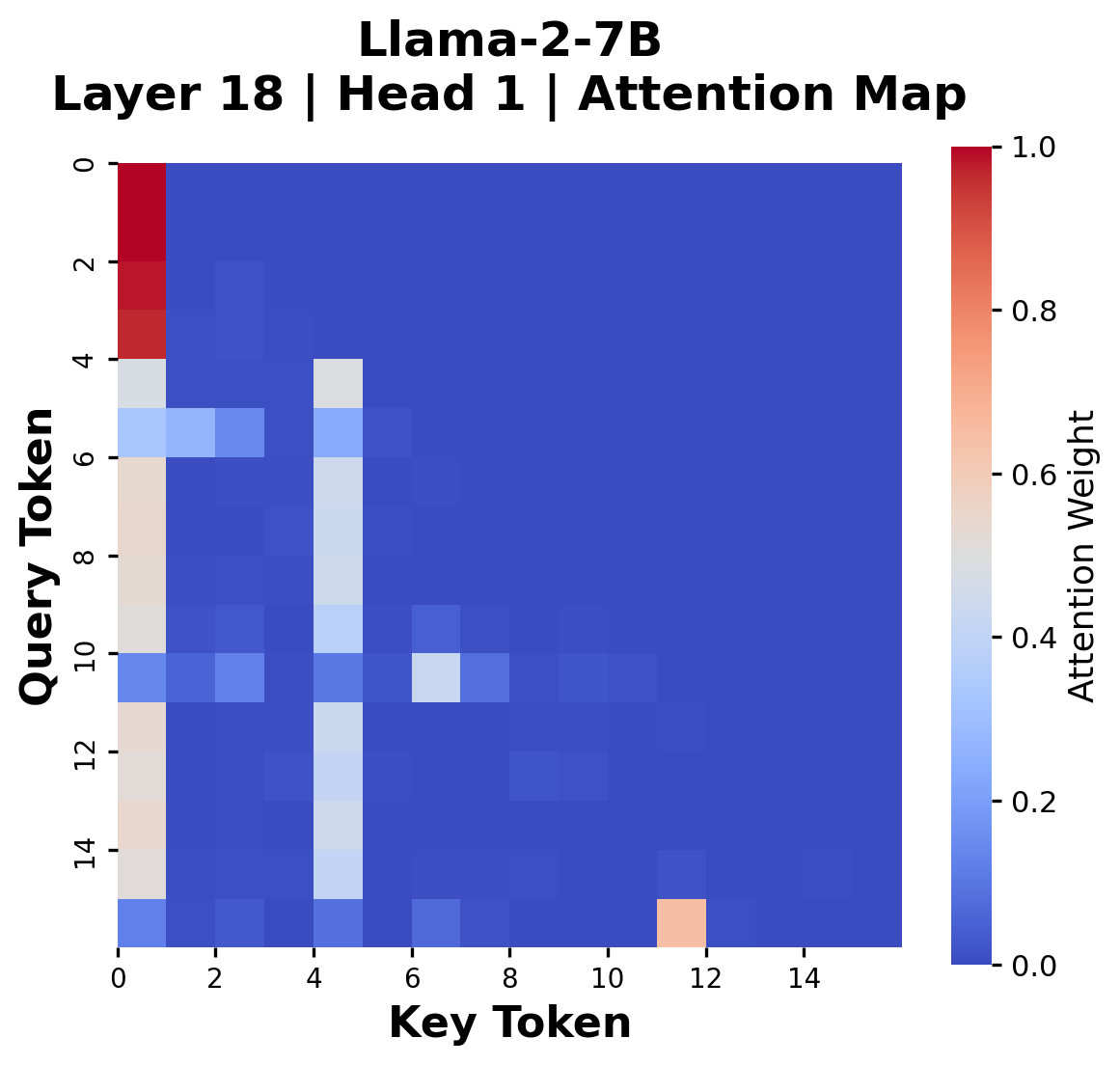}
        \caption{Attention map (Head 1)}
    \end{subfigure}
    \begin{subfigure}[b]{0.325\textwidth}
        \centering
        \includegraphics[width=\textwidth]{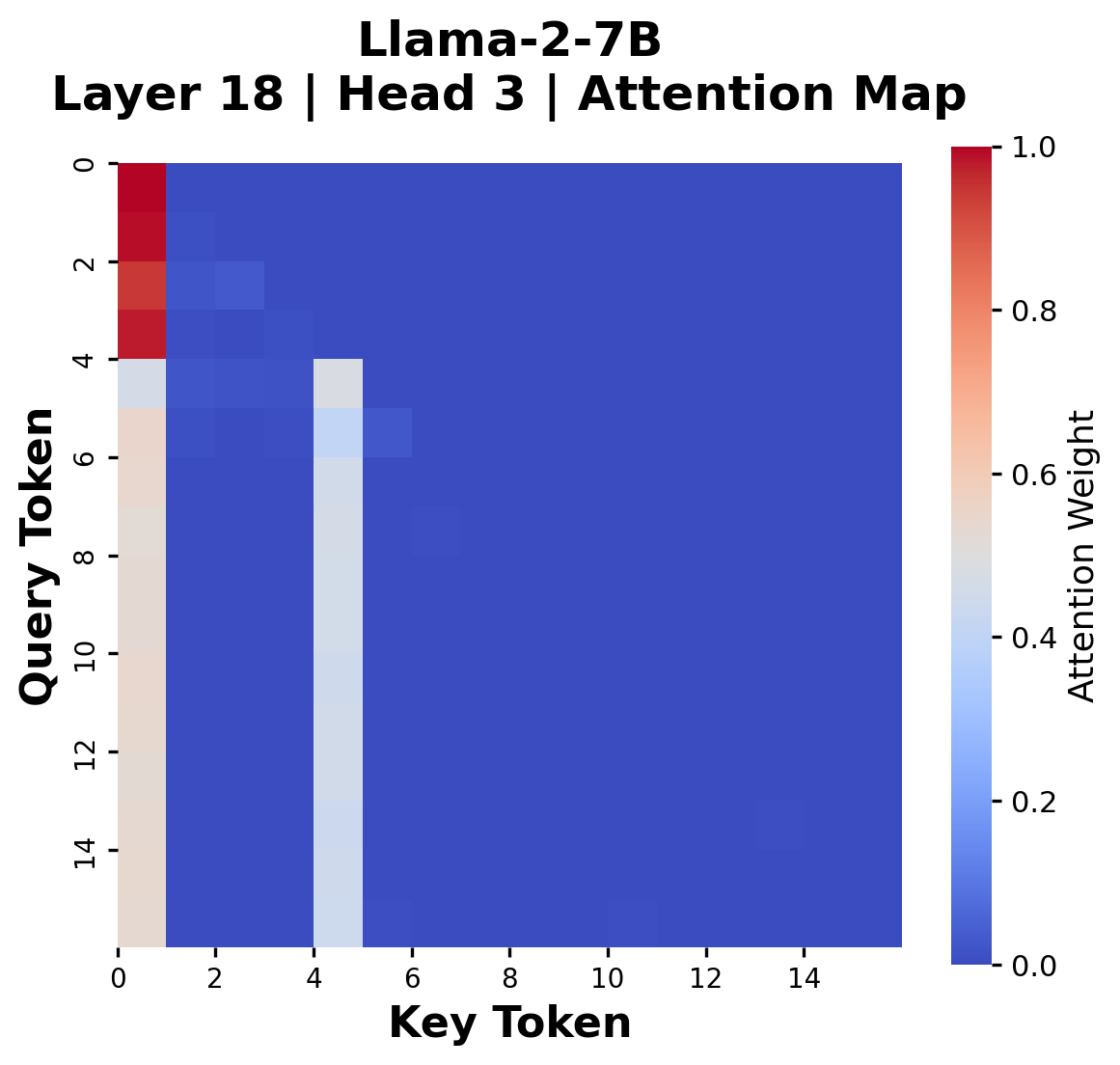}
        \caption{Attention map (Head 3)}
    \end{subfigure}
    
    \caption{L2 norm distributions (row 1), value heatmaps (row 2), and attention maps (row 3) of Query, Key, and Value states. Each attention state contains a sparse yet consistent subset of tokens with exceptionally low norms. These low-norm outlier tokens correspond exactly to the same token positions as Attention Sink tokens.}
    \label{fig:AS}
\end{figure}

As shown in Figure~\ref{fig:AS}, outlier tokens remain consistent not only across Query, Key, and Value states but also with Attention Sinks, corroborating prior studies~\citep{su2026attention,su2025kvsink}. A widely accepted explanation for this behavior is the \textit{softmax limitation and no-op theory} \citep{bondarenko2023quantizable}, which we briefly recapitulate below.

In standard attention, the sum-to-one softmax constraint requires that, for each query, the attention weights over all keys normalize to unity. When a query does not meaningfully align with any key in the context, the mechanism lacks a natural "null" option and is therefore forced to distribute attention mass to uninformative tokens. Consequently, attention heads learn to circumvent this constraint by adopting a no-op behavior.
Let $\mathcal{S}$ denote the set of sink tokens (e.g., \texttt{[SEP]}, punctuation, or background patches). The resulting attention pattern can be approximated as:
\begin{equation}
A_{ij} \approx 
\begin{cases}
1, & j \in \mathcal{S} \\[4pt]
0, & \text{otherwise}
\end{cases}
\qquad \text{with} \qquad \|V_{\mathcal{S}}\| \approx 0,
\end{equation}
where nearly all attention mass concentrates on sink tokens, whose Value vectors exhibit low norms, thereby producing minimal updates to the residual representation. This compression phenomenon also extends beyond Value states to Query and Key states~\citep{su2025kvsink}.

Studies demonstrate that Attention-Sink-related extreme tokens can compromise training stability and hinder low-precision deployment~\citep{qiu2025gated,liang2025tweo,su2025unveiling}. Moreover, misallocated attention to uninformative tokens inherently limits overall model capacity~\citep{kang2025see,tu2026attention}. Consequently, developing robust mitigation frameworks has emerged as a critical research frontier. Recent efforts have focused on systematically explaining and eliminating Attention Sinks, proposing approaches such as gated attention, modified softmax, and explicit attention bias~\citep{qiu2025gated,zuhri2025softpick,xiong2025dope,agarwal2025gpt}.

\section{Token Norm Imbalance in Multi-modal LLMs}
\label{app:mllm}

\begin{figure}[h]
    \centering
    
    \includegraphics[width=0.4\textwidth]{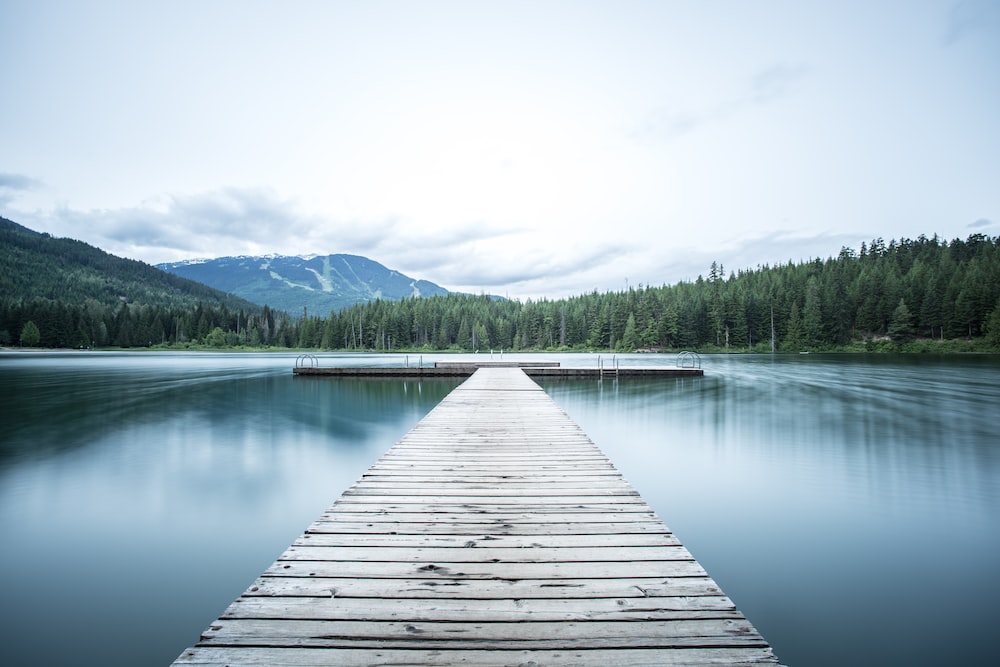}
    \caption{Example image used as visual input.}
    \label{fig:example_image}
    
\end{figure}

As discussed in Section~\ref{sec:revisiting}, we visualize the L2 norm distributions of Query, Key, and Value states in multi-modal LLMs. The input is formatted using the model's chat template with \texttt{add\_generation\_prompt=True}, resulting in the token sequence shown below, where \texttt{</td>} denotes the sequence of image patch tokens corresponding to the example image in Figure~\ref{fig:example_image}:

{\small
\begin{verbatim}
<|im_start|>user
<|vision_start|></td><|vision_end|>What is in this image?<|im_end|>
<|im_start|>assistant
\end{verbatim}
}

\begin{figure}[t]
    \centering
    \begin{subfigure}[b]{\textwidth}
        \includegraphics[width=\textwidth]{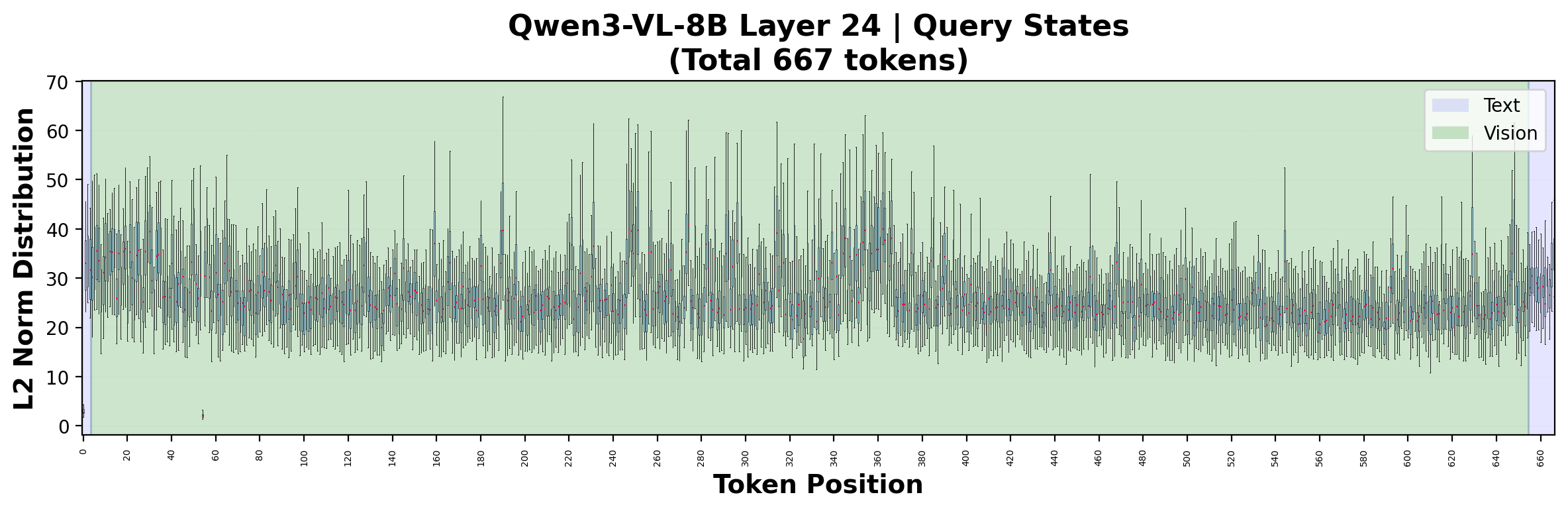}
        \caption{Query L2 norm distribution}
    \end{subfigure}
    \begin{subfigure}[b]{\textwidth}
        \includegraphics[width=\textwidth]{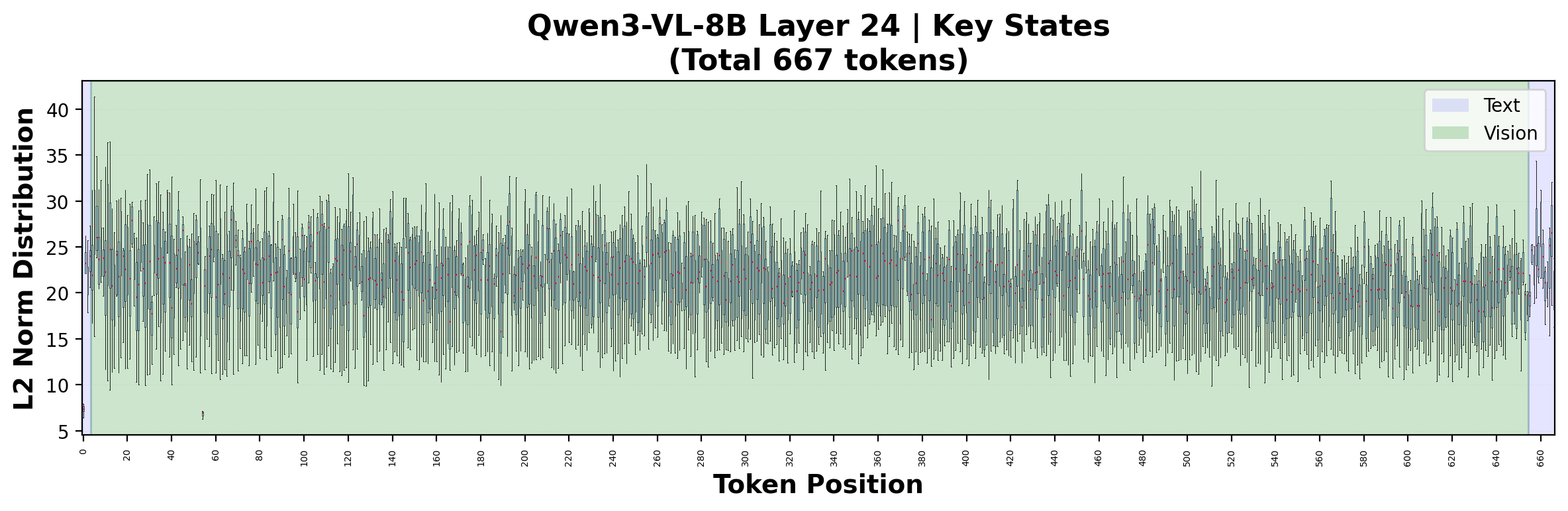}
        \caption{Key L2 norm distribution}
    \end{subfigure}
    \begin{subfigure}[b]{\textwidth}
        \includegraphics[width=\textwidth]{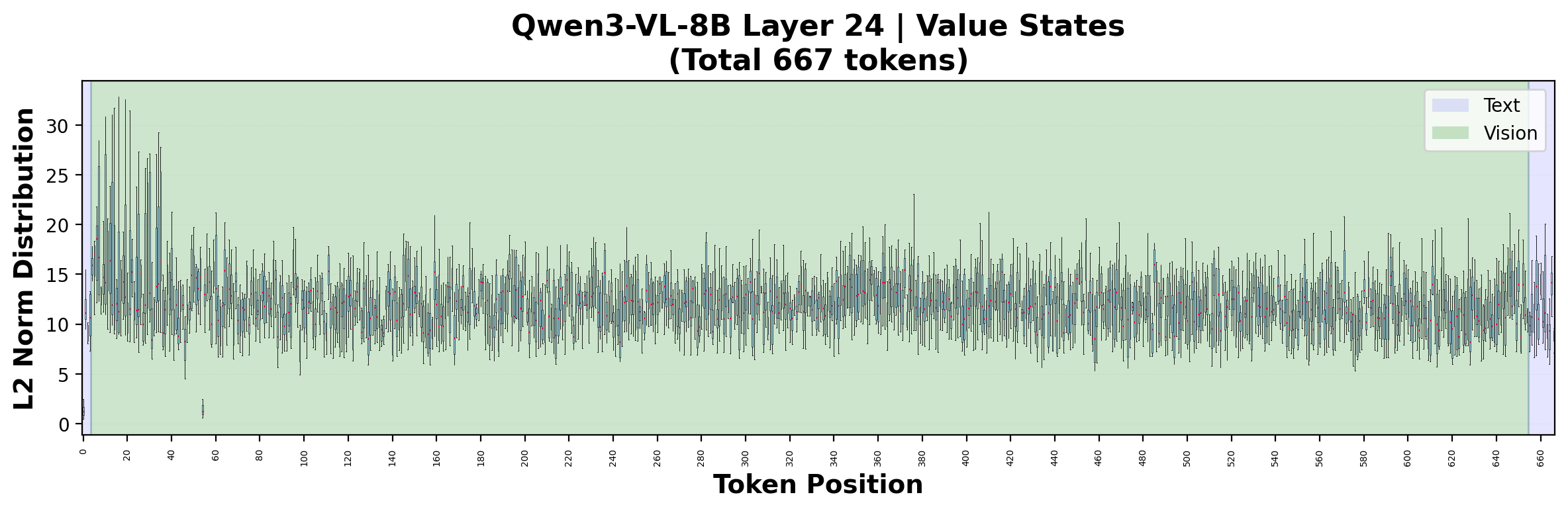}
        \caption{Value L2 norm distribution}
    \end{subfigure}
    
    \caption{L2 norm distributions of Query, Key, and Value states in Layer 24 of Qwen-3-VL-8B, showing broader token norm variation compared to text-only LLMs.}
    \label{fig:TNI-qwen-3-vl-8b-layer-24}
\end{figure}

\begin{figure}[t]
    \centering
    \begin{subfigure}[b]{\textwidth}
        \includegraphics[width=\textwidth]{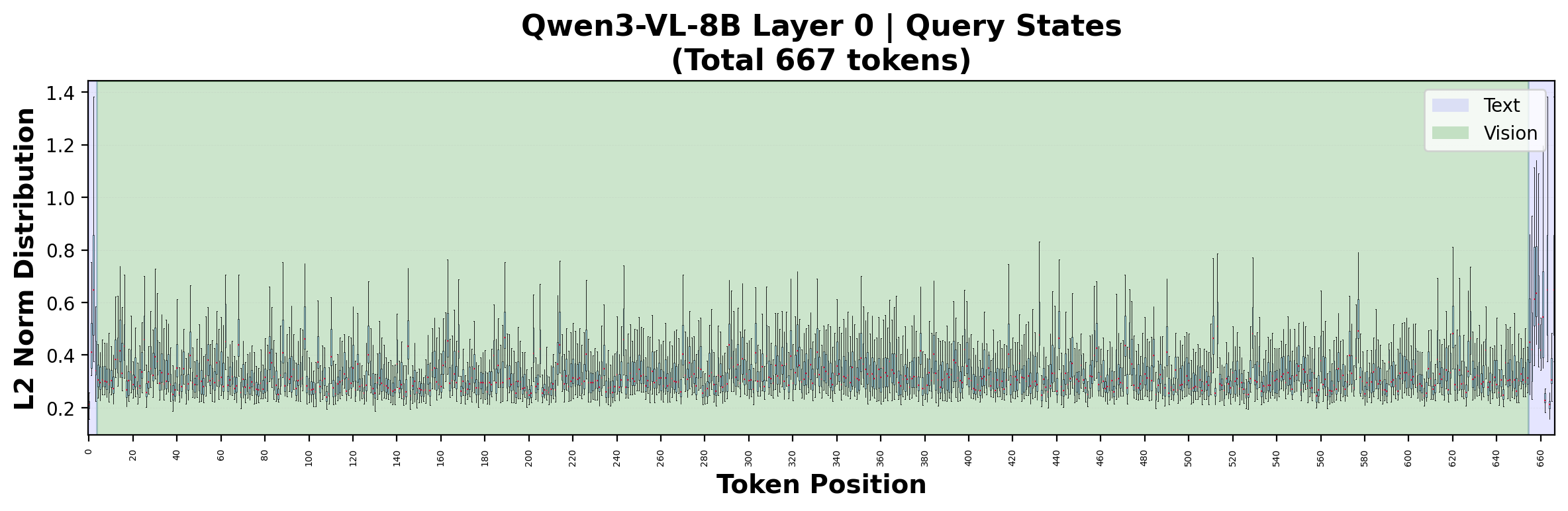}
        \caption{Query L2 norm distribution}
    \end{subfigure}
    \begin{subfigure}[b]{\textwidth}
        \includegraphics[width=\textwidth]{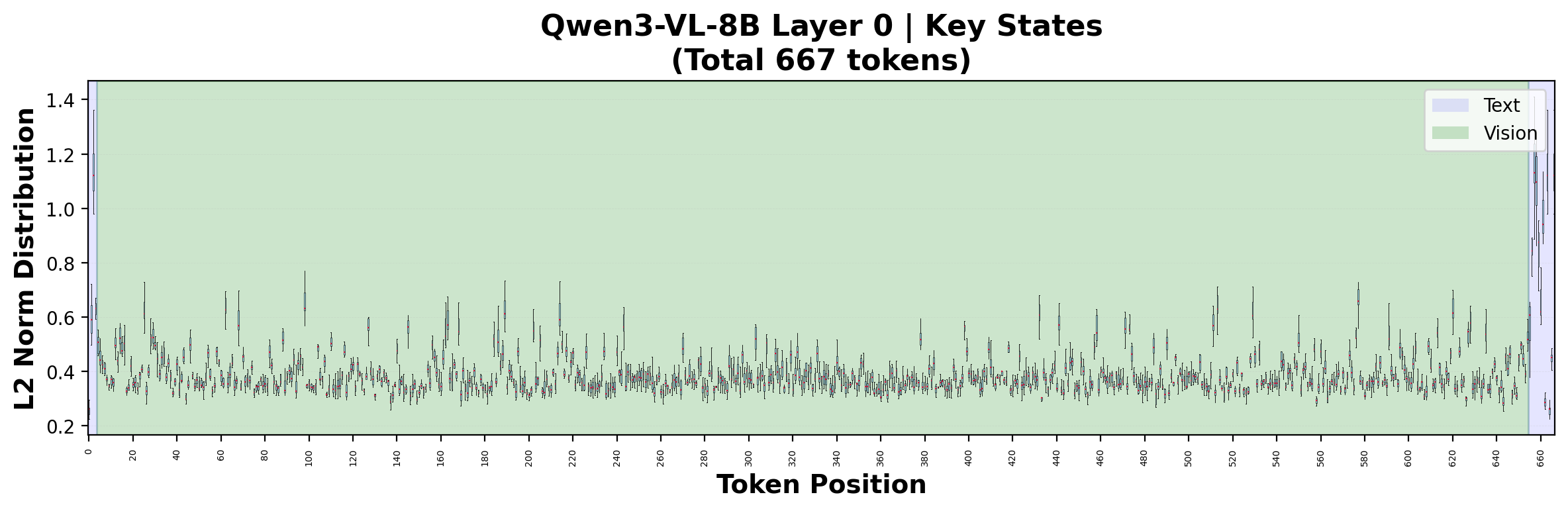}
        \caption{Key L2 norm distribution}
    \end{subfigure}
    \begin{subfigure}[b]{\textwidth}
        \includegraphics[width=\textwidth]{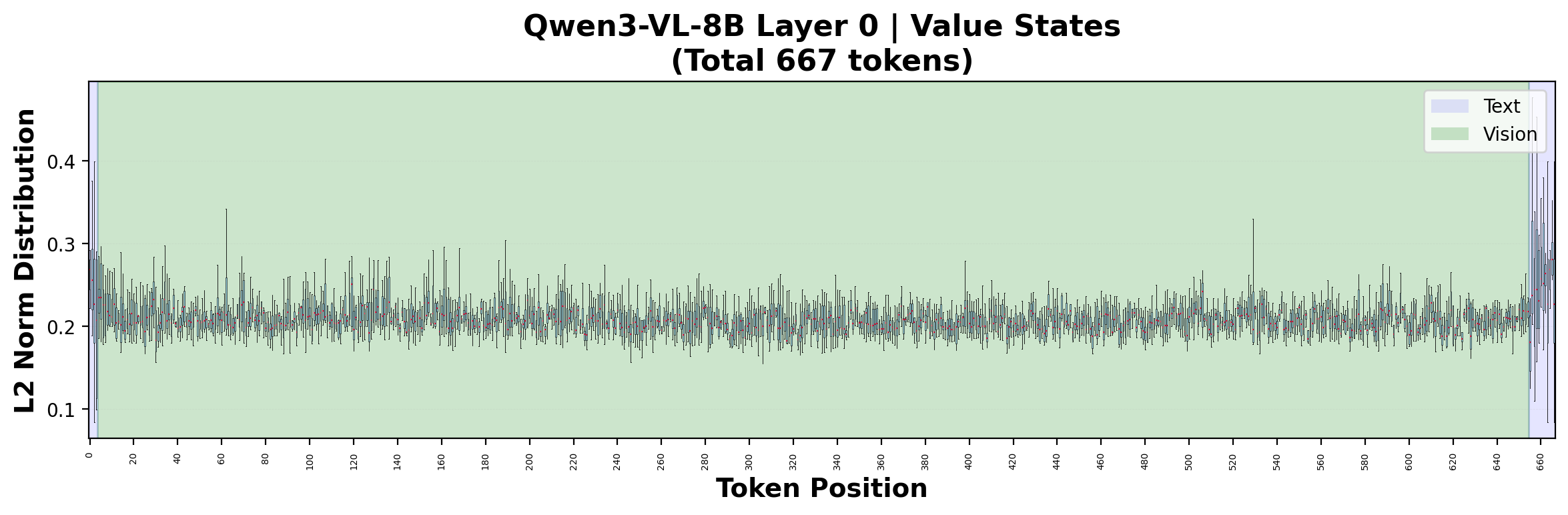}
        \caption{Value L2 norm distribution}
    \end{subfigure}
    \caption{L2 norm distributions of Query, Key, and Value states in Layer 0 of Qwen-3-VL-8B, revealing significant inter-modality norm disparities.}
    \label{fig:TNI-qwen-3-vl-8b-layer-0}
\end{figure}

\begin{figure}[t]
    \centering
    \begin{subfigure}[b]{\textwidth}
        \includegraphics[width=\textwidth]{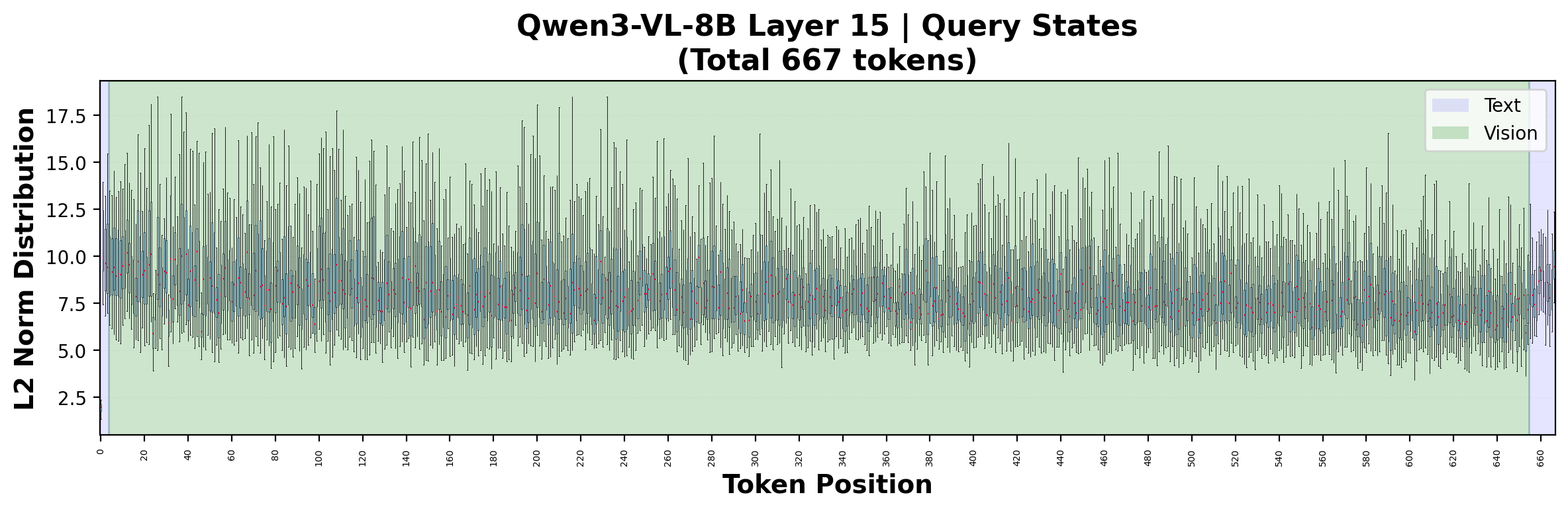}
        \caption{Query L2 norm distribution}
    \end{subfigure}
    \begin{subfigure}[b]{\textwidth}
        \includegraphics[width=\textwidth]{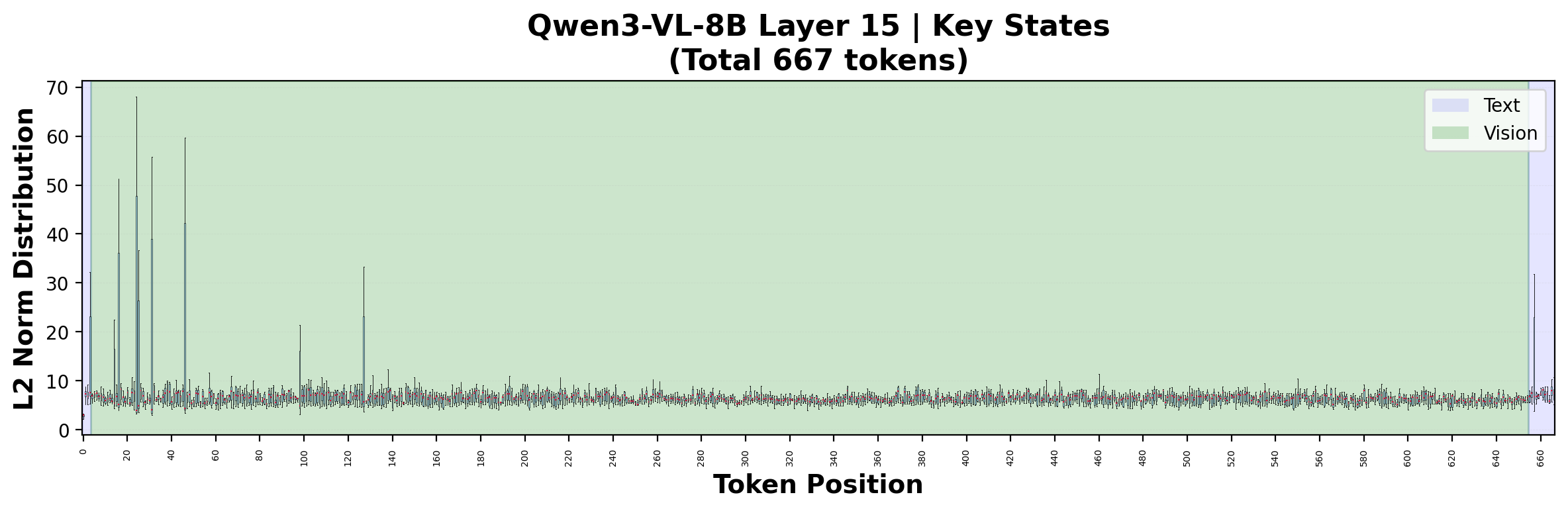}
        \caption{Key L2 norm distribution}
    \end{subfigure}
    \begin{subfigure}[b]{\textwidth}
        \includegraphics[width=\textwidth]{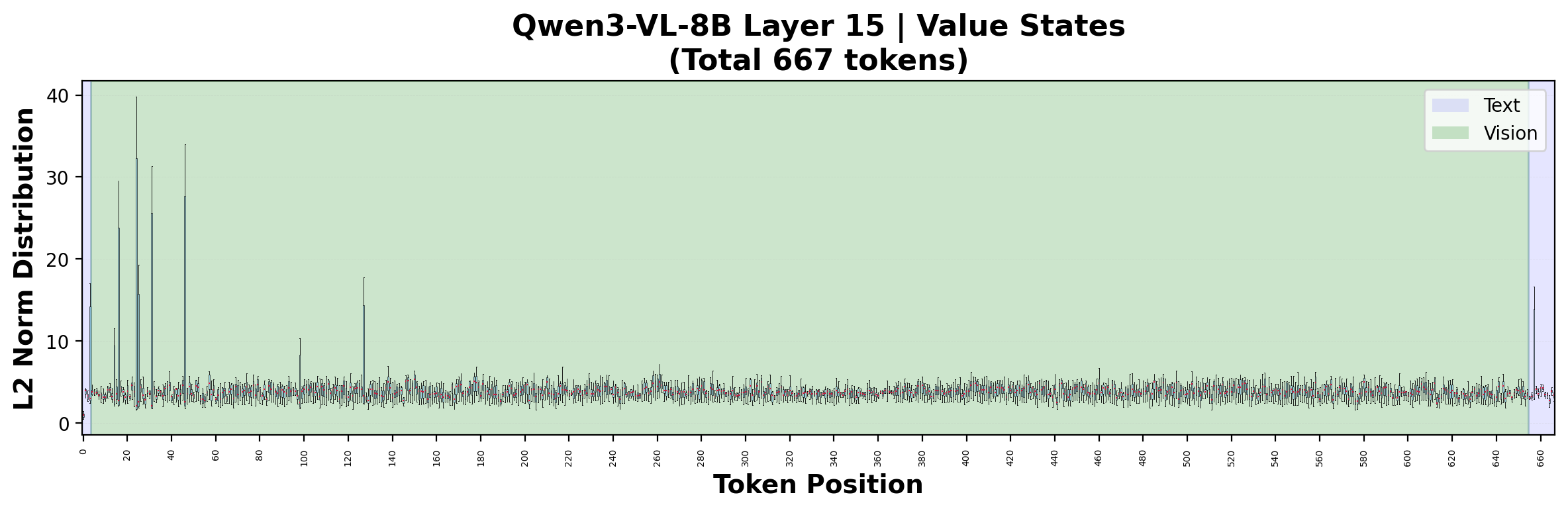}
        \caption{Value L2 norm distribution}
    \end{subfigure}
    
    \caption{L2 norm distributions of Query, Key, and Value states in Layer 15 of Qwen-3-VL-8B, revealing outlier tokens with exceptionally large norms.}
    \label{fig:TNI-qwen-3-vl-8b-layer-15}
\end{figure}

Representative results are presented in Figures~\ref{fig:TNI-qwen-3-vl-8b-layer-24}, \ref{fig:TNI-qwen-3-vl-8b-layer-0}, and \ref{fig:TNI-qwen-3-vl-8b-layer-15}, which respectively demonstrate three patterns of TNI in multi-modal LLMs beyond attention-sink-related low-norm tokens: (i) broader token norm variation relative to text-only LLMs (Figure~\ref{fig:TNI-qwen-3-vl-8b-layer-24}); (ii) inter-modality norm disparities, wherein norms remain smooth within each modality yet diverge substantially across modalities (Figure~\ref{fig:TNI-qwen-3-vl-8b-layer-0}); and (iii) exceptionally large-norm outlier tokens, which contrast with the low-norm attention sink tokens (Figure~\ref{fig:TNI-qwen-3-vl-8b-layer-15}).

\section{Theoretical Derivation of TNI-Induced Quantization Errors}
\label{app:theoretical_derivations}

Building on the empirical observations of TNI across X-LLMs in Section~\ref{sec:revisiting}, we now present a theoretical analysis of TNI-induced errors in per-channel quantization.

For a uniform $b$-bit quantizer, the quantization error $\epsilon$ is modeled as a random variable uniformly distributed over $[-\Delta_{j,g}/2, \Delta_{j,g}/2]$. The mean squared error (MSE) for the $j$-th channel in block $g$ is
\begin{equation}
\mathrm{MSE}_{j,g} = \mathbb{E}[\epsilon^2] = \frac{1}{\Delta_{j,g}} \int_{-\Delta_{j,g}/2}^{\Delta_{j,g}/2} \epsilon^2 \, d\epsilon = \frac{\Delta_{j,g}^2}{12}.
\label{eq:mse_exact}
\end{equation}

The quantization step size $\Delta_{j,g}$ is determined by the sample range $\mathcal{R}_{j,g} = \max_{t \in g} K_{t,j} - \min_{t \in g} K_{t,j}$ and the bit width, i.e., $\Delta_{j,g} = \mathcal{R}_{j,g}/(2^b-1)$. For any two tokens $u, v$ in the same block $g$, the range satisfies $\mathcal{R}_{j,g} \ge |K_{u,j} - K_{v,j}|$. Substituting this into Eq.~\eqref{eq:mse_exact} yields a pairwise lower bound:
\begin{equation}
\mathrm{MSE}_{j,g} \ge \frac{(K_{u,j} - K_{v,j})^2}{12(2^b-1)^2}.
\label{eq:mse_pair}
\end{equation}

Within block $g$, let $m = \arg\max_{t \in g} \|\mathbf{k}_t\|_2$ and $n = \arg\min_{t \in g} \|\mathbf{k}_t\|_2$ denote the tokens with the largest and smallest $\ell_2$ norms, respectively. Applying Eq.~\eqref{eq:mse_pair} and summing over all channels gives a conservative lower bound on the average reconstruction error:
\begin{equation}
\begin{aligned}
\overline{\mathrm{MSE}}_g
&\triangleq \frac{1}{|g|}\sum_{t \in g}\mathrm{MSE}_t
\;\gtrsim\;\frac{1}{12(2^b-1)^2}\sum_{j=1}^{d}(K_{m,j}-K_{n,j})^2 \\
&=\frac{\|\mathbf{k}_m-\mathbf{k}_n\|_2^2}{12(2^b-1)^2}
\ge\frac{\bigl(\|\mathbf{k}_m\|_2-\|\mathbf{k}_n\|_2\bigr)^2}{12(2^b-1)^2}.
\end{aligned}
\label{eq:core_bound}
\end{equation}
where $|g|$ denotes the block size and $\mathrm{MSE}_t = \sum_{j=1}^{d} \mathrm{MSE}_{j,g,t}$ represents the total quantization error across all channels for token $t$.

This inequality reveals that the reconstruction error of a per-channel quantization block is fundamentally governed by the range of token norms within the block. Thus, TNI-induced error amplification constitutes an intrinsic vulnerability of per-channel block-wise quantization.

Under the uniform-noise model, this bound connects per-channel reconstruction error to token norm disparity. The real-tensor calibration in Appendix~\ref{app:calibration} evaluates the approximation empirically; the bound is not a universal guarantee for deterministic low-bit quantization.

\subsection{Connection to attention-logit error}
\label{app:logit_error}
For a Key vector $\mathbf{k}$, let $\mathbf{e}=\widehat{\mathbf{k}}-\mathbf{k}$. The attention-logit perturbation for query $\mathbf{q}$ is exactly
\begin{equation}
\delta s=\frac{\mathbf{q}^{\top}\mathbf{e}}{\sqrt{d}}.
\end{equation}
Under the zero-mean uniform-noise approximation and uncorrelated channel errors, the conditional logit MSE is
\begin{equation}
\mathbb{E}[\delta s^2\mid\mathbf{q}]
=\frac{1}{12d}\sum_j q_j^2\Delta_{j,g}^{2}.
\label{eq:additional_logit_mse}
\end{equation}
Thus, expansion of the shared range enters the attention-logit error through the quantization step sizes, weighted by the query.
This relation connects the shared quantization range to attention-logit error under the stated noise assumptions. Downstream task accuracy also depends on attention patterns and nonlinear propagation; Table~\ref{tab:ablation_components} evaluates the coupled CR/OTS design through task accuracy.

\section{Quantitative Analysis of TNI-Induced Quantization Errors}
\label{app:quantitative_error_analysis}

\begin{table*}[h]
\centering
\resizebox{1\textwidth}{!}{
\begin{tabular}{@{}c|cccc|cccccccccc@{}}
\toprule
\multirow{4}{*}{\textbf{Bits}} & \multicolumn{4}{c|}{\textbf{w/ OTs vs. w/o OTs}} & \multicolumn{10}{c}{\textbf{Mixed-Modality vs. Single-Modality}} \\
 & \multicolumn{2}{c|}{\textbf{Per-Channel K}} & \multicolumn{2}{c|}{\textbf{Per-Token V}} & \multicolumn{5}{c|}{\textbf{Per-Channel K}} & \multicolumn{5}{c}{\textbf{Per-Token V}} \\ \cmidrule(l){2-15} 
 & \multirow{2}{*}{w/ OTs} & \multicolumn{1}{c|}{\multirow{2}{*}{w/o OTs}} & \multirow{2}{*}{w/ OTs} & \multirow{2}{*}{w/o OTs} & \multicolumn{3}{c|}{Mixed} & \multicolumn{2}{c|}{Single} & \multicolumn{3}{c|}{Mixed} & \multicolumn{2}{c}{Single} \\
 &  & \multicolumn{1}{c|}{} &  &  & overall & text & \multicolumn{1}{c|}{vision} & text & \multicolumn{1}{c|}{vision} & overall & text & \multicolumn{1}{c|}{vision} & text & vision \\ \midrule
4 & 0.23 & \multicolumn{1}{c|}{0.16} & 0.02 & 0.05 & 0.23 & 0.24 & \multicolumn{1}{c|}{0.24} & 0.29 & \multicolumn{1}{c|}{0.10} & 0.02 & 0.03 & \multicolumn{1}{c|}{0.02} & 0.03 & 0.02 \\
3 & 1.34 & \multicolumn{1}{c|}{0.78} & 0.11 & 0.14 & 1.34 & 1.50 & \multicolumn{1}{c|}{1.28} & 1.62 & \multicolumn{1}{c|}{0.62} & 0.11 & 0.15 & \multicolumn{1}{c|}{0.09} & 0.15 & 0.09 \\
2 & 5.92 & \multicolumn{1}{c|}{3.52} & 0.52 & 0.59 & 5.92 & 5.98 & \multicolumn{1}{c|}{5.87} & 6.17 & \multicolumn{1}{c|}{2.45} & 0.52 & 0.64 & \multicolumn{1}{c|}{0.40} & 0.65 & 0.41 \\ \bottomrule
\end{tabular}
}
\caption{Quantization error analysis. The left section compares quantization errors between groups with and without outlier tokens (OTs), while the right section compares errors between mixed-modality and single-modality groups. All values are scaled by a factor of 100.}
\label{tab:quantization_error}
\end{table*}

In Section~\ref{sec:revisiting} and Appendix~\ref{app:theoretical_derivations}, we presented empirical observations and theoretical analysis of TNI and its impact. Here, we conduct an empirical quantization error analysis under extreme KV cache compression to quantify these effects.
Our analysis centers on two TNI patterns:
\begin{itemize}
    \item \textbf{Impact of outlier tokens:} comparing quantization errors between groups that include low-norm outlier tokens and those from which these outliers have been removed.
    \item \textbf{Mixed-modality versus single-modality discrepancies:} comparing quantization errors between groups containing tokens from multiple modalities and those containing tokens from a single modality.
\end{itemize}
We conduct experiments on LLaVA-v1.5-7B. All experiments employ round-to-nearest (RTN) quantization with a group size of 32, and errors are measured using MSE. The results are summarized in Table~\ref{tab:quantization_error}. Below, we analyze the impact of TNI on both per-channel and per-token quantization.

\paragraph{Impact on Per-Channel Key Quantization}
As shown in Table~\ref{tab:quantization_error}, the presence of outlier tokens significantly amplifies per-channel Key quantization errors. Under 2-bit quantization, the error increases by approximately 68\% compared to groups from which these outliers are removed. Furthermore, mixed-modality groups exacerbate quantization errors relative to single-modality settings, leading to a 140\% increase for the visual component under the same 2-bit configuration.

\paragraph{Impact on Per-Token Value Quantization}
Value states lack channel-wise outliers, making per-token quantization the standard choice. Although TNI persists, per-token quantization confines norm variations to individual tokens and prevents cross-token interference. Consequently, the error amplification caused by TNI in per-channel schemes does not manifest under per-token quantization.

The above analysis confirms that TNI fundamentally undermines per-channel quantization while having negligible impact on per-token quantization, providing strong empirical validation for our assumption and theoretical derivations.

\subsection{Real-tensor calibration of the error model}
\label{app:calibration}
The approximation $\mathrm{MSE}_{\mathrm{pred}}(j,g)=\Delta_{j,g}^{2}/12$ isolates the effect of a shared per-channel quantization range. We evaluate post-RoPE, pre-CR/OTS Keys from Qwen3-8B Layer 18 on 16 sequences of 4K tokens from the Paul Graham Essays corpus, with group size 32 and the asymmetric round-to-nearest rule used in the paper. Correlations are computed per sequence and reported as mean $\pm$ standard deviation.

\begin{table}[htbp]
\centering
\caption{Predicted and measured Key reconstruction error on real tensors. Ratio denotes measured MSE divided by predicted MSE.}
\label{tab:real_tensor_calibration}
\small
\setlength{\tabcolsep}{4pt}
\begin{tabular}{@{}crrrcc@{}}
\toprule
Bits & Predicted MSE & Measured MSE & Ratio & Pearson (log MSE) & Spearman \\
\midrule
4 & 0.009105 & 0.009095 & 0.9989 & $0.9872\pm0.0002$ & $0.9792\pm0.0004$ \\
3 & 0.041810 & 0.041709 & 0.9976 & $0.9872\pm0.0002$ & $0.9792\pm0.0004$ \\
2 & 0.227634 & 0.226362 & 0.9944 & $0.9869\pm0.0002$ & $0.9785\pm0.0004$ \\
\bottomrule
\end{tabular}
\end{table}
The measured-to-predicted ratios in Table~\ref{tab:real_tensor_calibration} correspond to aggregate gaps of 0.11\%, 0.24\%, and 0.56\% at 4, 3, and 2 bits, respectively. The close errors and strong correlations support the shared-range approximation in this evaluated setting; they do not establish its validity across other models, layers, inputs, or quantizers.

\section{Detailed Analysis of Scaling-Induced Outlier Artifact}
\label{app:scaling_artifact}

As discussed in Section~\ref{sec:OScaR}, direct token-wise scaling, while effective in unifying token norms, introduces the \textit{Scaling-Induced Outlier Artifact} under per-channel quantization. We analyze this phenomenon through both mathematical formulation and concrete examples.

\paragraph{Mathematical Formulation}
Consider a normal token \(\mathbf{a} \in \mathbb{R}^d\) dominated by an outlier channel. Without loss of generality, assume channel \(d\) satisfies \(a_d \gg a_j\) for all \(j \neq d\). Let \(\mathbf{b} \in \mathbb{R}^d\) be a low-norm token with \(b_j \approx c\) (a small constant) for all \(j\). Their \(\ell_2\) norms are \(\|\mathbf{a}\|_2 \approx |a_d|\) and \(\|\mathbf{b}\|_2 \approx \sqrt{d} \cdot c\). Scaling both tokens to a target norm \(N\) yields scaling factors \(\alpha = N / \|\mathbf{b}\|_2\) and \(\beta = N / \|\mathbf{a}\|_2\). The resulting values are \(\mathbf{a}' = \beta \mathbf{a}\) and \(\mathbf{b}' = \alpha \mathbf{b}\).

The artifact condition captures \(\mathbf{b}'\) becoming an outlier in channels where \(\mathbf{a}'\) is small:
\begin{equation}
\alpha c \gg \beta \max_{j \neq d} a_j.
\end{equation}
Since \(\alpha / \beta = \|\mathbf{a}\|_2 / \|\mathbf{b}\|_2 \gg 1\) and \(c\) and \(\max_{j \neq d} a_j\) are of similar magnitude, the inequality holds, creating artificial outliers.

\paragraph{Direct Scaling: Numerical Demonstration}
A concrete example illustrates the failure. Set \(N = 1\), \(\mathbf{a} = [1, 1, 1, 100]\), and \(\mathbf{b} = [0.1, 0.1, 0.1, 0.1]\). Then \(\|\mathbf{a}\|_2 \approx 100.015\), \(\|\mathbf{b}\|_2 = 0.2\), giving \(\beta \approx 0.01\) and \(\alpha = 5\):
\begin{equation}
\mathbf{a}' \approx [0.01, 0.01, 0.01, 1.00], \qquad
\mathbf{b}' = [0.5, 0.5, 0.5, 0.5].
\end{equation}
Observe that \(\mathbf{b}'\) substantially exceeds \(\mathbf{a}'\) in channels 1--3 (0.5 vs. 0.01). In per-channel quantization, this expands the dynamic range of these channels from approximately 0.01 to 0.5, increasing the quantization step by a factor of 50 and severely degrading precision for normal tokens.

\subsection{Comparison with rotation and selective token protection}
\label{app:mechanism_positioning}
The distinction between balancing all token norms and protecting selected tokens is summarized in Table~\ref{tab:mechanism_comparison}.
\begin{table}[htbp]
\centering
\small
\setlength{\tabcolsep}{4pt}
\caption{Mechanistic distinction from representative alternatives. Selective high-precision token protection is distinct from the residual window used for online block quantization.}
\label{tab:mechanism_comparison}
\begin{tabular}{@{}>{\raggedright\arraybackslash}p{0.22\linewidth}>{\raggedright\arraybackslash}p{0.13\linewidth}>{\raggedright\arraybackslash}p{0.23\linewidth}>{\raggedright\arraybackslash}p{0.35\linewidth}@{}}
\toprule
Method & Key granularity & Channel treatment & Treatment of token disparity \\
\midrule
QuaRot~\citep{ashkboos2024quarot} & Per-token & Hadamard rotation & Separate token ranges; no all-token norm balancing \\
RotateKV~\citep{su2025rotatekv} & Per-token & Outlier-aware rotation & Selective high-precision sink-token protection \\
OTT~\citep{su2025accurate} & Per-channel & Channel-specific ranges & Tracing and high-precision protection of selected outlier tokens \\
OScaR & Per-channel & Canalized Rotation & Regular scaling of every Key token; no selective high-precision token pool \\
\bottomrule
\end{tabular}
\end{table}

Table~\ref{tab:mechanism_comparison} separates quantization granularity from token treatment. Per-token ranges avoid sharing a range across tokens, but still share a range across channels within each token. Selective protection targets a subset of exceptional tokens. OScaR instead conditions the distribution of every Key token before per-channel quantization, covering both sparse outliers and broader inter-modality norm gaps with one regular dataflow.

\section{Additional Visualizations of OScaR Processing Stages}
\label{app:oscar_vis}

\begin{figure}[t]
    \centering
    
    \begin{subfigure}[b]{0.245\textwidth}
        \centering
        \includegraphics[width=\textwidth]{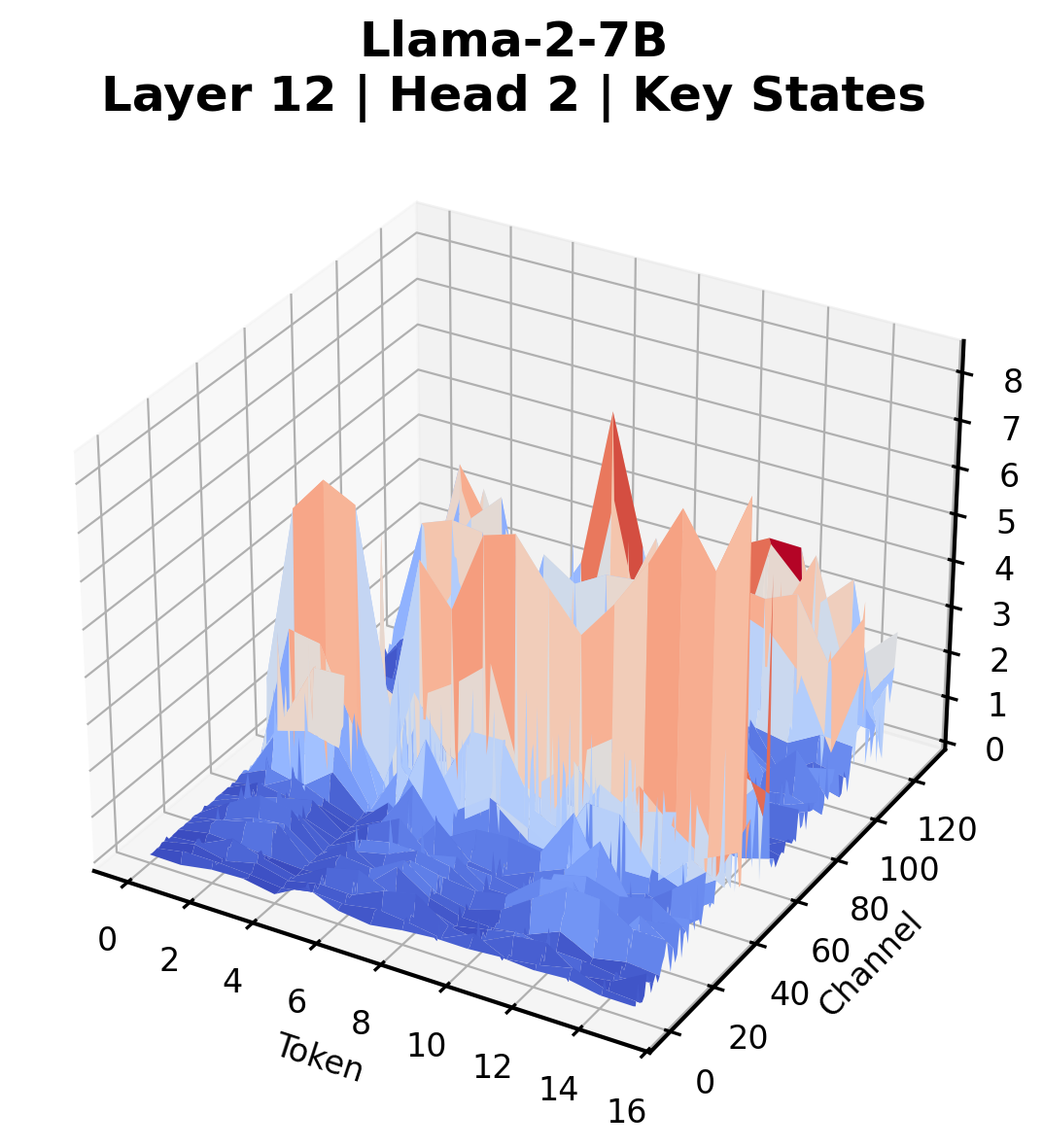}
    \end{subfigure}
    \begin{subfigure}[b]{0.245\textwidth}
        \centering
        \includegraphics[width=\textwidth]{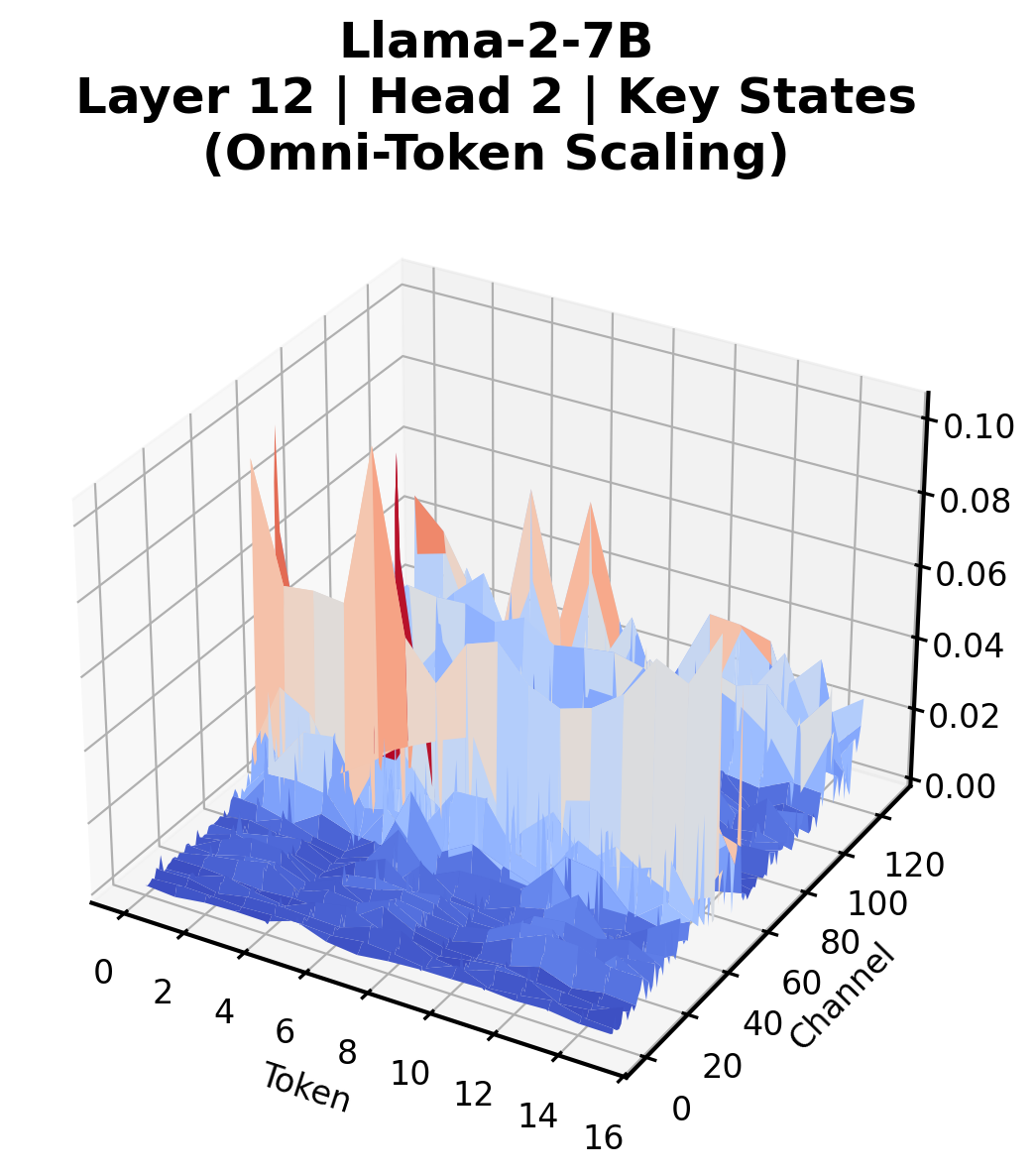}
    \end{subfigure}
    \begin{subfigure}[b]{0.245\textwidth}
        \centering
        \includegraphics[width=\textwidth]{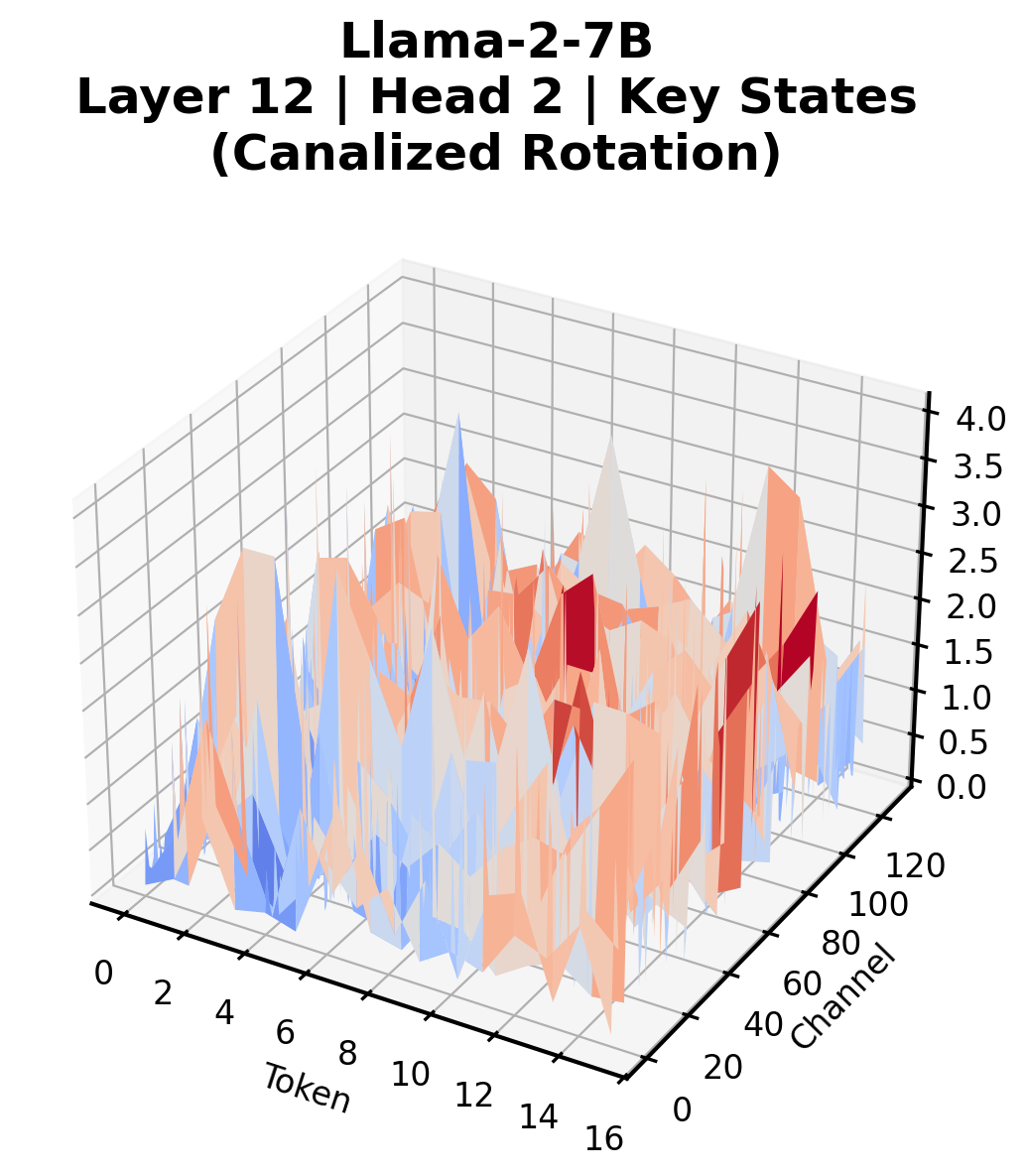}
    \end{subfigure}
    \begin{subfigure}[b]{0.245\textwidth}
        \centering
        \includegraphics[width=\textwidth]{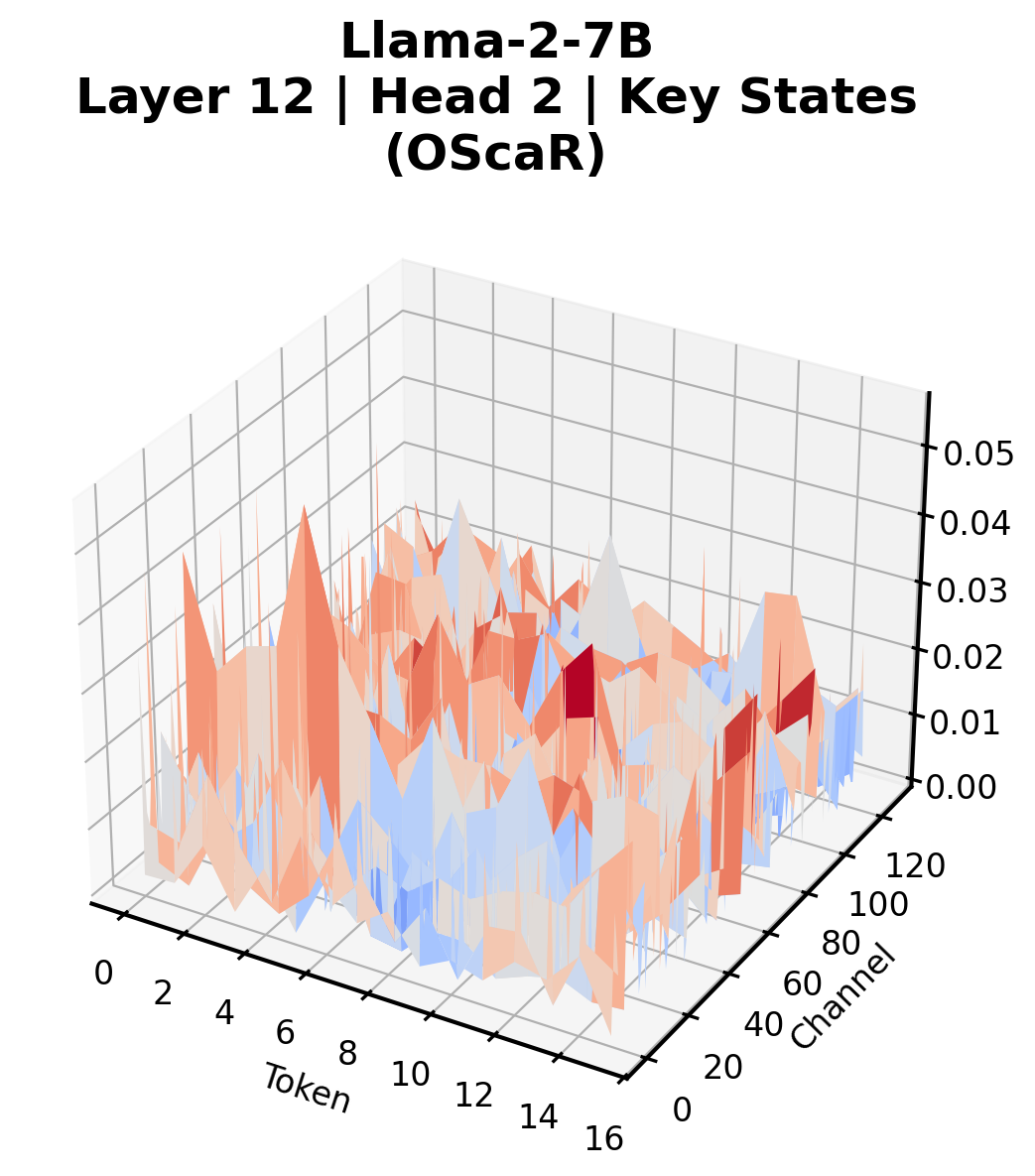}
    \end{subfigure}

    \begin{subfigure}[b]{0.245\textwidth}
        \centering
        \includegraphics[width=\textwidth]{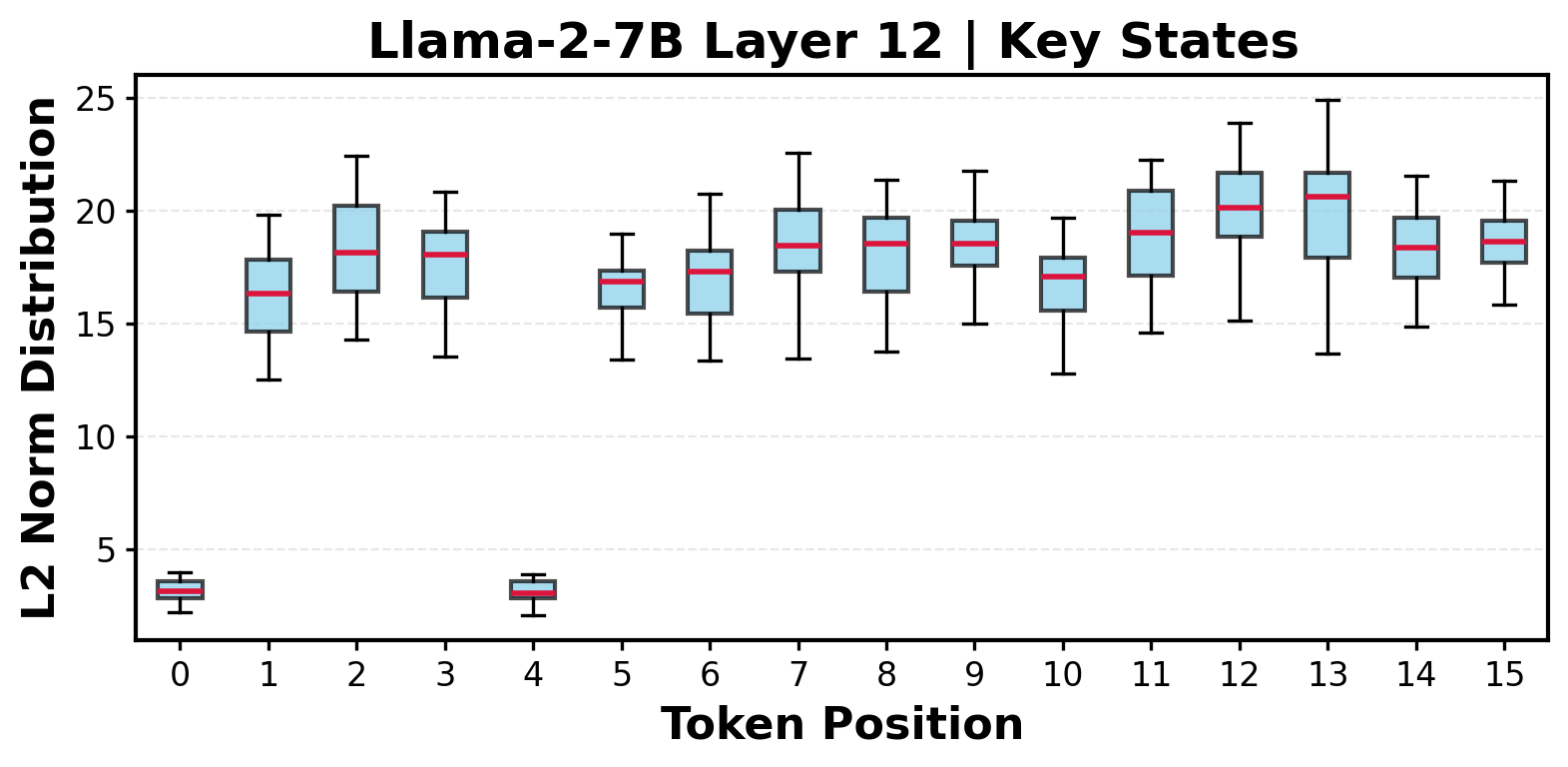}
    \end{subfigure}
    \begin{subfigure}[b]{0.245\textwidth}
        \centering
        \includegraphics[width=\textwidth]{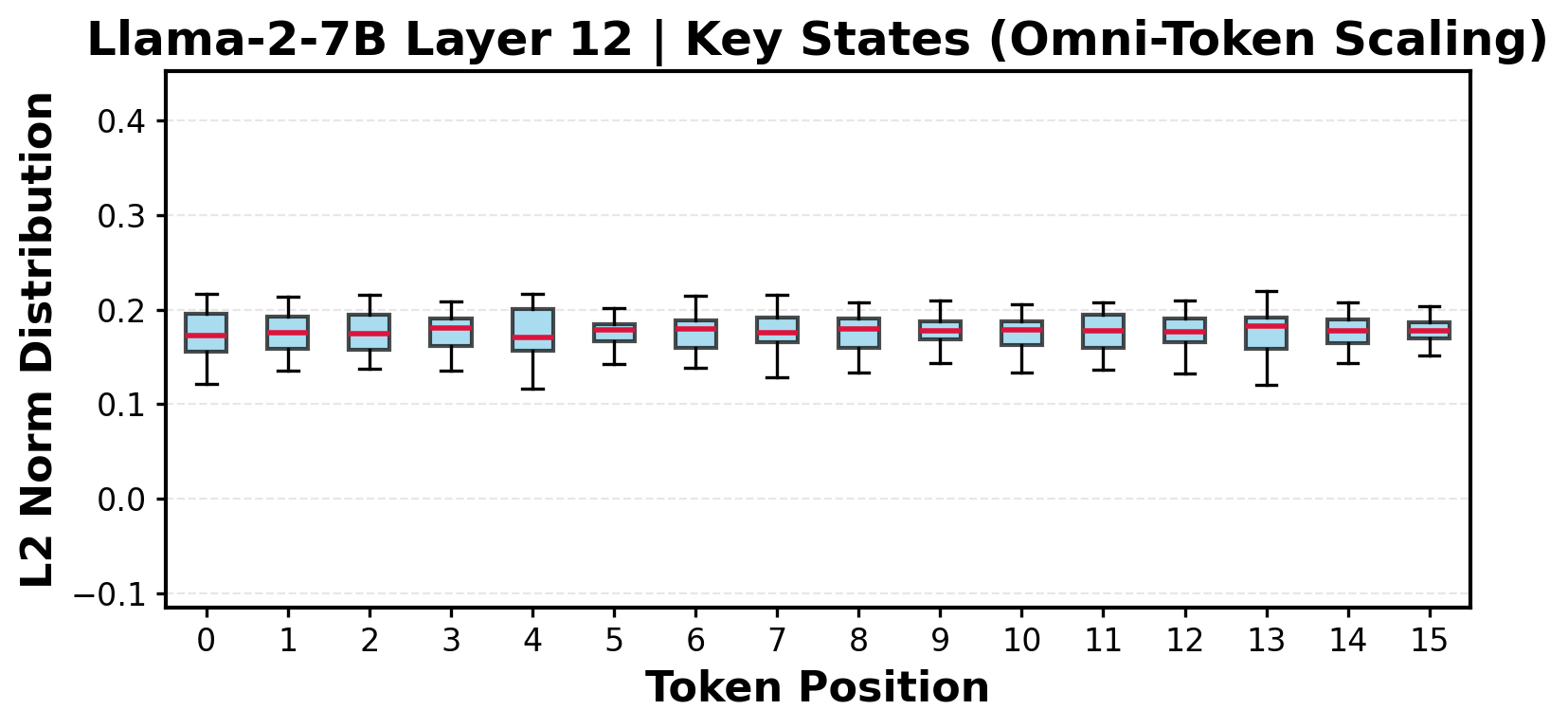}
    \end{subfigure}
    \begin{subfigure}[b]{0.245\textwidth}
        \centering
        \includegraphics[width=\textwidth]{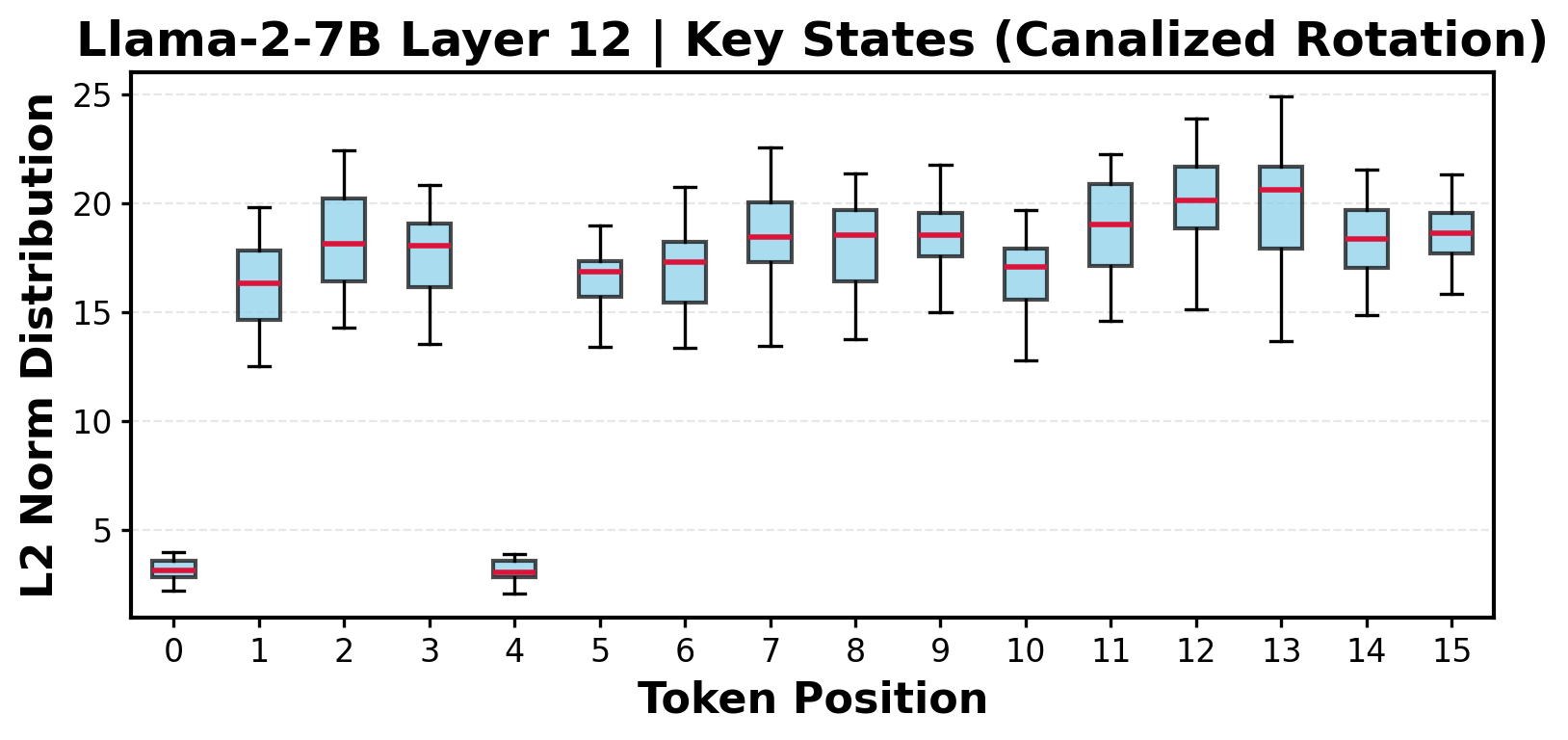}
    \end{subfigure}
    \begin{subfigure}[b]{0.245\textwidth}
        \centering
        \includegraphics[width=\textwidth]{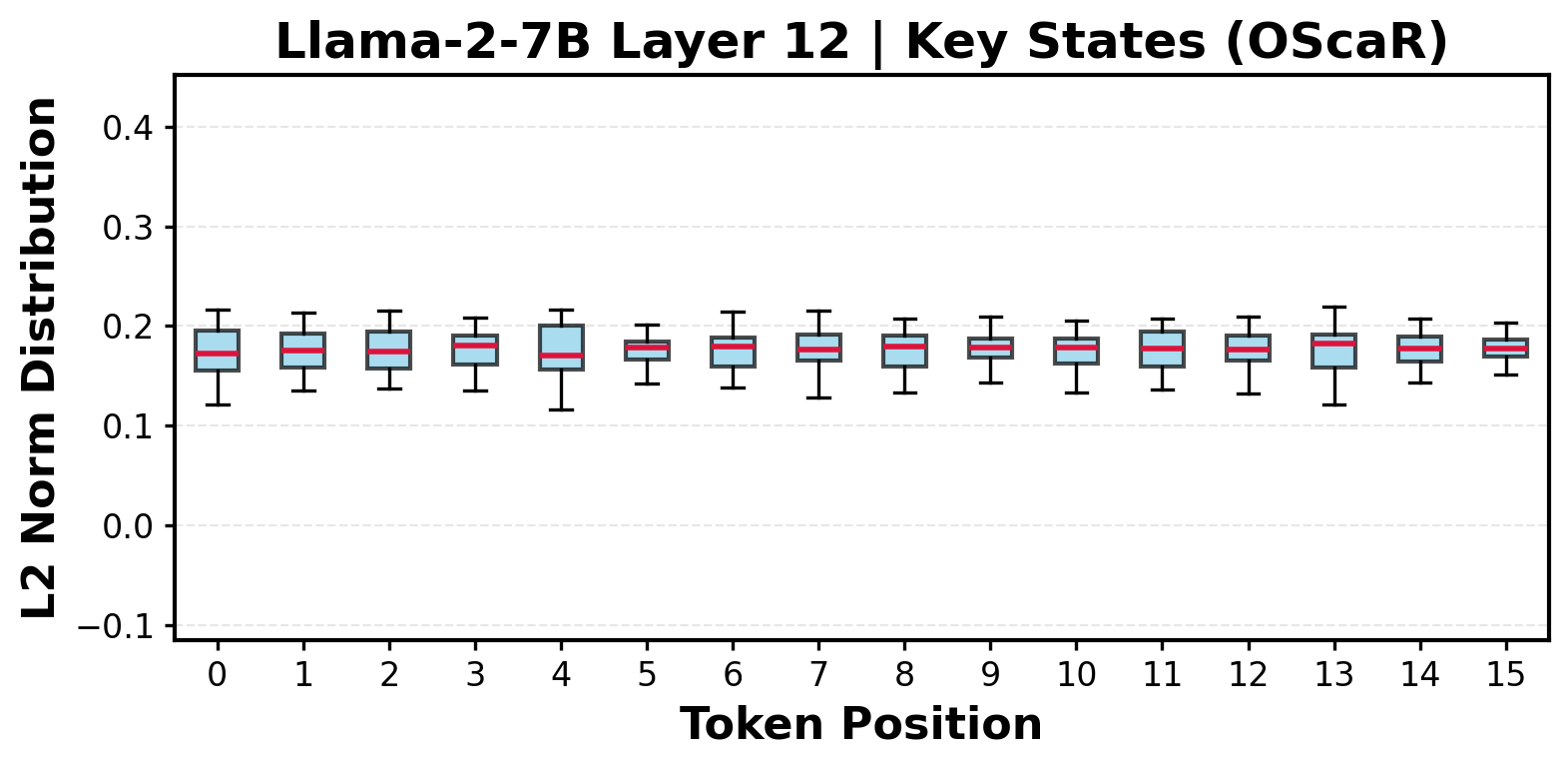}
    \end{subfigure}
    \caption{Key magnitude (top row) and L2 norm distribution (bottom row) across different processing stages. Results are shown for Llama-2-7B, Layer 12.}
    \label{fig:oscar_process_llama_layer_12}
\end{figure}

\begin{figure}[t]
    \centering
    
    \begin{subfigure}[b]{0.245\textwidth}
        \centering
        \includegraphics[width=\textwidth]{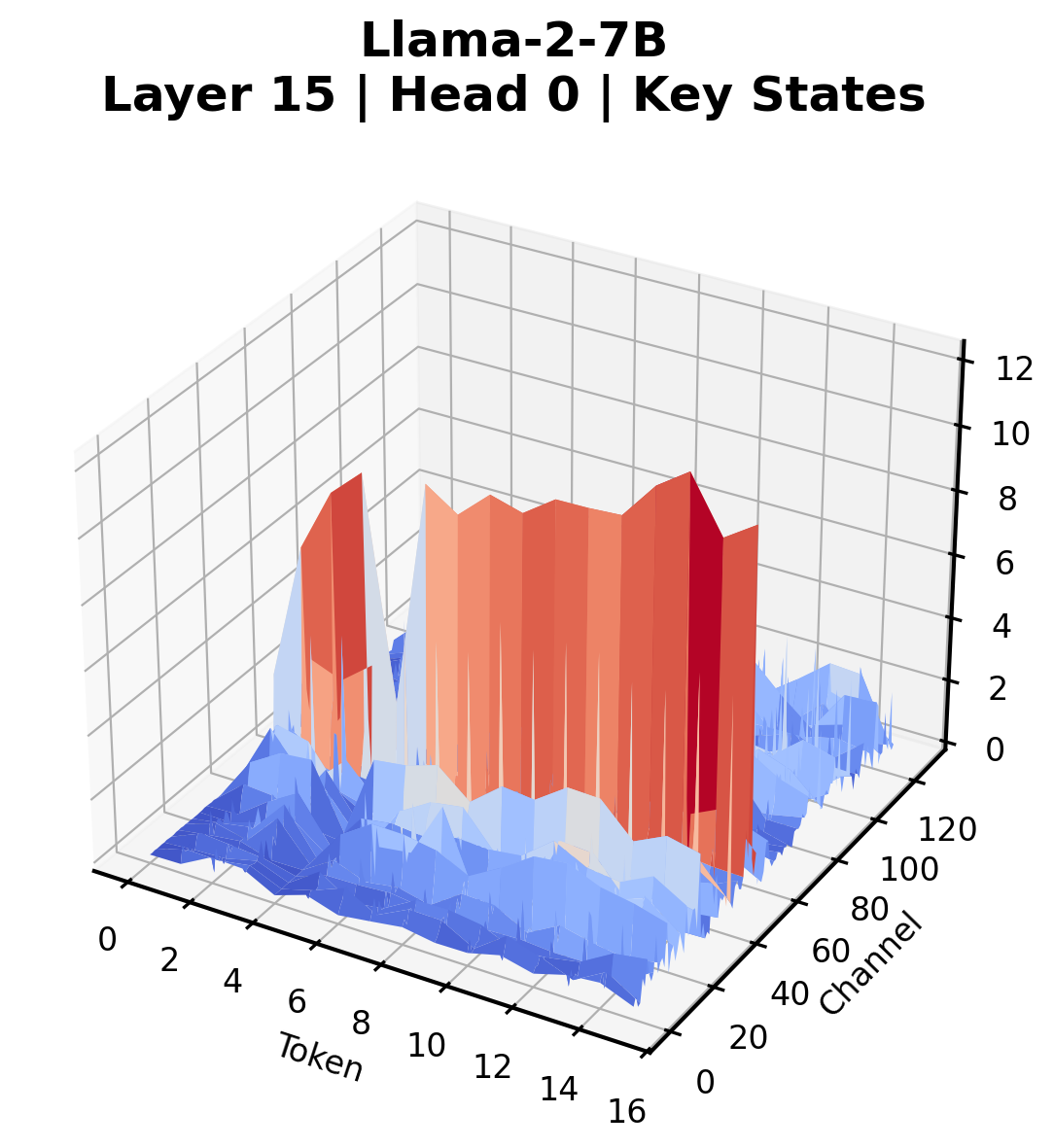}
    \end{subfigure}
    \begin{subfigure}[b]{0.245\textwidth}
        \centering
        \includegraphics[width=\textwidth]{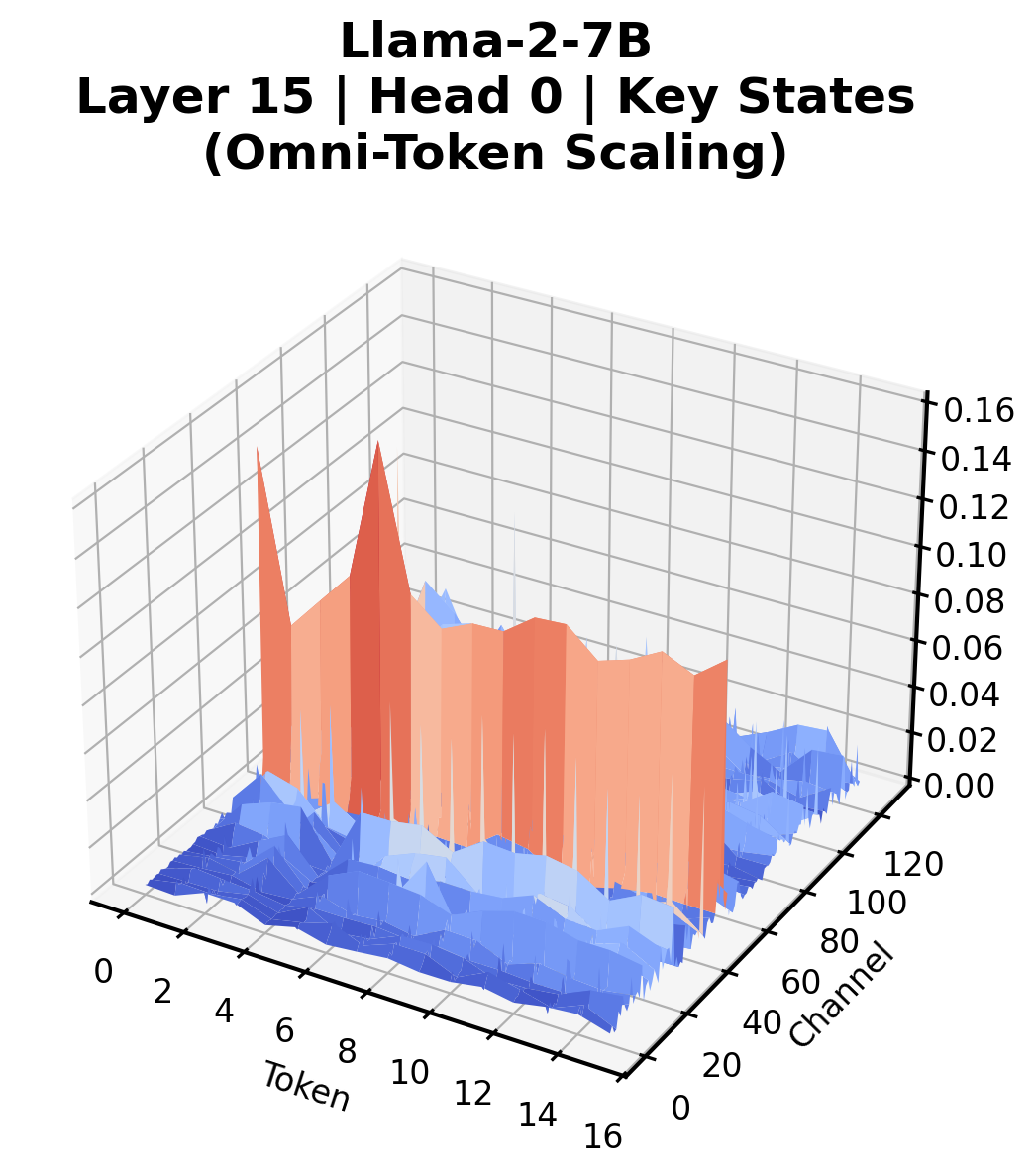}
    \end{subfigure}
    \begin{subfigure}[b]{0.245\textwidth}
        \centering
        \includegraphics[width=\textwidth]{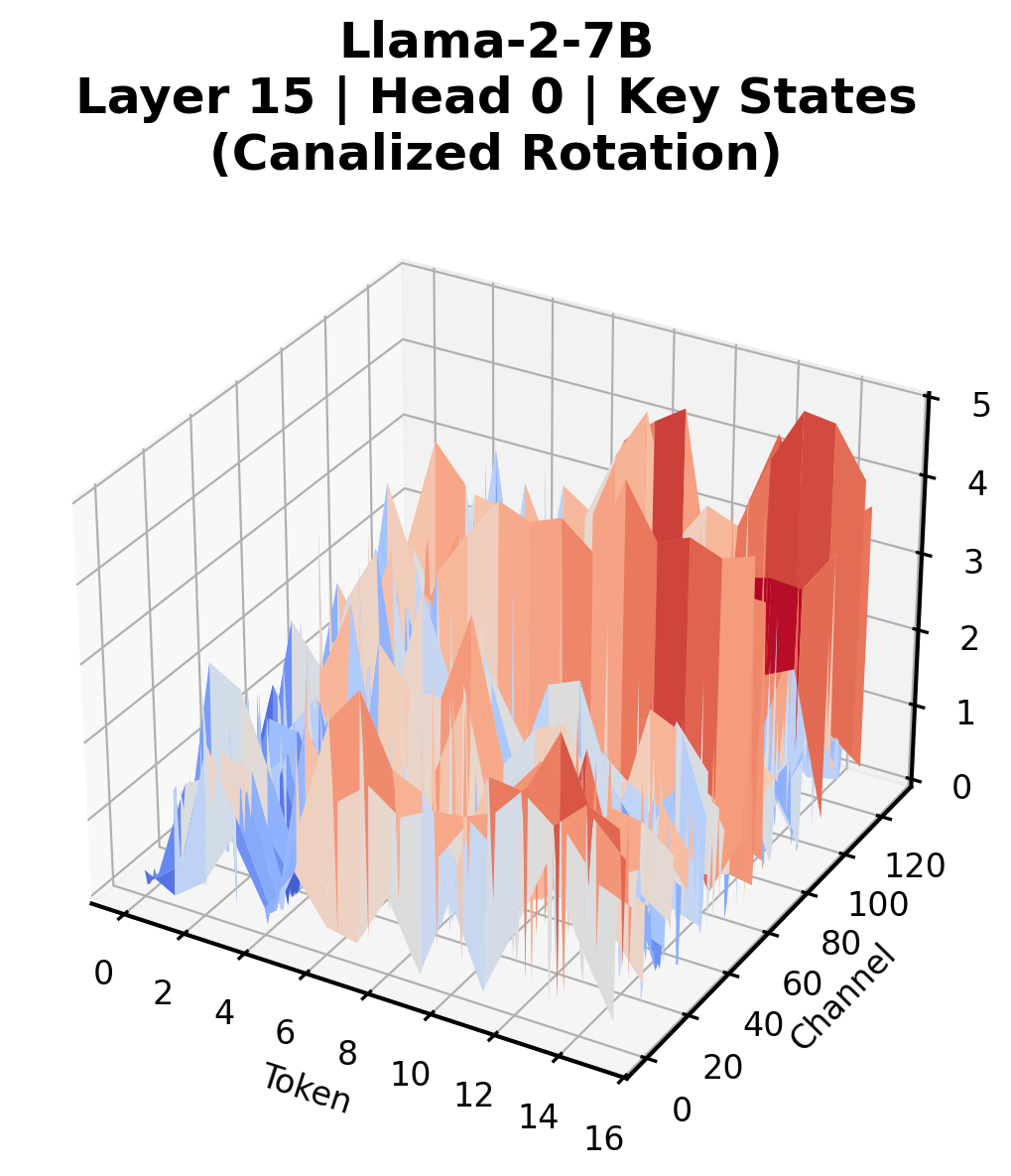}
    \end{subfigure}
    \begin{subfigure}[b]{0.245\textwidth}
        \centering
        \includegraphics[width=\textwidth]{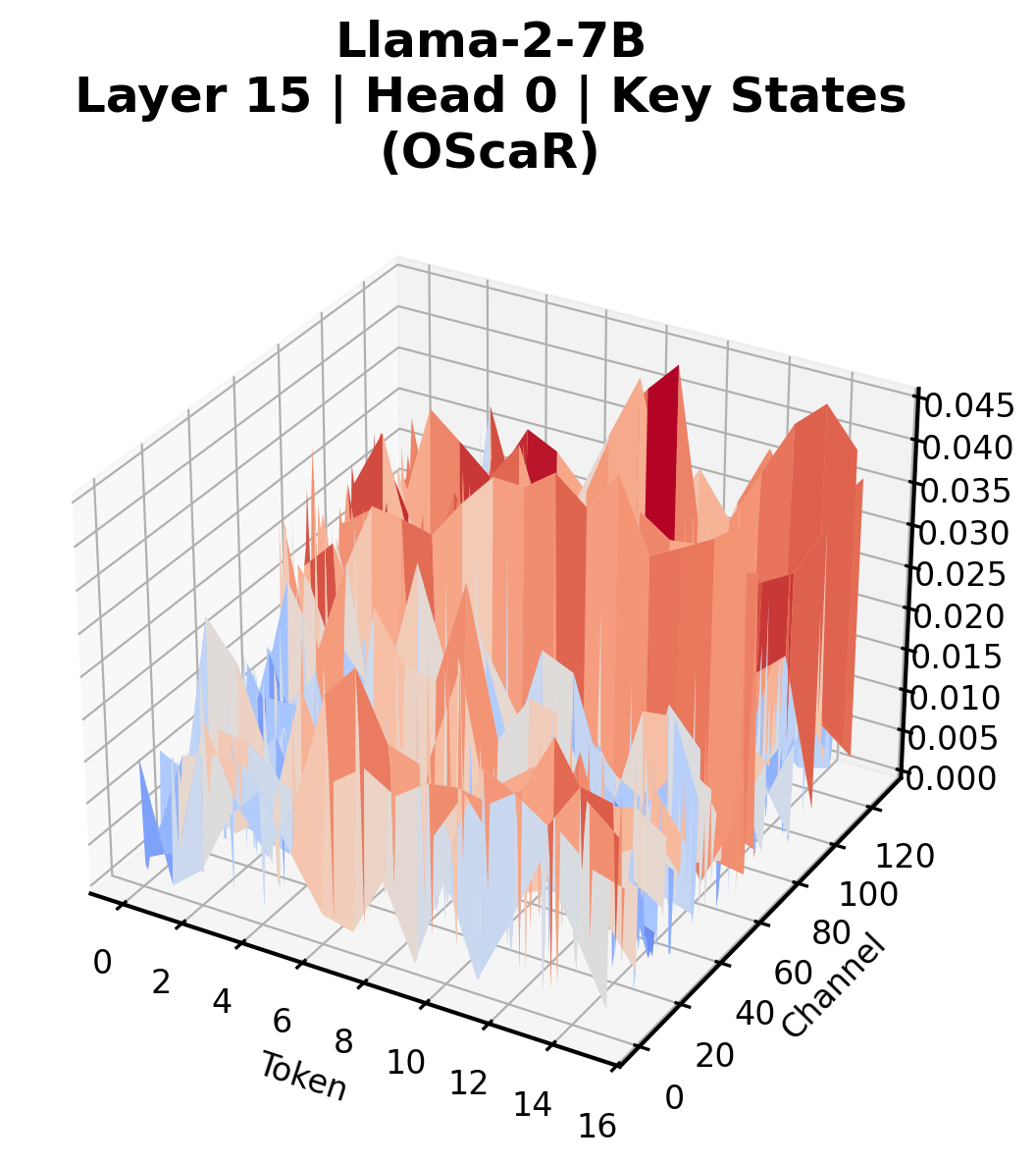}
    \end{subfigure}

    \begin{subfigure}[b]{0.245\textwidth}
        \centering
        \includegraphics[width=\textwidth]{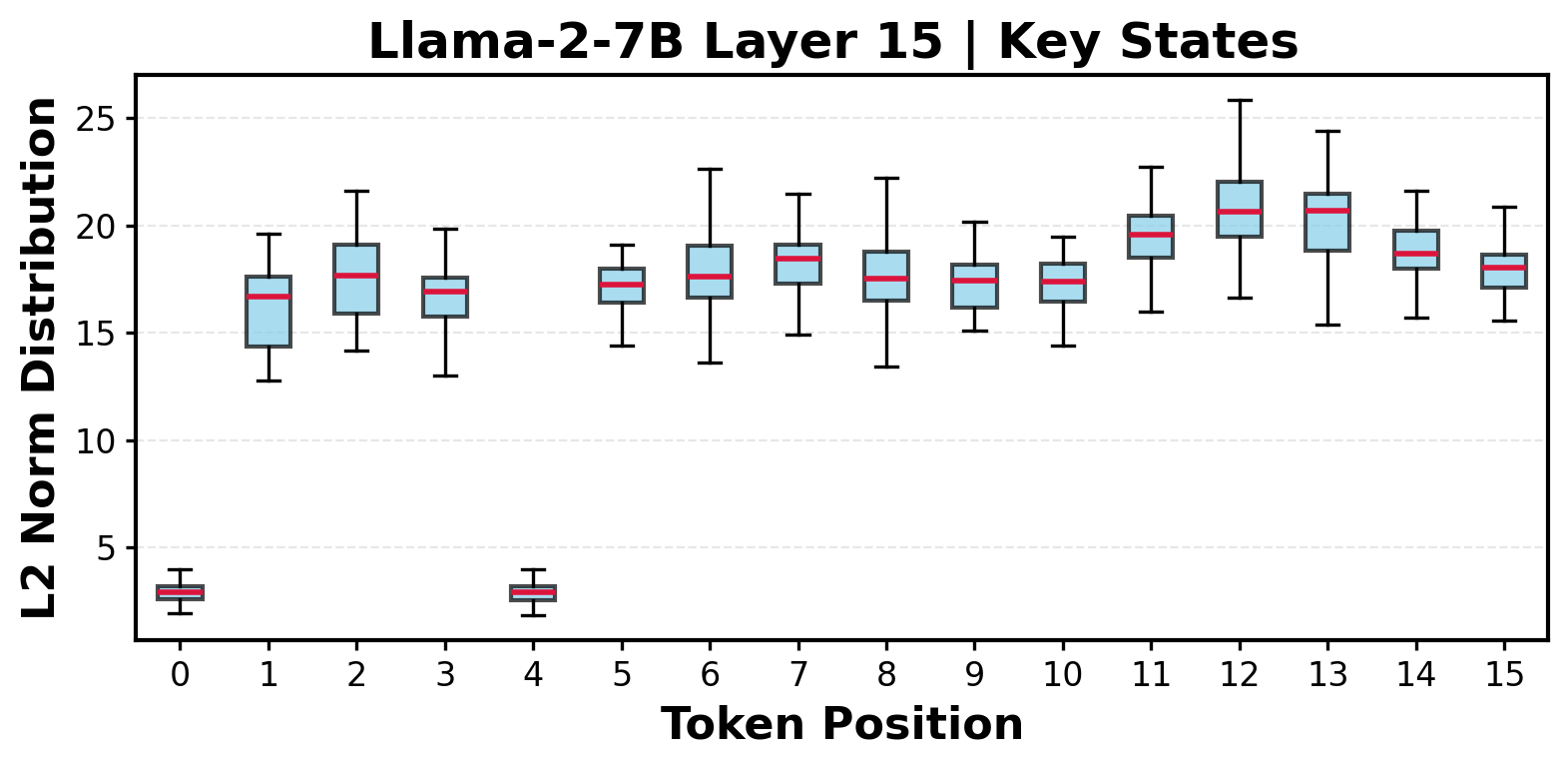}
    \end{subfigure}
    \begin{subfigure}[b]{0.245\textwidth}
        \centering
        \includegraphics[width=\textwidth]{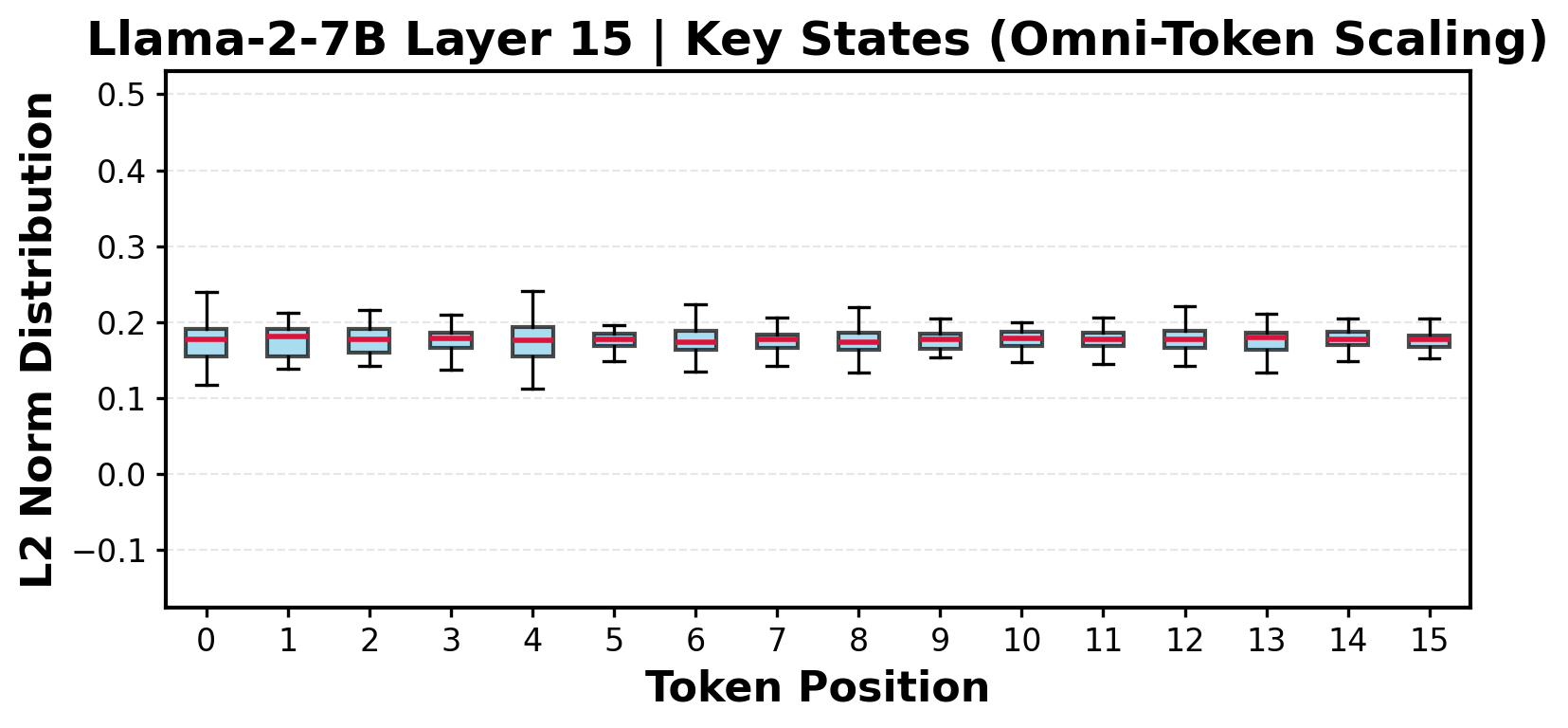}
    \end{subfigure}
    \begin{subfigure}[b]{0.245\textwidth}
        \centering
        \includegraphics[width=\textwidth]{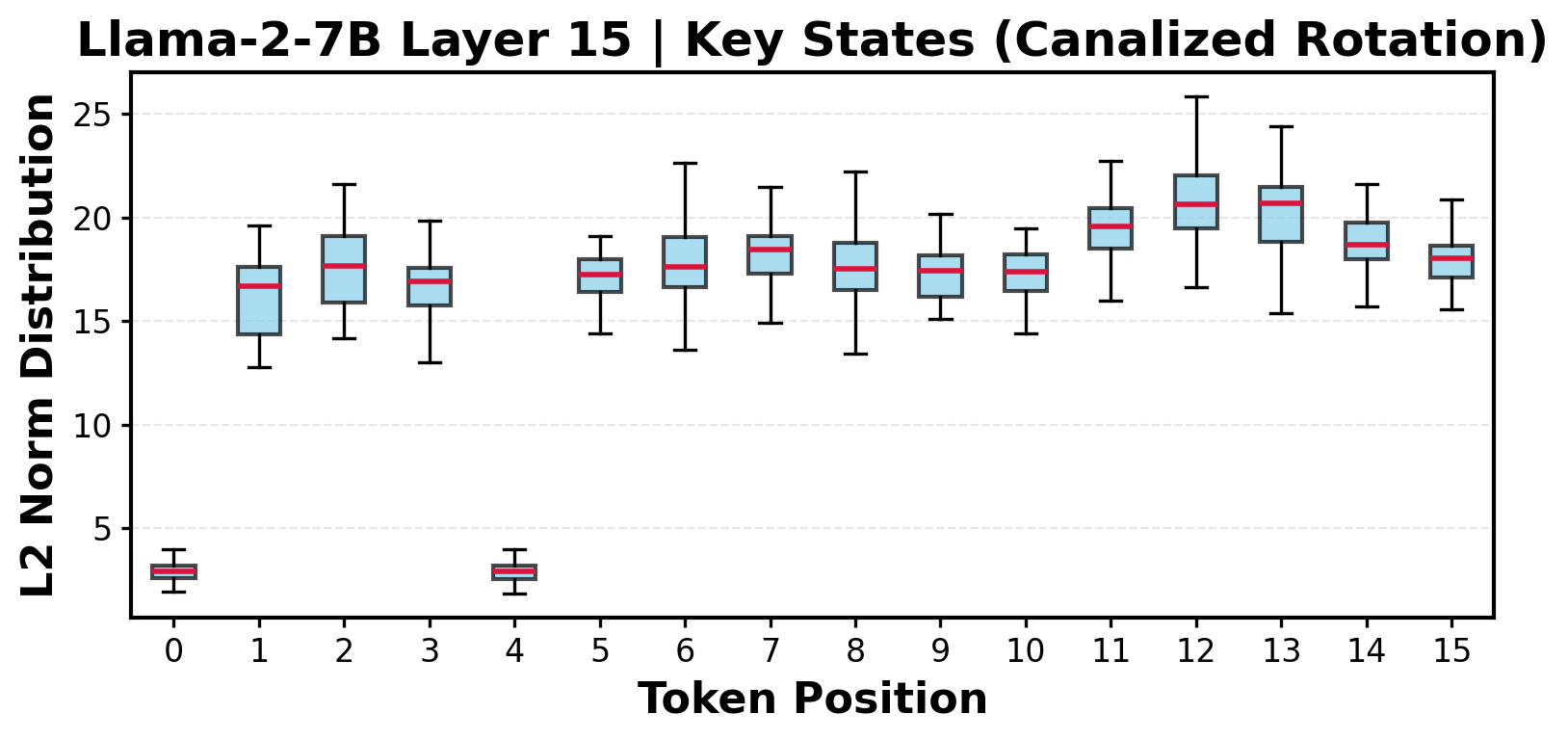}
    \end{subfigure}
    \begin{subfigure}[b]{0.245\textwidth}
        \centering
        \includegraphics[width=\textwidth]{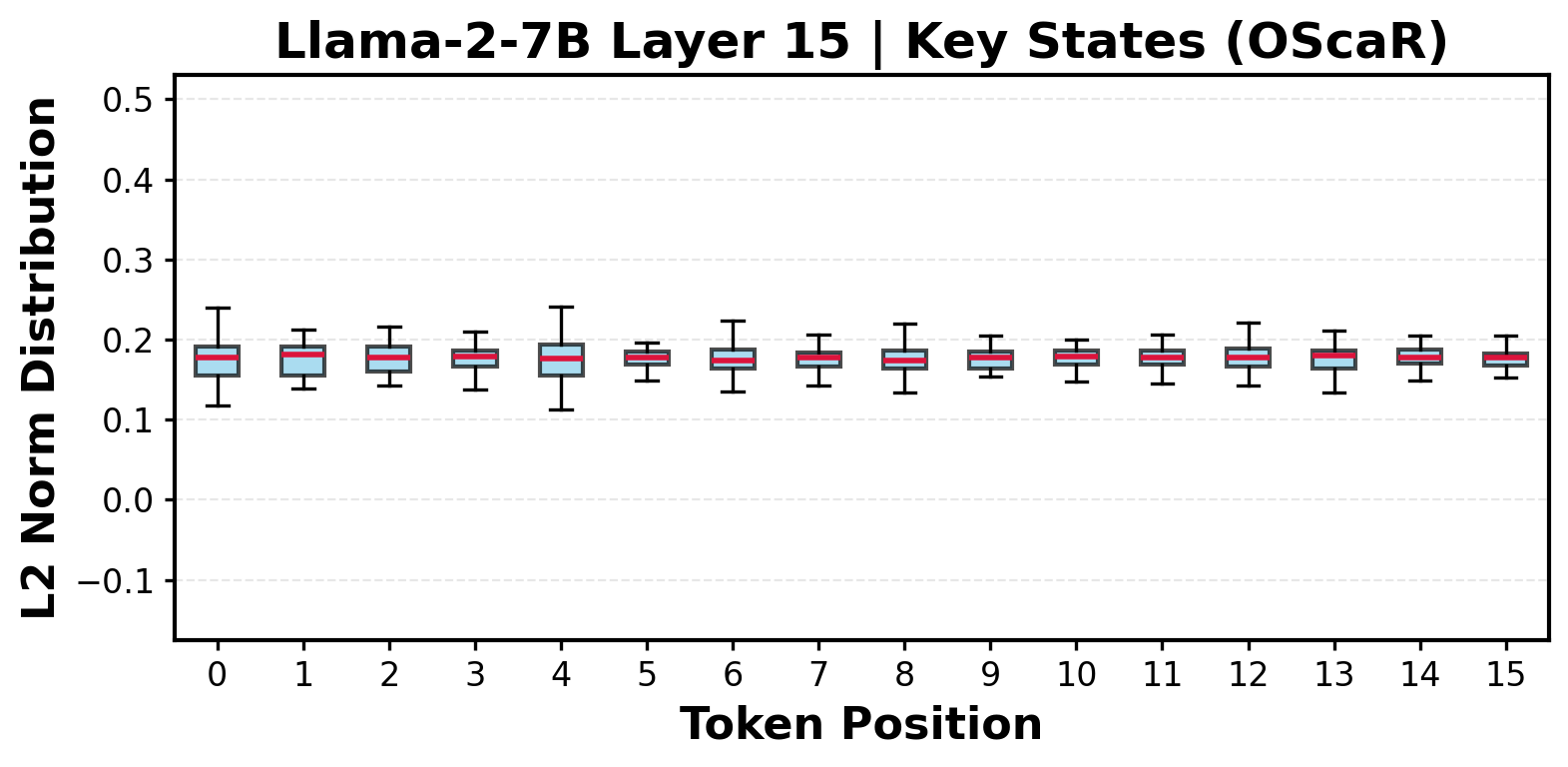}
    \end{subfigure}
    \caption{Key magnitude (top row) and L2 norm distribution (bottom row) across different processing stages. Results are shown for Llama-2-7B, Layer 18.}
    \label{fig:oscar_process_llama_layer_15}
\end{figure}

\begin{figure}[t]
    \centering
    
    \begin{subfigure}[b]{0.245\textwidth}
        \centering
        \includegraphics[width=\textwidth]{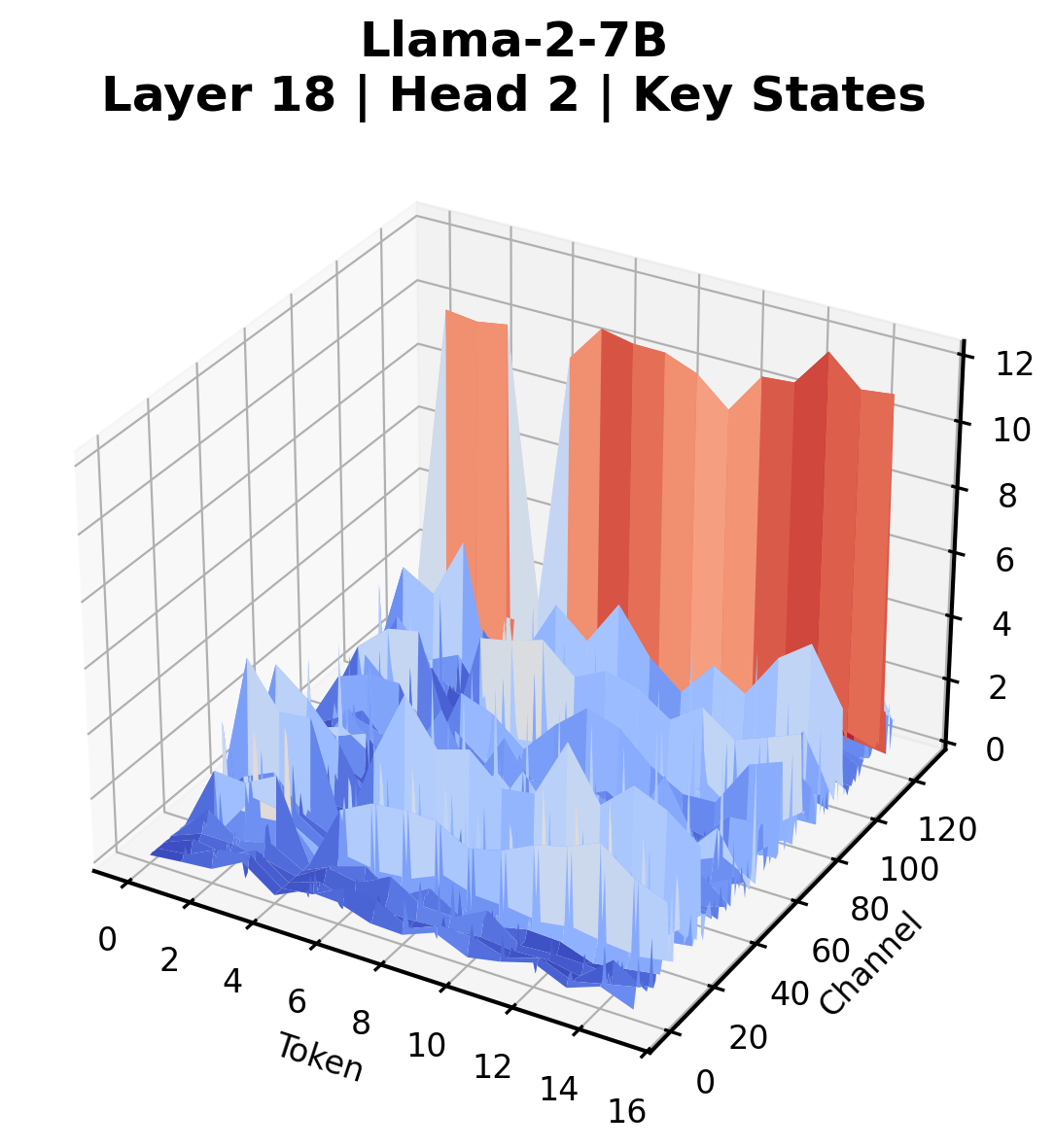}
        \label{fig:key_mesh_orig}
    \end{subfigure}
    \begin{subfigure}[b]{0.245\textwidth}
        \centering
        \includegraphics[width=\textwidth]{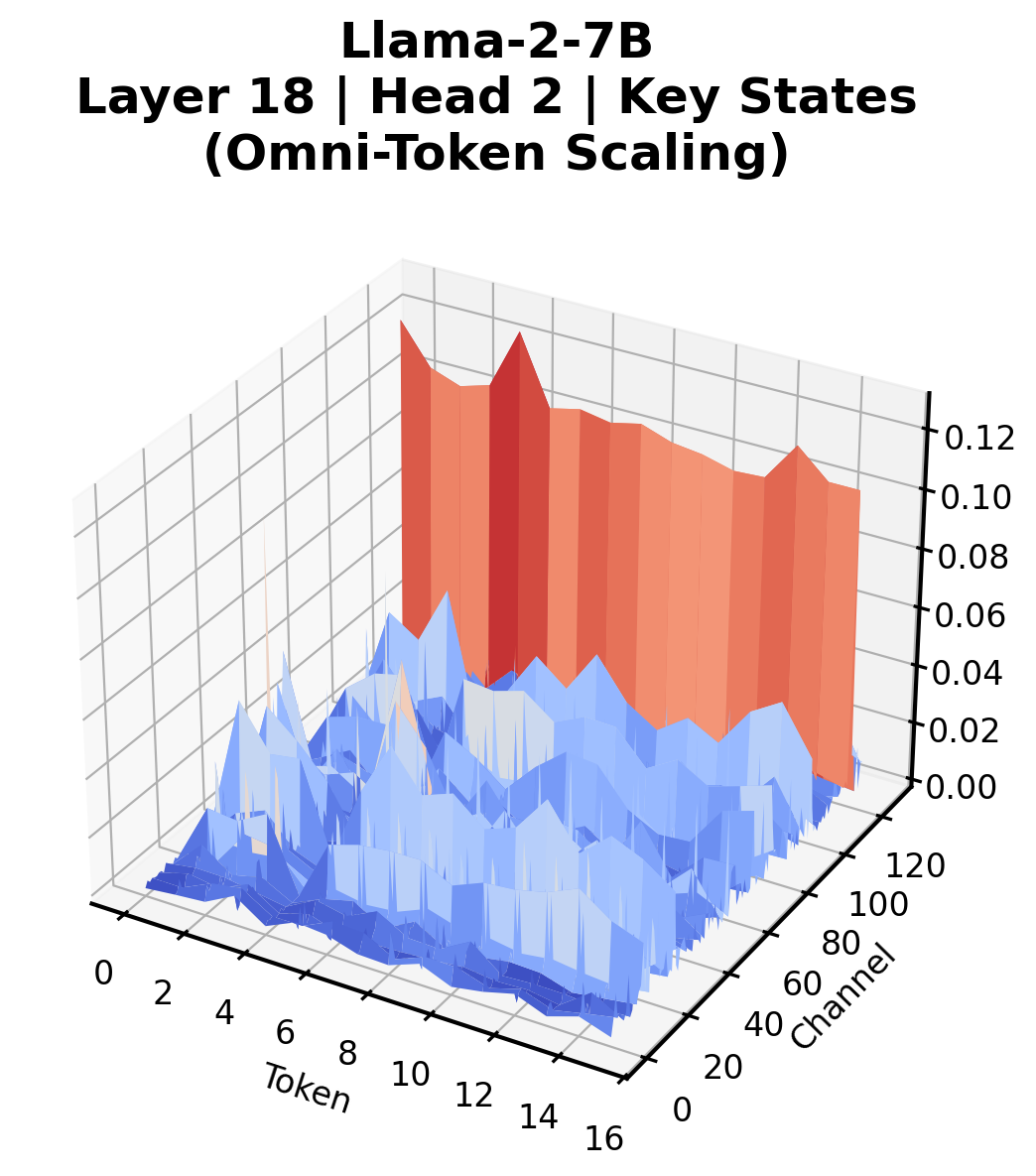}
    \end{subfigure}
    \begin{subfigure}[b]{0.245\textwidth}
        \centering
        \includegraphics[width=\textwidth]{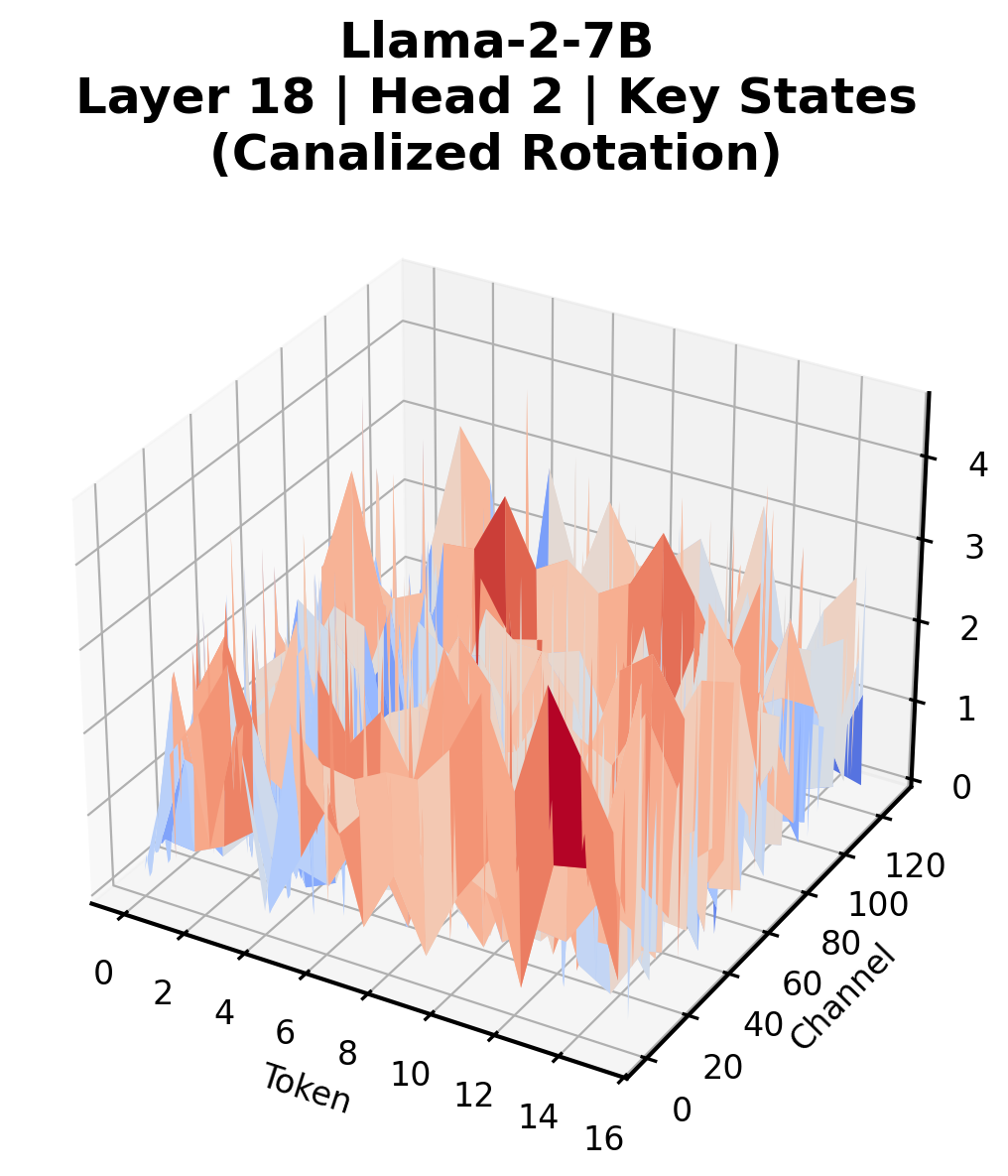}
    \end{subfigure}
    \begin{subfigure}[b]{0.245\textwidth}
        \centering
        \includegraphics[width=\textwidth]{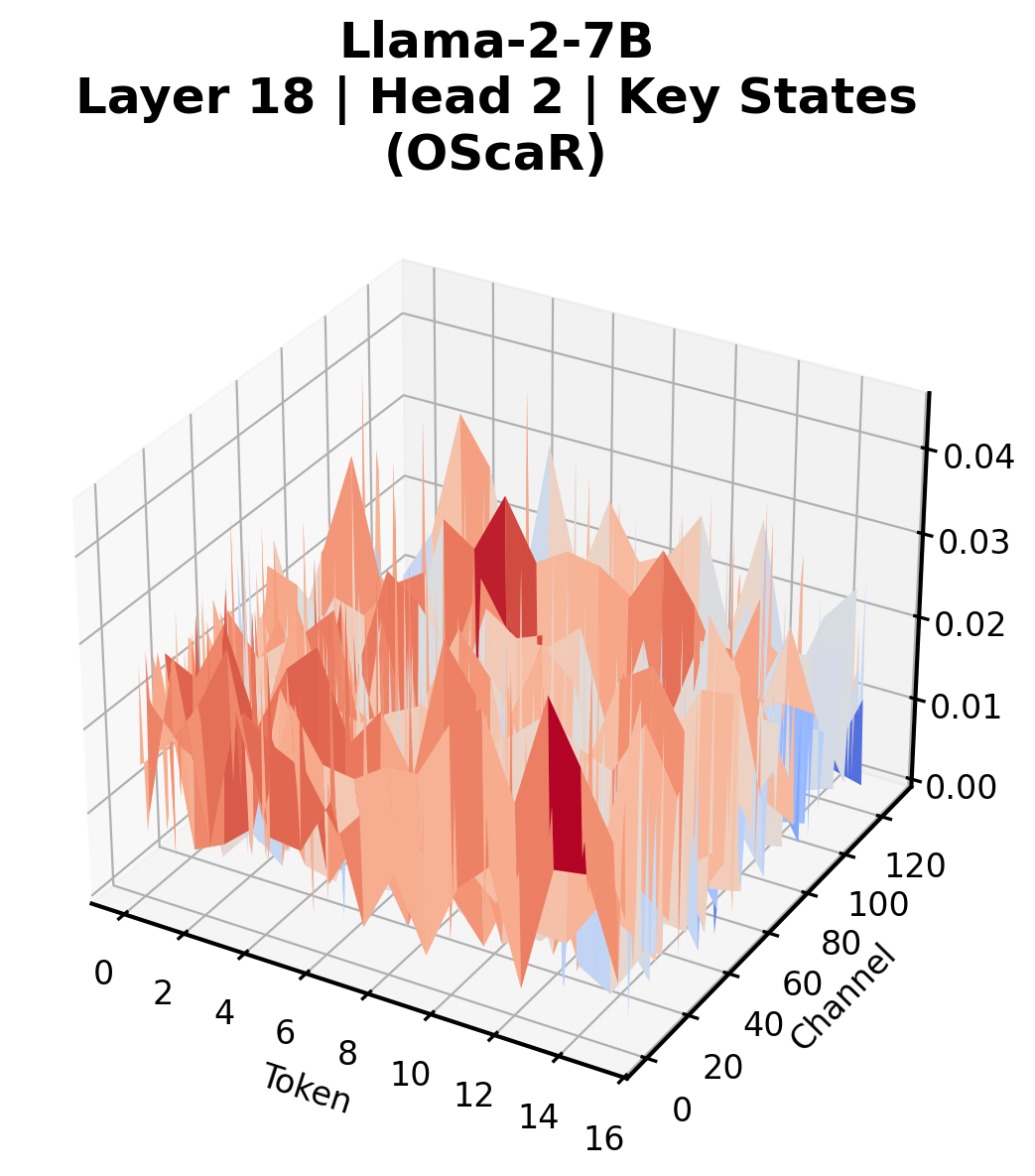}
    \end{subfigure}

    \begin{subfigure}[b]{0.245\textwidth}
        \centering
        \includegraphics[width=\textwidth]{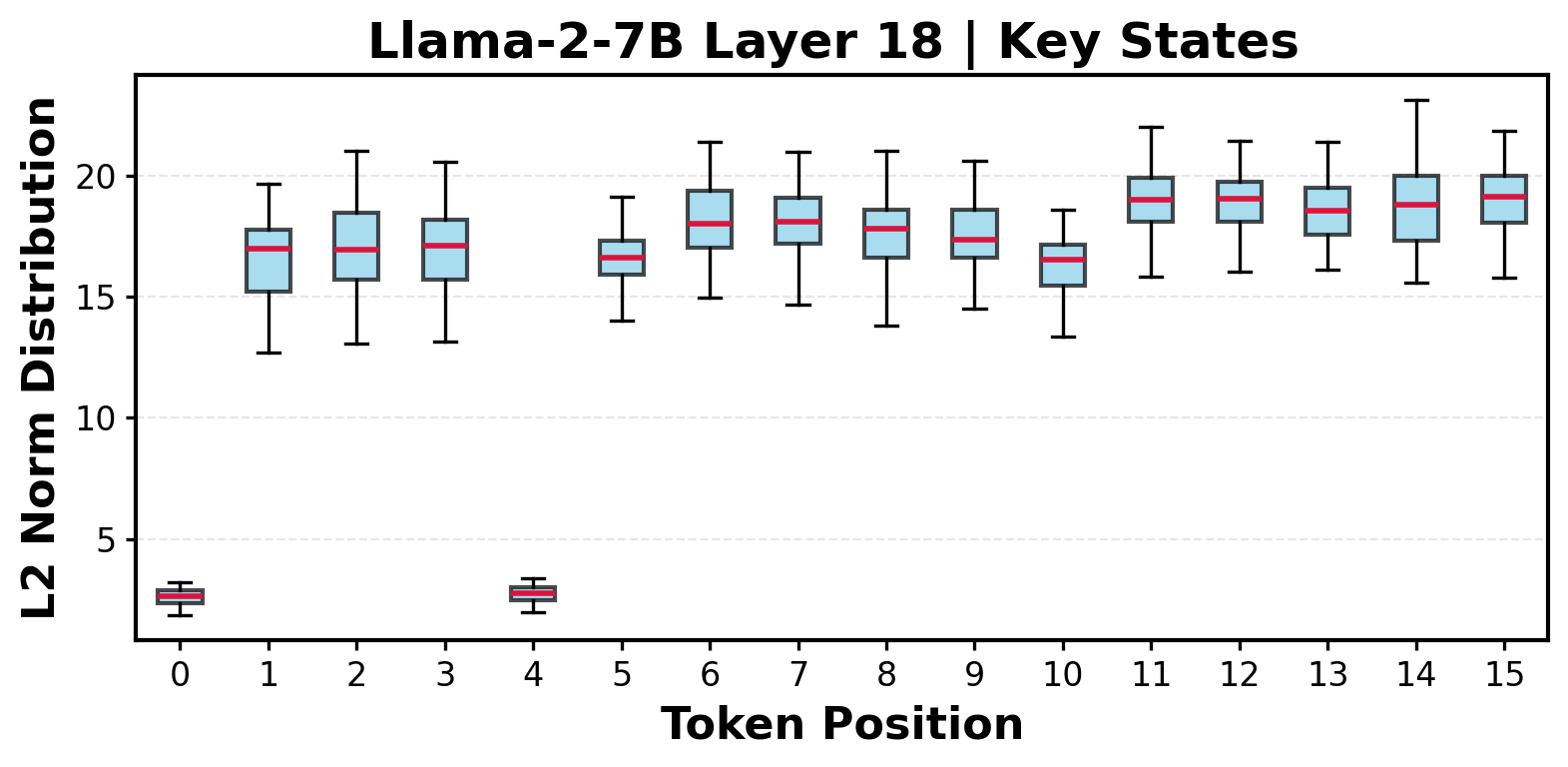}
    \end{subfigure}
    \begin{subfigure}[b]{0.245\textwidth}
        \centering
        \includegraphics[width=\textwidth]{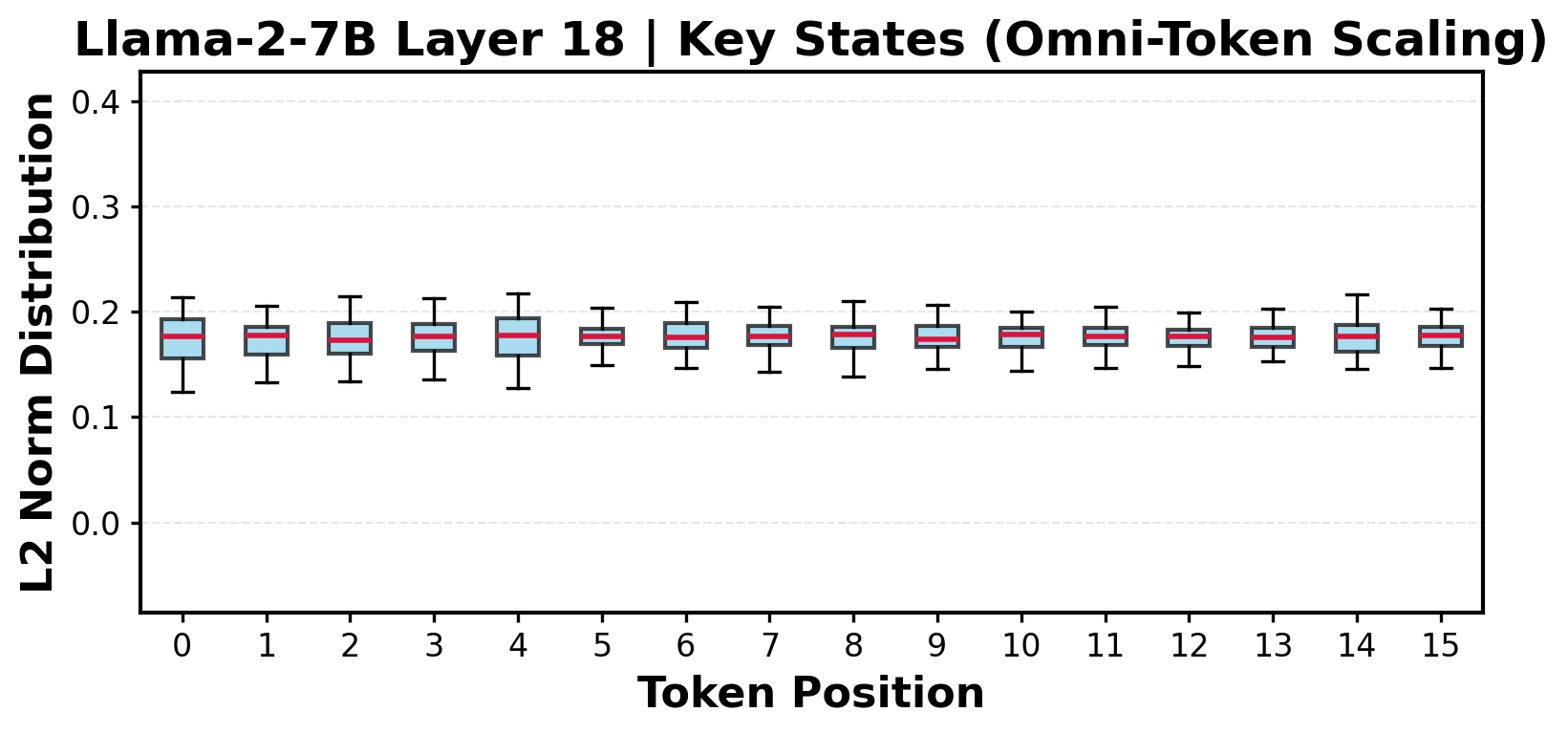}
    \end{subfigure}
    \begin{subfigure}[b]{0.245\textwidth}
        \centering
        \includegraphics[width=\textwidth]{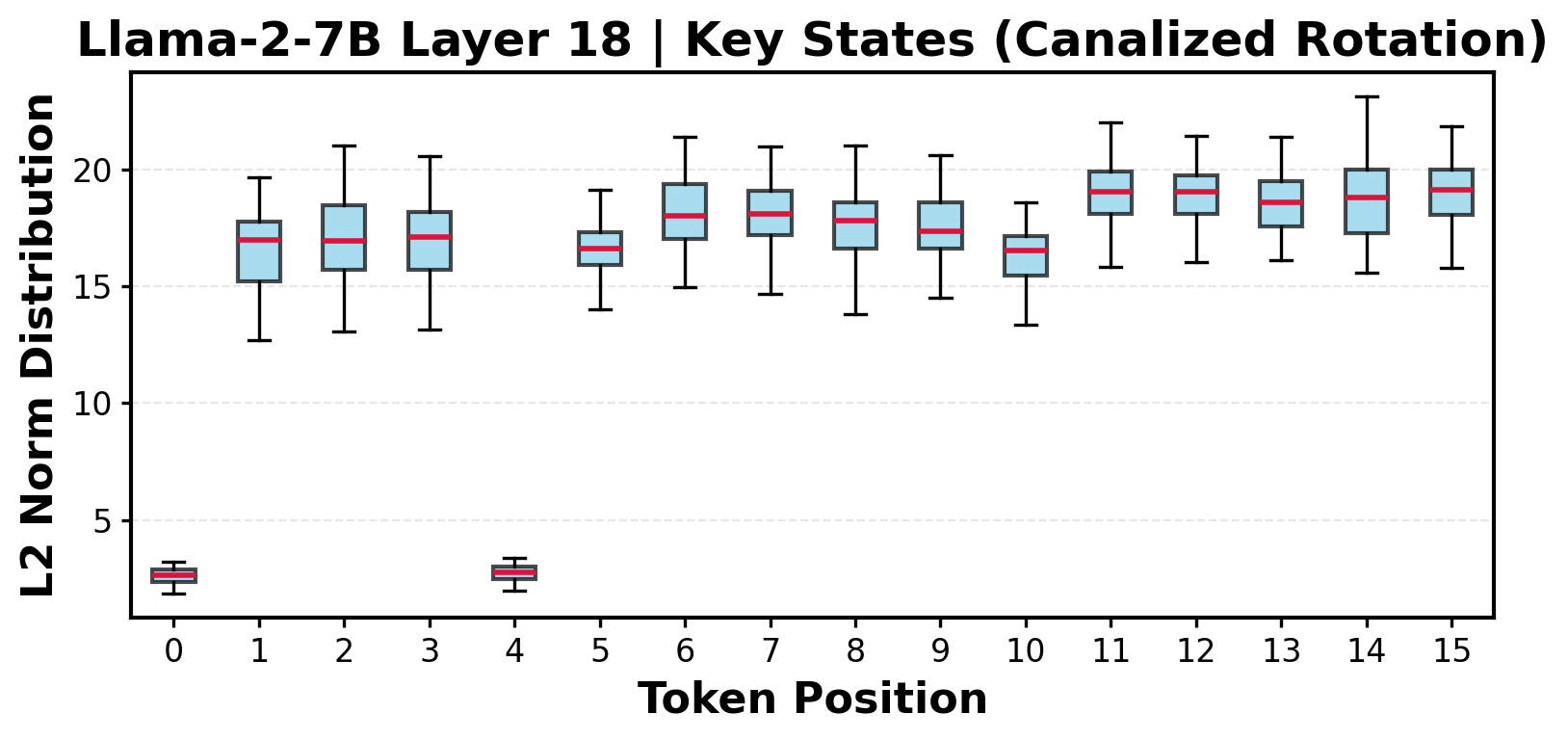}
    \end{subfigure}
    \begin{subfigure}[b]{0.245\textwidth}
        \centering
        \includegraphics[width=\textwidth]{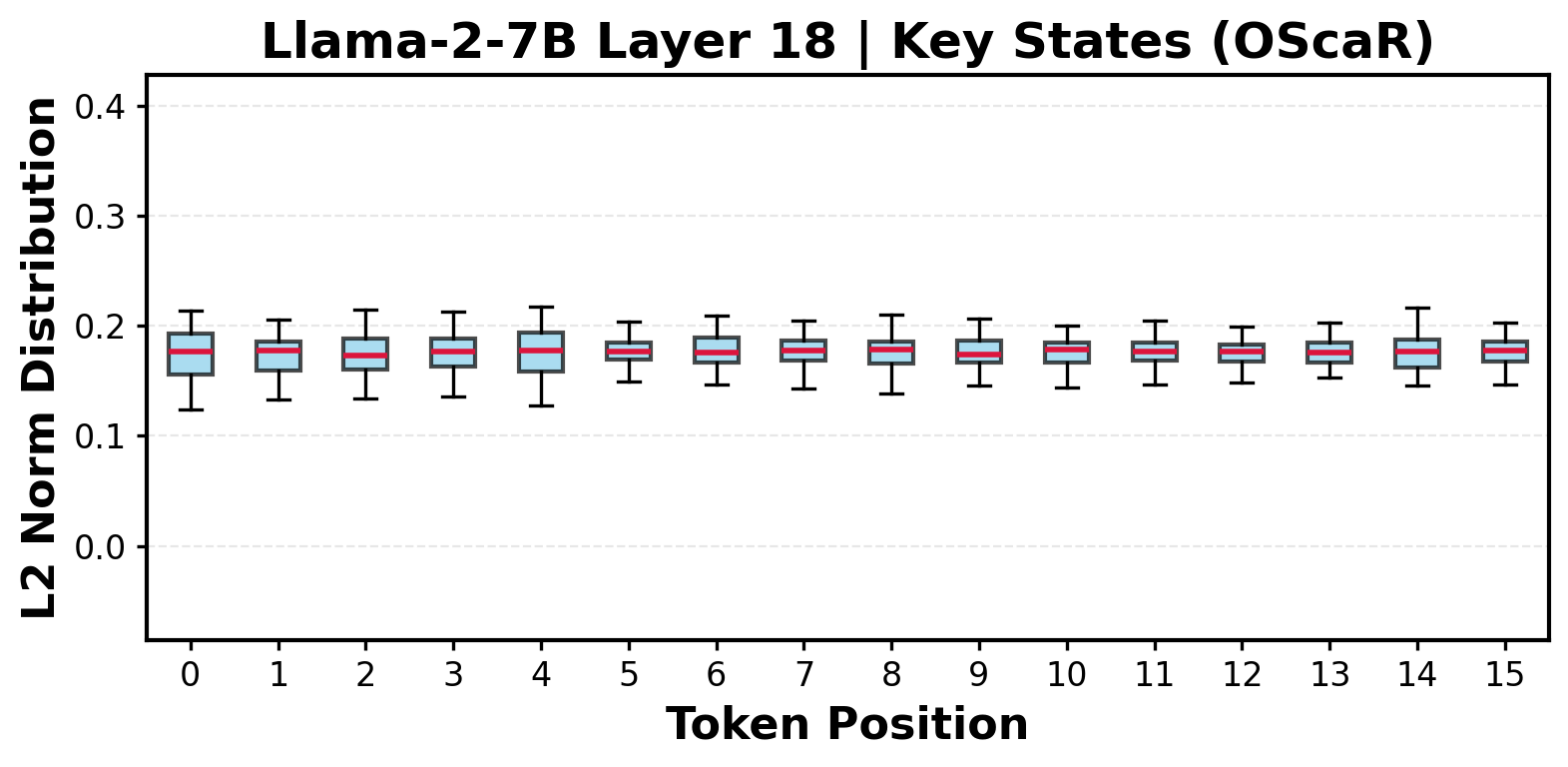}
    \end{subfigure}
    \caption{Key magnitude (top row) and L2 norm distribution (bottom row) across different processing stages. Results are shown for Llama-2-7B, Layer 18.}
    \label{fig:oscar_process_llama_layer_18}
\end{figure}

In this section, we provide additional visualizations of the OScaR processing stages. As shown in Figures~\ref{fig:oscar_process_llama_layer_12}, \ref{fig:oscar_process_llama_layer_15}, and \ref{fig:oscar_process_llama_layer_18}, direct scaling balances token norms but introduces the \textit{Scaling-Induced Outlier Artifact}. \textit{Canalized Rotation} alone fails to balance token norms. Only the complete OScaR framework successfully addresses TNI without incurring this artifact.
\section{Theoretical Complexity Analysis of KV Cache Quantization Methods}
\label{app:complexity}

\begin{table}[t]
\centering
\caption{Summary of raw operation counts (symbolic expressions).}
\label{tab:op_counts_summary}
\resizebox{0.85\textwidth}{!}{
\begin{tabular}{@{}lcccc@{}}
\toprule
\multirow{2}{*}{\textbf{Method}} & \multicolumn{2}{c}{\textbf{Prefill (per token)}} & \multicolumn{2}{c}{\textbf{Decode (per step)}} \\
\cmidrule(lr){2-3} \cmidrule(lr){4-5}
& \textbf{Arithmetic} & \textbf{Lookup} & \textbf{Arithmetic} & \textbf{Lookup} \\
\midrule
KIVI & $5d$ & $0$ & $5d + 2Ld$ & $0$ \\
QuaRot & $2d\log_2 h + 5d$ & $0$ & $2d\log_2 h + 5d + 2Ld$ & $0$ \\
OScaR (Ours) & $2d\log_2 h + 8d$ & $0$ & $2d\log_2 h + 8d + 3Ld$ & $0$ \\
TurboQuant (Original) & $6dh + 14.5d$ & $0$ & $6dh + 14.5d + Ld$ & $Ld$ \\
TurboQuant+ & $4dh + 5.25d$ & $0$ & $4dh + 5.25d + Ld$ & $Ld$ \\
\bottomrule
\end{tabular}
}
\end{table}

\begin{table}[t]
\centering
\caption{Effective cost in millions of units: total prefill for $L$ tokens and decode per step.}
\label{tab:complexity_numerical}
\resizebox{0.7\textwidth}{!}{
\begin{tabular}{@{}lcc@{}}
\toprule
\textbf{Method} & \textbf{Prefill total (M units)} & \textbf{Decode per step (M units)} \\
\midrule
KIVI & $204.8$ & $81.9$ \\
QuaRot & $778.2$ & $82.0$ \\
OScaR (Ours) & $901.1$ & $123.0$ \\
TurboQuant (Original) & $32{,}051.0$ & $249.0$ \\
TurboQuant+ & $21{,}187.0$ & $247.9$ \\
\bottomrule
\end{tabular}
}
\end{table}

We analyze the theoretical computational overhead introduced by five KV cache quantization methods. Symbolic costs are per token during prefill and per step during decode; the numerical table multiplies prefill costs by the prompt length $L$. Each arithmetic operation (multiplication, addition, comparison, rounding, square root) is counted as one operation. Table lookups (LUT) are accounted for separately. The analysis excludes attention computation to isolate the overhead incurred by quantization and its auxiliary transformations. We focus specifically on key processing, as this constitutes the core distinguishing factor among the compared methods.

We emphasize that while every effort has been made to ensure a fair and theoretically sound analysis, the resulting estimates may deviate from actual hardware efficiency due to factors such as memory bandwidth, parallelism, kernel launch overhead, and operator fusion. Therefore, this theoretical analysis should be interpreted as providing comparative insight rather than precise performance predictions. A comprehensive empirical evaluation is presented in Section~\ref{sec:efficiency}.

\subsection{Symbolic Operation Counts}

We derive symbolic expressions for the computational cost of each method in terms of hidden dimension $d$, head dimension $h$, and sequence length $L$. The resulting symbolic operation counts are summarized in Table~\ref{tab:op_counts_summary}.

\paragraph{KIVI~\citep{liu2024kivi}.}
KIVI employs per-channel uniform quantization without query or key pre-transformation. During prefill, processing one token involves scanning all $d$ dimensions to update per-channel min/max values, requiring $2d$ comparisons. Then, quantizing each element with $q = \text{round}((x - zero)/scale)$ incurs one subtraction, one division, and one rounding, totaling $3d$ operations. The prefill cost is therefore $5d$ arithmetic operations with no lookups. During decode, quantizing the newly generated key adds another $5d$ operations. Dequantizing the historical key cache of length $L$ requires one subtraction and one multiplication per element, amounting to $2Ld$ operations. The total decode cost per step is $5d + 2Ld$ arithmetic operations.

\paragraph{QuaRot~\citep{ashkboos2024quarot}.}
QuaRot applies an online Walsh-Hadamard Transform (WHT) to both query and key before quantization. For head dimension $h$, the WHT's butterfly structure requires $d\log_2 h$ additions per transformed tensor. During prefill, each token's query and key each undergo a WHT, contributing $2d\log_2 h$ additions, followed by the same $5d$ quantization as KIVI. The prefill cost is $2d\log_2 h + 5d$ arithmetic operations. During decode, the new key requires a WHT ($d\log_2 h$ adds) and quantization ($5d$ ops); the query requires a WHT ($d\log_2 h$ adds); and dequantizing the historical cache costs $2Ld$ ops. The total decode cost per step is $2d\log_2 h + 5d + 2Ld$ arithmetic operations.

\paragraph{OScaR (Ours).}
OScaR builds upon per-channel key quantization with two innovations: Canalized Rotation via online WHT and Omni-Token Scaling via token-wise L2 normalization. Canalized Rotation applies WHT to both query and key, requiring $d\log_2 h$ additions per tensor. Omni-Token Scaling normalizes each key token to unit length, involving three stages: (i) computing sum of squares across $d$ dimensions ($2d$ operations), (ii) square root (1 operation), and (iii) scaling each element by the reciprocal of the norm ($d$ multiplications). The total normalization cost is approximately $3d$ operations.
During prefill, each token incurs Canalized Rotation on Q and K ($2d\log_2 h$ additions), Omni-Token Scaling on K ($3d$ ops), and per-channel quantization ($5d$ ops). Total prefill cost: $2d\log_2 h + 8d$ arithmetic operations, no lookups.
During decode, the newly generated token requires: key rotation ($d\log_2 h$ adds), key scaling ($3d$ ops), key quantization ($5d$ ops), and query rotation ($d\log_2 h$ adds). Dequantizing the historical key cache costs $2Ld$ ops, and restoring key magnitudes via inverse scaling adds $Ld$ multiplications. Total decode cost: $2d\log_2 h + 8d + 3Ld$ arithmetic operations, no lookups.

\paragraph{TurboQuant~\citep{zandieh2025turboquant} (Original).}
TurboQuant uses a 2.5-bit mixed-precision scheme (32 outlier channels at 3-bit, 96 normal channels at 2-bit) and a dense Haar QR rotation matrix $\Pi \in \mathbb{R}^{h \times h}$. Rotating a $d$-dimensional vector with this dense matrix requires $2dh$ arithmetic operations. During prefill, each token incurs: dense rotation of Q ($2dh$) and K ($2dh$), L2 normalization ($3d$), brute-force Lloyd-Max quantization ($7.5d$), residual handling ($d$), QJL projection ($2dh + d$), and residual norm ($2d$). Total: $6dh + 14.5d$ arithmetic operations. During decode, the new key requires dense rotation and quantization ($4dh + 14.5d$ ops); the query requires dense rotation ($2dh$ ops). Dequantizing the historical cache costs $Ld$ arithmetic ops and $Ld$ lookups. Total decode cost: $6dh + 14.5d + Ld$ arithmetic ops and $Ld$ lookups.

\paragraph{TurboQuant+~\citep{turney2026turboquantplus}.}
This variant retains the 2.5-bit mixed-precision scheme, removes QJL, and replaces brute-force Lloyd-Max search with binary search, reducing quantization cost from $7.5d$ to $2.25d$ comparisons. No residual handling. During prefill: dense rotation of Q ($2dh$) and K ($2dh$), L2 normalization ($3d$), binary search quantization ($2.25d$). Total: $4dh + 5.25d$ arithmetic ops, no lookups. During decode: new key rotation and quantization ($2dh + 5.25d$ ops); query rotation ($2dh$ ops); dequantizing historical cache adds $Ld$ arithmetic ops and $Ld$ lookups. Total decode cost: $4dh + 5.25d + Ld$ arithmetic ops and $Ld$ lookups.

\subsection{Effective Cost Conversion}
\label{subsec:efc_conversion}

To enable fair comparison across methods with different operation types, we convert raw operation counts into weighted effective costs. We adopt a conservative weighting: one arithmetic operation costs 1 unit, and one random table lookup costs 5 units. This 1:5 ratio serves as a reasonable analytical compromise.
On modern GPUs, arithmetic operations are heavily pipelined and can be issued at high throughput when data is in registers. In contrast, table lookups require address computation, memory accesses, and often suffer from irregular access patterns. Even under ideal L1 cache locality, a lookup typically incurs higher latency due to dependency stalls, and lookups can cause warp divergence. These factors collectively make table lookups more expensive than arithmetic operations.

We emphasize that this weighting is an approximation intended solely for comparative analysis. In real GPU execution, the true cost of random table lookups is often substantially higher than five arithmetic units. Nevertheless, this conservative weighting offers a transparent and reproducible basis for theoretical comparison. Accordingly, for a method with $A$ arithmetic operations and $T$ table lookups, its effective cost is $A + 5T$.

\subsection{Numerical Results}
\label{subsec:numerical_results}

For numerical illustration, we use an equal-width Query/Key cost model with hidden dimension $d = 4096$, head dimension $h = 128$ ($\log_2 h = 7$), and sequence length $L = 10{,}000$. This is an analytical configuration, not an exact accounting of the Qwen3-8B grouped-query kernels: unequal Query and KV widths require separate counts. Applying the 1:5 weighting to Table~\ref{tab:op_counts_summary}, and multiplying per-token prefill costs by $L$, gives Table~\ref{tab:complexity_numerical}.

\subsection{Discussion}

OScaR introduces token-wise normalization during prefill and decode, incurring a decode cost of 123.0 million units under the assumed configuration ($d = 4096$, $L = 10{,}000$). This overhead is offset by two key advantages: it eliminates all table lookups, avoiding irregular memory accesses, and relies solely on efficient operations, including fast Hadamard transforms and the hardware-accelerated \texttt{rsqrt} instruction. These theoretical estimates are validated by our empirical efficiency analysis in Section~\ref{sec:efficiency}, where OScaR achieves substantial decode speedups while maintaining strong quantization fidelity with modest overhead.
\section{Implementation Details of OScaR's CUDA Kernels}
\label{app:cuda}

This section outlines the key implementation details of the OScaR inference system. Our implementation builds upon two prior frameworks: BitDecoding \citep{du2025bitdecoding} for the 2-bit quantization and cache management backbone, and HadaCore \citep{agarwal2024hadacore} for Tensor-Core accelerated Hadamard transforms. We extend both with fused Hadamard-norm preprocessing, integrated norm metadata, and residual-aware attention kernels.

\subsection{Overall Design}

OScaR compresses the KV cache to 2-bit representation to reduce memory footprint and bandwidth during long-context decoding. For Qwen3-8B, the core configuration is as follows: head dimension $d_h = 128$, group size $G = 32$ for per-channel key quantization, residual block size $R = 128$ for periodic cache flushing, and a K-side transformation consisting of Hadamard rotation followed by token-wise L2 normalization.
Mathematically, the key is transformed as:

$$
\mathbf{K}_h = \frac{\mathbf{H}(\mathbf{K})}{\sqrt{D}}, \qquad
n_k = \|\mathbf{K}_h\|_2, \qquad
\mathbf{K}_u = \frac{\mathbf{K}_h}{n_k},
$$

where $D = 128$, $\mathbf{H}$ is the Hadamard matrix, and $\mathbf{K}_u$ is the unit direction vector stored in the low-bit cache. The token-wise norm $n_k$ is stored as auxiliary metadata. During decoding, the query undergoes the same Hadamard rotation:

$$
\mathbf{Q}_h = \frac{\mathbf{H}(\mathbf{Q})}{\sqrt{D}}, \qquad
\text{logits} = \bigl(\mathbf{Q}_h \cdot \text{dequant}(\mathbf{K}_u)\bigr) \cdot n_k.
$$

Because the Hadamard transform is orthogonal, the inner product is preserved, and the norm metadata restores the original magnitude of each key token.

\subsection{Fused Hadamard-Norm Kernel}

The fused preprocessing kernel operates on key tensors of shape $[\text{batch}, \text{seqlen}, n_{\text{kv\_heads}}, d_h]$ with FP16 or BF16 data type. It outputs normalized keys and token-wise norms, combining the following steps into a single CUDA launch: reading all KV heads of a token, applying the normalized Hadamard transform to each head, accumulating squared sums across heads, computing the token-wise norm, and normalizing the transformed keys.

For generic implementations, one CUDA block processes one token with 128 threads, using butterfly Fast Hadamard Transform (FWHT) in shared memory followed by a reduction. For the fixed head dimension 128, we adopt the HadaCore-style Tensor Core optimization, which leverages the Kronecker decomposition $\mathbf{H}_{128} = \mathbf{H}_8 \otimes \mathbf{H}_{16}$. Specifically, we apply $\mathbf{H}_{16}$ via WMMA tiles and $\mathbf{H}_8$ via scalar butterfly, significantly reducing scalar instruction pressure compared to a naive FWHT.

The query side reuses the same kernel with Hadamard only (without normalization), ensuring that the inner product remains consistent with the original attention.

\subsection{Quantization Format and Cache Organization}

\textbf{Key Quantization.} Following BitDecoding, after Hadamard-norm preprocessing, $\mathbf{K}_u$ is quantized using per-channel grouping: each channel along the head dimension shares a scale and zero point across every $G = 32$ tokens. For 2-bit quantization, values range in $[0, 3]$, and eight values are packed into one uint16 word. A 128-token residual block contains exactly four quantization groups.

\textbf{Value Quantization.} Values are quantized via an offline Hadamard transform without requiring additional norm transformations, with per-token grouping along the head dimension, consistent with the original BitDecoding design. For $d_h = 128$ and $G = 32$, each token-head is partitioned into four quantization groups.

\textbf{Cache Organization.} The system maintains both packed and residual caches:

\begin{itemize}
    \item \textbf{Packed cache:} $\mathbf{K}_u$ 2-bit payload, K scale and zero point, V 2-bit payload, V scale and zero point, and K token-wise norms.
    \item \textbf{Residual cache:} $\mathbf{K}_u$ residual (FP16), V residual (FP16), and K norm residual (FP16).
\end{itemize}

\subsection{Prefill and Decode Workflows}

\textbf{Prefill.} Attention computation itself remains unchanged and uses FlashAttention-2. After attention, the prompt sequence is split: the prefix (length a multiple of $R$) is quantized and stored in the packed cache after applying Hadamard-norm; the tail (length less than $R$) is stored in the residual cache without quantization.

\textbf{Decode.} Each new token undergoes projection, RMSNorm, RoPE, and Hadamard (for Q) or Hadamard-norm (for K). The transformed key, value, and norm are appended to the residual cache. The attention kernel \texttt{fwd\_kvcache\_int} processes both packed and residual caches as follows:
\begin{itemize}
    \item For packed entries, keys are dequantized from their 2-bit representation and multiplied by their corresponding token-wise norms.
    \item For residual entries, keys are already in FP16 and are multiplied by their stored norms.
\end{itemize}
Both contributions are then combined in the logits before softmax and output projection. When the residual length reaches $R = 128$, the residual block is quantized and flushed to the packed cache.

\textbf{Residual Flush.} Flushing a 128-token block generates 16 packed rows ($128 / 8$) and 4 parameter rows ($128 / 32$). Since residual keys are already normalized, only quantization and packing are required, avoiding repeated Hadamard-norm computation.

\subsection{Summary of Key Differences from the Baseline}

Compared to the original BitDecoding, OScaR introduces the following key differences:

\begin{enumerate}
    \item Hadamard rotation and token-wise normalization for keys, building upon HadaCore's efficient transform primitive.
    \item Norm metadata incorporated into logits during attention.
    \item Hadamard rotation for queries to preserve inner product semantics.
    \item Removal of Hadamard and normalization on the value side.
    \item A fixed 128-token residual flush, aligned with the 2-bit K-channel quantization groups.
\end{enumerate}

\subsection{Packed-cache storage and norm metadata}
\label{app:storage_accounting}
For the packed-cache formats evaluated in Table~\ref{tab:longbench}, INT2 payloads with one FP16 scale and one FP16 zero point per 32 values cost $2+32/32=3.0$ bits/value. The evaluated TurboQuant+ format costs $2.5+2\times32/128=3.0$ bits/value for mixed indices and two FP32 subset norms per 128-dimensional vector. These are nominal packed-cache rates; context-dependent residual windows and protected-token buffers are additional.

OScaR's system implementation stores one FP32 Key norm per token per layer, shared across KV heads. With $H_{\mathrm{kv}}=8$ and $d_h=128$, this adds
\begin{equation}
\frac{32}{H_{\mathrm{kv}}d_h}=0.03125\ \text{bit/Key value},\qquad
\frac{32}{2H_{\mathrm{kv}}d_h}=0.015625\ \text{bit/KV value}.
\end{equation}
Thus, including this norm gives 3.015625 bits/value for OScaR in this configuration, before residual buffers. Norm extraction, metadata writes, and inverse scaling are fused without a separate kernel launch.

\begin{table}[htbp]
\centering
\caption{Nominal storage for the evaluated packed-cache formats, averaged over K/V. Residual windows and protected-token pools are excluded.}
\label{tab:additional_storage}
\small
\begin{tabular}{@{}llr@{}}
\toprule
Methods & Included metadata & Bits/value \\
\midrule
QuaRot / KIVI / OTT / RotateKV & FP16 scale and zero point, $G=32$ & 3.0 \\
TurboQuant+ & Two FP32 subset norms, $d_h=128$ & 3.0 \\
OScaR & Scale/zero point plus shared FP32 Key norm & 3.015625 \\
\bottomrule
\end{tabular}
\end{table}

\section{Details of Datasets and Benchmarks}
\label{app:dataset}

\subsection{Text-Only LLM Benchmarks}

\paragraph{LongBench-E.}
LongBench \citep{bai2024longbench} is a bilingual multitask benchmark for evaluating long-context understanding in large language models, covering both English and Chinese. It comprises 21 datasets across six task categories, with average lengths of 6,711 words for English and 13,386 characters for Chinese. The six categories are:

\begin{itemize}
    \item \textbf{Single-Document QA (4 tasks):} MultiFieldQA-en, MultiFieldQA-zh, NarrativeQA, and Qasper. Models answer questions based on a single long document, testing detailed comprehension and information retrieval.
    \item \textbf{Multi-Document QA (4 tasks):} HotpotQA, 2WikiMultihopQA, MuSiQue, and DuReader. Models answer questions by synthesizing information across multiple documents, testing multi-hop reasoning.
    \item \textbf{Summarization (4 tasks):} GovReport, QMSum, MultiNews, and VCSUM. Models generate concise summaries of long documents or meeting transcripts, testing content selection and abstraction.
    \item \textbf{Few-Shot Learning (4 tasks):} TriviaQA, SAMSum, TREC, and LSHT. Models answer questions with few-shot examples provided in context, testing in-context learning ability.
    \item \textbf{Synthetic Tasks (3 tasks):} PassageRetrieval-en, PassageCount, and PassageRetrieval-zh. Models perform artificial tasks such as passage retrieval and counting, testing length generalization.
    \item \textbf{Code Completion (2 tasks):} LCC and RepoBench-P. Models predict the next line of code given long code contexts, including cross-file dependencies.
\end{itemize}

LongBench-E is a uniformly sampled subset where context lengths are evenly distributed across three intervals: 0--4k, 4k--8k, and 8k+ tokens. This design enables systematic analysis of model performance degradation as sequence length increases.

\paragraph{Needle-in-a-Haystack.}
The Needle-in-a-Haystack test \citep{gkamradt2023needle} evaluates the in-context retrieval ability of long-context LLMs. A random fact (the needle) is inserted at varying depths within a long context window (the haystack), and the model is tasked with retrieving this specific statement. Performance is measured across context lengths ranging from 1K to 32K+ tokens and needle positions from 0\% to 100\% depth. The primary metric is retrieval accuracy, which reveals how well models maintain attention to relevant information buried in long sequences.

\subsection{Multi-modal LLM Benchmarks}

\paragraph{OCRBench.}
OCRBench \citep{liu2024ocrbench} is a comprehensive benchmark for evaluating optical character recognition capabilities in large multi-modal models. It comprises 29 datasets across five task categories, with a total of 1,000 human-annotated question-answer pairs:

\begin{itemize}
    \item \textbf{Text Recognition (300 samples):} Recognizes regular, irregular, non-semantic, digit strings, handwriting, and artistic text from images.
    \item \textbf{Scene Text-Centric VQA (200 samples):} Answers questions about text appearing naturally in scenes, such as street signs, billboards, and storefronts, requiring both visual understanding and text reading.
    \item \textbf{Document-Oriented VQA (200 samples):} Focuses on structured documents such as forms, invoices, reports, and letters, requiring understanding of document layout, tables, and formatting.
    \item \textbf{Key Information Extraction (200 samples):} Extracts specific information, such as total amounts, dates, and names, from structured documents.
    \item \textbf{Handwritten Mathematical Expression Recognition (100 samples):} Recognizes and transcribes handwritten mathematical formulas into LaTeX format.
\end{itemize}

\paragraph{DocVQA.}
DocVQA \citep{mathew2021docvqa} is a dataset for visual question answering on document images. It contains over 12,000 document images with 50,000 questions, covering diverse document types including forms, invoices, reports, letters, and tables. Questions require understanding of document structure, layout, and visual elements beyond simple text extraction. The validation set comprises 5,349 samples. The primary evaluation metric is ANLS (Average Normalized Levenshtein Similarity), which accounts for minor OCR and spelling variations.

\subsection{Omni-modal LLM Benchmark}

\paragraph{MMAU-Pro.}
MMAU-Pro \citep{kumar2026mmaupro} is a comprehensive benchmark for holistic evaluation of audio intelligence in AI systems. It contains 5,305 instances, each comprising one or more audio clips paired with human expert-generated question-answer pairs, spanning speech, non-speech sounds, music, and their combinations. The benchmark evaluates auditory intelligence across 49 unique skills and multiple complex dimensions, including long-form audio comprehension, spatial audio reasoning, and multi-audio understanding. All questions are designed to require deliberate multi-hop reasoning and include both multiple-choice and open-ended response formats. Notably, audio data is sourced directly from real-world environments (``from the wild'') rather than from existing datasets with known distributions.

For our evaluation, we focus on two challenging subsets that demand reasoning beyond standard multiple-choice: open-ended QA and audio instruction following.
The open-ended subset requires models to generate free-form responses. Following the MMAU-Pro protocol, we evaluate these responses using Qwen2.5-7B-Instruct as an LLM judge, which scores each response from 1 to 5 across four criteria: correctness, relevance, completeness, and clarity. The scores are then converted to percentages for consistent comparison with multiple-choice results.
The instruction-following subset comprises constraint instances drawn from 28 instruction types, with responses evaluated using deterministic scripts.
\section{Additional TurboQuant+ Implementation Details}
\label{app:turboquant}

\begin{table}[t]
\caption{OCRBench evaluation results comparing TurboQuant+ with and without QJL. TurboQuant+ uses a 2.5-bit setting, whereas OScaR employs INT2 quantization with a group size of 128.}
\label{tab:turboquant+qjl}
\centering
\resizebox{1\textwidth}{!}{
\begin{tabular}{@{}llcccccc@{}}
\toprule
\textbf{Model} & \textbf{Method} & \textbf{Recog.} & \textbf{\textit{VQA\textsuperscript{S}}} & \textbf{\textit{VQA\textsuperscript{D}}} & \textbf{KIE} & \textbf{HMER} & \textbf{Final Score} \\ \midrule
\textbf{Sample Size} &  & 300 & 200 & 200 & 200 & 100 & 1000 \\ \midrule
\multirow{4}{*}{LLaVA-v1.6-vicuna-7B} & 16bit & 187 & 160 & 82 & 107 & 0 & 536 \\
 & TurboQuant+ (w/ QJL) & 161 & 144 & 41 & 57 & 0 & 403 \\
 & TurboQuant+ (w/o QJL) & 173 & 156 & 73 & 99 & 0 & 501 \\
 & \textit{\textbf{OScaR (ours)}} & 178 & 158 & 82 & 101 & 0 & 519 \\ \midrule
\multirow{4}{*}{Qwen3-VL-8B} & 16bit & 270 & 181 & 174 & 181 & 52 & 858 \\
 & TurboQuant+ (w/ QJL) & 266 & 136 & 57 & 143 & 24 & 626 \\
 & TurboQuant+ (w/o QJL) & 269 & 178 & 169 & 182 & 49 & 847 \\
 & \textit{\textbf{OScaR (ours)}} & 270 & 180 & 170 & 183 & 53 & 856 \\ \bottomrule
\end{tabular}
}
\end{table}

Since TurboQuant does not provide an official code release, we adopt TurboQuant+ \citep{turney2026turboquantplus}, a widely used open-source implementation, for all evaluations. TurboQuant+ officially omits the QJL step \citep{zandieh2025qjl}, as it has been shown to increase variance that is subsequently amplified by the softmax function, ultimately degrading generation quality. To ensure a more challenging comparison, we also exclude the QJL step in our experiments, as it empirically reduces model performance (see Table~\ref{tab:turboquant+qjl}).

We also account for the average bit overhead introduced by quantization parameters in OScaR and other comparison methods. Specifically, TurboQuant+ employs a mixed-precision 2.5-bit quantization scheme: outlier channels are assigned higher bit-widths to preserve fidelity, while regular channels use 2 bits, yielding an average of approximately 2.5 bits per element. All other methods adopt INT2 quantization.
Furthermore, since most competing baselines apply the same quantization configuration across layers, we adopt a consistent setup for every layer in TurboQuant+, uniformly quantizing both Key and Value without layer-wise adaptive precision.

\section{Experimental Results and Analysis on Needle-in-a-Haystack}
\label{app:niah}

\begin{figure}[h]

\centering
\begin{subfigure}{1\textwidth}
    \centering
    \includegraphics[width=0.8\textwidth]{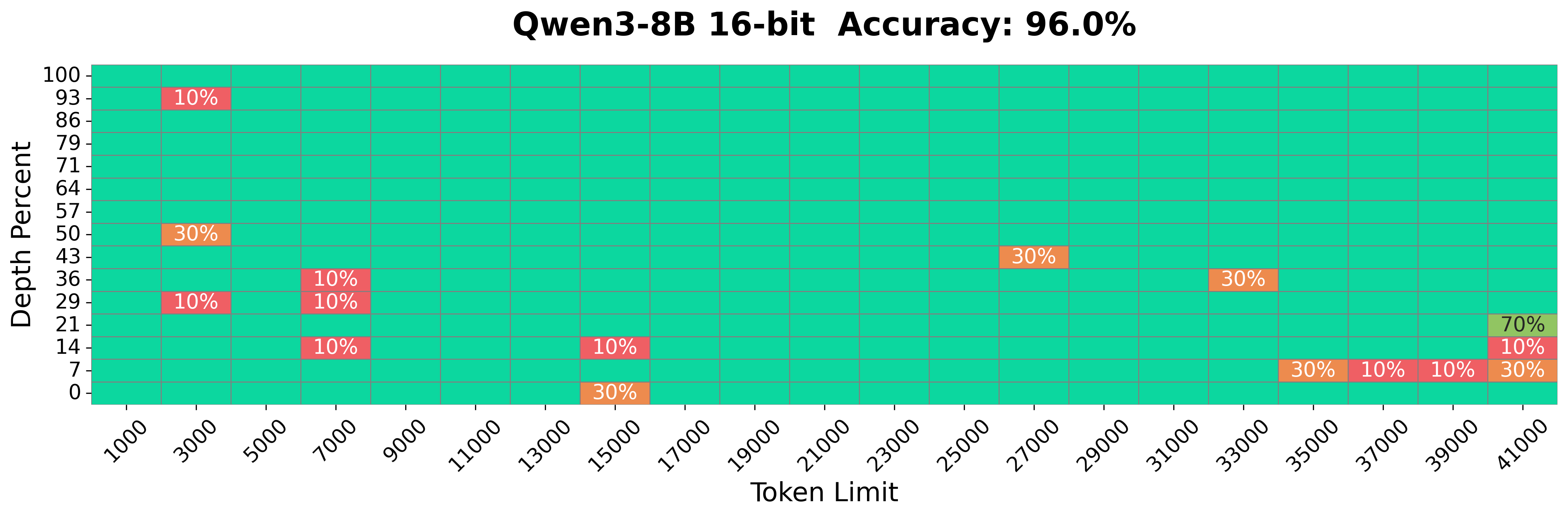}
    
    \caption{Full-precision baseline with 16-bit KV cache. Retrieval accuracy: 96.0\%.}
    \label{fig:16bit}
\end{subfigure}

\begin{subfigure}{1\textwidth}
    \centering
    \includegraphics[width=0.8\textwidth]{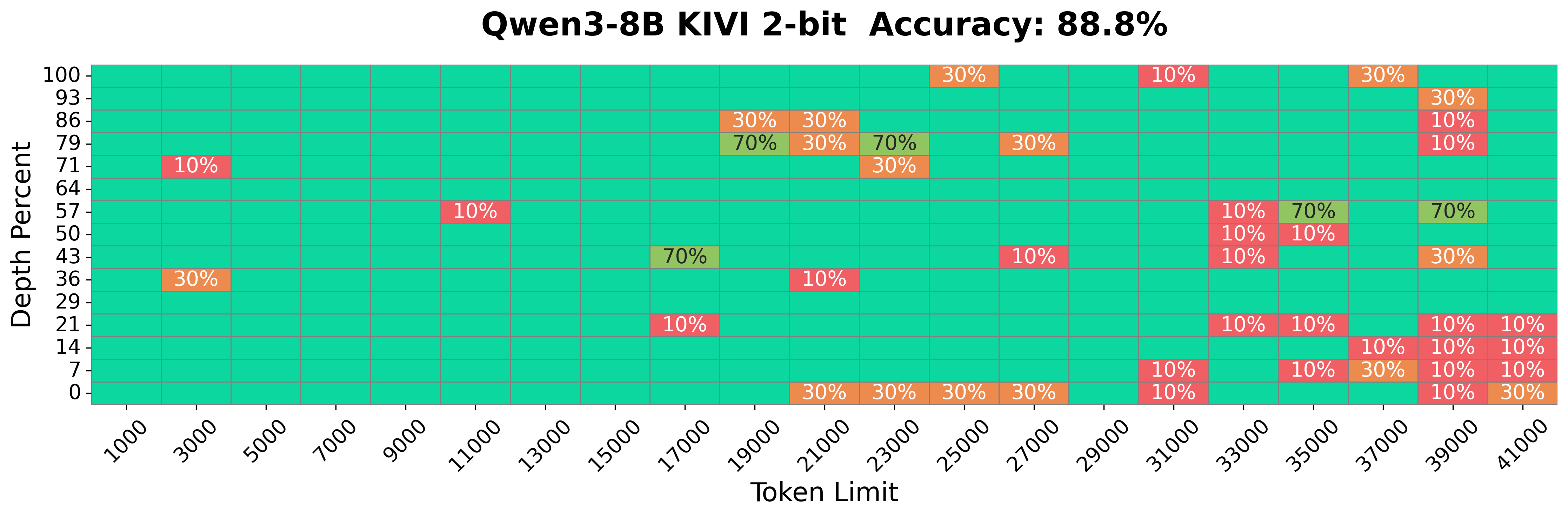}
    
    \caption{KIVI under 2-bit KV cache quantization. Retrieval accuracy: 88.8\%.}
    \label{fig:kivi}
\end{subfigure}

\begin{subfigure}{1\textwidth}
    \centering
    \includegraphics[width=0.8\textwidth]{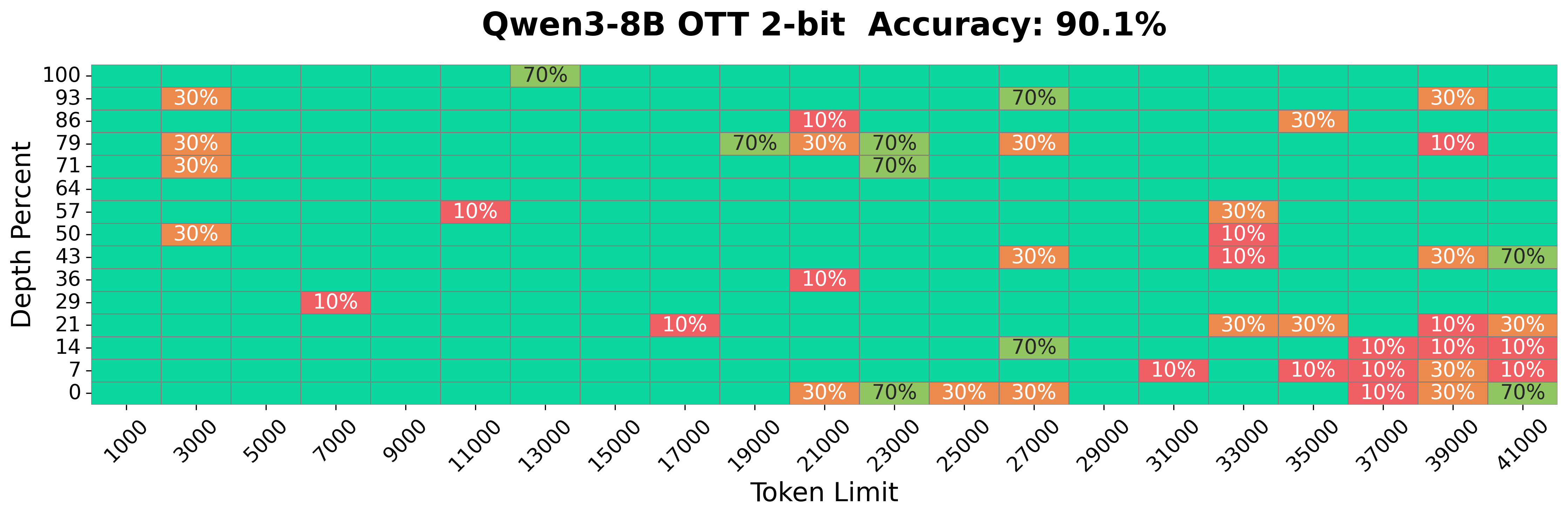}
    
    \caption{OTT under 2-bit KV cache quantization. Retrieval accuracy: 90.1\%.}
    \label{fig:ott}
\end{subfigure}

\begin{subfigure}{1\textwidth}
    \centering
    \includegraphics[width=0.8\textwidth]{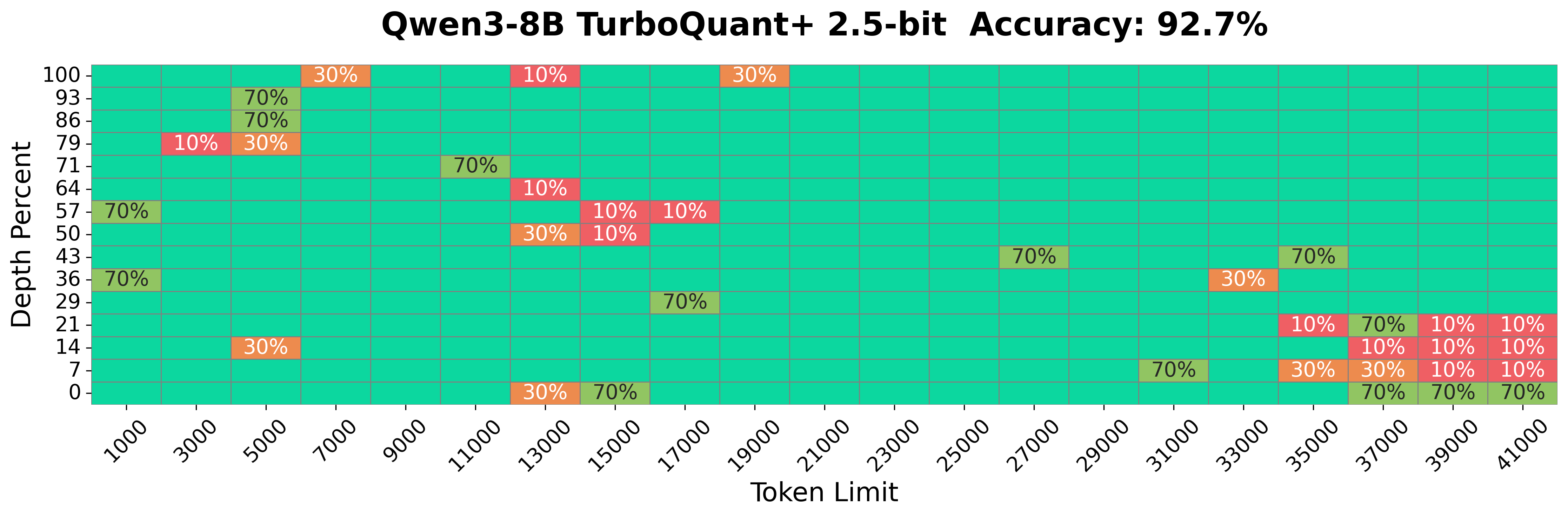}
    
    \caption{TurboQuant+ under 2.5-bit KV cache quantization. Retrieval accuracy: 92.7\%.}
    \label{fig:turboquant}
\end{subfigure}

\begin{subfigure}{1\textwidth}
    \centering
    \includegraphics[width=0.8\textwidth]{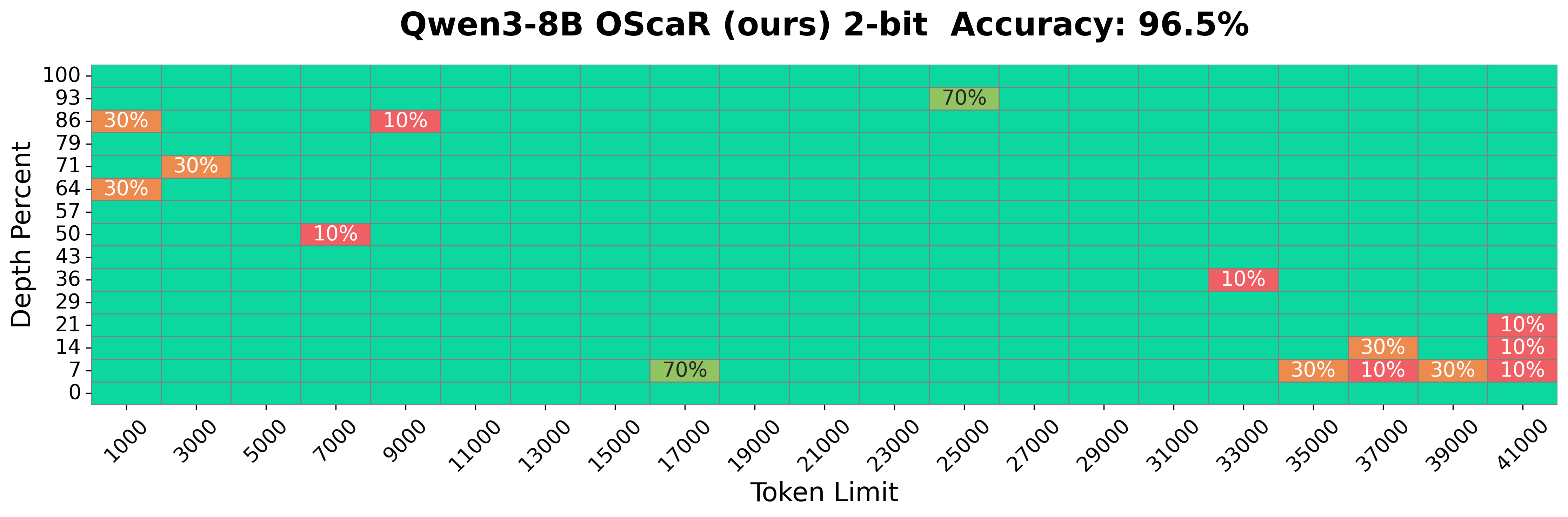}
    
    \caption{OScaR under 2-bit KV cache quantization (our method). Retrieval accuracy: 96.5\%.}
    \label{fig:oscar}
\end{subfigure}

\caption{NIAH evaluation results. All competing methods except TurboQuant+ are configured with INT2 quantization and a group size of 32. TurboQuant+ uses a 2.5-bit setting. TurboQuant is based on TurboQuant+ \citep{turney2026turboquantplus}; QJL is excluded as it degrades performance. See Appendix~\ref{app:turboquant} for details.}
\label{fig:needle_results}
\end{figure}

In this section, we present the Needle-in-a-Haystack experiment \citep{gkamradt2023needle}. Following the standard protocol, we evaluate retrieval accuracy across context lengths up to 42,000 tokens and at 15 different needle depth positions. A detailed description of the benchmarks is provided in Appendix~\ref{app:dataset}. The results are shown in Figure~\ref{fig:needle_results}.
The 16-bit full-precision baseline is provided for reference. OScaR achieves the highest retrieval accuracy among all quantized methods (96.5\%), slightly surpassing the 16-bit baseline (96.0\%) and outperforming the second-best quantized method (92.7\%).

\section{Experimental Results and Analysis on OCRBench}
\label{app:ocrbench}

\begin{table}[t]

\caption{OCRBench evaluation results across six task categories: Recognition (Recog.), Scene Text VQA (\textit{VQA\textsuperscript{S}}), Document Text VQA (\textit{VQA\textsuperscript{D}}), Key Information Extraction (KIE), Handwritten Mathematical Expression Recognition (HMER), and the final weighted score. The superscripts \textsuperscript{S} and \textsuperscript{D} denote scene text and document text, respectively. All competing methods except TurboQuant+ are configured with INT2 quantization and a group size of 128. TurboQuant+ uses a 2.5-bit setting.
TurboQuant is based on TurboQuant+ \citep{turney2026turboquantplus}; QJL is excluded as it degrades performance. See Appendix~\ref{app:turboquant} for details.}
\label{tab:ocrbench}
\centering
\resizebox{1\textwidth}{!}{
\begin{tabular}{@{}llcccccc@{}}
\toprule
\textbf{Model} & \textbf{Method} & \textbf{Recog.} & \textbf{\textit{VQA\textsuperscript{S}}} & \textbf{\textit{VQA\textsuperscript{D}}} & \textbf{KIE} & \textbf{HMER} & \textbf{Final Score} \\ \midrule
\textbf{Sample Size} &  & 300 & 200 & 200 & 200 & 100 & 1000 \\ \midrule
\multirow{7}{*}{LLaVA-v1.6-vicuna-7B} & 16bit & 187 & 160 & 82 & 107 & 0 & 536 \\ \cmidrule(l){2-8} 
 & QuaRot & 177 & 152 & 61 & 91 & 0 & 481 \\
 & RotateKV & 168 & 152 & 64 & 89 & 0 & 473 \\
 & KIVI & 173 & 157 & 69 & 89 & 0 & 488 \\
 & OTT & \textbf{187} & 157 & 72 & 97 & 0 & 513 \\
 & TurboQuant+ & 173 & 156 & 73 & 99 & 0 & 501 \\
 & \textit{\textbf{OScaR (ours)}} & 178 & \textbf{158} & \textbf{82} & \textbf{101} & 0 & \textbf{519} \\ \midrule
\multirow{7}{*}{Qwen3-VL-8B} & 16bit & 270 & 181 & 174 & 181 & 52 & 858 \\ \cmidrule(l){2-8} 
 & QuaRot & 266 & 171 & 104 & 150 & 31 & 722 \\
 & RotateKV & 267 & 163 & 127 & 162 & 35 & 754 \\
 & KIVI & 268 & 178 & \textbf{173} & 180 & 52 & 851 \\
 & OTT & \textbf{270} & 179 & 170 & 177 & 54 & 850 \\
 & TurboQuant+ & 269 & 178 & 169 & 182 & 49 & 847 \\
 & \textit{\textbf{OScaR (ours)}} & \textbf{270} & \textbf{180} & 170 & \textbf{183} & \textbf{53} & \textbf{856} \\ \midrule
\multirow{7}{*}{Qwen3-VL-4B} & 16bit & 261 & 183 & 171 & 186 & 51 & 852 \\ \cmidrule(l){2-8} 
 & QuaRot & 254 & 173 & 149 & 159 & 38 & 773 \\
 & RotateKV & 254 & 143 & 84 & 134 & 23 & 638 \\
 & KIVI & 255 & 177 & 160 & 175 & 46 & 813 \\
 & OTT & 256 & 180 & 167 & 177 & \textbf{51} & 831 \\
 & TurboQuant+ & \textbf{260} & \textbf{183} & \textbf{170} & 175 & 40 & 828 \\
 & \textit{\textbf{OScaR (ours)}} & 259 & 181 & 169 & \textbf{181} & 48 & \textbf{838} \\ \bottomrule
\end{tabular}
}
\end{table}

Table~\ref{tab:ocrbench} reports results on the OCRBench benchmark \citep{liu2024ocrbench}. The final score reflects the number of correctly answered samples. A detailed dataset description is provided in Appendix~\ref{app:dataset}.
OScaR consistently achieves the highest accuracy among all 2-bit quantization methods across all three evaluated models. On LLaVA-v1.6-vicuna-7B, OScaR attains 51.9\% accuracy, outperforming the second-best method by 0.6 percentage points. On Qwen3-VL-8B, OScaR achieves 85.6\% accuracy, coming within 0.2 percentage points of the 16-bit baseline (85.8\%). On Qwen3-VL-4B, OScaR scores 83.8\% accuracy, surpassing the best competing 2-bit method by 0.7 percentage points.
Overall, OScaR delivers robust performance on multi-modal document understanding tasks under INT2 KV cache quantization.
\section{Experimental Results and Analysis on DocVQA}
\label{app:docvqa}

Table~\ref{tab:docvqa} reports results on the DocVQA benchmark \citep{mathew2021docvqa}, which evaluates document understanding through visual question answering. A detailed description of the dataset is provided in Appendix~\ref{app:dataset}.

OScaR consistently achieves the highest accuracy among all 2-bit quantization methods across the three evaluated models. On Qwen3-VL-8B, it slightly surpasses the 16-bit baseline, demonstrating near-lossless performance under INT2 quantization. On Qwen3-VL-4B, OScaR trails the 16-bit baseline by only 0.4 percentage points while outperforming the strongest competing 2-bit method by 0.36 percentage points. On LLaVA-v1.6-vicuna-7B, it exceeds the best quantized baseline by 0.7 percentage points and remains within 1.1 percentage points of the 16-bit reference. These results confirm that OScaR's \textit{Canalized Rotation} and \textit{Omni-Token Scaling} generalize effectively across diverse multi-modal models and tasks.

\begin{table}[t]

\caption{DocVQA evaluation results. All competing methods except TurboQuant+ are configured with INT2 quantization and a group size of 128. TurboQuant+ uses a 2.5-bit setting.
TurboQuant is based on TurboQuant+ \citep{turney2026turboquantplus}; QJL is excluded as it degrades performance. See Appendix~\ref{app:turboquant} for details.}
\label{tab:docvqa}
\resizebox{\textwidth}{!}{
\begin{tabular}{@{}lccccccc@{}}
\toprule
\textbf{Method} & \textbf{16bit} & \textbf{QuaRot} & \textbf{RotateKV} & \textbf{KIVI} & \textbf{OTT} & \textbf{TurboQuant+} & \textbf{\begin{tabular}[c]{@{}c@{}}OScaR\\ (ours)\end{tabular}} \\ \midrule
LLaVA-v1.6-vicuna-7B & 68.57 & 65.55 & 65.26 & 65.34 & 66.11 & 67.09 & \textbf{67.79} \\
Qwen3-VL-8B & 94.93 & 93.93 & 92.76 & 94.70 & 94.86 & 94.51 & \textbf{95.01} \\
Qwen3-VL-4B & 94.23 & 89.89 & 76.12 & 93.49 & 93.24 & 93.54 & \textbf{93.85} \\ \bottomrule
\end{tabular}
}
\end{table}

\section{Experimental Results and Analysis on MMAU-Pro}
\label{app:mmau-pro}

For our evaluation on MMAU-Pro, we focus on two challenging subsets that demand reasoning beyond standard multiple-choice: open-ended QA and audio instruction following.

The open-ended subset requires models to generate free-form responses. Following the MMAU-Pro protocol, we evaluate these responses using Qwen2.5-7B-Instruct as an LLM judge, which scores each response from 1 to 5 across four criteria: correctness, relevance, completeness, and clarity. The scores are then converted to percentages for consistent comparison with multiple-choice results. We additionally report the "Good Rate," defined as the proportion of samples where the LLM judge assigns an overall score of 4.0 or higher, indicating high-quality responses.
The instruction-following subset comprises constraint instances drawn from 28 instruction types, with responses evaluated using deterministic scripts.

\begin{table}[h]
\caption{MMAU-Pro evaluation results. "Good Rate" denotes the proportion of open-ended responses with an overall score of 4 or higher (out of 5), and "AIF" stands for Audio Instruction Following accuracy.
All competing methods except TurboQuant+ are configured with INT2 quantization and a group size of 128. TurboQuant+ uses a 2.5-bit setting. TurboQuant is based on TurboQuant+ \citep{turney2026turboquantplus}; QJL is excluded as it degrades performance. See Appendix~\ref{app:turboquant} for details.}
\label{tab:mmaupro_results}
\centering
\resizebox{0.75\textwidth}{!}{
\begin{tabular}{@{}lccccc@{}}
\toprule
\textbf{Qwen3-Omni-30B-A3B} & \textbf{16bit} & \textbf{KIVI} & \textbf{OTT} & \textbf{TurboQuant+} & \textbf{OScaR (ours)} \\ \midrule
Open-ended & 66.2 & 65.8 & 65.8 & 66.6 & \textbf{67.4} \\
Good Rate & 27.8 & 27.0 & 26.9 & 27.0 & \textbf{29.8} \\
AIF & 87.4 & 78.2 & 83.9 & 79.3 & \textbf{88.5} \\ \bottomrule
\end{tabular}
}
\end{table}

As shown in Table~\ref{tab:mmaupro_results}, OScaR achieves the highest scores across all three evaluation metrics. On the open-ended subset, OScaR slightly surpasses the 16-bit baseline by 1.2 percentage points and outperforms all other quantized methods. More notably, OScaR achieves a Good Rate that exceeds the 16-bit baseline by 2.0 percentage points and outperforms the best competing quantized method by 2.8 percentage points, indicating that OScaR preserves high-quality response generation under extreme compression. For audio instruction following (AIF), OScaR surpasses the 16-bit baseline by 1.1 percentage points, demonstrating near-lossless performance.
These results demonstrate OScaR's strong generalization capability across omni-modal tasks and models, maintaining superior performance under INT2 quantization on the challenging MMAU-Pro benchmark.
\section{TNI Analysis Before and After OScaR}
\label{app:analysis}

\begin{figure}[t]
    \centering
    \begin{subfigure}[b]{0.48\textwidth}
        \centering
        \includegraphics[width=\textwidth]{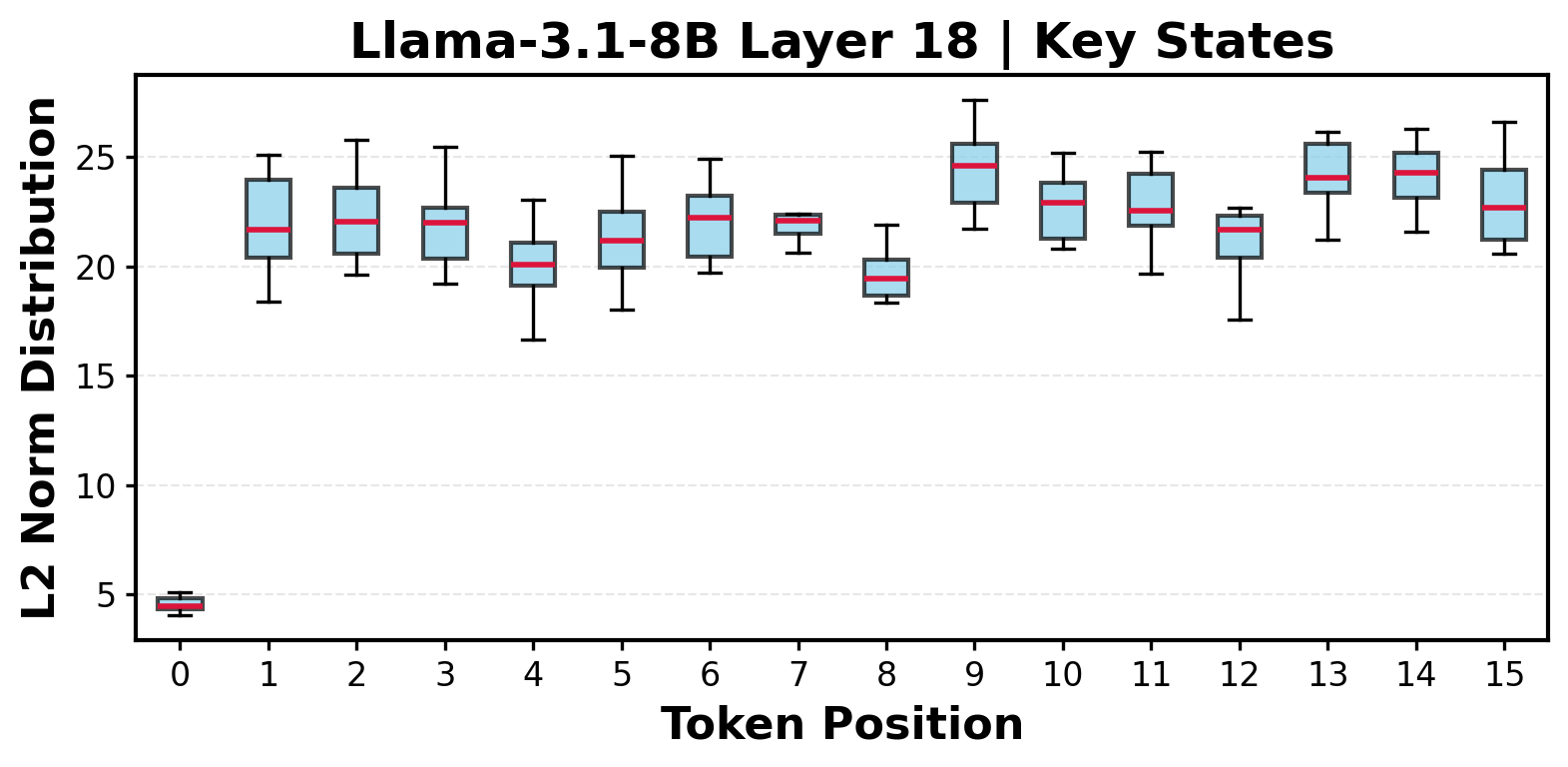}
        \caption{Llama-3.1-8B Layer 18 (before OScaR).}
    \end{subfigure}
    \begin{subfigure}[b]{0.48\textwidth}
        \centering
        \includegraphics[width=\textwidth]{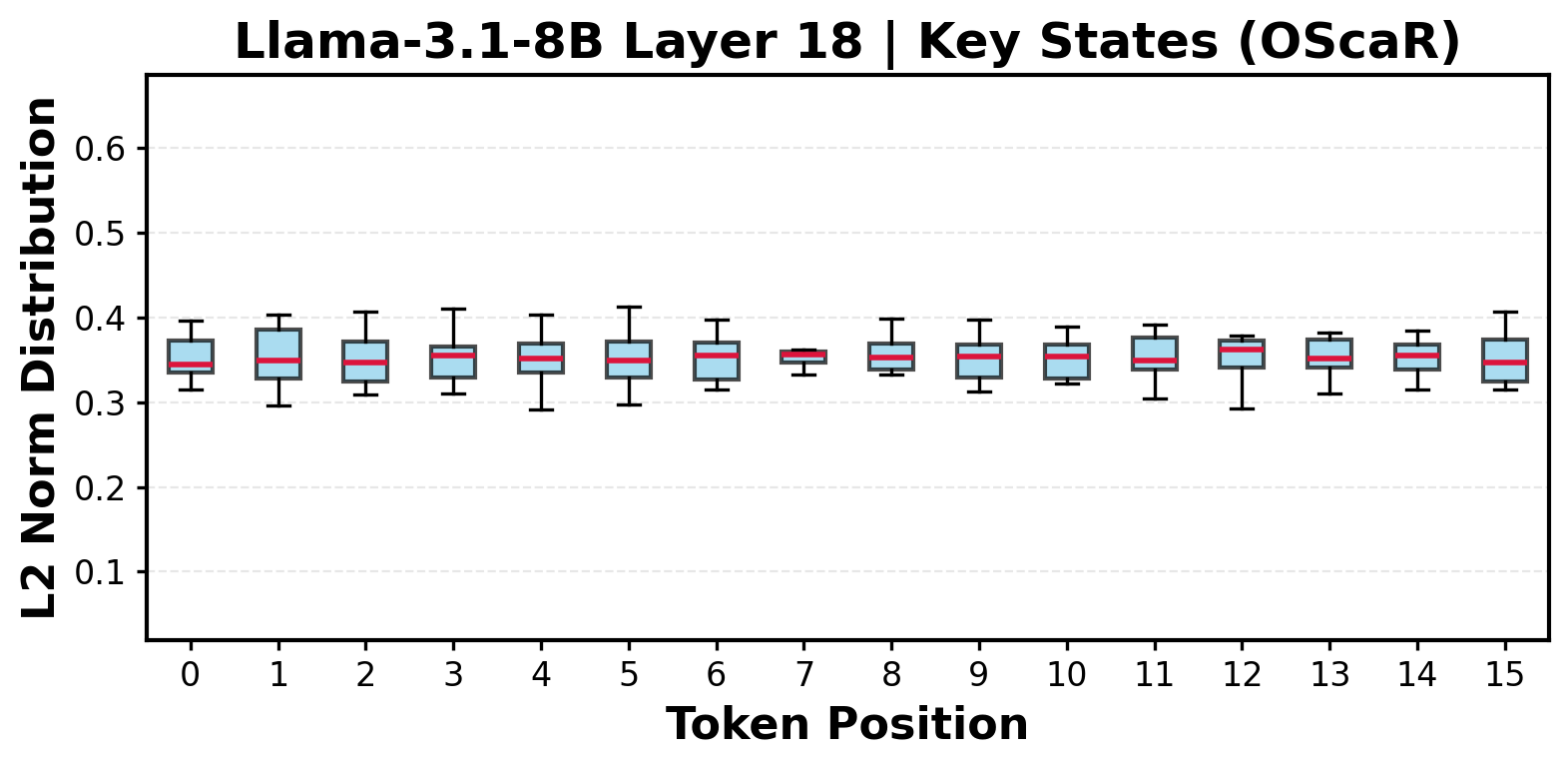}
        \caption{Llama-3.1-8B Layer 18 (after OScaR).}
    \end{subfigure}
    \begin{subfigure}[b]{0.48\textwidth}
        \centering
        \includegraphics[width=\textwidth]{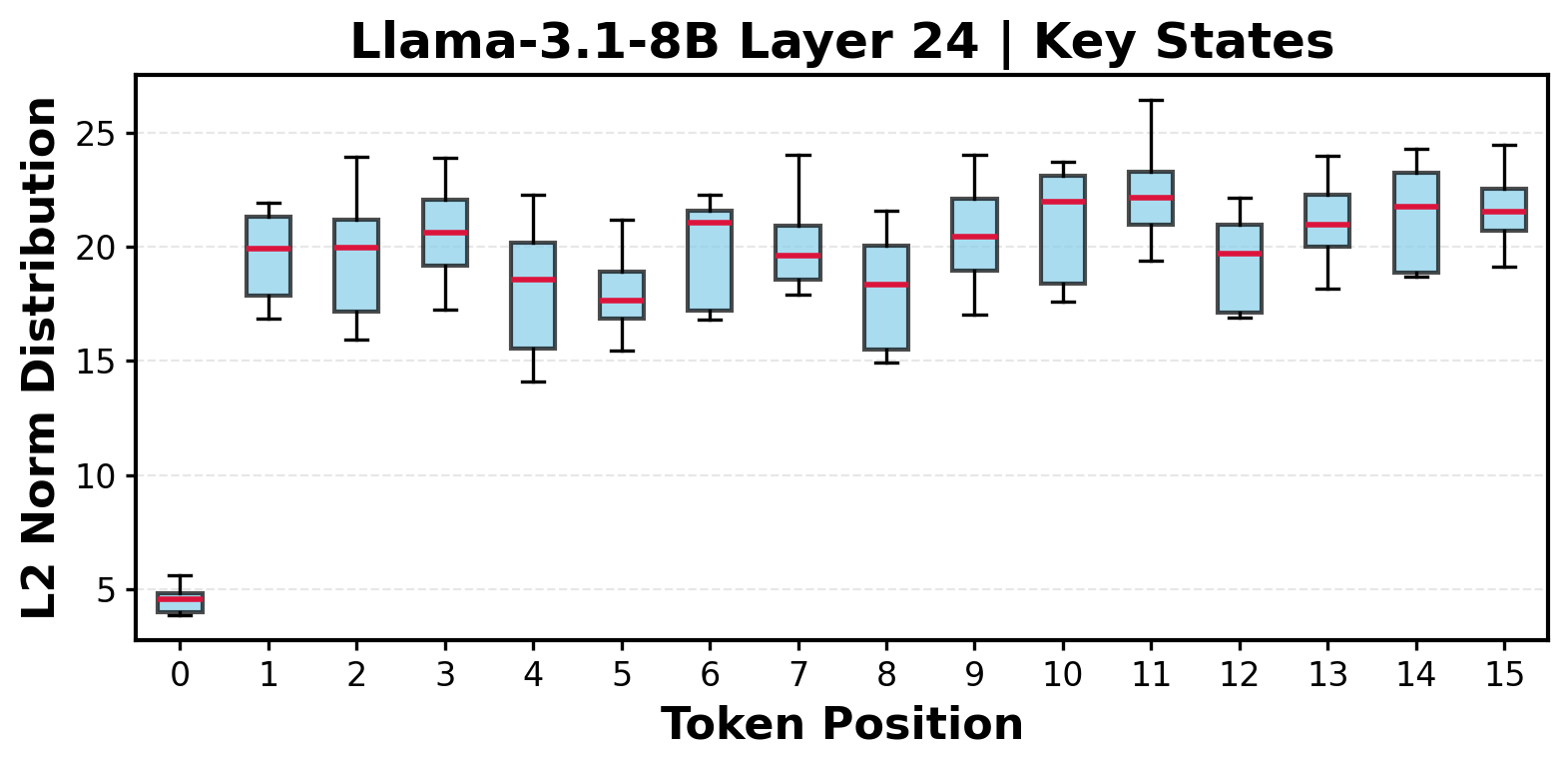}
        \caption{Llama-3.1-8B Layer 24 (before OScaR).}
    \end{subfigure}
    \begin{subfigure}[b]{0.48\textwidth}
        \centering
        \includegraphics[width=\textwidth]{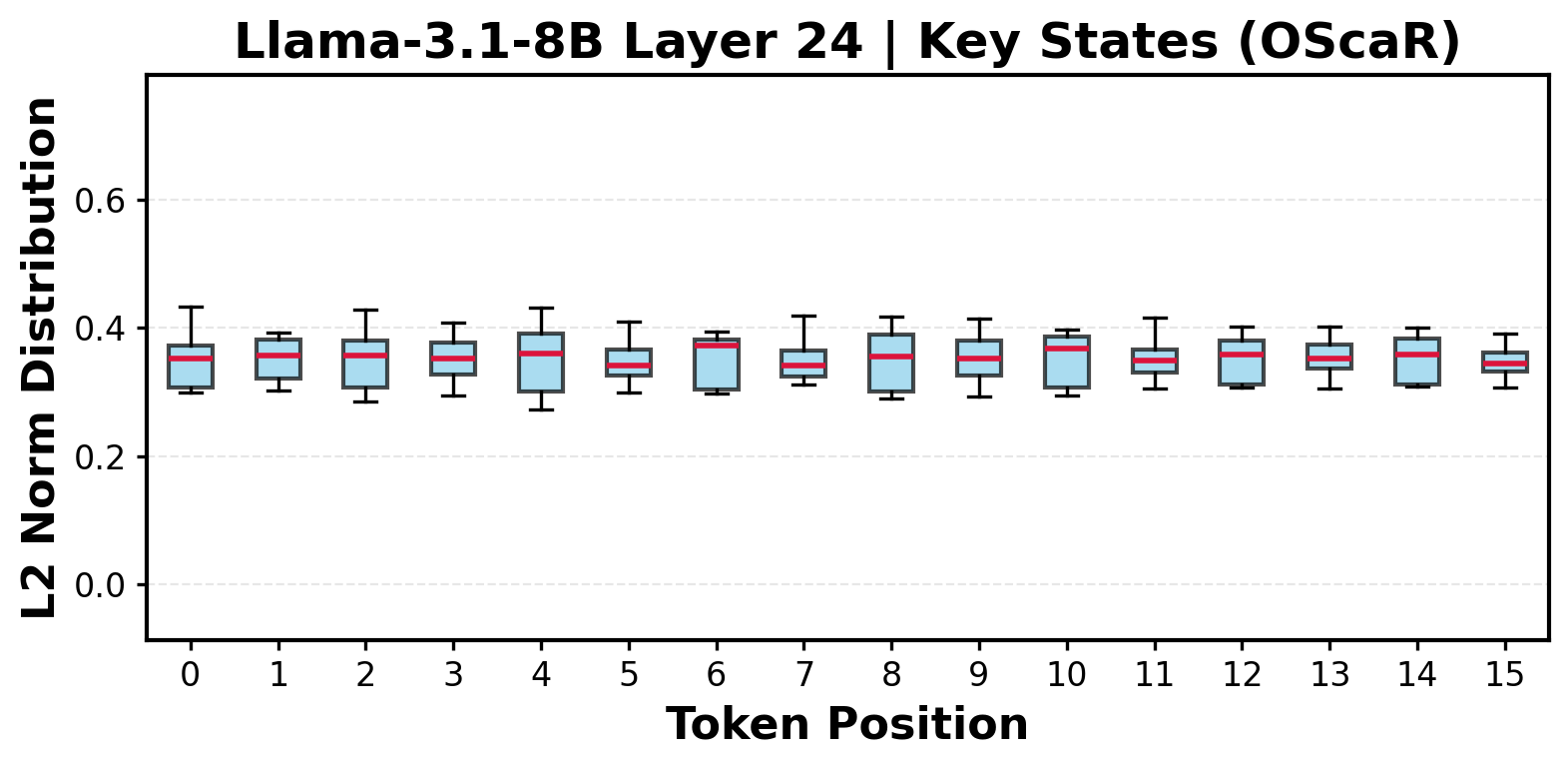}
        \caption{Llama-3.1-8B Layer 24 (after OScaR).}
    \end{subfigure}
    \caption{Token norm distribution on Llama-3.1-8B before and after applying OScaR.}
    \label{fig:ba_oscar_llama-3}
\end{figure}

\begin{figure}[t]
    \centering
    \begin{subfigure}[b]{1\textwidth}
        \centering
        \includegraphics[width=\textwidth]{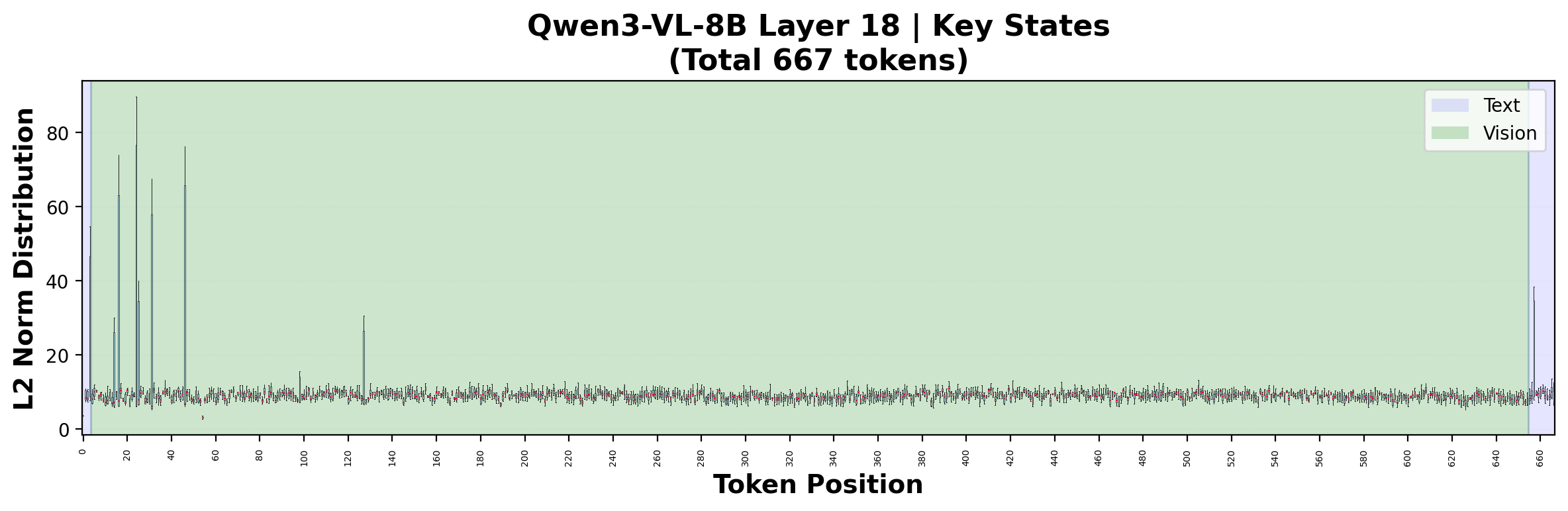}
        \caption{Qwen3-VL-8B (before OScaR).}
    \end{subfigure}
    \begin{subfigure}[b]{1\textwidth}
        \centering
        \includegraphics[width=\textwidth]{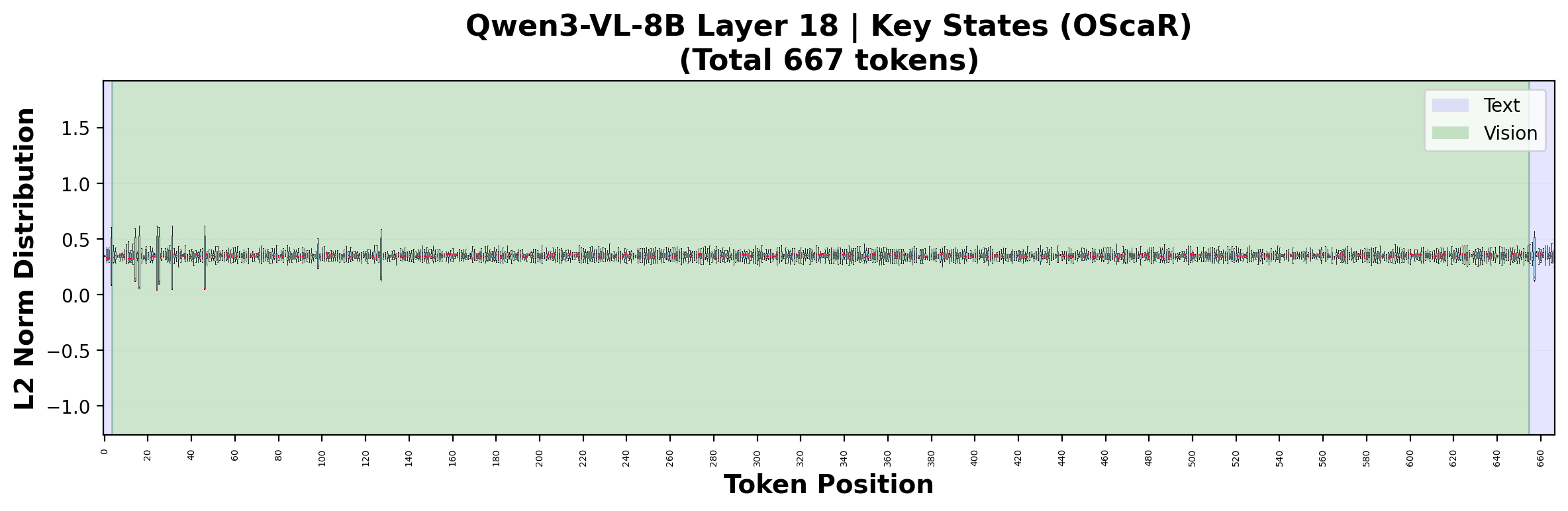}
        \caption{Qwen3-VL-8B (after OScaR).}
    \end{subfigure}
    \caption{Token norm distribution on Qwen3-VL-8B before and after applying OScaR.}
    \label{fig:ba_oscar_qwen_3_vl}
\end{figure}

\begin{figure}[t]
    \centering
    \begin{subfigure}[b]{0.48\textwidth}
        \centering
        \includegraphics[width=\textwidth]{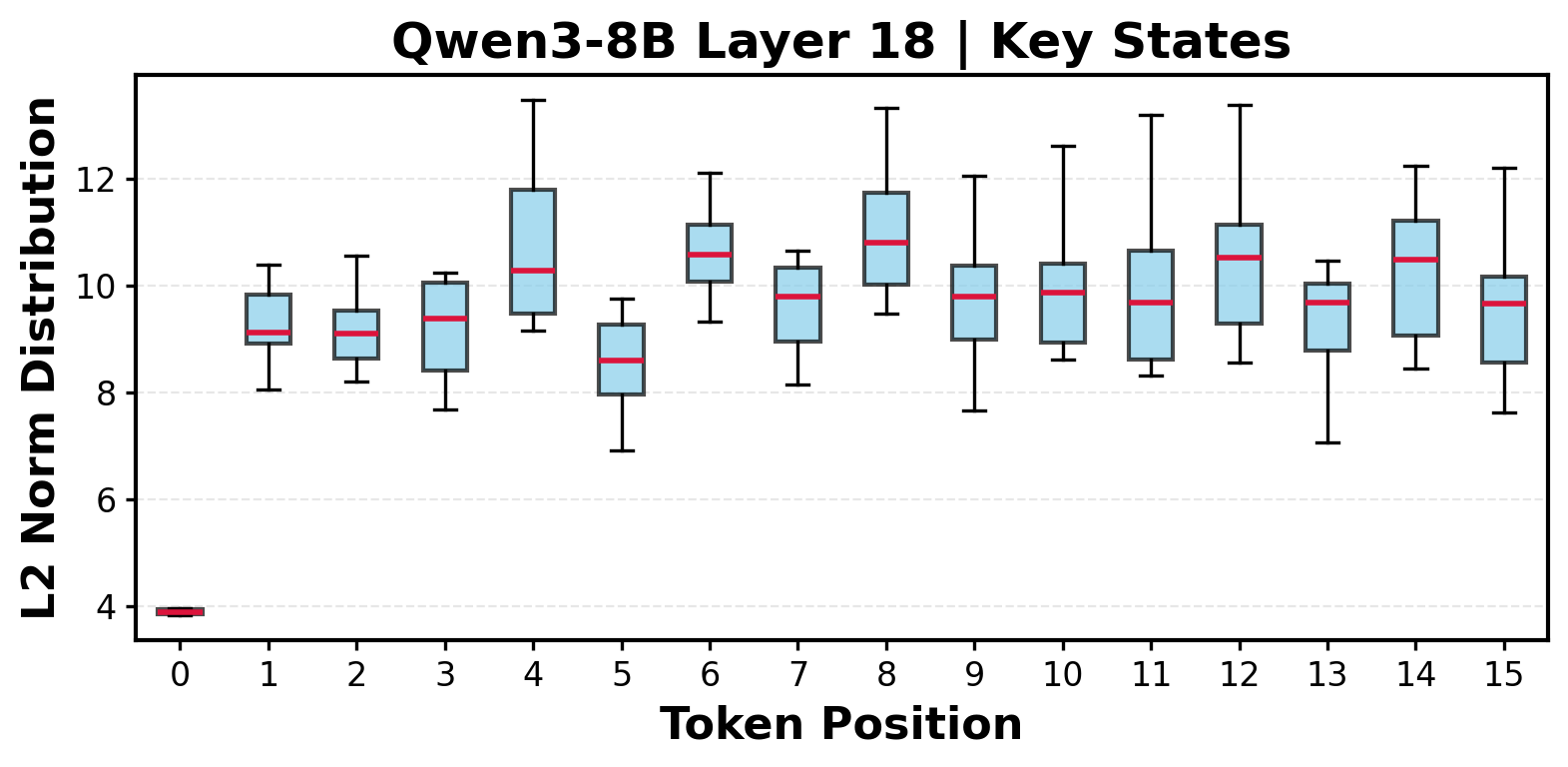}
        \caption{Qwen3-8B Layer 18 (before OScaR).}
    \end{subfigure}
    \begin{subfigure}[b]{0.48\textwidth}
        \centering
        \includegraphics[width=\textwidth]{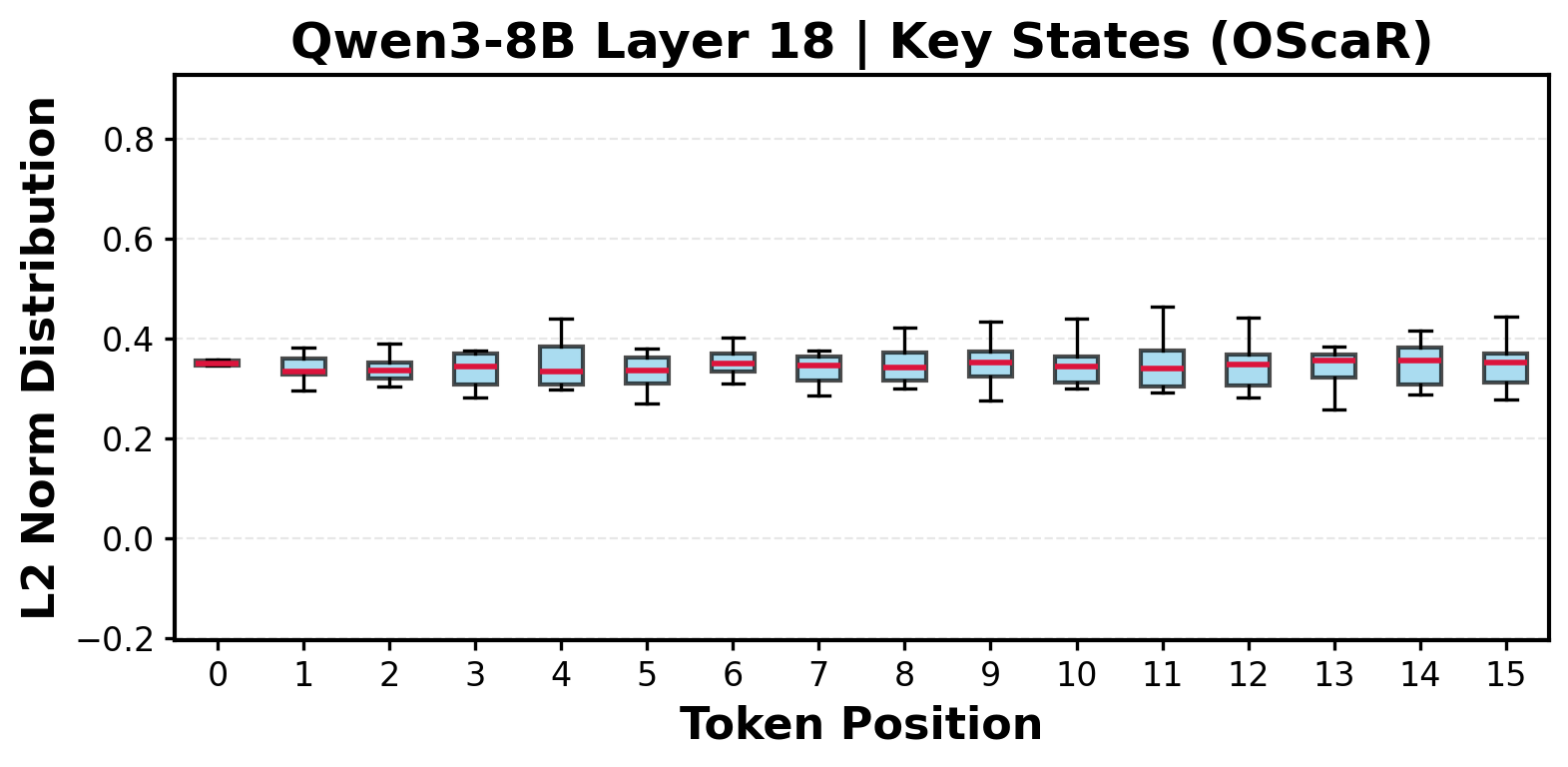}
        \caption{Qwen3-8B Layer 18 (after OScaR).}
    \end{subfigure}
    \begin{subfigure}[b]{0.48\textwidth}
        \centering
        \includegraphics[width=\textwidth]{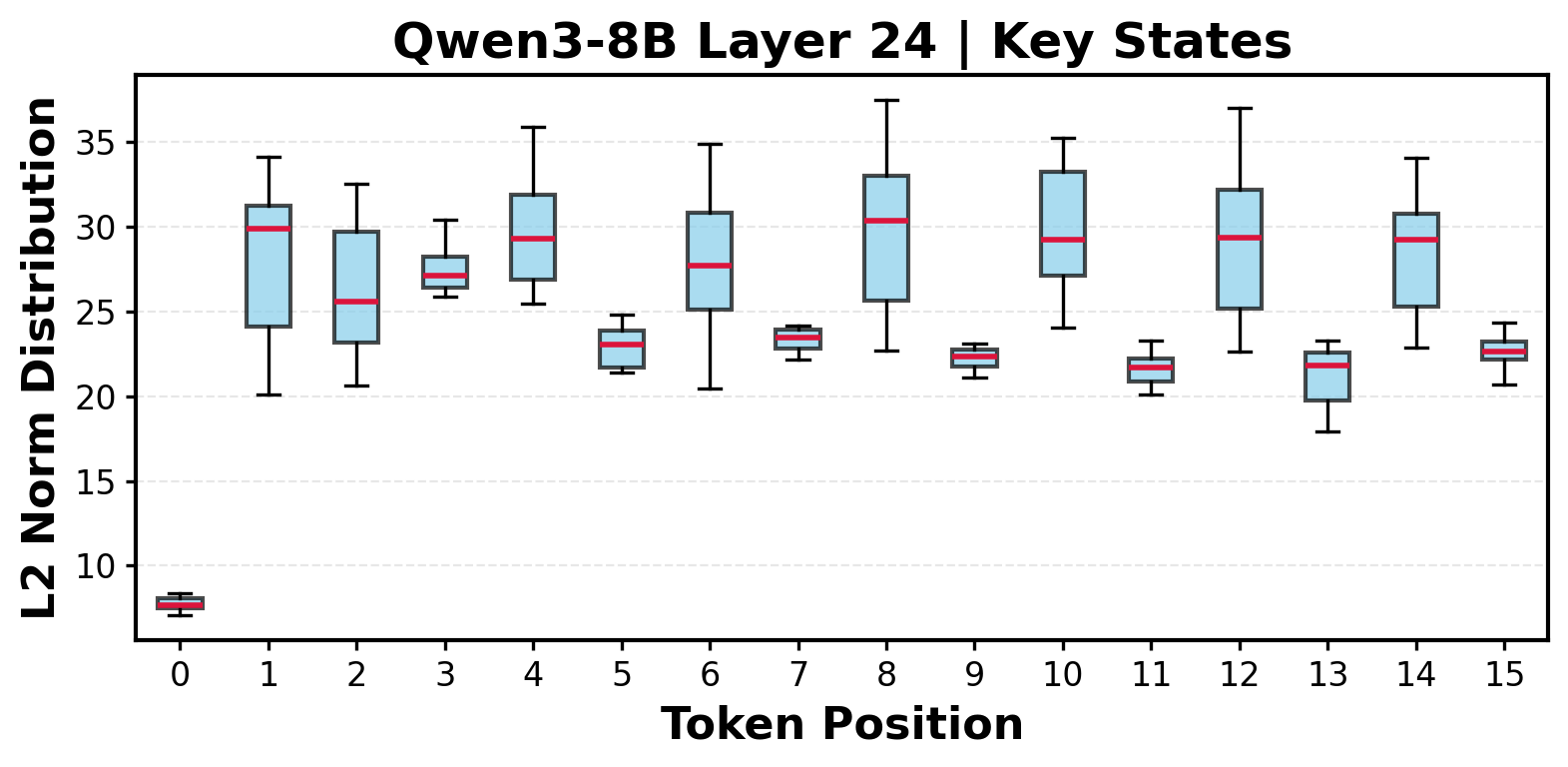}
        \caption{Qwen3-8B Layer 24 (before OScaR).}
    \end{subfigure}
    \begin{subfigure}[b]{0.48\textwidth}
        \centering
        \includegraphics[width=\textwidth]{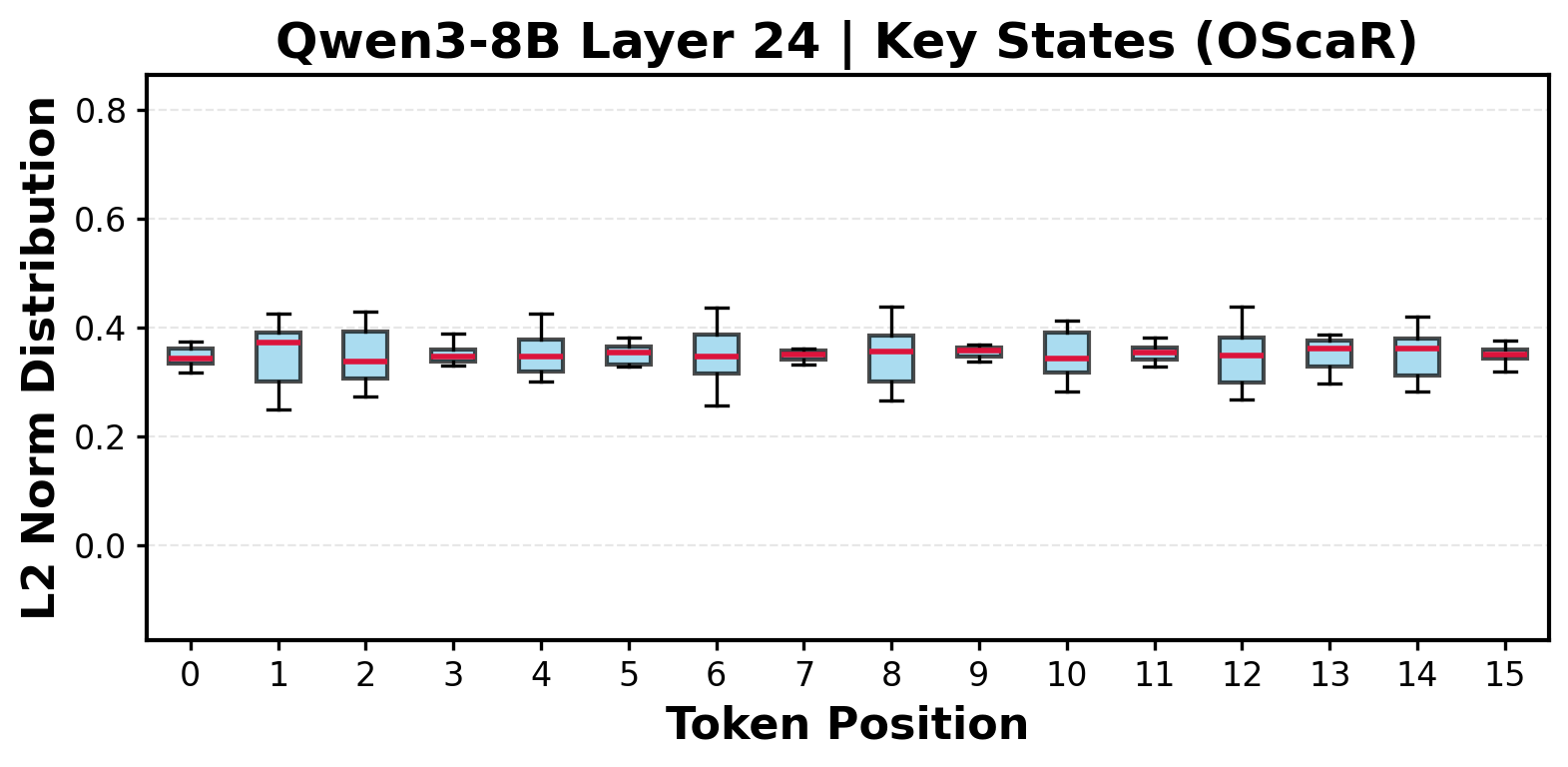}
        \caption{Qwen3-8B Layer 24 (after OScaR).}
    \end{subfigure}
    \caption{Token norm distribution on Qwen3-8B before and after applying OScaR.}
    \label{fig:ba_oscar_qwen_3}
\end{figure}

\begin{figure}[t]
    \centering
    \begin{subfigure}[b]{1\textwidth}
        \centering
        \includegraphics[width=\textwidth]{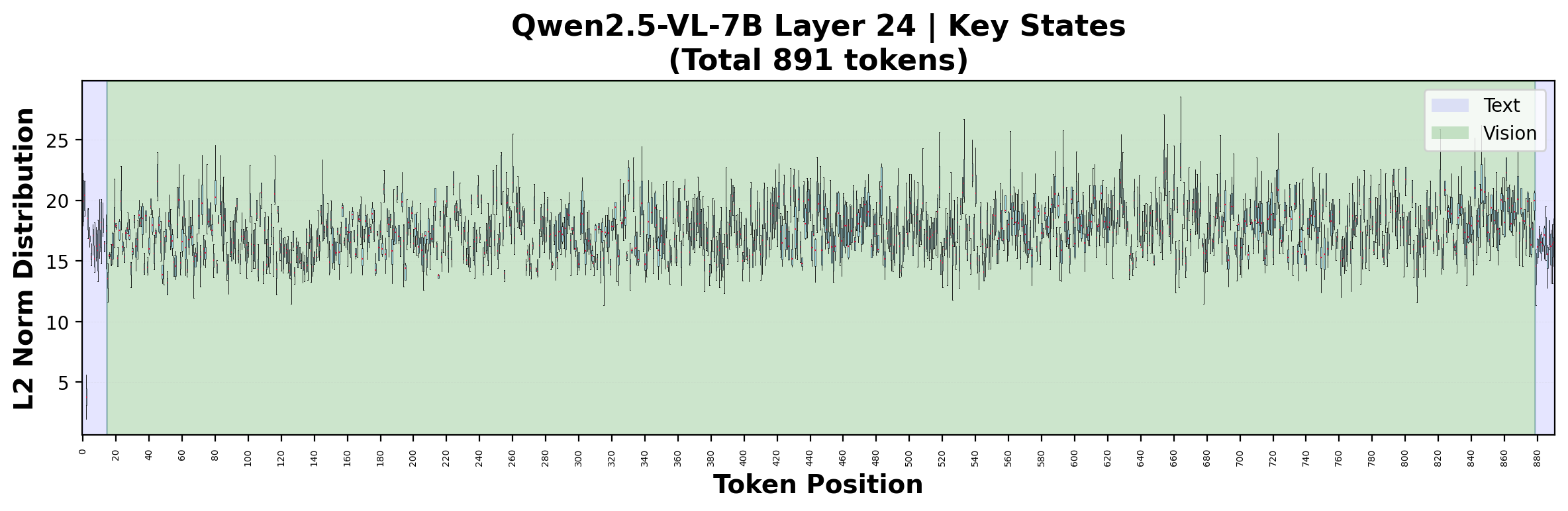}
        \caption{Qwen2.5-VL-7B (before OScaR).}
    \end{subfigure}
    \begin{subfigure}[b]{1\textwidth}
        \centering
        \includegraphics[width=\textwidth]{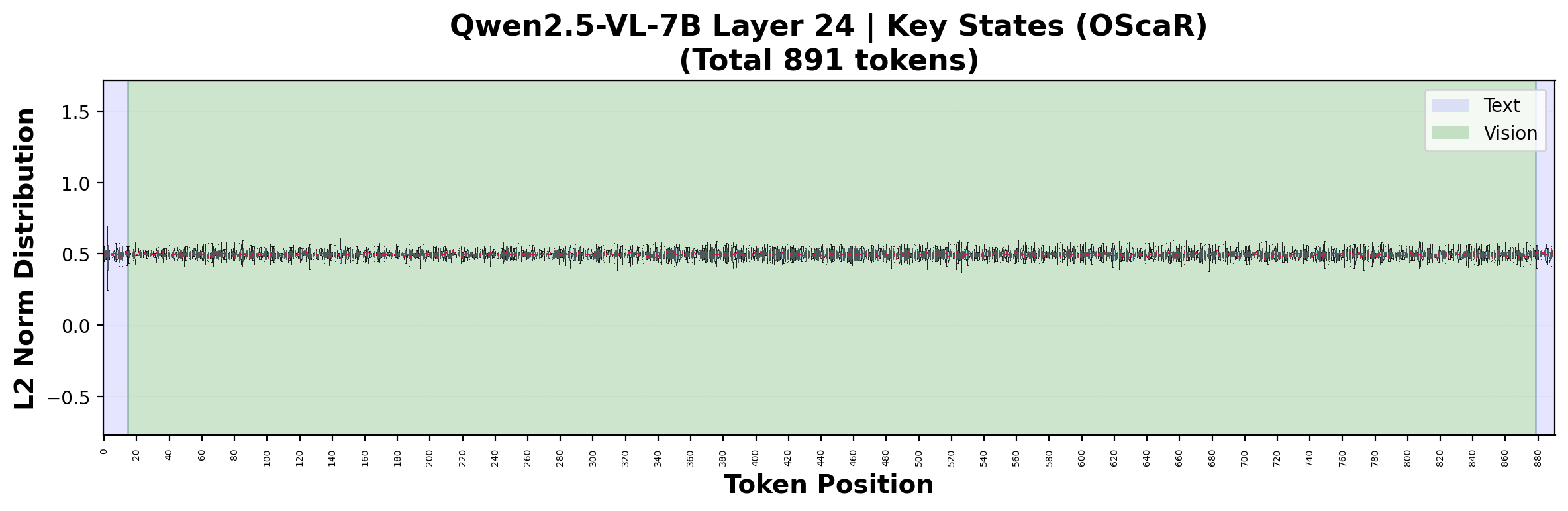}
        \caption{Qwen2.5-VL-7B (after OScaR).}
    \end{subfigure}
    \caption{Token norm distribution on Qwen2.5-VL-7B before and after applying OScaR.}
    \label{fig:ba_oscar_qwen_2_5_vl}
\end{figure}

To empirically verify the effectiveness of OScaR in mitigating TNI, we visualize token norm distributions across several models before and after applying OScaR. As shown in Figures~\ref{fig:ba_oscar_llama-3},~\ref{fig:ba_oscar_qwen_3_vl},~\ref{fig:ba_oscar_qwen_3} and~\ref{fig:ba_oscar_qwen_2_5_vl}, OScaR consistently alleviates TNI, transforming scattered norm distributions into more compact and balanced patterns across different models and modalities. These visualizations provide strong empirical support for OScaR's superior performance under extreme KV cache quantization.
\section{Ablation Study}
\label{app:ablation}

In this section, we present ablation studies examining the contributions of each proposed component. Table~\ref{tab:ablation_components} dissects the effect of our two core innovations: \textit{Canalized Rotation} and \textit{Omni-Token Scaling}. Applying \textit{Omni-Token Scaling} after \textit{Canalized Rotation} substantially recovers accuracy from the severely degraded INT2 baseline. In contrast, applying direct token-wise scaling alone without \textit{Canalized Rotation} further harms performance, aligning with the analysis in Section~\ref{sec:OScaR}.

\begin{table}[h]
\caption{Ablation of core components in OScaR on the WorldSense benchmark \citep{hong2025worldsense}. ``INT2 KCVT'' denotes per-channel Key and per-token Value quantization with 2-bit integers. 
The quantization group size is configured as 128.
The complete OScaR configuration (last row) achieves the best trade-off, recovering accuracy from the degraded INT2 baseline.}
\label{tab:ablation_components}
\resizebox{\textwidth}{!}{
\begin{tabular}{@{}lllcc@{}}
\toprule
\textbf{WorldSense} & \textbf{} & \textbf{} & \textbf{Qwen2.5-Omni-3B} & \textbf{Qwen2.5-Omni-7B} \\ \midrule
+ INT2 KCVT &  &  & 22.64 & 39.62 \\
+ INT2 KCVT &  & + Omni-Token Scaling & 17.61 & 35.22 \\
+ INT2 KCVT & + Canalized Rotation &  & 37.74 & 41.51 \\ \midrule
\textbf{+ INT2 KCVT} & \textbf{+ Canalized Rotation} & \textbf{+ Omni-Token Scaling} & \textbf{38.36} & \textbf{42.77} \\ \bottomrule
\end{tabular}
}
\end{table}

\paragraph{Conditional complementarity.}
On Qwen2.5-Omni-7B, OTS changes plain INT2 accuracy from 39.62 to 35.22, whereas adding OTS after CR changes 41.51 to 42.77. This reversal of OTS's marginal effect supports conditional complementarity: rotation preserves token norms, while scaling requires suitable channel geometry to avoid the artifact analyzed in Appendix~\ref{app:scaling_artifact}.

We further investigate alternative normalization strategies for computing the scaling coefficient in \textit{Omni-Token Scaling}: 
\begin{itemize}
    \item \textbf{$\ell_2$ norm:} uses the Euclidean norm of each token's hidden representation.
    \item \textbf{Rsqrt:} employs the hardware-accelerated reciprocal square root instruction to efficiently approximate $1 / \sqrt{\sum x_i^2}$, offering faster computation than explicit $\ell_2$ norm calculation.
    \item \textbf{Max:} uses the maximum absolute value across channels of each token.
    \item \textbf{Mean absolute value:} averages the absolute values across channels.
\end{itemize}

\begin{table}[h]
\caption{Comparison of alternative scaling coefficient strategies for \textit{Omni-Token Scaling} on the LongBench-E benchmark \citep{bai2024longbench}.}
\label{tab:scaling_strategies}
\centering
\resizebox{0.65\textwidth}{!}{
\begin{tabular}{@{}lcccc@{}}
\toprule
\textbf{LongBench-E} & \textbf{$\ell_2$ norm} & \textbf{Rsqrt} & \textbf{Max} & \textbf{Mean absolute value} \\ \midrule
Llama-2-7B & 29.59 & 29.57 & 29.23 & 29.68 \\
Llama-3-8B & 38.56 & 38.63 & 33.68 & 38.53 \\
Qwen2.5-7B & 42.57 & 42.83 & 14.47 & 41.95 \\
Qwen3-8B & 42.85 & 42.91 & 25.12 & 42.64 \\ \bottomrule
\end{tabular}
}
\end{table}

As shown in Table~\ref{tab:scaling_strategies}, the $\ell_2$ norm and its rsqrt approximation achieve the best and mutually comparable performance across all evaluated model families. Heuristics such as Max cause severe degradation: on Qwen2.5-7B, Max collapses accuracy to 14.47 compared to 42.57 with the $\ell_2$ norm. The mean absolute value, while not optimal, performs competitively and avoids such catastrophic drops. Given the negligible difference between $\ell_2$ norm and rsqrt, we adopt the rsqrt-based implementation in our final setup due to its superior hardware efficiency and lower latency.

\section{Accuracy-Efficiency Pareto Front Analysis}
\label{app:pareto}

In this section, we analyze the trade-off between computational efficiency and task accuracy using the theoretical decode cost per step from Table~\ref{tab:complexity_numerical} (Appendix~\ref{app:complexity}) and the LongBench-E scores of Qwen3-8B from Table~\ref{tab:longbench}. We compare OScaR against several state-of-the-art methods, including KIVI and TurboQuant+ (as detailed in Appendix~\ref{app:turboquant}). Figure~\ref{fig:pareto_frontier} visualizes the accuracy against decode cost for each method.

\begin{itemize}
    \item \textbf{KIVI} establishes the efficiency baseline. With the lowest decode cost among all methods, it incurs no rotation or normalization overhead, achieving an accuracy of 47.95. This reflects the inherent trade-off of minimal preprocessing.
    
    \item \textbf{TurboQuant+} has a decode cost of 247.9 million units, compared with 81.9 for KIVI and 123.0 for OScaR (Table~\ref{tab:complexity_numerical}). Its LongBench-E average is 47.56, below KIVI's 47.95, so KIVI dominates it in this analytical cost--accuracy comparison.
    
    \item \textbf{OScaR} achieves a favorable balance between efficiency and accuracy. Its decode cost is 1.5 times that of KIVI but less than half that of TurboQuant+, while delivering the highest accuracy among all quantized methods.
\end{itemize}

Overall, OScaR occupies a distinct and advantageous position on the Pareto front, offering a favorable combination of competitive computational cost and strong accuracy.

\begin{figure}[t]

\centering
\includegraphics[width=0.9\textwidth]{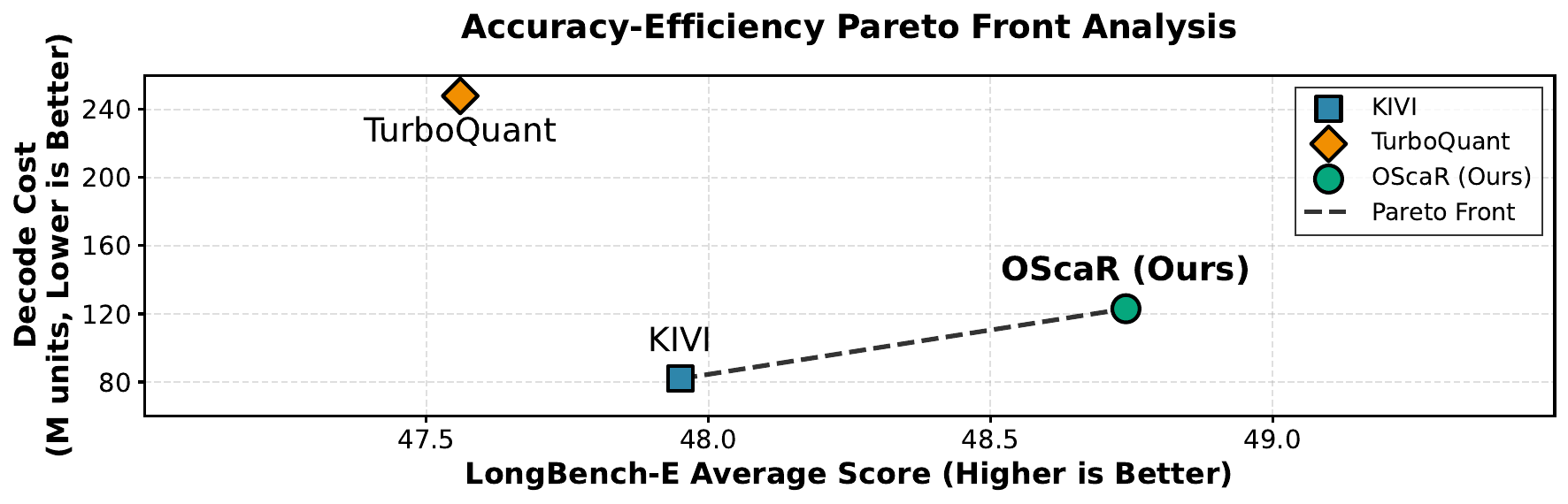}
\caption{Pareto front analysis of KV cache quantization methods on Qwen3-8B. The x-axis represents the average LongBench-E accuracy (higher is better), and the y-axis represents the decode cost in million units (lower is better). OScaR achieves the highest accuracy with competitive efficiency, occupying a distinct and advantageous position on the Pareto front.}
\label{fig:pareto_frontier}
\end{figure}

\section{Additional Decoding Efficiency Measurements}
\label{app:turboquant_decoding_efficiency}

\begin{table}[h]
\centering
\caption{Decode latency comparison on Qwen3-8B with a single H20 GPU (141GB), measured in milliseconds per token.}
\label{tab:decode_efficiency}
\resizebox{0.8\textwidth}{!}{
\begin{tabular}{@{}cccc@{}}
\toprule
\textbf{Context Length} & \textbf{FlashDecoding-v2 (ms/tok)} & \textbf{OScaR (ms/tok)} & \textbf{TurboQuant+ (ms/tok)} \\ \midrule
1K & 19.5 & 25.1 & 7.8 \\
2K & 19.8 & 26.4 & 8.5 \\
4K & 20.3 & 24.9 & 9.6 \\
8K & 23.8 & 24.9 & 11.7 \\
16K & 28.3 & 24.1 & 15.7 \\
32K & 38.0 & 25.8 & 23.9 \\
48K & 47.1 & 25.7 & 32.1 \\
64K & 56.3 & 25.3 & 40.2 \\
96K & 74.6 & 28.5 & 56.4 \\
128K & 92.9 & 30.9 & 72.9 \\ \bottomrule
\end{tabular}
}
\end{table}

Table~\ref{tab:decode_efficiency} reports the decoding latency of OScaR and TurboQuant+ alongside the \texttt{BF16 FlashDecoding-v2} baseline across context lengths ranging from 1K to 128K tokens. OScaR is implemented on the PyTorch runtime, while TurboQuant+ relies on \texttt{llama.cpp}; this comparison measures system performance and does not isolate algorithmic overhead. OScaR exhibits remarkably stable performance across the entire range, with latency increasing only modestly from 24.1 ms/tok at 16K to 30.9 ms/tok at 128K. In contrast, TurboQuant+ shows strong context dependence: while it achieves lower latency than OScaR at short contexts (e.g., 7.8 ms/tok at 1K), its latency grows rapidly, reaching 72.9 ms/tok at 128K.

OScaR is slower than BF16 at short contexts but becomes faster as KV-cache traffic grows. At 128K tokens, OScaR achieves a 3.0$\times$ speedup (92.9 ms/tok vs. 30.9 ms/tok), whereas TurboQuant+ attains only a modest 1.3$\times$ speedup at the same context length.

\begingroup
\setlength{\intextsep}{6pt plus 1pt minus 1pt}
\setlength{\textfloatsep}{10pt plus 2pt minus 2pt}
\setlength{\floatsep}{6pt plus 1pt minus 1pt}
\subsection{Controlled end-to-end and component-wise latency}
\label{app:controlled_latency}
We benchmark Qwen3-8B on one A800 80GB GPU at batch size 1 with identical inputs and positions, generating 320 tokens at 4K, 16K, 64K, and 128K contexts. Values are medians of five requests after warmup. Time to first token (TTFT) includes prefill, KV packing, the LM head, and first-token selection; time per output token (TPOT) averages the subsequent 319 steps, including cache update/flush and attention.

The KIVI-style INT2 control uses per-channel Keys and per-token Values on the same BitDecoding backend as OScaR, with CR and OTS disabled; it is not the official KIVI runtime. CR-only and OTS-only enable each component separately. OTS-only attributes cost; its accuracy limitation appears in Table~\ref{tab:ablation_components}.

\begin{table}[htbp]
\centering
\caption{Controlled end-to-end latency on an A800 80GB GPU. Each cell is TTFT / TPOT in milliseconds.}
\label{tab:controlled_end_to_end}
\small
\begin{tabular}{@{}lrrr@{}}
\toprule
Context & BF16 FlashDecoding & KIVI-style INT2 & OScaR INT2 \\
\midrule
4K & 366.30 / 28.94 & 396.38 / 32.25 & 400.25 / 34.09 \\
16K & 1,695.18 / 29.80 & 1,772.77 / 32.15 & 1,790.27 / 33.89 \\
64K & 11,649.19 / 54.29 & 11,881.14 / 34.36 & 11,882.33 / 35.38 \\
128K & 37,786.62 / 91.40 & 38,125.83 / 47.89 & 37,921.31 / 48.90 \\
\bottomrule
\end{tabular}
\end{table}

\begin{table}[htbp]
\centering
\caption{Component-wise latency on the same A800/BitDecoding setup. Each cell is TTFT / TPOT in milliseconds.}
\label{tab:controlled_components}
\small
\setlength{\tabcolsep}{4pt}
\begin{tabular}{@{}lrrrr@{}}
\toprule
Context & KIVI-style INT2 & CR-only & OTS-only & OScaR \\
\midrule
4K & 396.38 / 32.25 & 403.20 / 33.07 & 403.51 / 33.86 & 400.25 / 34.09 \\
16K & 1,772.77 / 32.15 & 1,817.04 / 33.13 & 1,820.35 / 33.61 & 1,790.27 / 33.89 \\
64K & 11,881.14 / 34.36 & 12,192.38 / 35.08 & 12,221.89 / 35.01 & 11,882.33 / 35.38 \\
128K & 38,125.83 / 47.89 & 39,580.81 / 48.56 & 39,073.33 / 48.40 & 37,921.31 / 48.90 \\
\bottomrule
\end{tabular}
\end{table}

We use TPOT to attribute online overhead because TTFT includes full prefill and varies more between runs. Relative to the KIVI-style control, CR-only adds 2.5\%/3.1\%/2.1\%/1.4\% TPOT at 4K/16K/64K/128K; OTS-only adds 5.0\%/4.5\%/1.9\%/1.1\%; and full OScaR adds 5.7\%/5.4\%/3.0\%/2.1\%. Shared fused paths mean isolated costs need not add linearly. OTS-only jointly measures normalization, metadata handling, and inverse scaling, rather than timing each operation separately.

At 4K/16K, the fixed INT2 kernel costs are not yet amortized. At 64K/128K, OScaR reduces TPOT relative to BF16 by 34.8\%/46.5\%; at 128K its TTFT is within 0.4\% of BF16. These A800 measurements are separate from the H20 measurements in Section~\ref{sec:efficiency} and Table~\ref{tab:decode_efficiency}.
\endgroup

\newpage
\begin{algorithm}[t]
\SetKwInput{KwParam}{parameter}
\KwParam{group size $G$, residual length $R$, head dimension $d_h$}
\SetKwProg{myProc}{Procedure}{\string:}{end}
\SetKwProg{myFunc}{Function}{\string:}{end}

\myProc{\FuncSty{OScaR-Preprocess}}{
  \KwIn{$\mathbf{W}_V \in \mathbb{R}^{d_h \times d_h}$, $\mathbf{W}_O \in \mathbb{R}^{d_h \times d_h}$}
  $\mathbf{H} \leftarrow$ Hadamard matrix of size $d_h \times d_h$ \\
  $\mathbf{W}_V \leftarrow \mathbf{W}_V \mathbf{H}$, $\quad \mathbf{W}_O \leftarrow \mathbf{H} \mathbf{W}_O$ \\
  \Return{$\mathbf{W}_V, \mathbf{W}_O$}
}

\myProc{\FuncSty{OScaR-Inference}}{
  \KwIn{Input $\mathbf{X}$, KV cache (empty for prefill)}
  \KwOut{Output $\mathbf{o}$}
  
  $\mathbf{X}_Q = \mathbf{X} \mathbf{W}_Q, \quad \mathbf{X}_K = \mathbf{X} \mathbf{W}_K, \quad \mathbf{X}_V = \mathbf{X} \mathbf{W}_V$ \\
  $\mathbf{X}_Q \leftarrow \text{FHT}(\mathbf{X}_Q), \quad \mathbf{X}_K \leftarrow \text{FHT}(\mathbf{X}_K)$ \\
  $\mathbf{s}_K \leftarrow \|\mathbf{X}_K\|_2,\quad \mathbf{X}_K \leftarrow \mathbf{X}_K / \mathbf{s}_K$ \\
  
  $Q(\mathbf{X}_{K}^{hist}), \mathbf{X}_{K_r}^{hist}, \mathbf{s}_{K_g}^{hist}, \mathbf{s}_{K_r}^{hist}, Q(\mathbf{X}_{V}^{hist}), \mathbf{X}_{V_r}^{hist} \leftarrow \texttt{KV cache}$ \\
  
  $\mathbf{K}_{all} \leftarrow \text{Concat}([\text{Dequant}(Q(\mathbf{X}_{K}^{hist})) \cdot \mathbf{s}_{K_g}^{hist},\; \mathbf{X}_{K_r}^{hist} \cdot \mathbf{s}_{K_r}^{hist},\; \mathbf{X}_K \cdot \mathbf{s}_K])$ \\
  $\mathbf{V}_{all} \leftarrow \text{Concat}([\text{Dequant}(Q(\mathbf{X}_{V}^{hist})),\; \mathbf{X}_{V_r}^{hist},\; \mathbf{X}_V])$ \\
  
  $\mathbf{o} \leftarrow \text{Attention}(\mathbf{X}_Q, \mathbf{K}_{all}, \mathbf{V}_{all}) \mathbf{W}_O$ \\
  
  $Q(\mathbf{X}_K^{curr}), \mathbf{X}_{K_r}^{curr}, \mathbf{s}_{K_g}^{curr}, \mathbf{s}_{K_r}^{curr} \leftarrow \FuncSty{BufferQuantK}(\mathbf{X}_K, \mathbf{s}_K, \mathbf{X}_{K_r}^{hist}, \mathbf{s}_{K_r}^{hist})$ \\
  $Q(\mathbf{X}_V^{curr}), \mathbf{X}_{V_r}^{curr} \leftarrow \FuncSty{BufferQuantV}(\mathbf{X}_V, \mathbf{X}_{V_r}^{hist})$ \\
  \texttt{KV cache} $\leftarrow Q(\mathbf{X}_K^{curr}), \mathbf{X}_{K_r}^{curr}, \mathbf{s}_{K_g}^{curr}, \mathbf{s}_{K_r}^{curr}, Q(\mathbf{X}_V^{curr}), \mathbf{X}_{V_r}^{curr}$ \\
  
  \Return{$\mathbf{o}$}
}

\myFunc{\FuncSty{BufferQuantK}($\mathbf{M}, \mathbf{s}, \mathbf{M}_r, \mathbf{s}_r$)}{
  \uIf{prefill (cache empty)}{
    $r = \text{len}(\mathbf{M}) \% R$ \\
    $\mathbf{M}_g \leftarrow \mathbf{M}[:-r], \quad \mathbf{M}_r \leftarrow \mathbf{M}[-r:]$ \\
    $\mathbf{s}_g \leftarrow \mathbf{s}[:-r], \quad \mathbf{s}_r \leftarrow \mathbf{s}[-r:]$ \\
    $\mathbf{M}_g^{quant} \leftarrow$ GroupQuant$(\mathbf{M}_g, \text{dim=channel}, \text{numGroup}=\text{len}(\mathbf{M}_g) // G)$ \\
    \Return{$\mathbf{M}_g^{quant}, \mathbf{M}_r, \mathbf{s}_g, \mathbf{s}_r$}
  }
  \uElse{
    Append $\mathbf{M}$ to $\mathbf{M}_r$, append $\mathbf{s}$ to $\mathbf{s}_r$ \\
    \If{$\text{len}(\mathbf{M}_r) = R$}{
      $\mathbf{M}_r^{quant} \leftarrow$ GroupQuant$(\mathbf{M}_r, \text{dim=channel}, \text{numGroup}=R // G)$ \\
      $\mathbf{M}_g \leftarrow \text{Concat}(\mathbf{M}_g, \mathbf{M}_r^{quant}), \quad \mathbf{s}_g \leftarrow \text{Concat}(\mathbf{s}_g, \mathbf{s}_r)$ \\
      $\mathbf{M}_r \leftarrow \emptyset, \quad \mathbf{s}_r \leftarrow \emptyset$
    }
    \Return{$\mathbf{M}_g, \mathbf{M}_r, \mathbf{s}_g, \mathbf{s}_r$}
  }
}

\myFunc{\FuncSty{BufferQuantV}($\mathbf{M}, \mathbf{M}_r$)}{
  \uIf{prefill (cache empty)}{
    $r = \text{len}(\mathbf{M}) \% R$ \\
    $\mathbf{M}_g \leftarrow \mathbf{M}[:-r], \quad \mathbf{M}_r \leftarrow \mathbf{M}[-r:]$ \\
    $\mathbf{M}_g^{quant} \leftarrow$ GroupQuant$(\mathbf{M}_g, \text{dim=token}, \text{numGroup}=d_h//G)$ \\
    \Return{$\mathbf{M}_g^{quant}, \mathbf{M}_r$}
  }
  \uElse{
    Append $\mathbf{M}$ to $\mathbf{M}_r$ \\
    \If{$\text{len}(\mathbf{M}_r) = R$}{
      $\mathbf{M}_r^{quant} \leftarrow$ GroupQuant$(\mathbf{M}_r, \text{dim=token}, \text{numGroup}=d_h//G)$ \\
      $\mathbf{M}_g \leftarrow \text{Concat}(\mathbf{M}_g, \mathbf{M}_r^{quant})$ \\
      $\mathbf{M}_r \leftarrow \emptyset$
    }
    \Return{$\mathbf{M}_g, \mathbf{M}_r$}
  }
}

\caption{The OScaR algorithm.}
\label{algo:oscar}
\end{algorithm}
\newpage
\clearpage

\newpage
\clearpage

\newpage

\newpage

\newpage
\clearpage

\newpage

\end{document}